\documentclass[twoside,11pt]{article}

\usepackage{blindtext}
\usepackage{tikz} % For drawing the color blocks
\usepackage{xcolor} % For defining custom colors
\usepackage{colortbl}

\definecolor{Gray}{HTML}{BAB0AC}         % A
\definecolor{CoralOrange}{HTML}{FF7043}  % B
\definecolor{VibrantGreen}{HTML}{66BB6A} % C
\definecolor{RichRed}{HTML}{E53935}      % D
\definecolor{SoftIndigo}{HTML}{5C6BC0}   % E
\definecolor{EmeraldTeal}{HTML}{009688}  % F
\definecolor{VividPurple}{HTML}{8E44AD}  % G
\definecolor{BrightCyan}{HTML}{26C6DA}   % H
\definecolor{FreshLime}{HTML}{AEE637}    % I
\definecolor{WarmAmber}{HTML}{F4B400}    % J

\usepackage{JMLR/jmlr2e}
\usepackage{booktabs}
\usepackage{longtable}
\usepackage{hyperref}
\usepackage{amssymb}
\usepackage{algorithm}
\usepackage{algorithmic}
\usepackage{textcomp}
\usepackage{microtype}
\usepackage{graphicx}
\usepackage{booktabs}
\usepackage{hyperref}
\usepackage{mathtools}
\usepackage[capitalize,noabbrev]{cleveref}
\usepackage[title]{appendix}
\usepackage{subcaption}
\usepackage{wrapfig}
\usepackage[utf8]{inputenc} % allow utf-8 input
\usepackage[T1]{fontenc}    % use 8-bit T1 fonts
\usepackage{url}            % simple URL typesetting
\usepackage{xspace}
\usepackage[export]{adjustbox}
\usepackage{nicefrac}       % compact symbols for 1/2, etc.
\usepackage{microtype}      % microtypography
\usepackage{wasysym}
\usepackage{enumitem}
\usepackage[american]{babel}
\usepackage{multirow}
\usepackage{adjustbox}
\usepackage{hyperref}       % hyperlinks
\usepackage{booktabs}       % professional-quality tables
\usepackage{xcolor}         % colors
\usepackage{longtable}
\usepackage{csquotes}
\usepackage[htt]{hyphenat}
\hyphenchar\font=`\-

\usepackage{lastpage}
\ShortHeadings{A Closer Look at Deep Learning Methods on Tabular Datasets
}{Ye, Liu, Cai, Zhou, and Zhan}
\firstpageno{1}

\makeatletter
\DeclareRobustCommand\onedot{\futurelet\@let@token\@onedot}
\def\@onedot{\ifx\@let@token.\else.\null\fi\xspace}

\def\eg{\emph{e.g}\onedot} 
\def\ie{\emph{i.e}\onedot}

\def\wrt{w.r.t\onedot} 

\makeatother

\begin{document}

\title{A Closer Look at Deep Learning Methods on Tabular Datasets}

\author{\name Han-Jia Ye \email yehj@lamda.nju.edu.cn\\
        \name Si-Yang Liu$^*$ \email liusy@lamda.nju.edu.cn \\
    	\name Hao-Run Cai$^*$ \email caihr@lamda.nju.edu.cn \\
    	\name Qi-Le Zhou \email zhouql@lamda.nju.edu.cn \\
     	\name De-Chuan Zhan \email zhandc@lamda.nju.edu.cn \\
        \addr School of Artificial Intelligence, Nanjing University, China \\
        \addr National Key Laboratory for Novel Software Technology, Nanjing University, 210023, China\\
}

\editor{My editor}

%%% load AMS-Latex Package

% \newcommand\figcaption{\def\@captype{figure}\caption}
% \newcommand\tabcaption{\def\@captype{table}\caption}

\newcommand{\twoline}[2]{\vtop{\hbox{\strut #1}\hbox{\strut #2}}}
\newcommand{\dc}{\textcolor{cyan}{\Delta_{\text{CL-FL}}}}
\newcommand{\dr}{\textcolor{magenta}{\Delta_{\text{rand}}}}

% define vector and matrix symbols
\newcommand{\vct}[1]{\boldsymbol{#1}} % vector
\newcommand{\mat}[1]{\boldsymbol{#1}} % matrix
\newcommand{\cst}[1]{\mathsf{#1}}  % constant

%%%% Special math symbols
\newcommand{\field}[1]{\mathbb{#1}}
\newcommand{\R}{\field{R}} % real domain
\newcommand{\C}{\field{C}} % complex domain
\newcommand{\F}{\field{F}} % functional domain
\newcommand{\I}{\field{I}} % functional domain
\newcommand{\T}{^{\textrm T}} % transpose

%% operator in linear algebra, functional analysis
\newcommand{\inner}[2]{#1\cdot #2}
\newcommand{\twonorm}[1]{\left\|#1\right\|_2^2}
\newcommand{\onenorm}[1]{\|#1\|_1}
\newcommand{\Map}[1]{\mathcal{#1}}  % operator in functions, maps such as M: domain1 --> domain 2

% operator in probability: expectation, covariance, 
\newcommand{\ProbOpr}[1]{\mathbb{#1}}
% independence
\newcommand\independent{\protect\mathpalette{\protect\independenT}{\perp}}
\def\independenT#1#2{\mathrel{\rlap{$#1#2$}\mkern2mu{#1#2}}}
\newcommand{\ind}[2]{{#1} \independent{#2}}
\newcommand{\cind}[3]{{#1} \independent{#2}\,|\,#3} % conditional independence
\newcommand{\expect}[2]{%
\ifthenelse{\equal{#2}{}}{\ProbOpr{E}_{#1}}
{\ifthenelse{\equal{#1}{}}{\ProbOpr{E}\left[#2\right]}{\ProbOpr{E}_{#1}\left[#2\right]}}} % Expectation: syntax: E{1}{2} = E_1[2], E{}{2}=E[2], E{1}{} = E_1
\newcommand{\cndexp}[2]{\ProbOpr{E}\,[ #1\,|\,#2\,]}  % conditional expectation

\newcommand{\name}{{\sc Talent}}

% special functions
%\newcommand{\trace}[1]{\operatornamewithlimits{tr}\left\{#1\right\}}
\newcommand{\diag}{\operatornamewithlimits{diag}}
\newcommand{\sign}{\operatornamewithlimits{sign}}
\newcommand{\const}{\operatornamewithlimits{const}}

% special display
\newcommand{\parde}[2]{\frac{\partial #1}{\partial  #2}}

% environment
% \newtheorem{thm}{Theorem}
% \newtheorem{theorem}{Theorem}
% \newtheorem{definition}{Definition}
% \newtheorem{lemma}[theorem]{Lemma}
% \newtheorem{conjecture}[theorem]{Conjecture}
% \newtheorem{proposition}[theorem]{Proposition}

% shorthand
\newcommand{\e}{{\vct{e}}}
\newcommand{\x}{{\vct{x}}}
\newcommand{\y}{\vct{y}}
\newcommand{\z}{{\vct{z}}}
\newcommand{\w}{\vct{w}}
\newcommand{\p}{\vct{p}}
\newcommand{\mphi}{{\boldsymbol{\phi}}}
\newcommand{\mgamma}{{\boldsymbol{\gamma}}}
\newcommand{\mGamma}{{\boldsymbol{\Gamma}}}
\newcommand{\mxi}{{\boldsymbol{\xi}}}
\newcommand{\mXi}{{\boldsymbol{\Xi}}}
\newcommand{\mtheta}{{\boldsymbol{\theta}}}
\newcommand{\mmu}{{\boldsymbol{\mu}}}

\newcommand{\vtheta}{\vct{\theta}}
\newcommand{\vTheta}{\vct{\Theta}}
\newcommand{\vmu}{\vct{\mu}}
\newcommand{\vlambda}{\vct{\lambda}}
\newcommand{\vc}{\vct{c}}
\newcommand{\vp}{\vct{p}}
\newcommand{\vb}{\vct{b}}
\newcommand{\vq}{\vct{q}}
\newcommand{\vo}{{\vct{o}}}
\newcommand{\vr}{\vct{r}}
\newcommand{\vt}{\vct{t}}
\newcommand{\vv}{\vct{v}}
\newcommand{\vzero}{\vct{0}}
\newcommand{\vm}{\vct{m}}
\newcommand{\vf}{\vct{f}}
\newcommand{\vh}{\vct{h}}
\newcommand{\vg}{\vct{g}}
\newcommand{\vphi}{\vct{\phi}}
\newcommand{\vpsi}{\vct{\psi}}
\newcommand{\ones}{\vct{1}}
\newcommand{\mU}{\mat{U}}
\newcommand{\mA}{\mat{A}}
\newcommand{\mB}{\mat{B}}
\newcommand{\mD}{\mat{D}}
\newcommand{\mE}{\mat{E}}
\newcommand{\mW}{\mat{W}}
\newcommand{\mH}{\mat{H}}
\newcommand{\mS}{\mat{S}}
\newcommand{\mJ}{\mat{J}}
\newcommand{\mM}{\mat{M}}
\newcommand{\mT}{\mat{T}}
\newcommand{\mZ}{\mat{Z}}
\newcommand{\mO}{\mat{O}}
\newcommand{\mY}{\mat{Y}}
\newcommand{\mL}{\mat{L}}
\newcommand{\mI}{\mat{I}}
\newcommand{\mK}{\mat{K}}
\newcommand{\mSigma}{\mat{\Sigma}}
\newcommand{\mOmega}{\mat{\Omega}}
\newcommand{\cC}{\cst{C}}
\newcommand{\cM}{\cst{M}}
\newcommand{\cN}{\cst{N}}
\newcommand{\cQ}{\cst{Q}}
\newcommand{\cD}{\cst{D}}
\newcommand{\cL}{\cst{L}}
\newcommand{\cK}{\cst{K}}
\newcommand{\cH}{\cst{H}}
\newcommand{\cR}{\cst{R}}
\newcommand{\cU}{\cst{U}}
\newcommand{\cS}{\cst{S}}
\newcommand{\sQ}{\mathcal{Q}}
\newcommand{\sS}{\mathcal{S}}
\newcommand{\sF}{\mathcal{F}}
\newcommand{\sC}{\mathcal{C}}
\newcommand{\sT}{\mathcal{T}}
\newcommand{\sD}{\mathcal{D}}
\newcommand{\sU}{\mathcal{U}}
\newcommand{\sL}{\mathcal{L}}
\newcommand{\sA}{\mathcal{A}}
\newcommand{\sN}{\mathcal{N}}
\newcommand{\sB}{\mathcal{B}}

\newcommand{\nx}{\tilde{x}}
\newcommand{\vnx}{{\tilde{\vx}}}
\newcommand{\vnz}{{\tilde{\vz}}}
\newcommand{\deltavx}{\delta_\vx}
\newcommand{\vmx}{\bar{\vx}}
\newcommand{\vmz}{\bar{\vz}}
\newcommand{\sigmax}{\mSigma_{\vx}}
\newcommand{\sigmaz}{\mSigma_{\vz}}
\newcommand{\no}{\tilde{o}}
\newcommand{\vno}{{\tilde{\vo}}}
\newcommand{\nell}{\tilde{\ell}}
\newcommand{\jacob}{\mat{J}}
\newcommand{\hess}{\mat{H}}
\newcommand{\mloss}{\hat{\ell}}

\newcommand{\eat}[1]{}
\newcommand{\method}[1]{\textsc{#1}}
\newcommand{\task}[1]{\textbf{#1}}

\newcommand{\bbR}{\mathbb{R}}

\newcommand{\bbE}{\mathbb{E}}
\newcommand{\bbV}{\mathbb{V}}

\newcommand{\mathI}{\mathcal{I}}

\newcommand{\mr}{{\mathrm{t}}}

\newcommand{\mP}{{\mathbf{P}}}

\maketitle
\def\thefootnote{*}\footnotetext{These authors contributed equally to this work}\def\thefootnote{\arabic{footnote}}
\begin{abstract}
Tabular data is prevalent across diverse domains in machine learning. With the rapid progress of deep tabular prediction methods, especially pretrained (foundation) models, there is a growing need to evaluate these methods systematically and to understand their behavior. We present an extensive study on {\name}, a collection of 300+ datasets spanning broad ranges of size, feature composition (numerical/categorical mixes), domains, and output types (binary, multi-class, regression).
Our evaluation shows that ensembling benefits both tree-based and neural approaches. Traditional gradient-boosted trees remain very strong baselines, yet recent pretrained tabular models now match or surpass them on many tasks, narrowing—but not eliminating—the historical advantage of tree ensembles. Despite architectural diversity, top performance concentrates within a small subset of models, providing practical guidance for method selection.
To explain these outcomes, we quantify dataset heterogeneity by learning from meta-features and early training dynamics to predict later validation behavior. This dynamics-aware analysis indicates that heterogeneity—such as the interplay of categorical and numerical attributes—largely determines which family of methods is favored.
Finally, we introduce a two-level design beyond the 300 common-size datasets: a compact {\name}-tiny core (45 datasets) for rapid, reproducible evaluation, and a {\name}-extension suite targeting high-dimensional, many-class, and very large-scale settings for stress testing. In summary, these results offer actionable insights into the strengths, limitations, and future directions for improving deep tabular learning.

\end{abstract}


\section{Introduction}
\label{sec:intro}
Machine learning systems are now deployed across a wide spectrum of real-world applications. Although raw data may arrive in varied and complex forms, it is commonly cast into vectorized representations through feature engineering or learned encoders. For example, image data can be converted into vectors using feature extractors such as SIFT~\citep{Szeliski2022Computer}, while modern approaches rely on convolutional layers to learn representations automatically~\citep{Goodfellow2016Deep}. 
In supervised learning, the objective is to map these vectors to labels—discrete for classification or continuous for regression—and to generalize to unseen instances drawn from the same distribution.

Among data modalities, \emph{tabular data} plays a central and pervasive role. It represents perhaps the most general and widely used form of supervised learning, organizing information as instances (rows) and attributes (columns), and it naturally arises in applications such as click‐through rate prediction~\citep{YanLXH14Coupled,JuanZCL16CTR,ZhangDW16Deep,GuoTYLH17DeepFM}, healthcare~\citep{HassanAHK20}, medical analysis~\citep{schwartz2007drug,subasi2012medical}, and e‐commerce~\citep{nederstigt2014floppies}. Its broad adoption stems from flexibility: scales and domains vary widely, attributes may be numerical or categorical (including binary or ordinal), and features often mix heterogeneous statistical behaviors.

Tabular machine learning methods have evolved significantly over time. Classical approaches, such as Logistic Regression (LR), Support Vector Machines (SVM), Multi-Layer Perceptrons (MLP), and decision trees, have served as the foundation for a wide range of algorithms~\citep{bishop2006pattern,HastieTF09ESL,Mohri2012FoML}. For practical applications, tree-based ensemble methods like Random Forest~\citep{Breiman01RandomForest}, XGBoost~\citep{chen2016xgboost}, LightGBM~\citep{ke2017lightgbm}, and CatBoost~\citep{Prokhorenkova2018Catboost} have demonstrated consistent advantages across various tasks. Inspired by the success of Deep Neural Networks (DNNs) in domains such as vision and natural language processing~\citep{simonyan2014very,vaswani2017attention,devlin2018bert}, recent efforts have adapted DNNs for tabular classification and regression tasks~\citep{WangFFW17DCN,SongS0DX0T19AutoInt,Badirli2020GrowNet,GorishniyRKB21Revisiting,Borisov2024Deep}.
Modern practice shows that carefully regularized and tuned MLPs can be highly competitive~\citep{Kadra2021Well,David2024RealMLP}, while tokenization/attention designs bring additional modeling capacity on mixed-type features~\citep{Huang2020TabTransformer,Chen2023Excel}.

A recent and influential development is the emergence of tabular foundation models. These methods pretrain on large collections of synthetic/real tasks and leverage in-context learning to enable fast adaptation to new datasets with minimal tuning~\citep{Hollmann2022TabPFN,Ma2024TabDPT,hollmann2025TabPFNv2,Qu2025TabICL,Zhang2025Limix}. In many scenarios, such models close a substantial portion of the historical performance gap between tree ensembles and deep architectures, while retaining attractive deployment properties (\eg, few-shot adaptation, fast inference). 
Understanding where and why these gains arise---relative to strong classical ensembles and modern deep baselines---requires evaluations that are both \emph{broad} (to avoid benchmark artifacts) and \emph{up-to-date} (to reflect the latest methods).

Unlike vision and language where widely adopted resources such as ImageNet enable consistent comparisons~\citep{DengDSLL009ImageNet}, the tabular domain lacks a unifying framework for systematic evaluation~\citep{Borisov2024Deep}. Existing tabular datasets are scattered across UCI~\citep{superconductivty}, OpenML~\citep{vanschoren2014openml}, and Kaggle, and they vary substantially in size, feature composition, and application domain. To reflect how tabular learning is used in practice over vectorized representations, evaluations must draw on datasets spanning diverse domains, feature types, and scales---rather than rely on narrow collections that risk benchmark artifacts. This need for broad coverage is further underscored by the ``no free lunch'' theorem,\footnote{Formally, the theorem states that if we average performance uniformly over all possible tasks with training and test sets independent, all algorithms perform equivalently on average~\citep{Wolpert1996NFL}. In practice, however, real-world applications focus on subsets of tasks that exhibit inductive biases and priors, within which certain algorithms may consistently outperform others.} which implies that empirical superiority only emerges within realistic, bounded task families. Consequently, collecting datasets that cover a wide range of real-world settings is essential to mimic practical conditions and to obtain meaningful insights into method behavior.

Prior studies also show that limited coverage and outdated baselines can yield biased conclusions~\citep{MaciaBOH13Learner}. While average rank remains a common summary across datasets~\citep{DelgadoCBA14,Grinsztajn2022Why,McElfreshKVCRGW23when}, complementary criteria have been advocated~\citep{DelgadoCBA14,McElfreshKVCRGW23when,David2024RealMLP,Yury2024TabM}, and recent work highlights challenges such as dataset aging~\citep{kohli2024towards} and reliance on expert feature engineering~\citep{Tschalzev2024DataCentric}. In this paper, we address these gaps with {\name}, a collection of 300+ datasets covering binary, multi-class, and regression tasks across domains including education, biology, chemistry, and finance. {\name} spans a wide range of sizes, feature types, and imbalance ratios to support fair, up-to-date, and comprehensive comparisons.  Based on this resource, we aim to answer three key questions:

\emph{Is there a consistent empirical picture across multiple tabular datasets?} We compare 40 representative tabular methods under a unified protocol using multiple criteria (average ranks, statistical tests, probability of best performance, aggregated errors). As the number of datasets grows, conclusions stabilize: while no single method dominates universally, top performance consistently concentrates within a small shortlist of models, and ensembling benefits both tree-based and deep tabular approaches. In the presence of recent foundation-style models, our results refine the long-standing ``trees vs.\ DNNs'' discussion.

\emph{How can we measure dataset heterogeneity, and how does it affect the behavior of deep tabular methods?} We quantify heterogeneity using meta-features and study their relation to model behavior by predicting later validation dynamics from meta-features and early training signals. This dynamics-aware view highlights the role of feature-space heterogeneity—especially the interplay between categorical and numerical attributes, sparsity, and entropy variance—in shaping when specific method families succeed or fail.

\emph{Can we support lightweight yet informative evaluation?} We design a two-level evaluation strategy: a compact \name\text{-tiny} (45 datasets, $\sim$15\% of the full suite) for rapid prototyping that balances tree-friendly and DNN-friendly cases under more stricter quality rules, and a supplemental \name\text{-extension} that stress-tests methods on high-dimensional, many-class, and very large-scale datasets. Together, they enable efficient experimentation and targeted analysis beyond the common-size regime.

\begin{figure}[t]
  \centering
   \begin{minipage}{0.45\linewidth}
    \includegraphics[width=\textwidth]{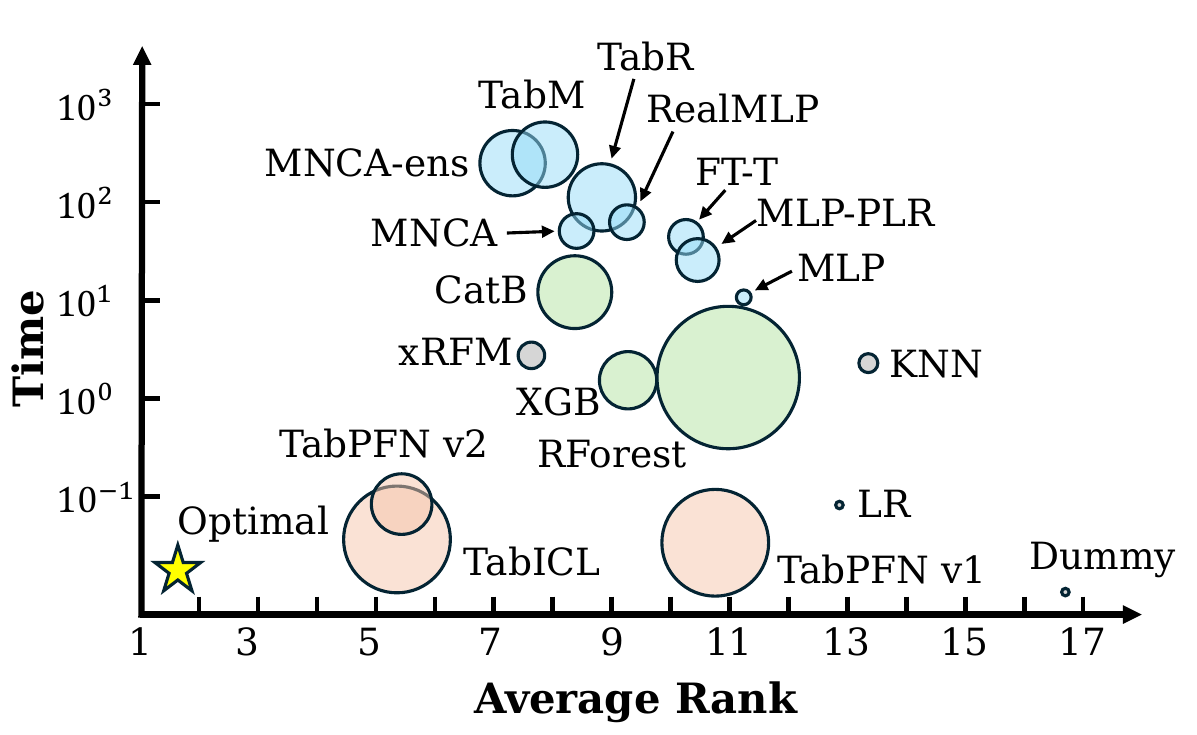}
    \centering
    {\small \mbox{(a) {Binary Classification}}}
    \end{minipage}
    \begin{minipage}{0.45\linewidth}
    \includegraphics[width=\textwidth]{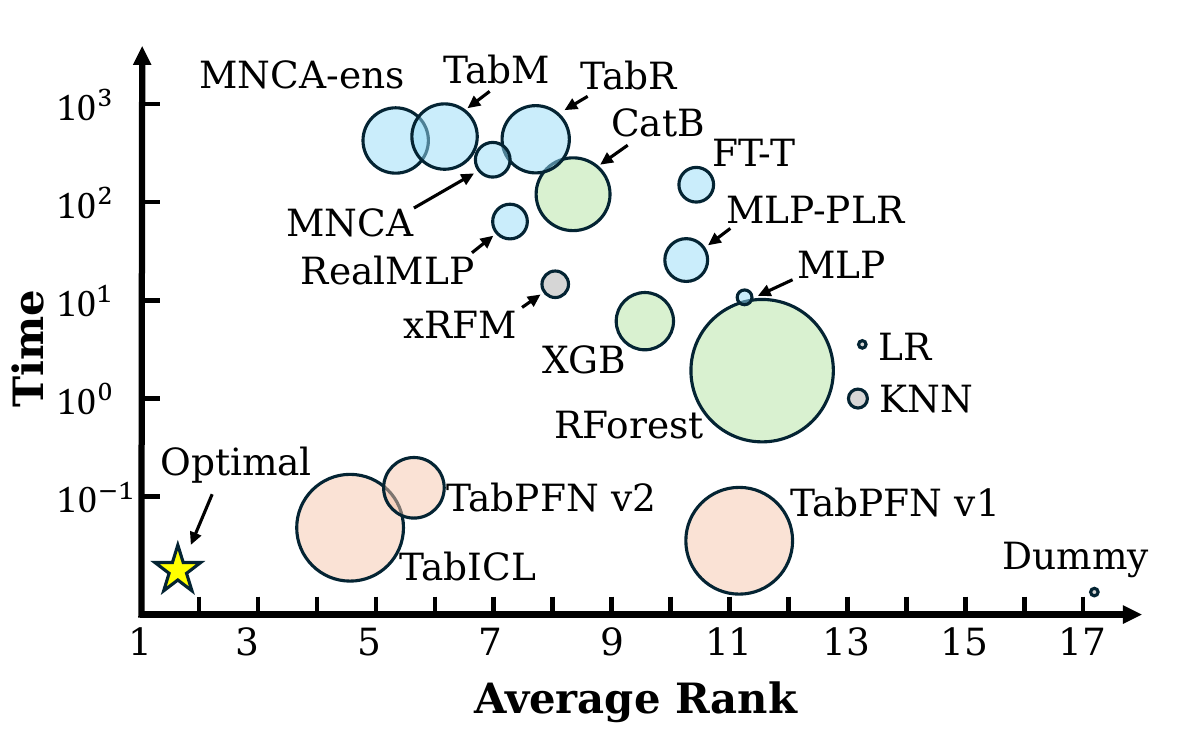}
    \centering
    {\small \mbox{(b) {Multi-Class Classification}}}
    \end{minipage}
    
    \begin{minipage}{0.45\linewidth}
    \includegraphics[width=\textwidth]{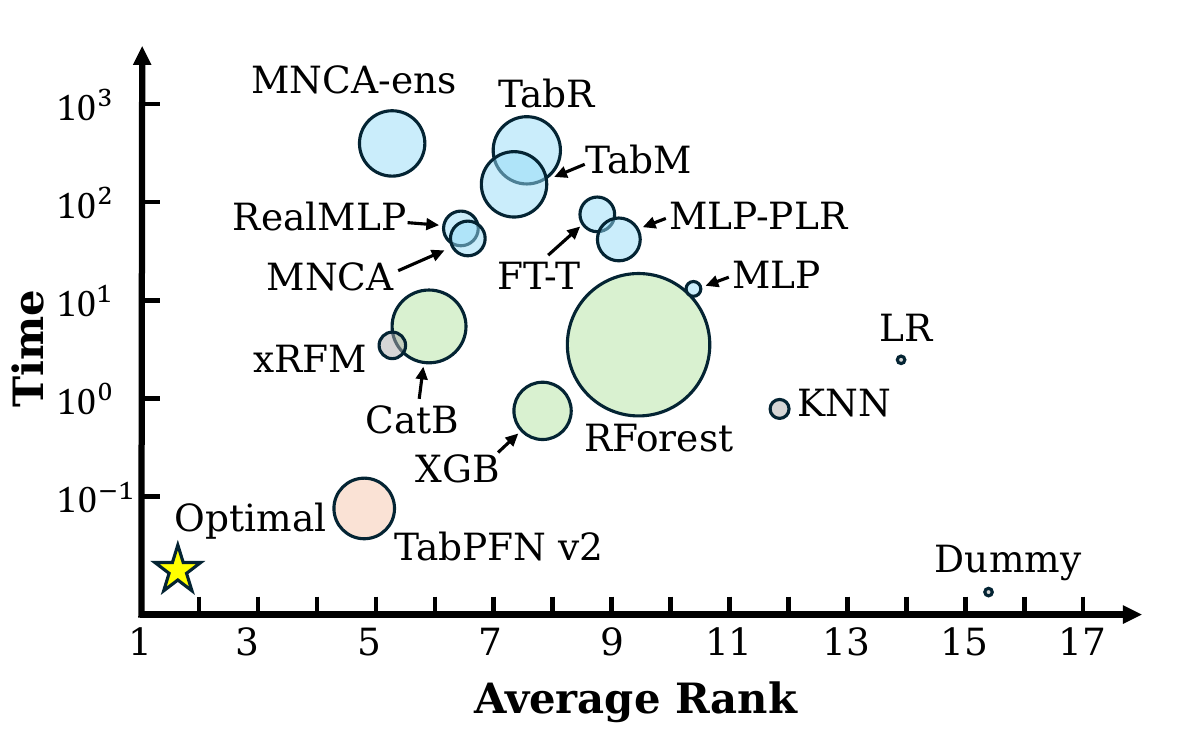}
    \centering
    {\small \mbox{(c) {Regression}}}
    \end{minipage}
    \begin{minipage}{0.45\linewidth}
    \includegraphics[width=\textwidth]{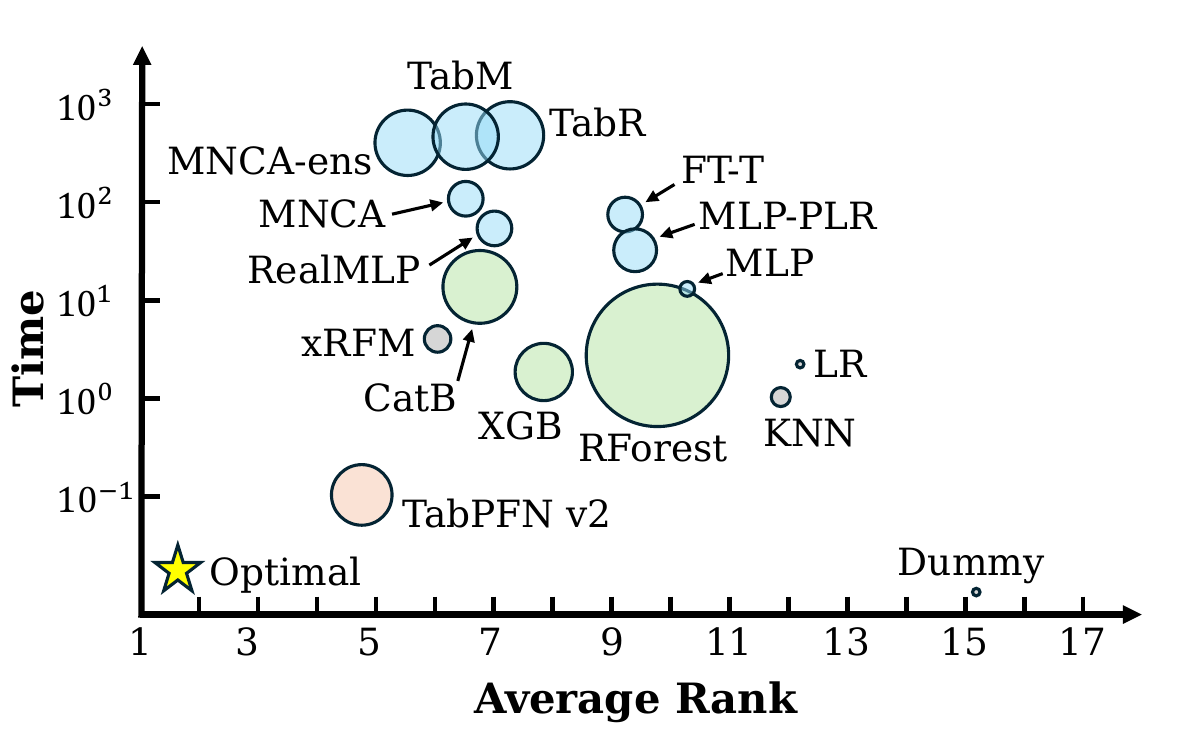}
    \centering
    {\small \mbox{(d) {All Tasks}}}
    \end{minipage}
  \caption{Performance–efficiency–size comparison of representative tabular methods on {\name} for (a) binary classification, (b) multi-class classification, (c) regression, and (d) all tasks. The performance is measured by the average rank of all methods (lower is better). The efficiency is measured by the average training time in seconds (lower is better). The model size is measured based on the average size of all models (the larger the radius, the larger the model).}
  \label{fig:teaser}
\end{figure}

\noindent The contributions of this paper are summarized as:\footnote{The code and the dataset link is available at \url{https://github.com/LAMDA-Tabular/TALENT}.
}
\begin{itemize}[noitemsep,topsep=0pt,leftmargin=*]
\item A large-scale, up-to-date evaluation of 40 tabular methods over 300+ datasets with multiple complementary criteria, showing that top performance concentrates within a small shortlist and that ensembling benefits both tree-based and DNN-based methods.
\item A dynamics-aware heterogeneity analysis that maps meta-features and early learning signals to later validation behavior, identifying the most predictive sources of heterogeneity (\eg, categorical–numerical interplay, sparsity, entropy variance).
\item A two-level evaluation design: \name\text{-tiny} ($\sim$15\% of \name) for fast, balanced comparisons under strict quality controls, and \name\text{-extension} for stress-testing on high-dimensional, many-class, and large-scale regimes.
\end{itemize}
\section{Related Work}
\subsection{Learning with Tabular Data}
Tabular data is a common format across various applications, such as click-through rate prediction~\citep{richardson2007predicting,ZhangDW16Deep} and time-series forecasting~\citep{ahmed2010empirical,Padhi2021Tabular}. 
The most common supervised settings are standard classification and regression, where models learn mappings from vectorized instances to discrete or continuous targets and are evaluated on i.i.d.\ test data~\citep{bishop2006pattern,HastieTF09ESL,Mohri2012FoML}. 
Tree-based methods, such as Random Forest~\citep{Breiman01RandomForest}, XGBoost~\citep{chen2016xgboost}, LightGBM~\citep{ke2017lightgbm}, and CatBoost~\citep{Prokhorenkova2018Catboost}, remain highly competitive due to their strong inductive bias for heterogeneous features and interactions. 
Because both model families and hyperparameters strongly influence generalization~\citep{DelgadoCBA14}, automated selection and tuning methods (\eg, AutoML) are widely used~\citep{FeurerKESBH15,guyon2019analysis}. Beyond supervised prediction, related tabular tasks include clustering~\citep{Rauf2024TableDC,SvirskyL24Interpretable}, anomaly detection~\citep{ShenkarW22Anomaly,Han2022ADBench,YinQZW024MCM}, data generation~\citep{XuSCV19TabGAN,HansenSSP23Reimagining,VeroBV24CuTS}, open-environment learning~\citep{YeZJZ21,HouFZH23Adaptive,HouGXQ23Incremental,XuTZHH23Label}, symbolic regression~\citep{Wilstrup2021Symbolic,CavaOBFVJKM21Contemporary}, and streaming scenarios~\citep{zhou2024core,Rubachev2024TabRed}.

\subsection{Deep Tabular Data Learning}
Deep models have been adapted to tabular prediction to learn representations directly from inputs and capture complex nonlinear interactions~\citep{Cheng2016Wide,GuoTYLH17DeepFM,PopovMB20Neural,ArikP21TabNet,Net-DNF,ChenLWCW22DAN}. Architectures commonly explored include residual MLPs and Transformer variants~\citep{GorishniyRKB21Revisiting,Hollmann2022TabPFN,Zhou2023TabToken,Chen2023Excel}, complemented by regularization and augmentation tailored for tabular data~\citep{UcarHE21SubTab,BahriJTM22Scarf,Rubachev2022revisiting}. A key observation is that carefully tuned, relatively simple networks can be highly competitive~\citep{Kadra2021Well,David2024RealMLP}.

\paragraph{Advantages of deep tabular models.}
DNNs excel at modeling higher-order interactions via nonlinear feature composition~\citep{WangFFW17DCN,WangSCJLHC21DCNv2}, support end-to-end multi-task learning and representation sharing~\citep{Somepalli2021SAINT,Wu2024SwitchTab}, and are trained by gradient-based optimization that flexibly accommodates new objectives with minimal redesign. They also integrate naturally into multi-modal systems combining tables with images, audio, or text~\citep{GorishniyRKB21Revisiting,JiangYW00Z24Tabular}. 

\paragraph{Design trends.}
Early neural approaches often mimicked tree workflows or emphasized feature-correlation modeling~\citep{Cheng2016Wide,GuoTYLH17DeepFM,PopovMB20Neural,Chang0G22NODEGAM}. Subsequent work refined MLPs with principled initialization, normalization, and regularization~\citep{GorishniyRKB21Revisiting,Kadra2021Well,David2024RealMLP}. Token/attention models adapt Transformer-style processing to heterogeneous columns~\citep{Huang2020TabTransformer,Chen2023Excel,Zhou2023TabToken}. Advanced ensemble strategies are also investigated and incorporated in deep tabular prediction~\citep{Yury2024TabM}.
Retrieval/neighborhood-based formulations (\eg, context-based prediction or exemplar conditioning) improve robustness and adaptation~\citep{gorishniy2023tabr,Ye2024ModernNCA}. 

\paragraph{Pretrained/foundation models.}
Recent work pretrains neural predictors on large collections of (often synthesized) tabular tasks and deploys them to novel datasets via \emph{in-context learning} without explicit gradient updates~\citep{Hollmann2022TabPFN,Ma2024TabDPT,Breugel2024Position,hollmann2025TabPFNv2,Qu2025TabICL,Zhang2025Limix}. Parameter- and data-efficient adaptation strategies further improve performance across regimes (\eg, lightweight fine-tuning and localized adapters)~\citep{Feuer2024TuneTable,Thomas2024LocalPFN,liu2025BETA}. Several studies have also evaluated and analyzed the behavior of recent foundation models such as TabPFN v2~\citep{Ye2025Closer,Rubachev2025On}. Overall, these foundation-style approaches substantially improve data efficiency and increasingly narrow the historical advantage of tree ensembles. Comprehensive surveys situate the field along a spectrum from task-specific to cross-task to general paradigms~\citep{Borisov2024Deep,Jiang2025Survey}.

\subsection{Tabular Prediction with Semantic Information and LLMs}
Recent work has begun to exploit the semantic information encoded in feature names, metadata, and textual descriptions to improve tabular prediction. One strategy is to transform features into embeddings (tokens), thereby mapping tables of varying sizes into a standardized token space. Pretrained models such as Transformers can then encode transferable knowledge that benefits downstream tasks~\citep{Yan2024Making,Ye2024Towards}.
Another line of research reformulates tabular inputs as natural language sequences, enabling large language models (LLMs) to directly learn feature–label relationships. For instance, LIFT~\citep{Dinh2022LIFT} and TabLLM~\citep{Hegselmann2022TabLLM} serialize tables into textual prompts for fine-tuning or few-shot prediction. UniPredict~\citep{Wang2023UniPredict} enhances this paradigm by enriching prompts with metadata and task-specific instructions, while IngesTables~\citep{Yak2024IngesTables} integrates external reasoning steps for multi-hop tabular inference. More recent efforts~\citep{Gardner2024Tabular8B,Wen2024GTL} propose specialized instruction-tuning techniques that further adapt LLMs for tabular contexts.
These approaches demonstrate the promise of leveraging prior knowledge embedded in LLMs for tabular tasks, particularly in low-data regimes. However, their effectiveness depends heavily on the richness of semantic information available (\eg, meaningful feature names or metadata) and can be limited by scalability issues when serializing high-dimensional tables into text.

\subsection{Tabular Methods Evaluations}
Comprehensive evaluations are essential for understanding how tabular methods behave before deployment. Several studies have attempted to benchmark tabular models, and differ in (i) the breadth and realism of their dataset coverage, (ii) the families of approaches they include, and (iii) the evaluation protocols they adopt.

\noindent\textbf{Dataset Coverage.}
Early benchmarks focused on relatively small or narrow collections of datasets. For example, \cite{DelgadoCBA14} evaluated 179 classifiers across 121 datasets, concluding that Random Forest variants were often the best performers, though later work by \cite{WainbergAF16Are} highlighted flaws in the evaluation protocol. More recent studies have expanded the coverage modestly: \cite{Kadra2021Well} studied MLPs on 40 classification datasets, while \cite{GorishniyRKB21Revisiting} examined MLPs, ResNets, and Transformer-based models on 11 datasets. \cite{Grinsztajn2022Why} used 45 datasets to investigate differences between tree-based and deep methods. A broader effort by \cite{McElfreshKVCRGW23when} included 176 classification datasets and 19 methods, but excluded regression tasks and applied strict limits on training data size and time, which may have disadvantaged deep models. Overall, most prior benchmarks underrepresent the diversity of real-world tabular tasks, particularly in regression, high-dimensional, and large-scale settings.

\noindent\textbf{Target Approaches.}
Benchmark scope also varies by method family. Classical evaluations emphasized tree ensembles and linear models; more recent work incorporates modern deep tabular methods. For instance, \cite{McElfreshKVCRGW23when} compared classical models with modern deep approaches, finding TabPFN~\citep{Hollmann2022TabPFN} to be a strong performer. Other studies have emphasized specific architectures, such as MLP variants~\citep{Kadra2021Well}, ResNets~\citep{GorishniyRKB21Revisiting}, or Transformers~\citep{Chen2023Excel}, but often in limited settings that make it difficult to generalize their findings.

\noindent\textbf{Evaluation Protocols.}
Benchmarking protocols also vary considerably. Some studies adopt uniform hyperparameter settings or limited tuning budgets, which can bias results against deep models that require careful tuning. For example, the strict time and trial limits in~\cite{McElfreshKVCRGW23when} may have led to suboptimal evaluations for complex neural architectures. Recent work has also explored alternative evaluation perspectives, including dataset quality~\citep{Erickson2025TabArena}, dataset age~\citep{kohli2024towards}, reliance on expert-crafted features~\citep{Tschalzev2024DataCentric}, temporal characteristics of tabular data~\citep{Rubachev2024TabRed,Tschalzev2024DataCentric}, and cross-validation as well as post-hoc ensemble strategies to boost performance~\citep{Erickson2025TabArena}. 

% \medskip
\noindent\textbf{Summary.}
Effective assessment requires datasets that span classification \emph{and} regression, cover diverse domains and feature types, and pair fair, well-tuned protocols with appropriate statistical comparisons. Achieving this breadth entails computational trade-offs. With rapid advances in deep tabular learning \citep{David2024RealMLP,Ye2024ModernNCA,Beaglehole2025xRFM}—especially pretrained foundation models \citep{hollmann2025TabPFNv2,Qu2025TabICL}—there is a pressing need for evaluations that balance wide coverage with rigor, enabling reliable, nuanced conclusions about the strengths and limitations of modern tabular methods.
\section{Preliminary}
\label{sec:pre}
% We first introduce the task learning with tabular data, followed by the descriptions of compared tabular methods.

\subsection{Learning with Tabular Data}
A tabular dataset $\sD=\{(\x_i, y_i)\}_{i=1}^N$ is formatted as $N$ examples and $d$ features (attributes), corresponding to $N$ rows and $d$ columns in a table. Each instance $\x_i \in \bbR^d$ is represented by its $d$ feature values. The $j$-th feature of instance $\x_i$, denoted $x_{i,j}$, may be numerical (continuous), $x_{i,j}^{\textit{\rm num}} \in \bbR$, or categorical (discrete), $x_{i,j}^{\textit{\rm cat}}$. Categorical features are typically transformed into numerical vectors using encoding strategies such as one-hot or target encoding~\citep{Hancock2020Categorical}. 

In a {\em supervised} prediction task, each instance is associated with a label $y_i$, where $y_i \in \{1,-1\}$ for binary classification, $y_i \in [C]=\{1,\ldots,C\}$ for multi-class classification, and $y_i \in \bbR$ for regression. Given $\sD=\{(\x_i, y_i)\}_{i=1}^N$, the goal is to learn a model $f$ by empirical risk minimization:
\begin{equation}
\min_f \; \sum_{(\x_i, y_i)\in\sD}  \ell(y_i, \;\hat{y}_i=f(\x_i)) + \Omega(f)\;, \label{eq:objective}
\end{equation}
where $\ell(\cdot,\cdot)$ measures the discrepancy between the predicted label $\hat{y}_i$ and the true label $y_i$ (\eg, cross-entropy for classification), and $\Omega(f)$ is a regularization term. The learned model $f$ is expected to generalize to unseen instances sampled from the same distribution as $\sD$. 

While this formulation captures the standard setting of supervised tabular learning, we note that it does not encompass all paradigms. Some methods adopt {\em unsupervised} or {\em self-supervised pretraining} objectives on tabular data, aiming to learn transferable representations before fine-tuning on supervised tasks~\citep{UcarHE21SubTab,Rubachev2022revisiting}. Others focus on {\em generative} modeling of tabular data~\citep{HansenSSP23Reimagining,VeroBV24CuTS}, or employ {\em ensemble strategies} that combine multiple predictions from different runs, seeds, or data splits~\citep{Erickson2025TabArena}. Since such strategies vary widely and often depend on additional design choices, in this work we restrict our evaluation to the {\em intrinsic supervised performance} of each model, while acknowledging that external ensembles or pretraining can further improve results.

\subsection{Representative Tabular Models}\label{sec:related_methods}
We consider several representative families of models for tabular prediction, including classical methods, tree-based methods, and deep neural network (DNN)-based methods.

\noindent{\bf Classical Methods.}
As a trivial baseline, we include the ``Dummy'' approach, which always predicts the majority class for classification or the mean of the target for regression. 
We further evaluate standard classical methods: Logistic Regression (LR), K-Nearest Neighbors (KNN), and Support Vector Machines (SVM). We also include Recursive Feature Machines (RFM)~\citep{Radhakrishnan2023RFM}, which enable kernel machines to learn features by recursively reweighting them via a gradient-inspired mechanism without backpropagation, and their extension xRFM~\citep{Beaglehole2025xRFM}.
For classification tasks, we also include Naive Bayes and the Nearest Class Mean (NCM)~\citep{tibshirani2002diagnosis}. 
For regression tasks, Linear Regression replaces LR, and we additionally consider DNNR~\citep{NaderSL22DNNR}.  \looseness=-1

\noindent{\bf Tree-based Methods.} Tree-based models are widely regarded as strong baselines for tabular learning~\cite{DelgadoCBA14}. We include Random Forest~\citep{Breiman01RandomForest}, as well as gradient-boosting ensembles~\citep{friedman2001greedy,friedman2002stochastic} such as XGBoost~\citep{chen2016xgboost}, LightGBM~\citep{ke2017lightgbm}, and CatBoost~\citep{Prokhorenkova2018Catboost}, all of which are established as highly competitive across tasks~\citep{Grinsztajn2022Why,McElfreshKVCRGW23when}.  \looseness=-1
% We also include Recursive Feature Machines (RFM)~\citep{Radhakrishnan2023RFM}, which enable kernel machines to learn features by recursively reweighting them via a gradient-inspired mechanism without backpropagation, and their extension xRFM~\citep{Beaglehole2025xRFM}, which integrates RFM into a tree-like architecture. 
%\footnote{Both RFM and xRFM are not classical decision trees or gradient-boosted trees. We group them under the ``tree-based'' category for reporting simplicity, as they are backpropagation-free, stage-wise learners and xRFM adopts a tree-style topology. This labeling choice does not affect per-method comparisons.}

\noindent{\bf DNN-based Methods.}
Deep tabular methods vary in their design principles and prediction strategies. We categorize them into the following groups:

\begin{itemize}[noitemsep,topsep=0pt,leftmargin=*]
\item {\bf MLP Variants.} 
Vanilla MLPs operate directly on raw features and, with careful tuning, can be competitive~\citep{Kadra2021Well,Chen2024Team}. 
Strong baselines include the implementation in~\citep{GorishniyRKB21Revisiting}, MLP-PLR with periodic activations~\citep{Gorishniy2022On}, and RealMLP with tailored modules/encodings and optimization~\citep{David2024RealMLP}. 
Other variants include SNN~\citep{KlambauerUMH17Self} and ResNet-style architectures~\citep{GorishniyRKB21Revisiting}.

\item {\bf Specially Designed Architectures.}
Several works design custom architectures to capture explicit feature interactions. 
DCNv2~\citep{WangSCJLHC21DCNv2} combines embeddings, cross layers, and deep networks. 
TabCaps~\citep{Chen2023TabCaps} encapsulates instance features into vectorial representations to enhance representation learning.

\item {\bf Token-based Methods.}
These methods map feature values into high-dimensional tokens, enabling attention mechanisms to model high-order interactions. 
Representative approaches include AutoInt~\citep{SongS0DX0T19AutoInt}, TabTransformer~\citep{Huang2020TabTransformer}, FT-Transformer (FT-T)~\citep{GorishniyRKB21Revisiting}, and ExcelFormer~\citep{Chen2023Excel}, which introduces semi-permeable attention and attentive feed-forward layers.

\item {\bf Regularization-based Methods.}
These methods enhance generalization through explicit regularization. 
TANGOS~\citep{jeffares2023tangos} enforces neuron specialization and orthogonality. 
SwitchTab~\citep{Wu2024SwitchTab} introduces a self-supervised encoder–decoder framework, while PTaRL~\citep{PTARL} calibrates features through prototypes.

\item {\bf Tree-mimic Methods.}
Inspired by decision trees, these architectures combine neural networks with tree-like structures. 
NODE~\citep{PopovMB20Neural} generalizes oblivious decision trees, GrowNet~\citep{Badirli2020GrowNet} embeds shallow networks within boosting, and TabNet~\citep{ArikP21TabNet} employs sequential attention for feature selection. 
DANets~\citep{ChenLWCW22DAN} group correlated features to produce higher-level abstractions.

\item {\bf Neighborhood-based Methods.}
These approaches make predictions by retrieving and weighting similar instances.
TabR~\citep{gorishniy2023tabr} augments a learned predictor with a KNN-style retrieval component, while ModernNCA~\citep{Ye2024ModernNCA} modernizes classic NCA~\citep{GoldbergerRHS04} for robust, retrieval-based tabular learning.

\item {\bf Ensemble-based Methods.}
Ensemble-style DNNs share parameters to train multiple predictors efficiently.
TabM~\citep{Yury2024TabM} builds on BatchEnsemble~\citep{Wen2020BatchEnsemble} with an MLP backbone; for analysis we also consider a BatchEnsemble-enhanced variant of ModernNCA (MNCA-ens).

\item {\bf Pretrained Foundation Models.}
Pretrained tabular transformers enable in-context inference on new datasets with little to no task-specific training or hyperparameter tuning.
TabPFN~\citep{Hollmann2022TabPFN} predicts labels by conditioning directly on the training set.
We also include stronger successors that scale to larger datasets: TabPFN v2~\citep{hollmann2025TabPFNv2} and TabICL~\citep{Qu2025TabICL}.
TabPFN and TabICL currently target \emph{classification}, whereas TabPFN v2 supports both \emph{classification} and \emph{regression}.
In addition, we evaluate adaptation methods built on TabPFN, including LocalPFN~\citep{Thomas2024LocalPFN}, TuneTables~\citep{Feuer2024TuneTable}, and BETA~\citep{liu2025BETA}.

\end{itemize}

\noindent{\bf Other Methods.}
Additional models such as Trompt~\citep{DBLP:conf/icml/Chen2023Trompt}, BiSHop~\citep{Xu2024BiSHop}, ProtoGate~\citep{Jiang2024ProtoGate}, and GRANDE~\citep{Marton2024GRANDE} fall outside our main categories.
We omit them here due to substantially longer training/inference times or because their goals (\eg, extreme efficiency) are orthogonal to our focus on predictive accuracy.
Similarly, pretrained variants such as HyperFast~\citep{BonetMGI2024HyperFast}, TabDPT~\citep{Ma2024TabDPT}, TabFlex~\citep{Zeng2025TabFlex}, and MotherNet~\citep{Mueller2020MotherNet} are not included, as stronger and more recent foundation models are already covered.
Finally, we exclude RealTabPFN~\citep{Garg2025RealTabPFN} to avoid potential data overlap: its continual pretraining corpus intersects with our benchmark, which could confound fair evaluation.
\section{A Comprehensive Tabular Data Benchmark}
\label{sec:benchmark}
This section describes how {\name} is constructed, characterizes the datasets it contains, details the quality controls we apply, and highlights why {\name} is well-suited for advancing research on tabular learning.

\subsection{Design Philosophy}
To meaningfully measure the ability of tabular methods across diverse scenarios, the guiding principle of the {\name} benchmark is to evaluate models under broad and realistic coverage that mirrors the heterogeneity of real-world applications. The benchmark is built around {\em two complementary layers of coverage}. First, we construct a large and diverse collection of \emph{common-size} datasets, which form the foundation for standard evaluation. This layer ensures that comparisons are made across a broad range of classical and deep models under typical conditions. Second, we incorporate \emph{specialized} settings—such as high-dimensional feature spaces, many-class classification problems, and very-large-scale datasets—in {\name}-extension that reflect more challenging but practically important scenarios. These are evaluated separately (see~\autoref{sec:talent_extension}), allowing us to analyze scalability and robustness under conditions that go beyond the common-size setting.

In addition to the coverage of dataset sizes, we adopt a {\em two-level dataset selection strategy} to balance inclusiveness with fairness. The general {\name} set is constructed with relatively weak filtering rules, excluding only datasets that present clear quality issues such as label leakage or annotation errors. This design maximizes breadth while minimizing evaluation bias. Complementing this, we curate a core set of datasets, \ie, {\name}-tiny, in~\autoref{sec:talent_tiny} that applies stricter rules to ensure balance across domains, feature types, and scales, and to include both tree-friendly and DNN-friendly tasks. This core set provides a controlled, reliable environment for detailed analysis and rapid iteration.

\subsection{Datasets Collection}\label{sec:benchmark_details}
{\name} aggregates datasets from UCI~\citep{superconductivty}, OpenML~\citep{vanschoren2014openml}, and Kaggle. We first construct the \emph{general} {\name} set using a set of filtering and preprocessing rules that emphasize inclusiveness while removing datasets with obvious quality problems. The stricter rules used to define the \emph{{\name}-tiny} core set follow the two-level strategy above and are described later.

\noindent\textbf{Initial filtering rules.}  
We begin with the following quality controls:
\begin{itemize}[noitemsep,topsep=0pt,leftmargin=*]
\item \textbf{Size.} Exclude datasets with fewer than 500 instances ($N<500$) or fewer than 5 features ($d<5$), which tend to yield unstable evaluations due to limited test coverage.
\item \textbf{Missing values.} Remove datasets with $>20\%$ missing values to avoid unreliable comparisons.
\item \textbf{Trivial classification.} Exclude classification datasets that are overly easy (\eg, a simple MLP exceeds $99\%$ accuracy) or dominated by a majority class.
\item \textbf{Attribute preprocessing.} Drop non-informative attributes (\eg, \texttt{id}, \texttt{index}, \texttt{timestamp}); ordinally encode categorical features~\citep{Borisov2024Deep,McElfreshKVCRGW23when}; and follow~\cite{GorishniyRKB21Revisiting} for the remaining preprocessing.
\item \textbf{Duplicates.} Remove subsets and near-duplicates from UCI/OpenML to ensure uniqueness.
\item \textbf{Task type correction.} Relabel 22 mis-specified regression datasets with only two unique targets as binary classification.
\end{itemize}

\noindent\textbf{Multi-version datasets.}  
Some datasets share an origin but differ in collection conditions, feature extraction, or augmentation (\eg, \texttt{forex-} at different sampling frequencies, \texttt{wine-quality-} under different standards, and \texttt{mfeat-} variants of handwritten digits). We retain such versions unless they completely overlap, as they reflect real-world application requirements and resource constraints. These variations provide an opportunity to evaluate how different tabular models handle such practical scenarios: the best method on one version is not always best on others. Removing them has little impact on aggregate rankings when many datasets are included, but retaining them provides valuable insights into consistency across related tasks.

\noindent\textbf{``Easy'' datasets revisited.}
Although the rules above filter out trivial cases, not all models achieve optimal performance on some seemingly easy datasets. For example, \texttt{mice\_protein\_expression} is solved by KNN and RealMLP~\citep{David2024RealMLP}, while many strong models underperform. We therefore retain such datasets when performance is not uniformly trivial, as they expose informative discrepancies.

% \noindent\textbf{Potential leakage in pretrained models.} Pretrained tabular models (\eg, TabPFN~\citep{Hollmann2022TabPFN,hollmann2025TabPFNv2}) may face evaluation leakage when early stopping is used on real datasets. We identified only limited overlaps (\eg, \texttt{PizzaCutter3}, \texttt{PieChart3}). Given their small number relative to the scale of {\name}, we retain these datasets; they are unlikely to bias overall conclusions.

\noindent\textbf{Datasets from other modalities.}  
{\name} includes 25 datasets whose features are extracted from images or audio (\eg, \texttt{optdigits}, \texttt{segment}, \texttt{phoneme}). While some recent analyses argue these are less relevant for tabular evaluation~\citep{kohli2024towards,Erickson2025TabArena}, in many practical scenarios, only extracted features—not raw modalities—are available due to resource or deployment constraints. We therefore keep them to reflect real-world usage and to test model robustness to such inputs.

\noindent\textbf{Inherent distribution shifts.}  
Some datasets contain implicit distribution shifts, for example, due to temporal splits in the collection where timestamps used during collection create non-stationary distributions~\citep{Tschalzev2024DataCentric,Rubachev2024TabRed}. 
\cite{Cai2025Temporal} show that the choice of split strategy between training and validation sets can significantly affect absolute performance values, even though the relative ranking of methods often remains stable. This suggests that while distribution shift complicates reliable generalization, comparative evaluations may still provide useful insights under consistent protocols.
In this paper, however, we focus on {\em standard tabular prediction tasks, where training and test instances are assumed to be drawn from the same underlying distribution}. This assumption allows us to establish a fair and controlled evaluation framework. Nevertheless, we acknowledge that distribution shifts are common in real-world applications, and developing benchmarks that explicitly address such scenarios is an important direction for future research.

\noindent\textbf{Datasets with known leakage.}  Recent analyses have revealed that a number of widely used tabular datasets suffer from data leakage, most often caused by the inclusion of features that directly or indirectly encode the target variable~\citep{Rubachev2024TabRed,Tschalzev2025Unreflected}. In our collection, we identify 13 datasets with potential leakage. While such datasets may compromise strict evaluation quality, we deliberately retain them in the general {\name} benchmark for two reasons: first, to maintain comparability with prior work, where these datasets have been widely used; and second, to assess the general ability of tabular methods to handle diverse real-world data, including imperfectly curated datasets that are commonly encountered in practice. For rigorous and bias-free evaluation, however, all datasets with identified leakage are excluded from the curated {\name}-tiny subset. The detailed list and explanations are provided in~\autoref{appendix:data_selection}.

\medskip
\noindent\textbf{From general to core sets.}  
The rules above define the \emph{general} {\name} benchmark, designed for inclusiveness while filtering out datasets with obvious issues. To provide a stricter, more balanced evaluation, we also construct the \emph{{\name}-tiny} core set by applying stronger rules: excluding datasets derived from other modalities and with leakages, removing trivial or duplicated variants, and focusing on tasks that balance tree- and DNN-friendly cases. This two-level design allows researchers to use {\name} for broad benchmarking while relying on {\name}-tiny for controlled, in-depth analysis.

\medskip
Ultimately, {\name} consists of 120 binary-classification, 80 multi-class, and 100 regression datasets, offering both breadth and depth for evaluating tabular learning. The complete list of {\name}, the selection for {\name}-tiny, and summary statistics for {\name}-extension are reported in~\autoref{tab:dataset_basic}.

\clearpage

{\tiny{
\tabcolsep 0.9pt
\begin{longtable}{clccccc|clccccc}
\caption{The list of datasets (including names and source URLs) in our proposed benchmark, along with the statistics for each dataset.} \\
\label{tab:dataset_basic} \\

% \toprule
% \multicolumn{14}{c}{{\footnotesize\name}} \\
% \midrule
\specialrule{\heavyrulewidth}{2pt}{0pt}
\rowcolor{gray!20}\multicolumn{14}{c}{\rule{0pt}{3ex}\footnotesize{\name}} \\
\specialrule{\lightrulewidth}{0pt}{3pt}

ID & \multicolumn{1}{c}{Name (Source)} & Task & Class & Sample & Feature & Tiny & ID & \multicolumn{1}{c}{Name (Source)} & Task & Class & Sample & Feature & Tiny \\
\midrule
  1 & \href{https://www.openml.org/search?type=data&status=active&id=1491&sort=runs}{100-plants-margin} & Cls & 100 &   1600 &  64 & \  & 151 & \href{https://www.openml.org/search?type=data&status=active&id=375&sort=runs}{JapaneseVowels} & Cls &   9 &   9961 &  14 & \  \\
  2 & \href{https://www.openml.org/search?type=data&status=active&id=1492&sort=runs}{100-plants-shape} & Cls & 100 &   1600 &  64 & \  & 152 & \href{https://www.openml.org/search?type=data&status=active&id=41143}{jasmine} & Cls &   2 &   2984 & 144 & \checkmark \\
  3 & \href{https://www.openml.org/search?type=data&status=active&id=1493&sort=runs}{100-plants-texture} & Cls & 100 &   1599 &  64 & \  & 153 & \href{https://www.openml.org/search?type=data&status=active&id=1053&sort=runs}{jm1} & Cls &   2 &  10885 &  21 & \  \\
  4 & \href{https://www.openml.org/search?type=data&status=active&id=43714&sort=runs}{1000-Cameras-Dataset} & Reg & $-$ &   1038 &  10 & \  & 154 & \href{https://www.openml.org/search?type=data&status=active&id=44311&sort=runs}{Job\_Profitability} & Reg & $-$ &  14480 &  28 & \  \\
  5 & \href{https://www.openml.org/search?type=data&status=active&id=215&sort=runs}{2dplanes} & Reg & $-$ &  40768 &  10 & \checkmark & 155 & \href{https://www.openml.org/search?type=data&status=active&id=41027&sort=runs}{jungle\_chess\_endgame} & Cls &   3 &  44819 &   6 & \checkmark \\
  6 & \href{https://www.openml.org/search?type=data&status=active&id=1557&sort=runs}{abalone} & Cls &   3 &   4177 &   8 & \  & 156 & \href{https://www.openml.org/search?type=data&status=active&id=1067&sort=runs}{kc1} & Cls &   2 &   2109 &  21 & \  \\
  7 & \href{https://archive.ics.uci.edu/dataset/1/abalone}{Abalone\_reg} & Reg & $-$ &   4177 &   8 & \checkmark & 157 & \href{https://www.openml.org/search?type=data&status=active&id=45075&sort=runs}{KDD} & Cls &   2 &   5032 &  45 & \  \\
  8 & \href{https://archive.ics.uci.edu/dataset/846/accelerometer}{accelerometer} & Cls &   4 & 153004 &   4 & \  & 158 & \href{https://www.openml.org/search?type=data&status=active&id=44124&sort=runs}{kdd\_ipums\_la\_97-small} & Cls &   2 &   5188 &  20 & \  \\
  9 & \href{https://www.openml.org/search?type=data&status=active&id=41156&sort=runs}{ada} & Cls &   2 &   4147 &  48 & \checkmark & 159 & \href{https://www.openml.org/search?type=data&status=active&id=44186&sort=runs}{KDDCup09\_upselling} & Cls &   2 &   5128 &  49 & \  \\
 10 & \href{https://www.openml.org/search?type=data&status=any&sort=runs&order=desc&id=1043}{ada\_agnostic} & Cls &   2 &   4562 &  48 & \  & 160 & \href{https://www.openml.org/search?type=data&status=active&id=189&sort=runs}{kin8nm} & Reg & $-$ &   8192 &   8 & \  \\
 11 & \href{https://www.openml.org/search?type=data&status=active&id=1037&sort=runs}{ada\_prior} & Cls &   2 &   4562 &  14 & \  & 161 & \href{https://www.openml.org/search?type=data&status=active&id=1481&sort=runs}{kr-vs-k} & Cls &  18 &  28056 &   6 & \  \\
 12 & \href{https://archive.ics.uci.edu/dataset/2/adult}{adult} & Cls &   2 &  48842 &  14 & \  & 162 & \href{https://www.openml.org/search?type=data&status=active&id=3}{kr-vs-kp} & Cls &   2 &   3196 &  36 & \  \\
 13 & \href{https://www.openml.org/search?type=data&status=active&id=296&sort=runs}{Ailerons} & Reg & $-$ &  13750 &  40 & \  & 163 & \href{https://www.openml.org/search?type=data&status=active&id=184&sort=runs}{kropt} & Cls &  18 &  28056 &   6 & \  \\
 14 & \href{https://www.openml.org/search?type=data&status=active&id=43919&sort=runs}{airfoil\_self\_noise} & Reg & $-$ &   1503 &   5 & \  & 164 & \href{https://www.kaggle.com/datasets/talhabarkaatahmad/laptop-prices-dataset-october-2023}{Laptop\_Prices\_Dataset} & Reg & $-$ &   4441 &   8 & \  \\
 15 & \href{https://www.kaggle.com/datasets/yakhyojon/customer-satisfaction-in-airline}{airline\_satisfaction} & Cls &   2 & 129880 &  21 & \  & 165 & \href{https://www.openml.org/search?type=data&status=active&id=43889&sort=runs}{law-school-admission} & Cls &   2 &  20800 &  11 & \checkmark \\
 16 & \href{https://www.openml.org/search?type=data&status=active&id=44528&sort=runs}{airlines\_2000} & Cls &   2 &   2000 &   7 & \checkmark & 166 & \href{https://www.openml.org/search?type=data&status=active&id=40677&sort=runs}{led24} & Cls &  10 &   3200 &  24 & \  \\
 17 & \href{https://www.openml.org/search?type=data&status=active&id=40707&sort=runs}{allbp} & Cls &   3 &   3772 &  29 & \checkmark & 167 & \href{https://www.openml.org/search?type=data&status=active&id=40678&sort=runs}{led7} & Cls &  10 &   3200 &   7 & \  \\
 18 & \href{https://www.openml.org/search?type=data&status=active&id=40708&sort=runs}{allrep} & Cls &   4 &   3772 &  29 & \  & 168 & \href{https://www.openml.org/search?type=data&status=active&id=6&sort=runs}{letter} & Cls &  26 &  20000 &  15 & \  \\
 19 & \href{https://www.openml.org/search?type=data&status=active&id=4135&sort=runs}{Amazon\_employee\_access} & Cls &   2 &  32769 &   7 & \  & 169 & \href{https://www.openml.org/search?type=data&status=active&id=42636&sort=runs}{Long} & Cls &   2 &   4477 &  19 & \  \\
 20 & \href{https://www.openml.org/search?type=data&status=active&id=458&sort=runs}{analcatdata\_authorship} & Cls &   4 &    841 &  69 & \  & 170 & \href{https://www.openml.org/search?type=data&status=active&id=43892&sort=runs}{longitudinal-survey} & Cls &   2 &   4908 &  16 & \checkmark \\
 21 & \href{https://www.openml.org/search?type=data&status=active&id=504&sort=runs}{analcatdata\_supreme} & Reg & $-$ &   4052 &   7 & \checkmark & 171 & \href{https://www.openml.org/search?type=data&status=active&id=41144&sort=runs}{madeline} & Cls &   2 &   3140 & 259 & \  \\
 22 & \href{https://www.kaggle.com/datasets/yasserh/wine-quality-dataset}{archive2} & Reg & $-$ &   1143 &  12 & \  & 172 & \href{https://www.openml.org/search?type=data&status=active&id=1120&sort=runs}{MagicTelescope} & Cls &   2 &  19020 &   9 & \  \\
 23 & \href{https://www.kaggle.com/datasets/whenamancodes/student-performance/data}{archive\_r56\_Portuguese} & Reg & $-$ &    651 &  30 & \  & 173 & \href{https://www.openml.org/search?type=data&status=active&id=310&sort=runs}{mammography} & Cls &   2 &  11183 &   6 & \  \\
 24 & \href{https://www.openml.org/search?type=data&status=active&id=1459&sort=runs}{artificial-characters} & Cls &  10 &  10218 &   7 & \  & 174 & \href{https://www.kaggle.com/datasets/rodsaldanha/arketing-campaign}{Marketing\_Campaign} & Cls &   2 &   2240 &  27 & \  \\
 25 & \href{https://www.openml.org/search?type=data&status=active&id=41705&sort=runs}{ASP-POTASSCO} & Cls &  11 &   1294 & 141 & \checkmark & 175 & \href{https://archive.ics.uci.edu/dataset/863/maternal+health+risk}{maternal\_health\_risk} & Cls &   3 &   1014 &   6 & \  \\
 26 & \href{https://archive.ics.uci.edu/dataset/713/auction+verification}{auction\_verification} & Reg & $-$ &   2043 &   7 & \  & 176 & \href{https://www.openml.org/search?type=data&status=active&id=41187&sort=runs}{mauna-loa-atmospheric} & Reg & $-$ &   2225 &   6 & \  \\
 27 & \href{https://www.openml.org/search?type=data&status=active&id=1548&sort=runs}{autoUniv-au4-2500} & Cls &   3 &   2500 & 100 & \  & 177 & \href{https://www.openml.org/search?type=data&status=active&id=12&sort=runs}{mfeat-factors} & Cls &  10 &   2000 & 216 & \  \\
 28 & \href{https://www.openml.org/search?type=data&status=active&id=1552&sort=runs}{autoUniv-au7-1100} & Cls &   5 &   1100 &  12 & \checkmark & 178 & \href{https://www.openml.org/search?type=data&status=active&id=14&sort=runs}{mfeat-fourier} & Cls &  10 &   2000 &  76 & \  \\
 29 & \href{https://www.openml.org/search?type=data&status=active&id=43927&sort=runs}{avocado\_sales} & Reg & $-$ &  18249 &  13 & \  & 179 & \href{https://www.openml.org/search?type=data&status=active&id=16&sort=runs}{mfeat-karhunen} & Cls &  10 &   2000 &  64 & \  \\
 30 & \href{https://archive.ics.uci.edu/ml/datasets/Bank+Marketing}{bank} & Cls &   2 &  45211 &  16 & \  & 180 & \href{https://www.openml.org/search?type=data&status=active&id=18&sort=runs}{mfeat-morphological} & Cls &  10 &   2000 &   6 & \  \\
 31 & \href{https://www.openml.org/search?type=data&status=active&id=558&sort=runs}{bank32nh} & Reg & $-$ &   8192 &  32 & \  & 181 & \href{https://www.openml.org/search?type=data&status=active&id=20&sort=runs}{mfeat-pixel} & Cls &  10 &   2000 & 240 & \  \\
 32 & \href{https://www.openml.org/search?type=data&status=active&id=572&sort=runs}{bank8FM} & Reg & $-$ &   8192 &   8 & \  & 182 & \href{https://www.openml.org/search?type=data&status=active&id=22&sort=runs}{mfeat-zernike} & Cls &  10 &   2000 &  47 & \  \\
 33 & \href{https://www.kaggle.com/datasets/gauravtopre/bank-customer-churn-dataset}{Bank\_Customer\_Churn} & Cls &   2 &  10000 &  10 & \  & 183 & \href{https://www.openml.org/search?type=data&status=active&id=43093&sort=runs}{MiamiHousing2016} & Reg & $-$ &  13932 &  16 & \  \\
 34 & \href{https://archive.ics.uci.edu/dataset/267/banknote+authentication}{banknote\_authentication} & Cls &   2 &   1372 &   4 & \  & 184 & \href{https://www.openml.org/search?type=data&status=active&id=45648&sort=runs}{MIC} & Cls &   2 &   1649 & 104 & \  \\
 35 & \href{https://www.openml.org/search?type=data&status=active&id=185&sort=runs}{baseball} & Cls &   3 &   1340 &  16 & \  & 185 & \href{https://archive.ics.uci.edu/dataset/342/mice+protein+expression}{mice\_protein\_expression} & Cls &   8 &   1080 &  75 & \  \\
 36 & \href{https://www.kaggle.com/datasets/yakhyojon/national-basketball-association-nba}{Basketball\_c} & Cls &   2 &   1340 &  11 & \  & 186 & \href{https://www.openml.org/search?type=data&status=active&id=41671&sort=runs}{microaggregation2} & Cls &   5 &  20000 &  20 & \checkmark \\
 37 & \href{https://archive.ics.uci.edu/dataset/514/bias+correction+of+numerical+prediction+model+temperature+forecast}{Bias\_correction\_r} & Reg & $-$ &   7725 &  21 & \  & 187 & \href{https://www.openml.org/search?type=data&status=any&id=43071}{MIP-2016-regression} & Reg & $-$ &   1090 & 144 & \  \\
 38 & \href{https://archive.ics.uci.edu/dataset/514/bias+correction+of+numerical+prediction+model+temperature+forecast}{Bias\_correction\_r\_2} & Reg & $-$ &   7725 &  21 & \  & 188 & \href{https://archive.ics.uci.edu/dataset/755/accelerometer+gyro+mobile+phone+dataset}{mobile\_c36\_oversampling} & Cls &   2 &  51760 &   6 & \  \\
 39 & \href{https://www.openml.org/search?type=data&status=active&id=1414&sort=runs}{bike\_sharing\_demand} & Reg & $-$ &  10886 &   9 & \  & 189 & \href{https://www.kaggle.com/datasets/redpen12/mobile-phone-market-in-ghana}{Mobile\_Phone\_in\_Ghana} & Reg & $-$ &   3600 &  14 & \  \\
 40 & \href{https://archive.ics.uci.edu/dataset/586/ble+rssi+dataset+for+indoor+localization}{BLE\_RSSI\_localization} & Cls &   3 &   9984 &   3 & \  & 190 & \href{https://www.kaggle.com/datasets/iabhishekofficial/mobile-price-classification}{Mobile\_Price} & Cls &   4 &   2000 &  20 & \  \\
 41 & \href{https://www.openml.org/search?type=data&status=active&id=251&sort=runs}{BNG(breast-w)} & Cls &   2 &  39366 &   9 & \  & 191 & \href{https://www.openml.org/search?type=data&status=active&id=41021}{Moneyball} & Reg & $-$ &   1232 &  14 & \  \\
 42 & \href{https://www.openml.org/search?type=data&status=active&id=255&sort=runs}{BNG(cmc)} & Cls &   3 &  55296 &   9 & \  & 192 & \href{https://www.openml.org/search?type=data&status=active&id=1046&sort=runs}{mozilla4} & Cls &   2 &  15545 &   4 & \  \\
 43 & \href{https://www.openml.org/search?type=data&status=active&id=1199&sort=runs}{BNG(echoMonths)} & Reg & $-$ &  17496 &   8 & \checkmark & 193 & \href{https://www.openml.org/search?type=data&status=active&id=344&sort=runs}{mv} & Reg & $-$ &  40768 &  10 & \  \\
 44 & \href{https://www.openml.org/search?type=data&status=active&id=1193&sort=runs}{BNG(lowbwt)} & Reg & $-$ &  31104 &   9 & \  & 194 & \href{https://www.openml.org/search?type=data&status=any&sort=runs&order=desc&qualities.NumberOfClasses=lte\_1&id=42821}{NASA\_PHM2008} & Reg & $-$ &  45918 &  21 & \  \\
 45 & \href{https://www.openml.org/search?type=data&status=active&id=1213&sort=runs}{BNG(mv)} & Reg & $-$ &  78732 &  10 & \checkmark & 195 & \href{https://archive.ics.uci.edu/dataset/722/naticusdroid+android+permissions+dataset}{naticusdroid\_permissions} & Cls &   2 &  29332 &  86 & \  \\
 46 & \href{https://www.openml.org/search?type=data&status=active&id=1200&sort=runs}{BNG(stock)} & Reg & $-$ &  59049 &   9 & \  & 196 & \href{https://archive.ics.uci.edu/dataset/887/national+health+and+nutrition+health+survey+2013-2014+(nhanes)+age+prediction+subset}{NHANES\_age\_prediction} & Reg & $-$ &   2277 &   7 & \  \\
 47 & \href{https://www.openml.org/search?type=data&status=active&id=137&sort=runs}{BNG(tic-tac-toe)} & Cls &   2 &  39366 &   9 & \  & 197 & \href{https://archive.ics.uci.edu/dataset/887/national+health+and+nutrition+health+survey+2013-2014+(nhanes)+age+prediction+subset}{Nutrition\_Health\_Survey} & Cls &   2 &   2278 &   7 & \  \\
 48 & \href{https://www.openml.org/search?type=data&status=active&id=531}{boston} & Reg & $-$ &    506 &  13 & \  & 198 & \href{https://www.openml.org/search?type=data&status=active&id=45067&sort=runs}{okcupid\_stem} & Cls &   3 &  26677 &  13 & \checkmark \\
 49 & \href{https://www.openml.org/search?type=data&status=active&id=44152&sort=runs}{Brazilian\_houses} & Reg & $-$ &  10692 &   8 & \  & 199 & \href{https://www.openml.org/search?type=data&status=active&id=45060&sort=runs}{online\_shoppers} & Cls &   2 &  12330 &  14 & \checkmark \\
 50 & \href{https://www.openml.org/search?type=data&status=active&id=45578&sort=runs}{California-Housing-Cls} & Cls &   2 &  20640 &   8 & \  & 200 & \href{https://www.openml.org/search?type=data&status=active&id=4545}{OnlineNewsPopularity} & Reg & $-$ &  39644 &  59 & \  \\
 51 & \href{https://www.openml.org/search?type=data&status=active&id=40664&sort=runs}{car-evaluation} & Cls &   4 &   1728 &  21 & \  & 201 & \href{https://www.openml.org/search?type=data&status=active&id=28&sort=runs}{optdigits} & Cls &  10 &   5620 &  64 & \  \\
 52 & \href{https://www.openml.org/search?type=data&status=active&id=45547&sort=runs}{Cardiovascular-Disease} & Cls &   2 &  70000 &  11 & \  & 202 & \href{https://www.openml.org/search?type=data&status=active&id=1487&sort=runs}{ozone-level-8hr} & Cls &   2 &   2534 &  72 & \  \\
 53 & \href{https://www.openml.org/search?type=data&status=active&id=703}{chscase\_foot} & Reg & $-$ &    526 &   5 & \  & 203 & \href{https://www.openml.org/search?type=data&status=active&id=301&sort=runs}{ozone\_level} & Cls &   2 &   2536 &  36 & \checkmark \\
 54 & \href{https://www.openml.org/search?type=data&status=active&id=40701&sort=runs}{churn} & Cls &   2 &   5000 &  20 & \  & 204 & \href{https://www.openml.org/search?type=data&status=active&id=30&sort=runs}{page-blocks} & Cls &   5 &   5473 &  10 & \  \\
 55 & \href{https://www.openml.org/search?type=data&status=any&sort=runs&order=desc&id=1220}{Click\_prediction} & Cls &   2 &  39948 &   3 & \  & 205 & \href{https://archive.ics.uci.edu/dataset/301/parkinson+speech+dataset+with+multiple+types+of+sound+recordings}{Parkinson\_Sound\_Record} & Reg & $-$ &   1040 &  26 & \  \\
 56 & \href{https://www.openml.org/search?type=data&status=active&id=23&sort=runs}{cmc} & Cls &   3 &   1473 &   9 & \  & 206 & \href{https://archive.ics.uci.edu/dataset/189/parkinsons+telemonitoring}{Parkinson\_Telemonitor} & Reg & $-$ &   5875 &  19 & \  \\
 57 & \href{https://www.openml.org/search?type=data&status=active&id=42727}{colleges} & Reg & $-$ &   7063 &  44 & \  & 207 & \href{https://www.openml.org/search?type=data&status=active&id=1068&sort=runs}{pc1} & Cls &   2 &   1109 &  21 & \  \\
 58 & \href{https://archive.ics.uci.edu/dataset/294/combined+cycle+power+plant}{combined\_cycle\_plant} & Reg & $-$ &   9568 &   4 & \  & 208 & \href{https://www.openml.org/search?type=data&status=active&id=1050&sort=runs}{pc3} & Cls &   2 &   1563 &  37 & \  \\
 59 & \href{https://archive.ics.uci.edu/dataset/183/communities+and+crime}{communities\_and\_crime} & Reg & $-$ &   1994 & 102 & \  & 209 & \href{https://www.openml.org/search?type=data&status=active&id=1049&sort=runs}{pc4} & Cls &   2 &   1458 &  37 & \checkmark \\
 60 & \href{https://www.kaggle.com/datasets/fedesoriano/company-bankruptcy-prediction}{company\_bankruptcy} & Cls &   2 &   6819 &  95 & \checkmark & 210 & \href{https://www.openml.org/search?type=data&status=active&id=32&sort=runs}{pendigits} & Cls &  10 &  10992 &  16 & \  \\
 61 & \href{https://www.openml.org/search?type=data&status=active&id=44162&sort=runs}{compass} & Cls &   2 &  16644 &  17 & \  & 211 & \href{https://www.openml.org/search?type=data&status=active&id=43812&sort=runs}{Performance-Prediction} & Cls &   2 &   1340 &  19 & \  \\
 62 & \href{https://archive.ics.uci.edu/dataset/165/concrete+compressive+strength}{compressive\_strength} & Reg & $-$ &   1030 &   8 & \checkmark & 212 & \href{https://www.openml.org/search?type=data&status=active&id=41145}{philippine} & Cls &   2 &   5832 & 308 & \  \\
 63 & \href{https://www.openml.org/search?type=data&status=active&id=40668}{connect-4} & Cls &   3 &  67557 &  42 & \  & 213 & \href{https://www.openml.org/search?type=data&status=active&id=4534&sort=runs}{PhishingWebsites} & Cls &   2 &  11055 &  30 & \checkmark \\
 64 & \href{https://www.openml.org/search?type=data&status=active&id=45538&sort=runs}{Contaminant-10.0GHz} & Cls &   2 &   2400 &  30 & \  & 214 & \href{https://www.openml.org/search?type=data&status=active&id=1489&sort=runs}{phoneme} & Cls &   2 &   5404 &   5 & \  \\
 65 & \href{https://www.openml.org/search?type=data&status=active&id=45539&sort=runs}{Contaminant-10.5GHz} & Cls &   2 &   2400 &  30 & \  & 215 & \href{https://archive.ics.uci.edu/dataset/265/physicochemical+properties+of+protein+tertiary+structure}{Physicochemical\_r} & Reg & $-$ &  45730 &   9 & \checkmark \\
 66 & \href{https://www.openml.org/search?type=data&status=active&id=45540&sort=runs}{Contaminant-11.0GHz} & Cls &   2 &   2400 &  30 & \  & 216 & \href{https://www.openml.org/search?type=data&status=active&id=1453&sort=runs}{PieChart3} & Cls &   2 &   1077 &  37 & \  \\
 67 & \href{https://www.openml.org/search?type=data&status=active&id=45536&sort=runs}{Contaminant-9.0GHz} & Cls &   2 &   2400 &  30 & \  & 217 & \href{https://www.kaggle.com/datasets/uciml/pima-indians-diabetes-database}{Pima\_Indians\_Diabetes} & Cls &   2 &    768 &   8 & \  \\
 68 & \href{https://www.openml.org/search?type=data&status=active&id=45537&sort=runs}{Contaminant-9.5GHz} & Cls &   2 &   2400 &  30 & \  & 218 & \href{https://www.openml.org/search?type=data&status=active&id=1444&sort=runs}{PizzaCutter3} & Cls &   2 &   1043 &  37 & \  \\
 69 & \href{https://archive.ics.uci.edu/dataset/30/contraceptive+method+choice}{contraceptive\_method} & Cls &   3 &   1473 &   9 & \  & 219 & \href{https://www.openml.org/search?type=data&status=active&id=44122&sort=runs}{pol} & Cls &   2 &  10082 &  26 & \  \\
 70 & \href{https://www.openml.org/search?type=data&status=active&id=45744&sort=runs}{CookbookReviews} & Reg & $-$ &  18182 &   7 & \  & 220 & \href{https://www.openml.org/search?type=data&status=active&id=201&sort=runs}{pol\_reg} & Reg & $-$ &  15000 &  48 & \  \\
 71 & \href{https://www.openml.org/search?type=data&status=active&id=41700&sort=runs}{CPMP-2015-regression} & Reg & $-$ &   2108 &  25 & \  & 221 & \href{None}{pole} & Reg & $-$ &  14998 &  26 & \checkmark \\
 72 & \href{https://www.openml.org/search?type=data&status=active&id=43963&sort=runs}{CPS1988} & Reg & $-$ &  28155 &   6 & \  & 222 & \href{https://archive.ics.uci.edu/dataset/697/predict+students+dropout+and+academic+success}{predict\_students\_dropout} & Cls &   3 &   4424 &  34 & \  \\
 73 & \href{https://www.openml.org/search?type=data&status=active&id=197&sort=runs}{cpu\_act} & Reg & $-$ &   8192 &  21 & \checkmark & 223 & \href{https://www.openml.org/search?type=data&status=active&id=308&sort=runs}{puma32H} & Reg & $-$ &   8192 &  32 & \  \\
 74 & \href{https://www.openml.org/search?type=data&status=active&id=562&sort=runs}{cpu\_small} & Reg & $-$ &   8192 &  12 & \  & 224 & \href{https://www.openml.org/search?type=data&status=active&id=225&sort=runs}{puma8NH} & Reg & $-$ &   8192 &   8 & \  \\
 75 & \href{https://www.openml.org/search?type=data&status=active&id=44089&sort=runs}{credit} & Cls &   2 &  16714 &  10 & \  & 225 & \href{https://www.kaggle.com/datasets/muratkokludataset/pumpkin-seeds-dataset}{Pumpkin\_Seeds} & Cls &   2 &   2500 &  12 & \  \\
 76 & \href{https://www.openml.org/search?type=data&status=active&id=31}{credit-g} & Cls &   2 &   1000 &  20 & \  & 226 & \href{https://archive.ics.uci.edu/dataset/505/qsar+aquatic+toxicity}{qsar\_aquatic\_toxicity} & Reg & $-$ &    546 &   8 & \checkmark \\
 77 & \href{https://www.kaggle.com/code/clkmuhammed/credit-score-classification-part-1-data-cleaning#B.-Numeric-Column-NaN-Values:-Reassign-Group-Min-Max}{Credit\_c} & Cls &   3 & 100000 &  22 & \  & 227 & \href{https://archive.ics.uci.edu/dataset/254/qsar+biodegradation}{QSAR\_biodegradation} & Cls &   2 &   1054 &  41 & \  \\
 78 & \href{https://archive.ics.uci.edu/dataset/350/default+of+credit+card+clients}{credit\_card\_defaults} & Cls &   2 &  30000 &  23 & \  & 228 & \href{https://archive.ics.uci.edu/dataset/504/qsar+fish+toxicity}{qsar\_fish\_toxicity} & Reg & $-$ &    908 &   6 & \  \\
 79 & \href{https://www.kaggle.com/datasets/imakash3011/customer-personality-analysis}{Customer\_Personality} & Cls &   2 &   2240 &  24 & \  & 229 & \href{https://www.kaggle.com/datasets/jsphyg/weather-dataset-rattle-package}{Rain\_in\_Australia} & Cls &   3 & 145460 &  18 & \  \\
 80 & \href{https://archive.ics.uci.edu/dataset/296/diabetes+130-us+hospitals+for+years+1999-2008}{dabetes\_us\_hospitals} & Cls &   2 & 101766 &  20 & \  & 230 & \href{https://archive.ics.uci.edu/dataset/329/diabetic+retinopathy+debrecen}{Retinopathy\_Debrecen} & Cls &   2 &   1151 &  19 & \  \\
 81 & \href{https://www.kaggle.com/datasets/arnabchaki/data-science-salaries-2023/data}{Data\_Science\_Salaries} & Reg & $-$ &   3755 &   5 & \  & 231 & \href{https://archive.ics.uci.edu/dataset/545/rice+cammeo+and+osmancik}{rice\_cammeo\_and\_osmancik} & Cls &   2 &   3810 &   7 & \checkmark \\
 82 & \href{https://www.openml.org/search?type=data&status=active&id=42183&sort=runs}{dataset\_sales} & Reg & $-$ &  10738 &  10 & \  & 232 & \href{https://www.openml.org/search?type=data&status=active&id=1496&sort=runs}{ringnorm} & Cls &   2 &   7400 &  20 & \  \\
 83 & \href{https://www.openml.org/search?type=data&status=active&id=23516&sort=runs}{debutanizer} & Reg & $-$ &   2394 &   7 & \  & 233 & \href{https://www.openml.org/search?type=data&status=active&id=44160&sort=runs}{rl} & Cls &   2 &   4970 &  12 & \  \\
 84 & \href{https://www.openml.org/search?type=data&status=active&id=803&sort=runs}{delta\_ailerons} & Cls &   2 &   7129 &   5 & \  & 234 & \href{https://www.openml.org/search?type=data&status=active&id=45718&sort=runs}{RSSI\_Estimation} & Reg & $-$ &   5760 &   6 & \checkmark \\
 85 & \href{https://www.openml.org/search?type=data&status=active&id=198&sort=runs}{delta\_elevators} & Reg & $-$ &   9517 &   6 & \  & 235 & \href{https://www.openml.org/search?type=data&status=active&id=45720&sort=runs}{RSSI\_Estimation\_1} & Reg & $-$ &  14400 &  12 & \  \\
 86 & \href{https://www.kaggle.com/datasets/joebeachcapital/diamonds}{Diamonds} & Reg & $-$ &  53940 &   9 & \  & 236 & \href{https://www.openml.org/search?type=data&status=any&id=41980}{SAT11-HAND-regression} & Reg & $-$ &   4440 & 116 & \  \\
 87 & \href{https://www.openml.org/search?type=data&status=active&id=40713&sort=runs}{dis} & Cls &   2 &   3772 &  29 & \  & 237 & \href{https://www.openml.org/search?type=data&status=active&id=294&sort=runs}{satellite\_image} & Reg & $-$ &   6435 &  36 & \  \\
 88 & \href{https://www.openml.org/search?type=data&status=active&id=40670&sort=runs}{dna} & Cls &   3 &   3186 & 180 & \  & 238 & \href{https://www.openml.org/search?type=data&status=active&id=182&sort=runs}{satimage} & Cls &   6 &   6430 &  36 & \  \\
 89 & \href{https://archive.ics.uci.edu/dataset/373/drug+consumption+quantified}{drug\_consumption} & Cls &   7 &   1884 &  12 & \  & 239 & \href{https://www.kaggle.com/datasets/fedesoriano/stellar-classification-dataset-sdss17}{SDSS17} & Cls &   3 & 100000 &  12 & \  \\
 90 & \href{https://archive.ics.uci.edu/dataset/602/dry+bean+dataset}{dry\_bean\_dataset} & Cls &   7 &  13611 &  16 & \  & 240 & \href{https://www.openml.org/search?type=data&status=active&id=36&sort=runs}{segment} & Cls &   7 &   2310 &  17 & \  \\
 91 & \href{https://www.kaggle.com/datasets/prachi13/customer-analytics}{E-CommereShippingData} & Cls &   2 &  10999 &  10 & \  & 241 & \href{https://archive.ics.uci.edu/dataset/266/seismic+bumps}{seismic+bumps} & Cls &   2 &   2584 &  18 & \  \\
 92 & \href{https://www.openml.org/search?type=data&status=active&id=1471&sort=runs}{eeg-eye-state} & Cls &   2 &  14980 &  14 & \  & 242 & \href{https://www.openml.org/search?type=data&status=active&id=1501&sort=runs}{semeion} & Cls &  10 &   1593 & 256 & \  \\
 93 & \href{https://www.openml.org/search?type=data&status=active&id=151&sort=runs}{electricity} & Cls &   2 &  45312 &   8 & \  & 243 & \href{https://www.openml.org/search?type=data&status=any&sort=runs&order=desc&qualities.NumberOfClasses=lte\_1&id=546}{sensory} & Reg & $-$ &    576 &  11 & \  \\
 94 & \href{https://www.openml.org/search?type=data&status=active&id=216&sort=runs}{elevators} & Reg & $-$ &  16599 &  18 & \  & 244 & \href{https://www.openml.org/search?type=data&status=active&id=42889&sort=runs}{shill-bidding} & Cls &   2 &   6321 &   3 & \checkmark \\
 95 & \href{https://www.kaggle.com/datasets/tawfikelmetwally/employee-dataset}{Employee} & Cls &   2 &   4653 &   8 & \  & 245 & \href{https://www.openml.org/search?type=data&status=active&id=45074&sort=runs}{Shipping} & Cls &   2 &  10999 &   9 & \checkmark \\
 96 & \href{https://archive.ics.uci.edu/dataset/544/estimation+of+obesity+levels+based+on+eating+habits+and+physical+condition}{estimation\_of\_obesity} & Cls &   7 &   2111 &  16 & \  & 246 & \href{https://www.kaggle.com/datasets/datascientistanna/customers-dataset}{Shop\_Customer\_Data} & Reg & $-$ &   2000 &   6 & \  \\
 97 & \href{https://www.openml.org/search?type=data&status=active&id=188}{eucalyptus} & Cls &   5 &    736 &  19 & \checkmark & 247 & \href{https://www.openml.org/search?type=data&status=active&id=45062&sort=runs}{shrutime} & Cls &   2 &  10000 &  10 & \  \\
 98 & \href{https://www.openml.org/search?type=data&status=active&id=1044&sort=runs}{eye\_movements} & Cls &   3 &  10936 &  27 & \  & 248 & \href{https://www.openml.org/search?type=data&status=active&id=40685&sort=runs}{shuttle} & Cls &   7 &  58000 &   9 & \  \\
 99 & \href{https://www.openml.org/search?type=data&status=active&id=44130&sort=runs}{eye\_movements\_bin} & Cls &   2 &   7608 &  20 & \  & 249 & \href{https://www.openml.org/search?type=data&status=active&id=541&sort=runs}{socmob} & Reg & $-$ &   1156 &   5 & \  \\
100 & \href{https://archive.ics.uci.edu/dataset/363/facebook+comment+volume+dataset}{Facebook\_Comment\_Volume} & Reg & $-$ &  40949 &  53 & \  & 250 & \href{https://www.openml.org/search?type=data&status=active&id=507&sort=runs}{space\_ga} & Reg & $-$ &   3107 &   6 & \  \\
101 & \href{https://www.openml.org/search?type=data&status=active&id=43828&sort=runs}{Fiat} & Reg & $-$ &   1538 &   6 & \  & 251 & \href{https://www.openml.org/search?type=data&status=active&id=44&sort=runs}{spambase} & Cls &   2 &   4601 &  57 & \  \\
102 & \href{https://www.openml.org/search?type=data&status=active&id=45553&sort=runs}{FICO-HELOC-cleaned} & Cls &   2 &   9871 &  23 & \  & 252 & \href{https://www.openml.org/search?type=data&status=active&id=46&sort=runs}{splice} & Cls &   3 &   3190 &  60 & \  \\
103 & \href{https://www.openml.org/search?type=data&status=active&id=44026&sort=runs}{fifa} & Reg & $-$ &  18063 &   5 & \checkmark & 253 & \href{https://archive.ics.uci.edu/dataset/450/sports+articles+for+objectivity+analysis}{sports\_articles} & Cls &   2 &   1000 &  59 & \  \\
104 & \href{https://archive.ics.uci.edu/dataset/324/firm+teacher+clave+direction+classification}{Firm-Teacher-Direction} & Cls &   4 &  10800 &  16 & \  & 254 & \href{https://archive.ics.uci.edu/dataset/144/statlog+german+credit+data}{statlog} & Cls &   2 &   1000 &  20 & \checkmark \\
105 & \href{https://www.openml.org/search?type=data&status=active&id=1475&sort=runs}{first-order-theorem} & Cls &   6 &   6118 &  51 & \  & 255 & \href{https://archive.ics.uci.edu/dataset/851/steel+industry+energy+consumption}{steel\_industry\_energy} & Reg & $-$ &  35040 &  10 & \  \\
106 & \href{https://www.kaggle.com/datasets/ddosad/datacamps-data-science-associate-certification}{Fitness\_Club\_c} & Cls &   2 &   1500 &   6 & \  & 256 & \href{https://archive.ics.uci.edu/dataset/198/steel+plates+faults}{steel\_plates\_faults} & Cls &   7 &   1941 &  27 & \  \\
107 & \href{https://www.kaggle.com/datasets/rajatkumar30/food-delivery-time}{Food\_Delivery\_Time} & Reg & $-$ &  45593 &   8 & \  & 257 & \href{https://www.openml.org/search?type=data&status=any&sort=runs&order=desc&qualities.NumberOfClasses=lte\_1&id=223}{stock} & Reg & $-$ &    950 &   9 & \  \\
108 & \href{https://www.openml.org/search?type=data&status=active&id=41721&sort=runs}{FOREX\_audcad-day-High} & Cls &   2 &   1834 &  10 & \  & 258 & \href{https://www.openml.org/search?type=data&status=active&id=42545&sort=runs}{stock\_fardamento02} & Reg & $-$ &   6277 &   6 & \  \\
109 & \href{https://www.openml.org/search?type=data&status=active&id=41763&sort=runs}{FOREX\_audcad-hour-High} & Cls &   2 &  43825 &  10 & \  & 259 & \href{https://www.openml.org/search?type=data&status=active&id=23515&sort=runs}{sulfur} & Reg & $-$ &  10081 &   6 & \  \\
110 & \href{https://www.openml.org/search?type=data&status=active&id=41875&sort=runs}{FOREX\_audchf-day-High} & Cls &   2 &   1833 &  10 & \  & 260 & \href{https://archive.ics.uci.edu/dataset/464/superconductivty+data}{Superconductivty} & Reg & $-$ &  21197 &  81 & \  \\
111 & \href{https://www.openml.org/search?type=data&status=active&id=41882&sort=runs}{FOREX\_audjpy-day-High} & Cls &   2 &   1832 &  10 & \  & 261 & \href{https://www.openml.org/search?type=data&status=active&id=1589&sort=runs}{svmguide3} & Cls &   2 &   1243 &  22 & \  \\
112 & \href{https://www.openml.org/search?type=data&status=active&id=41718&sort=runs}{FOREX\_audjpy-hour-High} & Cls &   2 &  43825 &  10 & \  & 262 & \href{https://www.openml.org/search?type=data&status=active&id=41146&sort=runs}{sylvine} & Cls &   2 &   5124 &  20 & \  \\
113 & \href{https://www.openml.org/search?type=data&status=active&id=41865&sort=runs}{FOREX\_audsgd-hour-High} & Cls &   2 &  43825 &  10 & \  & 263 & \href{https://archive.ics.uci.edu/dataset/572/taiwanese+bankruptcy+prediction}{taiwanese\_bankruptcy} & Cls &   2 &   6819 &  95 & \  \\
114 & \href{https://www.openml.org/search?type=data&status=active&id=41843&sort=runs}{FOREX\_audusd-hour-High} & Cls &   2 &  43825 &  10 & \  & 264 & \href{https://www.openml.org/search?type=data&status=active&id=42178&sort=runs}{telco-customer-churn} & Cls &   2 &   7043 &  18 & \  \\
115 & \href{https://www.openml.org/search?type=data&status=active&id=41839&sort=runs}{FOREX\_cadjpy-day-High} & Cls &   2 &   1834 &  10 & \  & 265 & \href{https://www.kaggle.com/datasets/mnassrib/telecom-churn-datasets}{Telecom\_Churn\_Dataset} & Cls &   2 &   3333 &  17 & \  \\
116 & \href{https://www.openml.org/search?type=data&status=active&id=41833&sort=runs}{FOREX\_cadjpy-hour-High} & Cls &   2 &  43825 &  10 & \  & 266 & \href{https://www.openml.org/search?type=data&status=active&id=40499&sort=runs}{texture} & Cls &  11 &   5500 &  40 & \  \\
117 & \href{https://www.openml.org/search?type=data&status=active&id=564&sort=runs}{fried} & Reg & $-$ &  40768 &  10 & \  & 267 & \href{https://archive.ics.uci.edu/dataset/102/thyroid+disease}{thyroid} & Cls &   3 &   7200 &  21 & \checkmark \\
118 & \href{https://www.openml.org/search?type=data&status=active&id=40646&sort=runs}{GAMETES\_Epistasis} & Cls &   2 &   1600 &  20 & \  & 268 & \href{https://www.openml.org/search?type=data&status=active&id=40497&sort=runs}{thyroid-ann} & Cls &   3 &   3772 &  21 & \  \\
119 & \href{https://www.openml.org/search?type=data&status=active&id=40649&sort=runs}{GAMETES\_Heterogeneity} & Cls &   2 &   1600 &  20 & \  & 269 & \href{https://www.openml.org/search?type=data&status=active&id=40478&sort=runs}{thyroid-dis} & Cls &   5 &   2800 &  26 & \  \\
120 & \href{https://archive.ics.uci.edu/dataset/597/productivity+prediction+of+garment+employees}{garments\_productivity} & Reg & $-$ &   1197 &  13 & \  & 270 & \href{https://www.openml.org/search?type=data&status=any&sort=runs&order=desc&qualities.NumberOfClasses=lte\_1&id=422}{topo\_2\_1} & Reg & $-$ &   8885 & 266 & \  \\
121 & \href{https://www.openml.org/search?type=data&status=active&id=1476&sort=runs}{gas-drift} & Cls &   6 &  13910 & 128 & \  & 271 & \href{https://www.openml.org/search?type=data&status=active&id=42367&sort=runs}{treasury} & Reg & $-$ &   1049 &  15 & \  \\
122 & \href{https://archive.ics.uci.edu/dataset/551/gas+turbine+co+and+nox+emission+data+set}{gas\_turbine\_emission} & Reg & $-$ &  36733 &  10 & \  & 272 & \href{https://archive.ics.uci.edu/dataset/262/turkiye+student+evaluation}{turiye\_student} & Cls &   5 &   5820 &  32 & \  \\
123 & \href{https://archive.ics.uci.edu/dataset/852/gender+gap+in+spanish+wp}{Gender\_Gap\_in\_Spanish} & Cls &   3 &   4746 &  13 & \checkmark & 273 & \href{https://www.openml.org/search?type=data&status=active&id=1507&sort=runs}{twonorm} & Cls &   2 &   7400 &  20 & \  \\
124 & \href{https://www.openml.org/search?type=data&status=active&id=4538&sort=runs}{Gesture\_Phase\_Segment} & Cls &   5 &   9873 &  32 & \  & 274 & \href{https://archive.ics.uci.edu/dataset/160/uji+pen+characters}{UJI\_Pen\_Characters} & Cls &  35 &   1364 &  80 & \  \\
125 & \href{https://www.kaggle.com/datasets/samybaladram/golf-play-extended}{golf\_play\_extended} & Cls &   2 &   1095 &   9 & \  & 275 & \href{https://www.openml.org/search?type=data&status=active&id=42730}{us\_crime} & Reg & $-$ &   1994 & 126 & \  \\
126 & \href{https://www.openml.org/search?type=data&status=active&id=43785&sort=runs}{Goodreads-Computer-Book} & Reg & $-$ &   1234 &   5 & \checkmark & 276 & \href{https://www.openml.org/search?type=data&status=active&id=54&sort=runs}{vehicle} & Cls &   4 &    846 &  18 & \  \\
127 & \href{https://www.kaggle.com/datasets/arunjangir245/healthcare-insurance-expenses/}{healthcare\_expenses} & Reg & $-$ &   1338 &   6 & \  & 277 & \href{https://archive.ics.uci.edu/dataset/363/facebook+comment+volume+dataset}{volume} & Reg & $-$ &  50993 &  53 & \  \\
128 & \href{https://www.openml.org/search?type=data&status=active&id=43672&sort=runs}{Heart-Disease-Dataset} & Cls &   2 &   1190 &  11 & \  & 278 & \href{https://www.openml.org/search?type=data&status=active&id=44150&sort=runs}{VulNoneVul} & Cls &   2 &   5692 &  16 & \  \\
129 & \href{https://www.openml.org/search?type=data&status=active&id=41169}{helena} & Cls & 100 &  65196 &  27 & \  & 279 & \href{https://www.openml.org/search?type=data&status=active&id=1509&sort=runs}{walking-activity} & Cls &  22 & 149332 &   4 & \  \\
130 & \href{https://www.openml.org/search?type=data&status=active&id=45023&sort=runs}{heloc} & Cls &   2 &  10000 &  22 & \  & 280 & \href{https://www.openml.org/search?type=data&status=active&id=1497&sort=runs}{wall-robot-navigation} & Cls &   4 &   5456 &  24 & \  \\
131 & \href{https://www.openml.org/search?type=data&status=active&id=1479&sort=runs}{hill-valley} & Cls &   2 &   1212 & 100 & \checkmark & 281 & \href{https://www.kaggle.com/datasets/uom190346a/water-quality-and-potability}{Water\_Potability} & Cls &   2 &   3276 &   8 & \  \\
132 & \href{https://www.openml.org/search?type=data&status=active&id=44123&sort=runs}{house\_16H} & Cls &   2 &  13488 &  16 & \checkmark & 282 & \href{https://www.kaggle.com/datasets/mssmartypants/water-quality}{water\_quality} & Cls &   2 &   7996 &  20 & \  \\
133 & \href{https://www.openml.org/search?type=data&status=active&id=574&sort=runs}{house\_16H\_reg} & Reg & $-$ &  22784 &  16 & \  & 283 & \href{https://www.openml.org/search?type=data&status=active&id=42464&sort=runs}{Waterstress} & Cls &   2 &   1188 &  22 & \  \\
134 & \href{https://www.openml.org/search?type=data&status=active&id=218&sort=runs}{house\_8L} & Reg & $-$ &  22784 &   8 & \checkmark & 284 & \href{https://archive.ics.uci.edu/dataset/882/large-scale+wave+energy+farm}{Wave\_Energy\_Perth\_100} & Reg & $-$ &   7277 & 201 & \  \\
135 & \href{https://www.openml.org/search?type=data&status=any&id=42563}{house\_prices\_nominal} & Reg & $-$ &   1460 &  79 & \  & 285 & \href{https://archive.ics.uci.edu/dataset/882/large-scale+wave+energy+farm}{Wave\_Energy\_Sydney\_100} & Reg & $-$ &   2318 & 201 & \  \\
136 & \href{https://www.openml.org/search?type=data&status=active&id=42635&sort=runs}{house\_sales\_reduced} & Reg & $-$ &  21613 &  18 & \  & 286 & \href{https://archive.ics.uci.edu/dataset/882/large-scale+wave+energy+farm}{Wave\_Energy\_Sydney\_49} & Reg & $-$ &  17964 &  99 & \  \\
137 & \href{https://www.openml.org/search?type=data&status=active&id=537&sort=runs}{houses} & Reg & $-$ &  20640 &   8 & \  & 287 & \href{https://archive.ics.uci.edu/dataset/107/waveform+database+generator+version+1}{waveform-v1} & Cls &   3 &   5000 &  21 & \checkmark \\
138 & \href{https://www.kaggle.com/datasets/harishkumardatalab/housing-price-prediction}{housing\_price\_prediction} & Reg & $-$ &    545 &  12 & \  & 288 & \href{https://www.openml.org/search?type=data&status=active&id=60&sort=runs}{waveform-v2} & Cls &   3 &   5000 &  40 & \  \\
139 & \href{https://www.kaggle.com/datasets/arashnic/hr-analytics-job-change-of-data-scientists}{HR\_Analytics} & Cls &   2 &  19158 &  13 & \  & 289 & \href{https://www.openml.org/search?type=data&status=active&id=42369&sort=runs}{weather\_izmir} & Reg & $-$ &   1461 &   9 & \  \\
140 & \href{https://archive.ics.uci.edu/dataset/372/htru2}{htru} & Cls &   2 &  17898 &   8 & \  & 290 & \href{https://archive.ics.uci.edu/dataset/379/website+phishing}{website\_phishing} & Cls &   3 &   1353 &   9 & \  \\
141 & \href{https://www.openml.org/search?type=data&status=active&id=43895&sort=runs}{ibm-employee-performance} & Cls &   2 &   1470 &  30 & \checkmark & 291 & \href{https://www.openml.org/search?type=data&id=44489}{Wilt} & Cls &   2 &   4821 &   5 & \  \\
142 & \href{https://www.openml.org/search?type=data&status=active&id=43180&sort=runs}{IEEE80211aa-GATS} & Reg & $-$ &   4046 &  27 & \checkmark & 292 & \href{https://www.openml.org/search?type=data&status=active&id=503&sort=runs}{wind} & Reg & $-$ &   6574 &  14 & \  \\
143 & \href{https://archive.ics.uci.edu/dataset/603/in+vehicle+coupon+recommendation}{in\_vehicle\_coupon} & Cls &   2 &  12684 &  21 & \  & 293 & \href{https://www.openml.org/search?type=data&status=active&id=44091&sort=runs}{wine} & Cls &   2 &   2554 &   4 & \  \\
144 & \href{https://www.openml.org/search?type=data&status=active&id=41972&sort=runs}{Indian\_pines} & Cls &   8 &   9144 & 220 & \  & 294 & \href{https://archive.ics.uci.edu/dataset/186/wine+quality}{wine+quality} & Reg & $-$ &   6497 &  11 & \  \\
145 & \href{https://www.kaggle.com/datasets/mariyamalshatta/inn-hotels-group}{INNHotelsGroup} & Cls &   2 &  36275 &  17 & \checkmark & 295 & \href{https://www.openml.org/search?type=data&status=active&id=40691&sort=runs}{wine-quality-red} & Cls &   6 &   1599 &   4 & \checkmark \\
146 & \href{https://www.openml.org/search?type=data&status=active&id=45064&sort=runs}{Insurance} & Cls &   2 &  23548 &  10 & \  & 296 & \href{https://www.openml.org/search?type=data&status=active&id=40498&sort=runs}{wine-quality-white} & Cls &   7 &   4898 &  11 & \  \\
147 & \href{https://archive.ics.uci.edu/dataset/542/internet+firewall+data}{internet\_firewall} & Cls &   4 &  65532 &   7 & \checkmark & 297 & \href{https://archive.ics.uci.edu/dataset/186/wine+quality}{Wine\_Quality\_red} & Reg & $-$ &   1599 &  11 & \  \\
148 & \href{https://www.openml.org/search?type=data&status=active&id=372&sort=runs}{internet\_usage} & Cls &  46 &  10108 &  70 & \  & 298 & \href{https://archive.ics.uci.edu/dataset/186/wine+quality}{Wine\_Quality\_white} & Reg & $-$ &   4898 &  11 & \  \\
149 & \href{https://www.openml.org/search?type=data&status=active&id=45040&sort=runs}{Intersectional-Bias} & Cls &   2 &  11000 &  19 & \  & 299 & \href{https://www.openml.org/search?type=data&status=active&id=181&sort=runs}{yeast} & Cls &  10 &   1484 &   8 & \  \\
150 & \href{https://www.openml.org/search?type=data&status=active&id=43442&sort=runs}{Is-this-a-good-customer} & Cls &   2 &   1723 &  13 & \  & 300 & \href{https://www.openml.org/search?type=data&status=active&id=416}{yprop\_4\_1} & Reg & $-$ &   8885 & 251 & \  \\

% \midrule
% \multicolumn{14}{c}{{\footnotesize{\name}-Extension (High-dimension)}} \\
% \midrule
\specialrule{\lightrulewidth}{2pt}{0pt}
\rowcolor{gray!20}\multicolumn{14}{c}{\rule{0pt}{3ex}\footnotesize{\name}-Extension (high-dimensional)} \\
\specialrule{\lightrulewidth}{0pt}{3pt}

ID & \multicolumn{1}{c}{Name (Source)} & Task & Class & Sample & Feature &  & ID & \multicolumn{1}{c}{Name (Source)} & Task & Class & Sample & Feature &  \\
\midrule
  1 & \href{https://jundongl.github.io/scikit-feature/files/datasets/ALLAML.mat}{ALLAML} & Cls &   2 &     72 & 7129 &  &  10 & \href{https://jundongl.github.io/scikit-feature/files/datasets/lung.mat}{lung} & Cls &   5 &    203 & 3312 &  \\
  2 & \href{https://jundongl.github.io/scikit-feature/files/datasets/arcene.mat}{arcene} & Cls &   2 &    200 & 10000 &  &  11 & \href{https://jundongl.github.io/scikit-feature/files/datasets/orlraws10P.mat}{orlraws10P} & Cls &  10 &    100 & 10304 &  \\
  3 & \href{https://jundongl.github.io/scikit-feature/files/datasets/BASEHOCK.mat}{BASEHOCK} & Cls &   2 &   1993 & 4862 &  &  12 & \href{https://jundongl.github.io/scikit-feature/files/datasets/PCMAC.mat}{PCMAC} & Cls &   2 &   1943 & 3289 &  \\
  4 & \href{https://jundongl.github.io/scikit-feature/files/datasets/CLL_SUB_111.mat}{CLL\_SUB\_111} & Cls &   3 &    111 & 11340 &  &  13 & \href{https://jundongl.github.io/scikit-feature/files/datasets/Prostate_GE.mat}{Prostate\_GE} & Cls &   2 &    102 & 5966 &  \\
  5 & \href{https://jundongl.github.io/scikit-feature/files/datasets/colon.mat}{colon} & Cls &   2 &     62 & 2000 &  &  14 & \href{https://jundongl.github.io/scikit-feature/files/datasets/RELATHE.mat}{RELATHE} & Cls &   2 &   1427 & 4322 &  \\
  6 & \href{https://jundongl.github.io/scikit-feature/files/datasets/gisette.mat}{gisette} & Cls &   2 &   7000 & 5000 &  &  15 & \href{https://jundongl.github.io/scikit-feature/files/datasets/SMK_CAN_187.mat}{SMK\_CAN\_187} & Cls &   2 &    187 & 19993 &  \\
  7 & \href{https://jundongl.github.io/scikit-feature/files/datasets/GLI_85.mat}{GLI\_85} & Cls &   2 &     85 & 22283 &  &  16 & \href{https://jundongl.github.io/scikit-feature/files/datasets/TOX_171.mat}{TOX\_171} & Cls &   4 &    171 & 5748 &  \\
  8 & \href{https://jundongl.github.io/scikit-feature/files/datasets/GLIOMA.mat}{GLIOMA} & Cls &   4 &     50 & 4434 &  &  17 & \href{https://jundongl.github.io/scikit-feature/files/datasets/warpAR10P.mat}{warpAR10P} & Cls &  10 &    130 & 2400 &  \\
  9 & \href{https://jundongl.github.io/scikit-feature/files/datasets/leukemia.mat}{leukemia} & Cls &   2 &     72 & 7070 &  &  18 & \href{https://jundongl.github.io/scikit-feature/files/datasets/warpPIE10P.mat}{warpPIE10P} & Cls &  10 &    210 & 2420 &  \\

% \midrule
% \multicolumn{14}{c}{{\footnotesize{\name}-Extension (Many-class)}} \\
% \midrule
\specialrule{\lightrulewidth}{2pt}{0pt}
\rowcolor{gray!20}\multicolumn{14}{c}{\rule{0pt}{3ex}\footnotesize{\name}-Extension (many-class)} \\
\specialrule{\lightrulewidth}{0pt}{3pt}

ID & \multicolumn{1}{c}{Name (Source)} & Task & Class & Sample & Feature &  & ID & \multicolumn{1}{c}{Name (Source)} & Task & Class & Sample & Feature &  \\
\midrule
  1 & \href{https://www.openml.org/search?type=data&status=active&id=1592}{aloi} & Cls & 1000 & 108000 & 128 &   &  4 & \href{https://www.openml.org/search?type=data&status=active&id=41167}{dionis} & Cls & 355 & 416188 &  60 & \\
  2 & \href{https://www.openml.org/search?type=data&status=active&id=4552}{BachChoralHarmony} & Cls & 102 &   5665 &  15 &   &  5 & \href{https://www.openml.org/search?type=data&status=active&id=45049}{MD\_MIX\_Mini\_Copy} & Cls & 706 &  28240 &  31 & \\
  3 & \href{https://www.openml.org/search?type=data&status=active&id=42087}{beer\_reviews} & Cls & 104 & 1586614 &  12 &   &  6 & \href{https://www.openml.org/search?type=data&status=active&id=41960}{seattlecrime6} & Cls & 135 & 523577 &   7 & \\

% \midrule
% \multicolumn{14}{c}{{\footnotesize{\name}-Extension (Large-scale)}} \\
% \midrule
\specialrule{\lightrulewidth}{2pt}{0pt}
\rowcolor{gray!20}\multicolumn{14}{c}{\rule{0pt}{3ex}\footnotesize{\name}-Extension (very-large-scale)} \\
\specialrule{\lightrulewidth}{0pt}{3pt}

ID & \multicolumn{1}{c}{Name (Source)} & Task & Class & Sample & Feature &  & ID & \multicolumn{1}{c}{Name (Source)} & Task & Class & Sample & Feature &  \\
\midrule
  1 & \href{https://www.openml.org/search?type=data&status=active&id=42728}{Airlines\_DepDelay\_10M} & Reg & $-$ & 10000000 &   9 &  &  12 & \href{https://www.openml.org/search?type=data&status=active&id=41168}{jannis} & Cls &   4 &  83733 &  54 &  \\
  2 & \href{https://archive.ics.uci.edu/dataset/304/blogfeedback}{blogfeedback} & Reg & $-$ &  60021 & 276 &  &  13 & \href{https://www.openml.org/search?type=data&status=active&id=42746}{KDDCup99} & Cls &  23 & 4898431 &  41 &  \\
  3 & \href{https://www.openml.org/search?type=data&status=active&id=258}{BNG(credit-a)} & Cls &   2 & 1000000 &  15 &  &  14 & \href{https://www.openml.org/search?type=data&status=active&id=45579}{microsoft} & Reg & $-$ & 1200192 & 136 &  \\
  4 & \href{https://archive.ics.uci.edu/dataset/891/cdc+diabetes+health+indicators}{CDC\_Diabetes\_Health} & Cls &   2 & 253680 &  21 &  &  15 & \href{https://www.openml.org/search?type=data&status=active&id=1486}{nomao} & Cls &   2 &  34465 & 118 &  \\
  5 & \href{https://www.openml.org/search?type=data&status=active&id=150}{covertype} & Cls &   7 & 581012 &  54 &  &  16 & \href{https://www.openml.org/search?type=data&status=active&id=1567}{poker-hand} & Cls &  10 & 1025009 &  10 &  \\
  6 & \href{https://www.kaggle.com/datasets/kiva/data-science-for-good-kiva-crowdfunding}{Data\_Science\_Good\_Kiva} & Cls &   4 & 671205 &  11 &  &  17 & \href{https://www.openml.org/search?type=data&status=active&id=42344}{sf-police-incidents} & Cls &   2 & 2215023 &   8 &  \\
  7 & \href{https://www.openml.org/search?type=data&status=active&id=41163}{dilbert} & Cls &   5 &  10000 & 2000 &  &  18 & \href{https://www.kaggle.com/datasets/sooyoungher/smoking-drinking-dataset/data}{Smoking\_and\_Drinking} & Cls &   2 & 991346 &  23 &  \\
  8 & \href{https://www.openml.org/search?type=data&status=active&id=41164}{fabert} & Cls &   7 &   8237 & 800 &  &  19 & \href{https://archive.ics.uci.edu/dataset/310/ujiindoorloc}{UJIndoorLoc} & Reg & $-$ &  21048 & 520 &  \\
  9 & \href{https://www.openml.org/search?type=data&status=active&id=40996}{Fashion-MNIST} & Cls &  10 &  70000 & 784 &  &  20 & \href{https://www.openml.org/search?type=data&status=active&id=41166&sort=runs}{volkert} & Cls &  10 &  58310 & 180 &  \\
 10 & \href{https://www.openml.org/search?type=data&status=any&sort=runs&order=desc&id=1038}{gina\_agnostic} & Cls &   2 &   3468 & 970 &  &  21 & \href{https://archive.ics.uci.edu/dataset/882/large-scale+wave+energy+farm}{Wave\_Energy\_Perth\_49} & Reg & $-$ &  36043 &  99 &  \\
 11 & \href{https://archive.ics.uci.edu/dataset/280/higgs}{Higgs} & Cls &   2 & 1000000 &  28 &  &  22 & \href{https://webscope.sandbox.yahoo.com/catalog.php?datatype=c&guccounter=1&guce\_referrer=aHR0cHM6Ly93d3cuZ29vZ2xlLmNvbS5oay8&guce\_referrer\_sig=AQAAAG9yp-KLlry\_y1FwAHjGXEpNTR0uGEVXv-xZVSxDprnGdXGuD8sIHrUO2znvYPzNo3pTnHcDWwfON0mElgpTdswbhDpWXy68Jxy3F-FbxUd8AOv9OEA5SvMBjn6ET1JFc8aIZYxvySrBcKeeBgLxSQg7ZRGUL6S\_JzBlhPz\_kAg5}{yahoo} & Reg & $-$ & 709877 & 699 &  \\
\bottomrule
\end{longtable}
}}

\subsection{Dataset Splits and Evaluation Criteria}
\noindent{\bf Implementation details.}  
We evaluate all methods described in~\autoref{sec:related_methods}. Because tabular models are sensitive to hyperparameters, we adopt a uniform tuning protocol to ensure fairness. Full $K$-fold cross-validation would be robust but computationally prohibitive at our scale; applying CV only to small datasets would introduce inconsistent selection pressure due to arbitrary size thresholds. We therefore use a single, fixed hold-out protocol for the main benchmark and study CV+ensembling separately on {\name}-tiny (see~\autoref{sec:tiny-bench}).

\noindent\textit{Data splits and tuning.}
Each dataset is randomly split into train/val/test with proportions $64\%/16\%/20\%$ following the setup in~\citep{GorishniyRKB21Revisiting,gorishniy2023tabr}. Hyperparameters are selected on the validation split, and early stopping is triggered by the task metric on validation (accuracy for classification; RMSE for regression). We use Optuna~\citep{akiba2019optuna} with a fixed budget of 100 trials per method–dataset pair. After selecting the best configuration, we retrain and evaluate each model with 15 random seeds and report the mean across seeds. In~\autoref{sec:tiny-bench} we compare this protocol to CV+ensembles and show that, while CV can improve absolute scores, the relative ordering of methods remains largely unchanged.

\noindent\textit{Preprocessing.}
We follow the pipeline of~\citep{GorishniyRKB21Revisiting}. Numerical features are imputed by column means and standardized (zero mean, unit variance). Categorical features are ordinally encoded; missing categories are mapped to a dedicated token ``$-1$''. For non-deep methods (except CatBoost) and for deep methods without an explicit categorical module, we apply one-hot encoding after the ordinal step.

\noindent\textit{Method-specific settings.}
For gradient boosting, we explicitly pass feature types to CatBoost (native categorical handling). For all deep methods, we use AdamW~\citep{LoshchilovH19AdamW} and a batch size of 1024 unless noted otherwise. Pretrained tabular models are evaluated from their latest public checkpoints with default inference hyperparameters (i.e., no per-dataset tuning). The complete search spaces and per-method training options are available at \url{https://github.com/LAMDA-Tabular/TALENT/tree/main/TALENT/configs}.

\noindent{\bf Evaluation criteria.}
For classification tasks, we evaluate models using accuracy (higher is better) as the primary metric and use Root Mean Square Error (RMSE, lower is better) for regression tasks to select the best-performing model during training on the validation set. Additionally, for classification tasks, we record F1 and AUC scores, which are especially valuable for imbalanced datasets. For regression tasks, we also compute MAE and R\(^2\) to provide complementary evaluations of test set performance.

To aggregate per-dataset performance and provide a holistic evaluation across all datasets, we adopt several criteria:
\begin{itemize}[noitemsep,topsep=0pt,leftmargin=*]
    \item \textbf{Average Rank.} Following~\citet{DelgadoCBA14,McElfreshKVCRGW23when}, we report the average performance rank across all methods and datasets (lower is better). 
    \item \textbf{Statistical Comparison.} To assess significant differences between methods, we plot critical difference diagrams via Wilcoxon-Holm test~\citep{Demsar06Statistical,McElfreshKVCRGW23when} and paired t-test heatmaps to illustrate statistical comparisons.
    \item \textbf{Average relative improvement.} Following~\citet{Yury2024TabM}, we calculate the relative improvement of a tabular method \wrt the performance of a tabular baseline, \eg, MLP,  and report the average value across all methods and datasets (higher is better). 
    \item \textbf{Aggregated Performance.} We aggregate per-dataset results using the Shifted Geometric Mean (SGM)~\citep{David2024RealMLP}. For classification tasks, we use classification error (1 – accuracy), and for regression tasks, we use normalized RMSE (nRMSE).
    \item \textbf{PAMA (Probability of Achieving the Best Accuracy).} The fraction of datasets on which a method attains the best performance among all contenders~\citep{DelgadoCBA14}. Although originally proposed for classification, we extend it to regression by defining ``best'' as the lowest error (\eg, RMSE), and retain the name for consistency.
\end{itemize}
These evaluation metrics ensure both robust performance aggregation and statistically sound comparisons across diverse tabular datasets.

\begin{figure}[t]
  \centering
   \begin{minipage}{0.48\linewidth}
    \includegraphics[width=\textwidth]{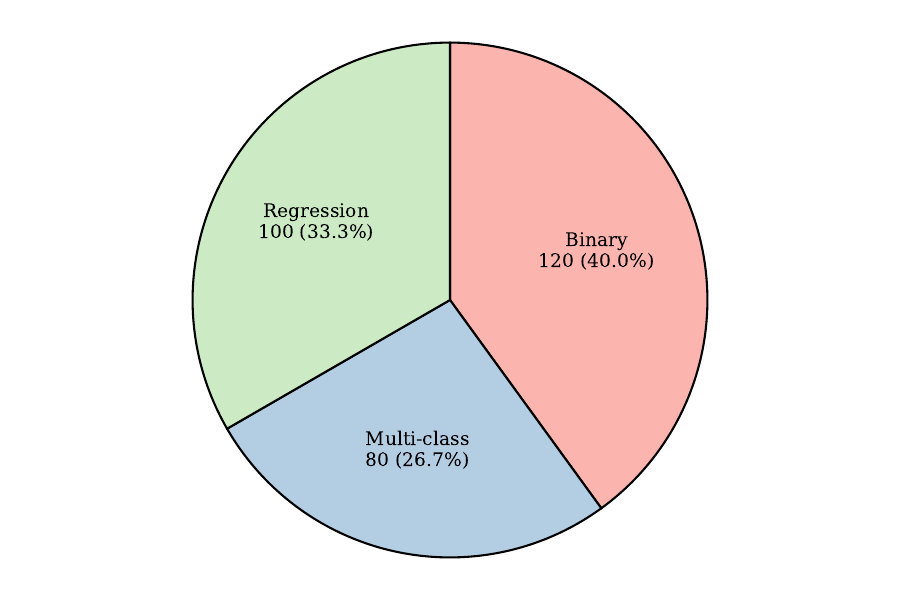}
    \centering
    {\small \mbox{(a) {Statistics over task types}}}
    \end{minipage}
    \begin{minipage}{0.48\linewidth}
    \includegraphics[width=\textwidth]{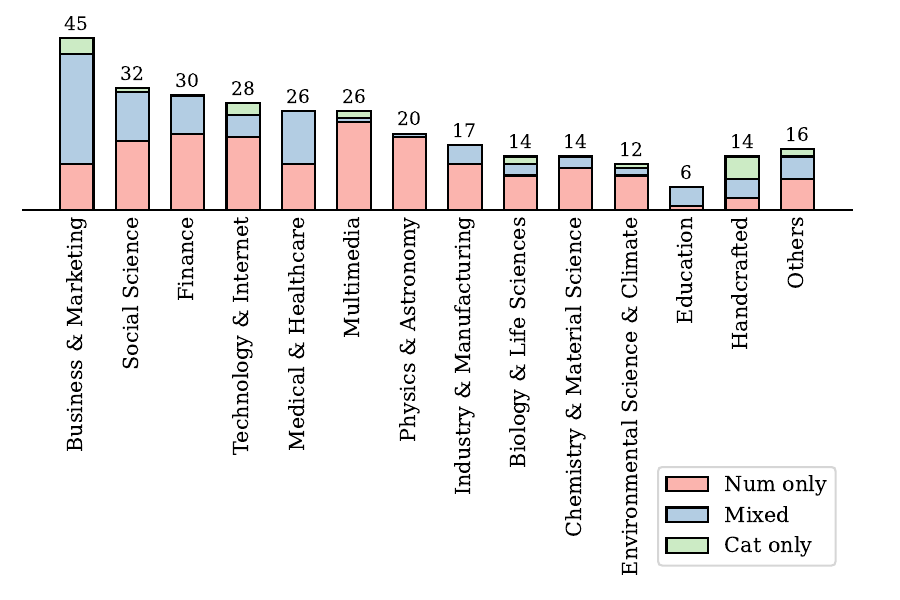}
    \centering
    {\small \mbox{(b) {Statistics over domains}}}
    \end{minipage}
    
    \begin{minipage}{0.48\linewidth}
    \includegraphics[width=\textwidth]{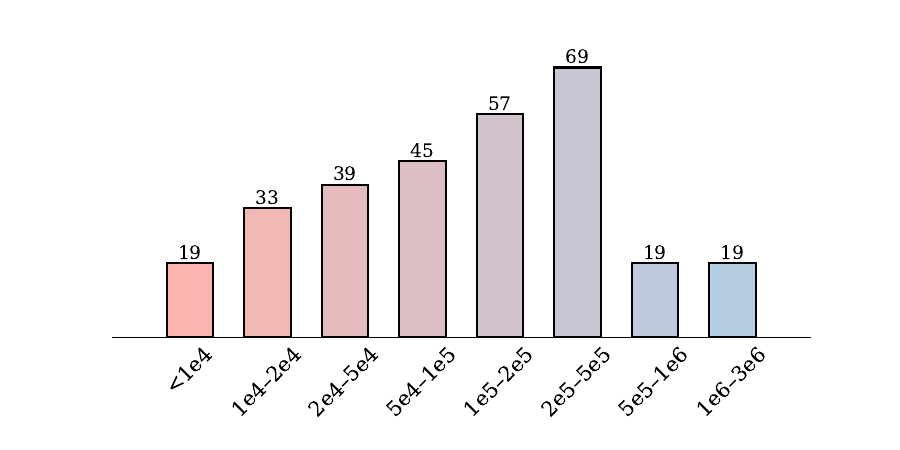}
    \centering
    {\small \mbox{(c) {Statistics over data sizes}}}
    \end{minipage}
    \begin{minipage}{0.48\linewidth}
    \includegraphics[width=\textwidth]{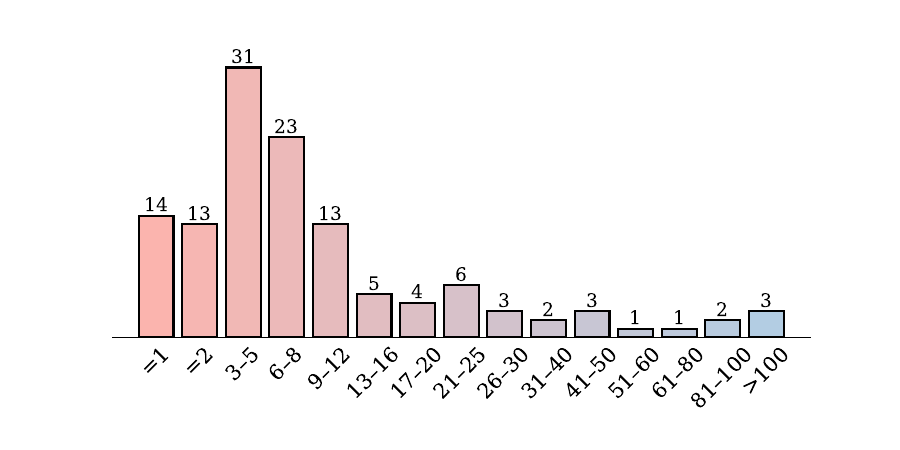}
    \centering
    {\small \mbox{(d) {Statistics of the number of categorical features}}}
    \end{minipage}
    
    \begin{minipage}{0.48\linewidth}
    \includegraphics[width=\textwidth]{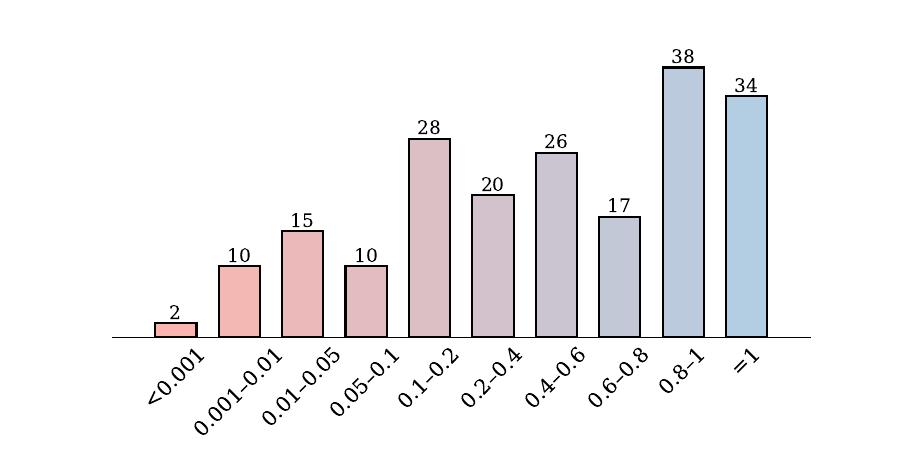}
    \centering
    {\small \mbox{(e) {Statistics over imbalance rate}}}
    \end{minipage}
    \begin{minipage}{0.48\linewidth}
    \includegraphics[width=\textwidth]{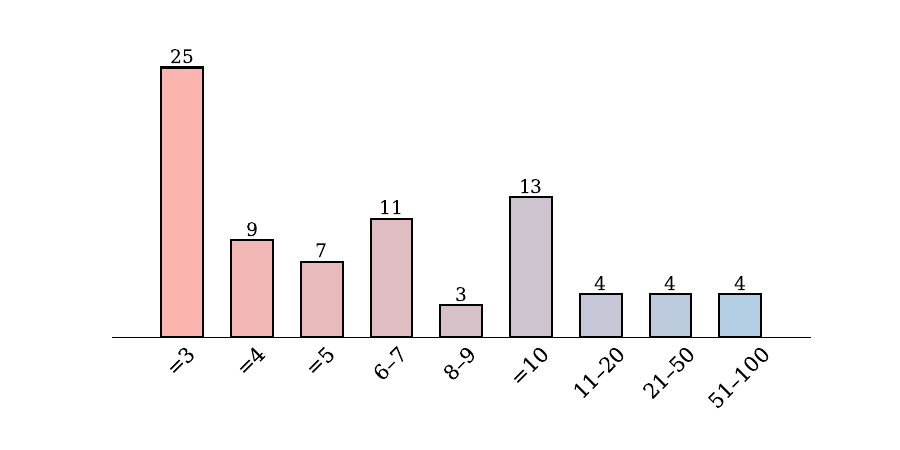}
    \centering
    {\small \mbox{(f) {Statistics of classes in multi-class tasks}}}
    \end{minipage}
  \caption{Advantages of the proposed benchmark. (a) shows the number of datasets for three tabular prediction tasks. (b) shows the histogram of datasets across various domains, as well as the types of attributes. (c) shows the number of datasets along with the change of their sizes ($N\times d$). (d) shows the histogram of the number of categorical features in datasets with categorical features. (e) shows the histogram of the imbalance rate for classification datasets. (f) shows the histogram of the number of classes for multi-class classification datasets.
  }
  \label{fig:statistics}
\end{figure}

\subsection{Advantages of Our Benchmark}
Our benchmark is designed to evaluate tabular methods under broad, realistic coverage and to surface behaviors that previous studies may miss. \autoref{fig:statistics} summarizes the key properties of {\name}; together they highlight three advantages: substantially broader task/domain/feature coverage, a more balanced size distribution, and explicit attention to real-world difficulty factors (imbalance and class cardinality).

\noindent\textbf{Coverage of tasks.}
Unlike benchmarks that focus only on classification~\citep{McElfreshKVCRGW23when}, {\name} spans \emph{all three} standard tabular settings: 120 binary, 80 multi-class, and 100 regression datasets (\autoref{fig:statistics}a). This breadth is important because many modern tabular models are intended to handle both classification and regression with a single design.

\noindent\textbf{Coverage of domains \& feature types.}
We curate datasets from 13 application areas—including \textsf{business \& marketing}, \textsf{social science}, \textsf{finance}, \textsf{technology \& internet}, \textsf{medical \& healthcare}, \textsf{multimedia}, \textsf{physics \& astronomy}, \textsf{industry \& manufacturing}, \textsf{biology \& life sciences}, \textsf{chemistry \& materials}, \textsf{environmental science \& climate}, \textsf{education}, and \textsf{handcrafted}. This diversity enables us to assess whether tabular methods can generalize across applications from varied fields.
Within each domain, we retain mixtures of feature types (numeric only, categorical only, and mixed), as shown by the stacked bars in \autoref{fig:statistics}b. This combination in our benchmark stresses models to cope with heterogeneous attributes rather than a single, homogeneous regime.

\noindent\textbf{Coverage of data sizes.}
To avoid size-driven artifacts, we target a \emph{more uniform} spread over dataset complexity, measured by $N\times d$ (instances $\times$ features). \autoref{fig:statistics}c shows that {\name} allocates substantial mass to small, medium, and moderately large problems alike, yielding fairer aggregate conclusions than skewed size distributions.

\noindent\textbf{Coverage of categorical structure.}
Because categorical attributes are central to tabular heterogeneity, we report both the number of categorical features per dataset (\autoref{fig:statistics}d) and the number of classes in multi-class tasks (\autoref{fig:statistics}f). The resulting distributions are intentionally broad—spanning datasets with very few to many categorical columns, and tasks with small to large class cardinalities. This diversity ensures that evaluations meaningfully stress models’ ability to handle categorical encodings, tokenization or embedding strategies, and class-aware objectives across a wide spectrum of practical scenarios.

\noindent\textbf{Coverage of imbalance.}
Real deployments often face skewed label distributions. \autoref{fig:statistics}e shows the imbalance-ratio histogram across our classification sets: while many datasets are near-balanced, a substantial tail is moderately to strongly imbalanced. We evaluate with vanilla training (no task-specific rebalancing), and additionally report F1/AUC to ensure fair comparison on skewed test sets.

\medskip
\noindent\textbf{Summary.} Compared with prior benchmarks, {\name} offers (i) richer task coverage (including regression), (ii) multi-domain, mixed-feature datasets within each domain, (iii) a more even spread over $N\times d$, and (iv) explicit variation in categorical count, class cardinality, and imbalance. This breadth and balance make {\name} a stronger proxy for real-world tabular challenges and a more reliable basis for comparing deep and tree-based methods.
\section{Comparison Results among Datasets}\label{sec:main_results}

We compare tabular methods across 300 datasets using multiple evaluation criteria. 
To provide a concise and clear analysis, we report only the aggregated performance metrics, such as average rank, across the entire dataset collection. 
Detailed per-dataset results are available in the online supplementary document at \url{https://github.com/LAMDA-Tabular/TALENT/tree/main/results}.

In the figures presented in this section, we use distinct colors to represent different categories of methods, ensuring clarity and ease of comparison. 
% Specifically, \textbf{Dummy} is represented by white (\raisebox{-0.3mm}{\tikz{\draw[white,fill=white] (0,0) rectangle (0.3,0.3);}}), while \textbf{classical methods} are denoted by coral orange (\raisebox{-0.3mm}{\tikz{\draw[white,fill=CoralOrange] (0,0) rectangle (0.3,0.3);}}). \textbf{Tree-based methods} are visualized using vibrant green (\raisebox{-0.3mm}{\tikz{\draw[white,fill=VibrantGreen] (0,0) rectangle (0.3,0.3);}}), and \textbf{MLP variants} are shown in rich red (\raisebox{-0.3mm}{\tikz{\draw[white,fill=RichRed] (0,0) rectangle (0.3,0.3);}}). For methods with \textbf{specially designed architectures}, we use soft indigo (\raisebox{-0.3mm}{\tikz{\draw[white,fill=SoftIndigo] (0,0) rectangle (0.3,0.3);}}), while \textbf{tree-mimic methods}  are represented by emerald teal (\raisebox{-0.3mm}{\tikz{\draw[white,fill=EmeraldTeal] (0,0) rectangle (0.3,0.3);}}). \textbf{Regularization-based methods} are depicted in fresh lime (\raisebox{-0.3mm}{\tikz{\draw[white,fill=FreshLime] (0,0) rectangle (0.3,0.3);}}), \textbf{token-based methods} in bright cyan (\raisebox{-0.3mm}{\tikz{\draw[white,fill=BrightCyan] (0,0) rectangle (0.3,0.3);}}), and \textbf{context-based methods} in vivid purple (\raisebox{-0.3mm}{\tikz{\draw[white,fill=VividPurple] (0,0) rectangle (0.3,0.3);}}).
Specifically, \textbf{Dummy} is represented by gray \raisebox{-0.3mm}{\tikz{\draw[black,fill=Gray] (0,0) rectangle (0.3,0.3);}}, 
while \textbf{classical methods} are denoted by coral orange \raisebox{-0.3mm}{\tikz{\draw[black,fill=CoralOrange] (0,0) rectangle (0.3,0.3);}}. 
\textbf{Tree-based methods} are visualized using vibrant green \raisebox{-0.3mm}{\tikz{\draw[black,fill=VibrantGreen] (0,0) rectangle (0.3,0.3);}}, 
and \textbf{MLP variants} are shown in rich red \raisebox{-0.3mm}{\tikz{\draw[black,fill=RichRed] (0,0) rectangle (0.3,0.3);}}. 
For methods with \textbf{specially designed architectures}, we use soft indigo \raisebox{-0.3mm}{\tikz{\draw[black,fill=SoftIndigo] (0,0) rectangle (0.3,0.3);}}, 
while \textbf{tree-mimic methods} are represented by emerald teal \raisebox{-0.3mm}{\tikz{\draw[black,fill=EmeraldTeal] (0,0) rectangle (0.3,0.3);}}. 
\textbf{Neighborhood-based methods} are depicted in vivid purple \raisebox{-0.3mm}{\tikz{\draw[black,fill=VividPurple] (0,0) rectangle (0.3,0.3);}},
\textbf{token-based methods} in bright cyan \raisebox{-0.3mm}{\tikz{\draw[black,fill=BrightCyan] (0,0) rectangle (0.3,0.3);}}, 
\textbf{regularization-based methods} in fresh lime \raisebox{-0.3mm}{\tikz{\draw[black,fill=FreshLime] (0,0) rectangle (0.3,0.3);}}, 
and \textbf{pretrained foundation models} in warm amber \raisebox{-0.3mm}{\tikz{\draw[black,fill=WarmAmber] (0,0) rectangle (0.3,0.3);}}.

It is notable that some methods are limited to specific types of tasks. For example, TabPFN v1 and TabICL are designed exclusively for classification tasks and cannot handle regression, while DNNR is tailored for regression and cannot be applied to classification. When showing the performance over all datasets, we only present the results of methods capable of addressing both classification and regression tasks.

\begin{figure}[t]
  \centering
   \begin{minipage}{0.45\linewidth}
    \includegraphics[width=\textwidth]{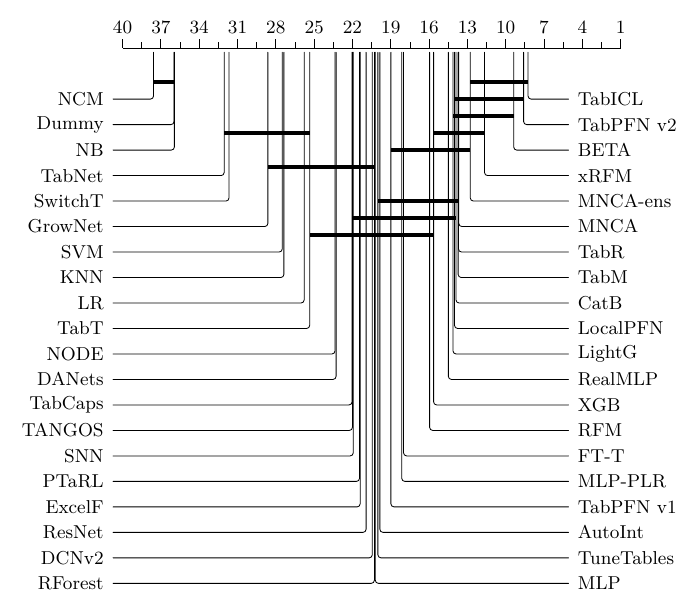}
    \centering
    {\small \mbox{(a) {Binary Classification}}}
    \end{minipage}
    \begin{minipage}{0.45\linewidth}
    \includegraphics[width=\textwidth]{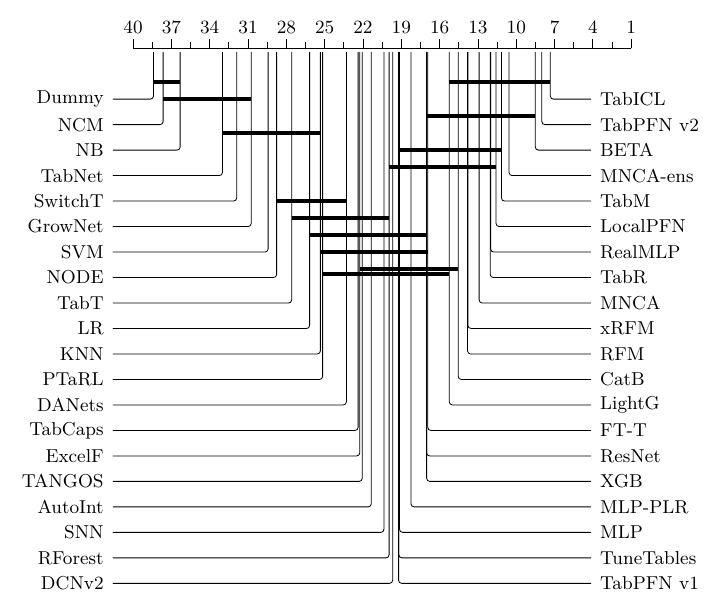}
    \centering
    {\small \mbox{(b) {Multi-Class Classification}}}
    \end{minipage}
    
    \begin{minipage}{0.45\linewidth}
    \includegraphics[width=\textwidth]{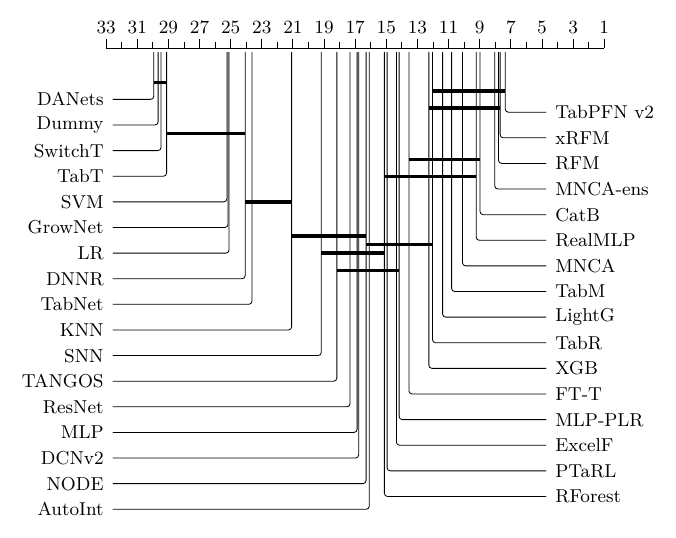}
    \centering
    {\small \mbox{(c) {Regression}}}
    \end{minipage}
    \begin{minipage}{0.45\linewidth}
    \includegraphics[width=\textwidth]{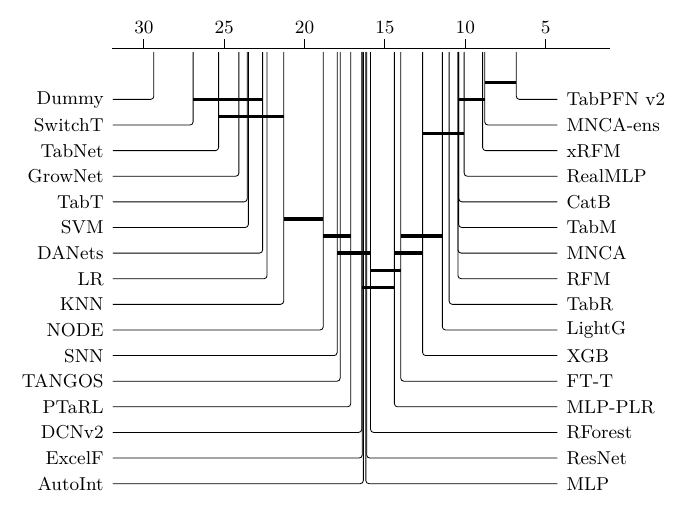}
    \centering
    {\small \mbox{(d) {All Tasks}}}
    \end{minipage}
  \caption{Critical difference of all methods via the Wilcoxon-Holm test with a significance level of 0.05. The lower the rank value, the better the performance. }
  \label{fig:critical_difference}
\end{figure}

\subsection{On Average Performance}
We compare 40 representative tabular methods across 300 datasets, reporting average performance ranks and conducting statistical significance tests with the Wilcoxon--Holm procedure at a 0.05 level~\citep{Demsar06Statistical}. The critical difference diagrams are shown in~\autoref{fig:critical_difference}, while detailed rank values and pairwise heatmaps are deferred to the appendix. \autoref{fig:teaser} further contextualizes representative methods in terms of efficiency and model size.\footnote{The average ranks in \autoref{fig:teaser} are computed on a representative subset and may differ slightly from the full-rank results.}

The most striking results come from \emph{pretrained foundation models}. Across different task types, TabPFN v2 and TabICL consistently rank among the best-performing methods. In many cases, they significantly outperform classical ensembles and tuned deep models, highlighting the benefits of pretraining and in-context learning for tabular data. While the earlier TabPFN v1 lags behind, its successors—especially TabPFN v2—extend to regression tasks and display clear generalization advantages.

Beyond foundation models, several consistent patterns emerge. Tree-based ensembles remain highly competitive: Random Forest and XGBoost provide reliable baselines, while CatBoost and LightGBM often achieve top-tier ranks, especially in regression tasks. The Wilcoxon--Holm analysis shows no significant differences among these gradient boosting methods, underscoring their maturity and robustness. Recursive Feature Machines (RFM) and its extension xRFM also achieve performance close to the strongest ensembles, occupying similar rank intervals in both binary and regression settings. The results indicate that tree-like structures may form an effective hybrid paradigm.

For deep methods, vanilla MLPs are generally weak, but tuned implementations such as MLP-PLR and RealMLP close the gap substantially. RealMLP, in particular, achieves competitive performance across tasks and is statistically stronger than many ResNet-style or regularization-based variants. Token-based approaches (\eg, FT-T, ExcelFormer, AutoInt) achieve robust results, especially in classification, but significance tests show they often cluster with ensembles in the same equivalence group, indicating that attention-based tokenization provides stability but not decisive superiority.

Tree-mimic networks such as NODE and TabNet generally underperform relative to ensembles. In contrast, neighborhood-based models such as ModernNCA achieve excellent results and are often statistically comparable to CatBoost and LightGBM, highlighting the promise of retrieval-based learning. Interestingly, \emph{ensembling within deep methods} further improves performance—TabM outperforms base MLPs, and MNCA-ens consistently surpasses ModernNCA—indicating that ensemble effects remain beneficial even for neural models.

Despite these advances, the Wilcoxon--Holm tests show that foundation models, ensembles, and top DNNs (RealMLP, ModernNCA) often remain statistically tied, suggesting that universal superiority has not yet been achieved. Scalability and computational cost also remain open challenges for foundation models. Results from BETA further indicate that fine-tuning strategies (\eg, task-specific adaptation of pretrained TabPFN) can yield improvements, suggesting a promising direction for enhancing current foundation models.

\textbf{Overall, these results yield several key observations:}
\begin{itemize}[noitemsep,topsep=0pt,leftmargin=*]
    \item Pretrained foundation models (TabPFN v2, TabICL) deliver state-of-the-art performance across many datasets and task types, substantially advancing over earlier versions. The results of foundation models significantly narrow—but not entirely close—the gap between tree-based and DNN-based paradigms.
    \item Tree-based ensembles (CatBoost, LightGBM, XGBoost) remain strong, reliable, and statistically robust baselines.  
    \item Carefully optimized DNNs, especially RealMLP and ModernNCA, can rival or surpass ensembles, showing robustness across both classification and regression tasks.  
    \item Token-based transformers (FT-T, ExcelFormer, AutoInt) provide stable and competitive results, but their advantages are not statistically decisive over ensembles.  
    \item Ensemble-style strategies (\eg, TabM, MNCA-ens) demonstrate consistent gains over their base variants, suggesting that ensembling remains an effective principle even in modern deep tabular learning.  
    \item The Wilcoxon--Holm tests highlight large equivalence groups: many methods, while different in design, are statistically indistinguishable. This indicates that progress often comes from incremental but robust improvements, rather than single universally dominant architectures, reflecting the growing maturity of the tabular learning ecosystem.
\end{itemize}

\begin{figure}[p]
  \centering
  \begin{minipage}[b]{0.3\linewidth} % bottom-aligned
    \vfill
    \includegraphics[width=\textwidth]{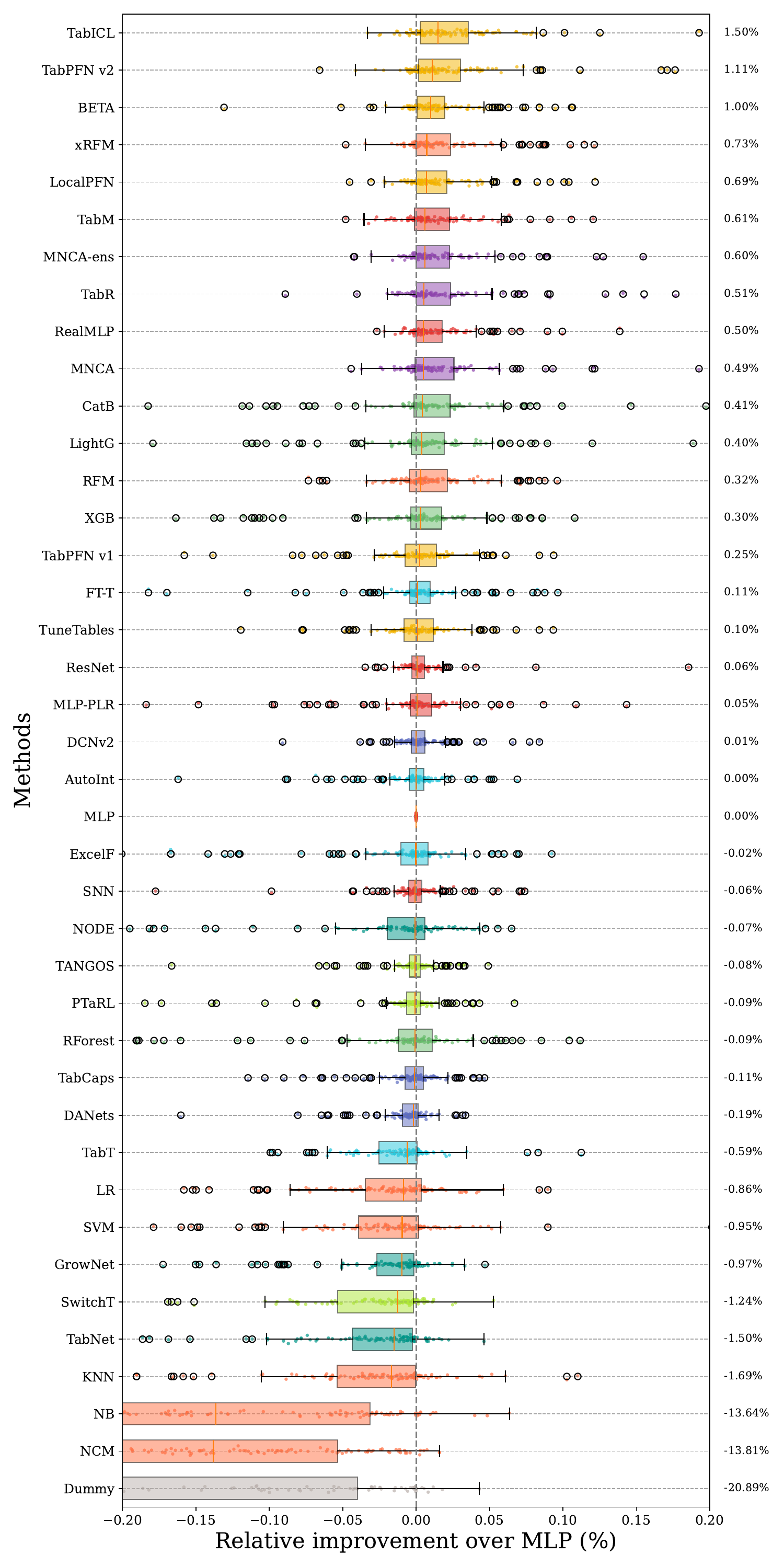}
    \centering
    {\small \mbox{(a) {Binary Classification}}}
  \end{minipage}
  \begin{minipage}[b]{0.3\linewidth} % bottom-aligned
    \vfill
    \includegraphics[width=\textwidth]{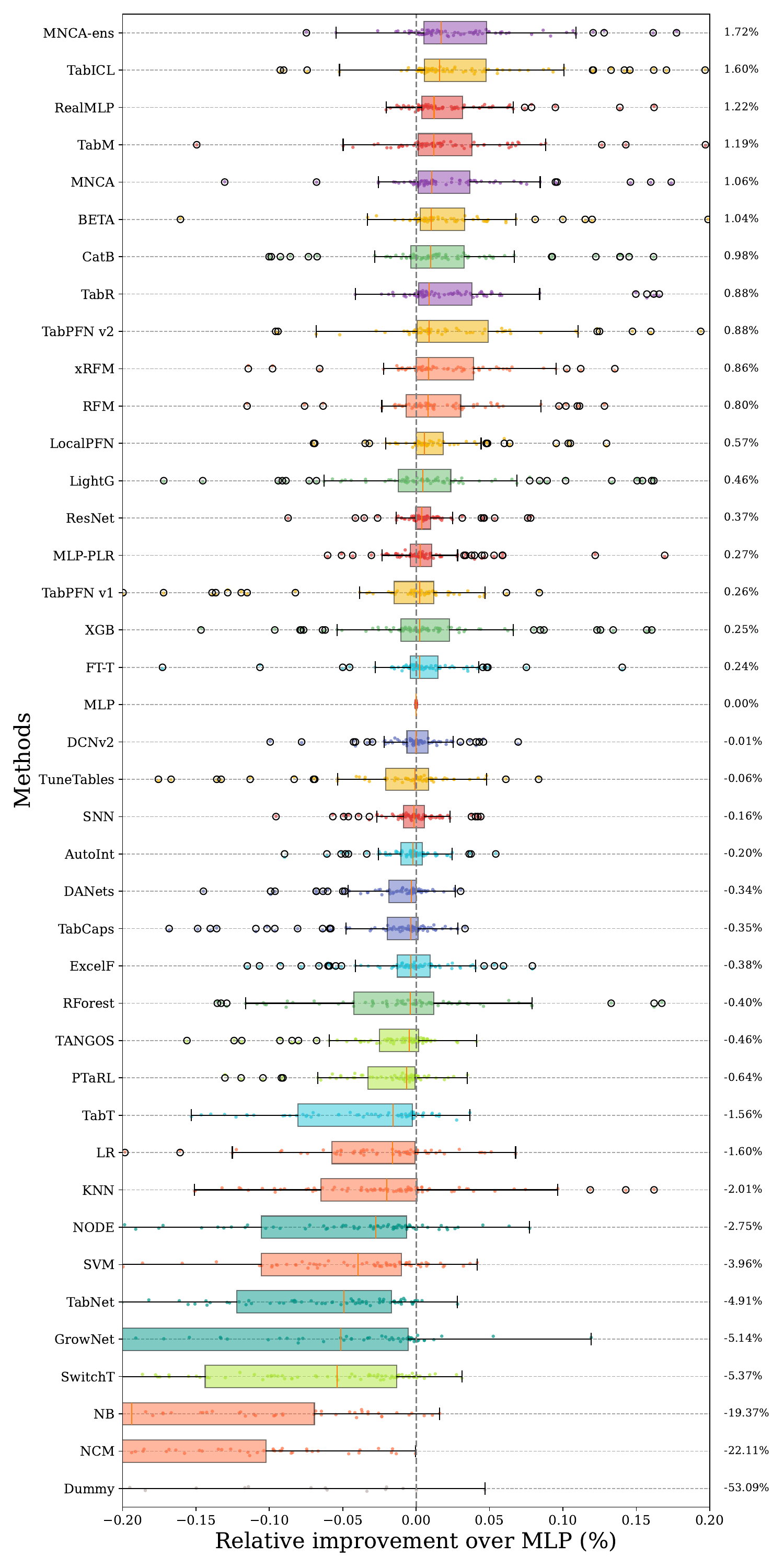}
    \centering
    {\small \mbox{(b) {Multi-Class Classification}}}
  \end{minipage}
  
  \begin{minipage}[b]{0.3\linewidth} % bottom-aligned
    \vfill
    \includegraphics[width=\textwidth]{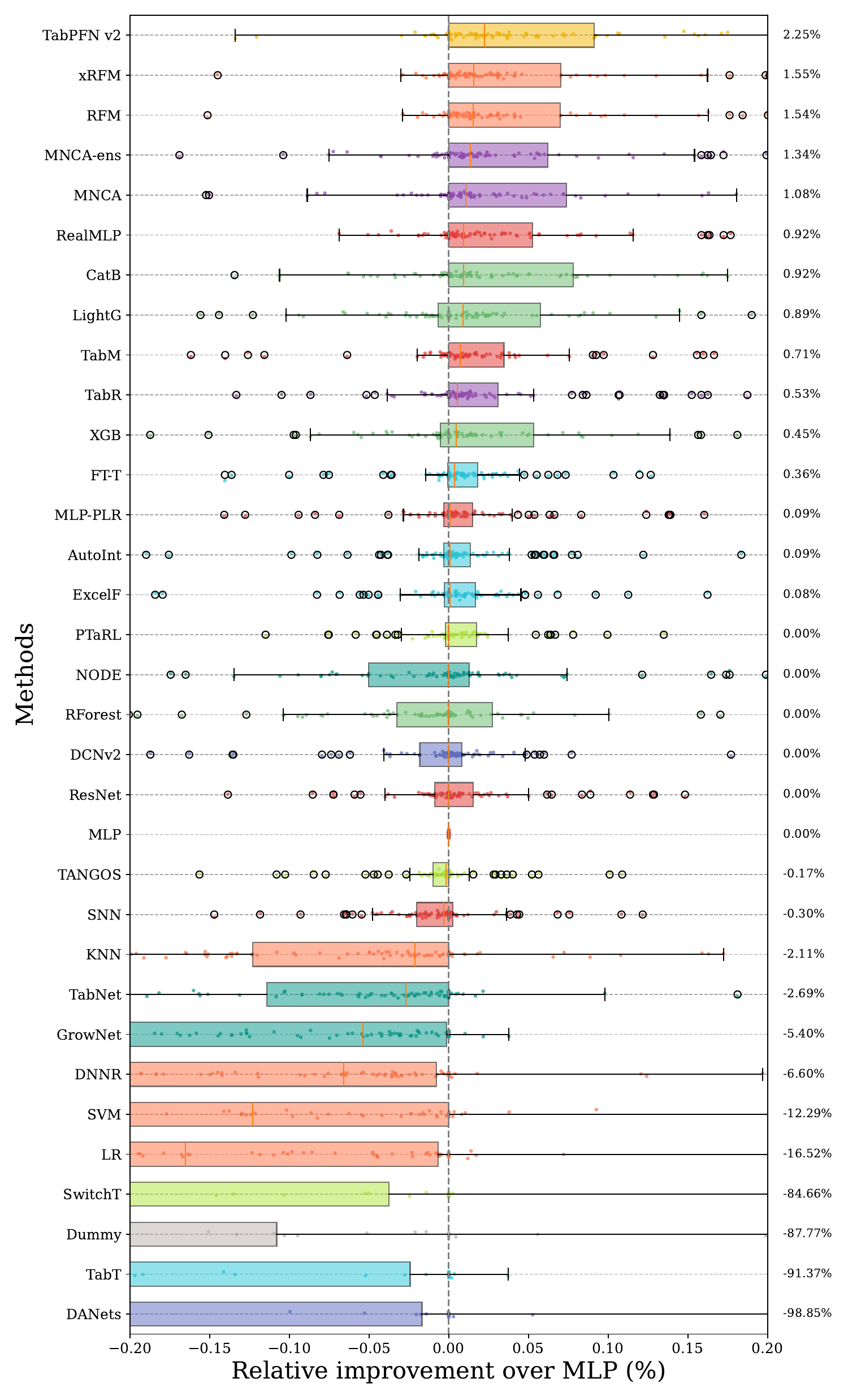}
    \centering
    {\small \mbox{(c) {Regression}}}
  \end{minipage}
  \begin{minipage}[b]{0.3\linewidth} % bottom-aligned
    \vfill
    \includegraphics[width=\textwidth]{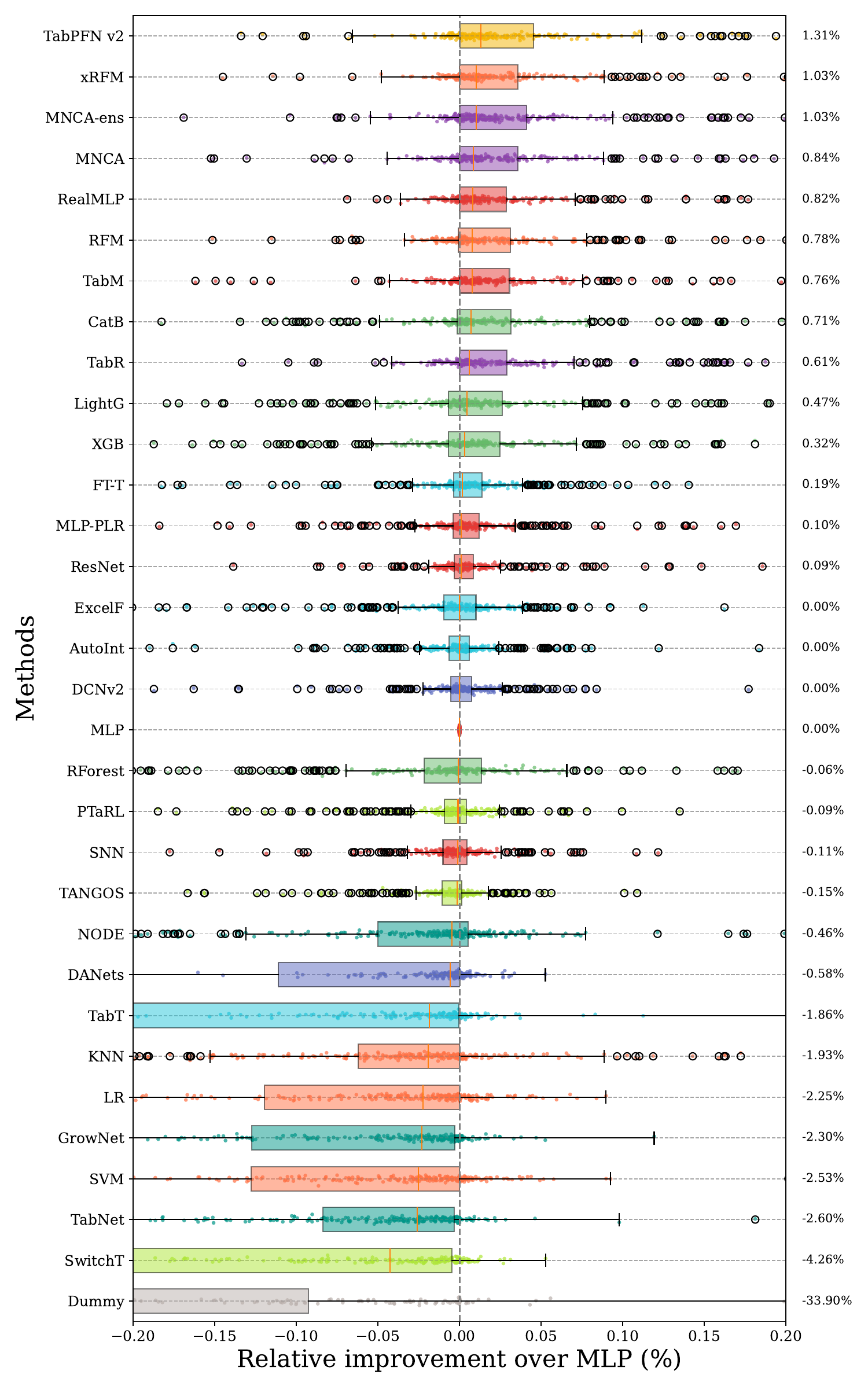}
    \centering
    {\small \mbox{(d) {All Tasks}}}
  \end{minipage}
  \caption{The Box-Plot of relative performance improvements of tabular methods over the MLP baseline across binary classification, multi-class classification, and regression tasks. The relative improvement is calculated for each dataset, where larger values indicate stronger performance relative to the MLP baseline. The box plots show the median, interquartile range (IQR), and outliers for each method. Methods with narrower IQRs demonstrate greater stability, while wider distributions suggest variability in performance. 
  }
  \label{fig:box_plot}
\end{figure}

\subsection{Relative Improvements over Tabular Baselines}
Well-tuned MLP is widely regarded as a strong baseline for tabular prediction tasks. To assess robustness, we evaluate each method by its relative improvement over MLP following~\cite{Yury2024TabM}. Formally, for a method \( m \) on dataset \( d \), the relative improvement is defined as
\[
\Delta R_{m, d} = \frac{R_{m, d} - R_{\text{MLP}, d}}{R_{\text{MLP}, d}},
\]
where \( R \) represents accuracy for classification and the min–max scaled negative RMSE for regression. Box plots in~\autoref{fig:box_plot} summarize improvements across tasks, with medians reflecting typical gains and interquartile ranges (IQRs) indicating stability.

Overall, the results confirm earlier statistical findings while highlighting additional nuances.  
Tree ensembles (CatBoost, LightGBM, XGBoost) not only achieve strong median gains over MLP but also exhibit narrow IQRs, underscoring their stability. Among deep methods, RealMLP, TabR, and ModernNCA consistently improve upon MLP, with ModernNCA showing particularly high and robust gains in regression. In contrast, models such as SwitchTab, GrowNet, and TabNet show wide distributions with negative medians in some settings, reflecting instability and lack of robustness.

Pretrained foundation models (TabPFN v2, TabICL) outperform MLP across nearly all datasets, though their relative margins are often modest, suggesting broad consistency rather than large per-dataset gains. Ensemble-enhanced approaches (\eg, MNCA-ens, TabM) also reliably surpass MLP, confirming that ensembling deep methods improves stability. Variants such as MLP-PLR provide small but systematic improvements over the vanilla MLP, validating the importance of encoding refinements.

Across tasks, binary classification shows the most consistent gains for ensembles and tuned DNNs, while multi-class classification is more challenging, with greater variance and some ensembles underperforming MLP on subsets of datasets. Regression highlights the relative strength of retrieval-based methods (ModernNCA, TabR) and pretrained models, which achieve both higher medians and broader coverage of positive gains.

In summary, outperforming a strong MLP baseline remains non-trivial. Only a subset of methods—gradient boosting ensembles, carefully tuned MLP variants, retrieval-based methods, and pretrained foundation models—achieve consistent and stable improvements, validating them as reliable baselines for future research.

\begin{figure}[t]
  \centering
   \begin{minipage}{0.45\linewidth}
    \includegraphics[width=\textwidth]{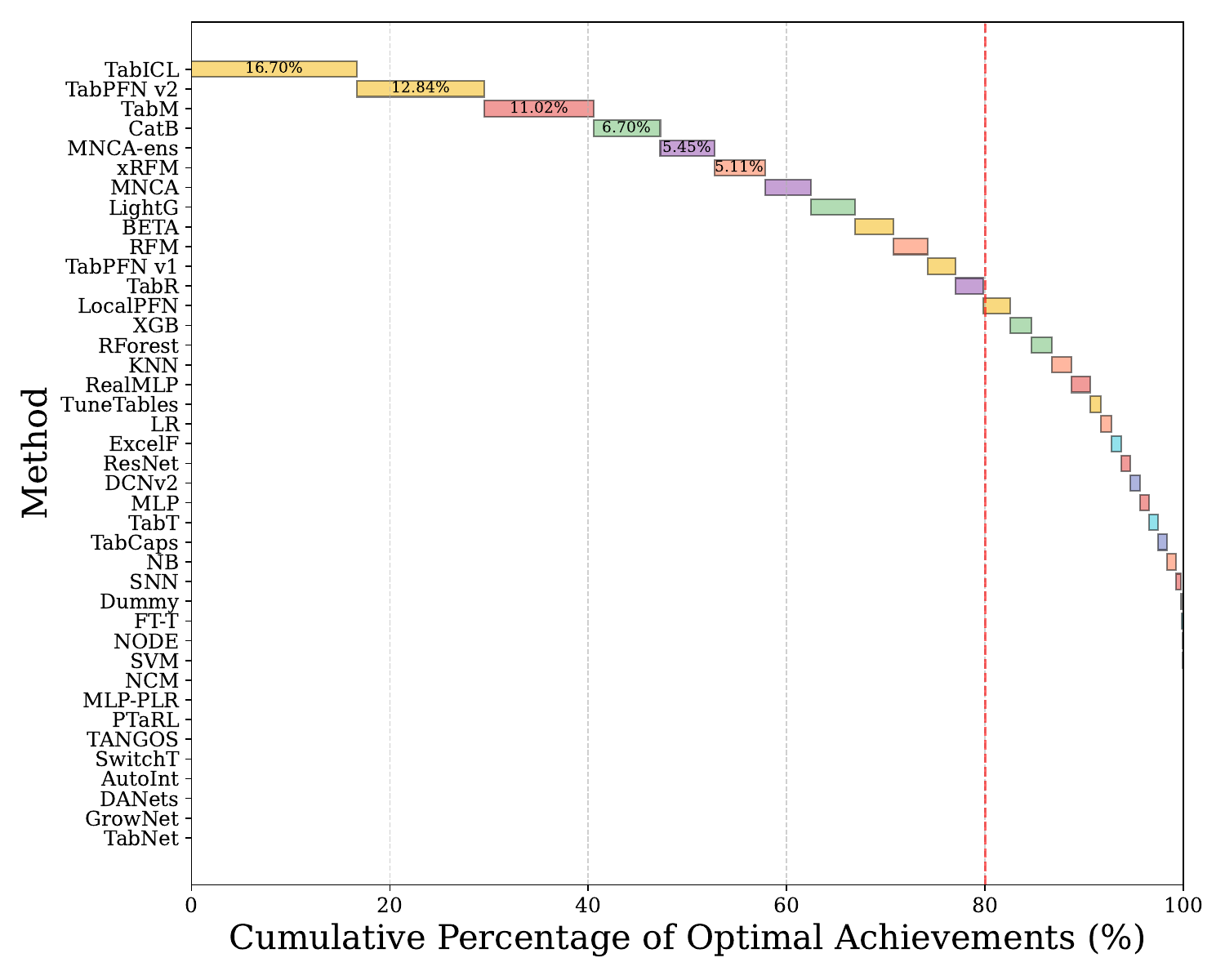}
    \centering
    {\small \mbox{(a) {Binary Classification}}}
    \end{minipage}
    \begin{minipage}{0.45\linewidth}
    \includegraphics[width=\textwidth]{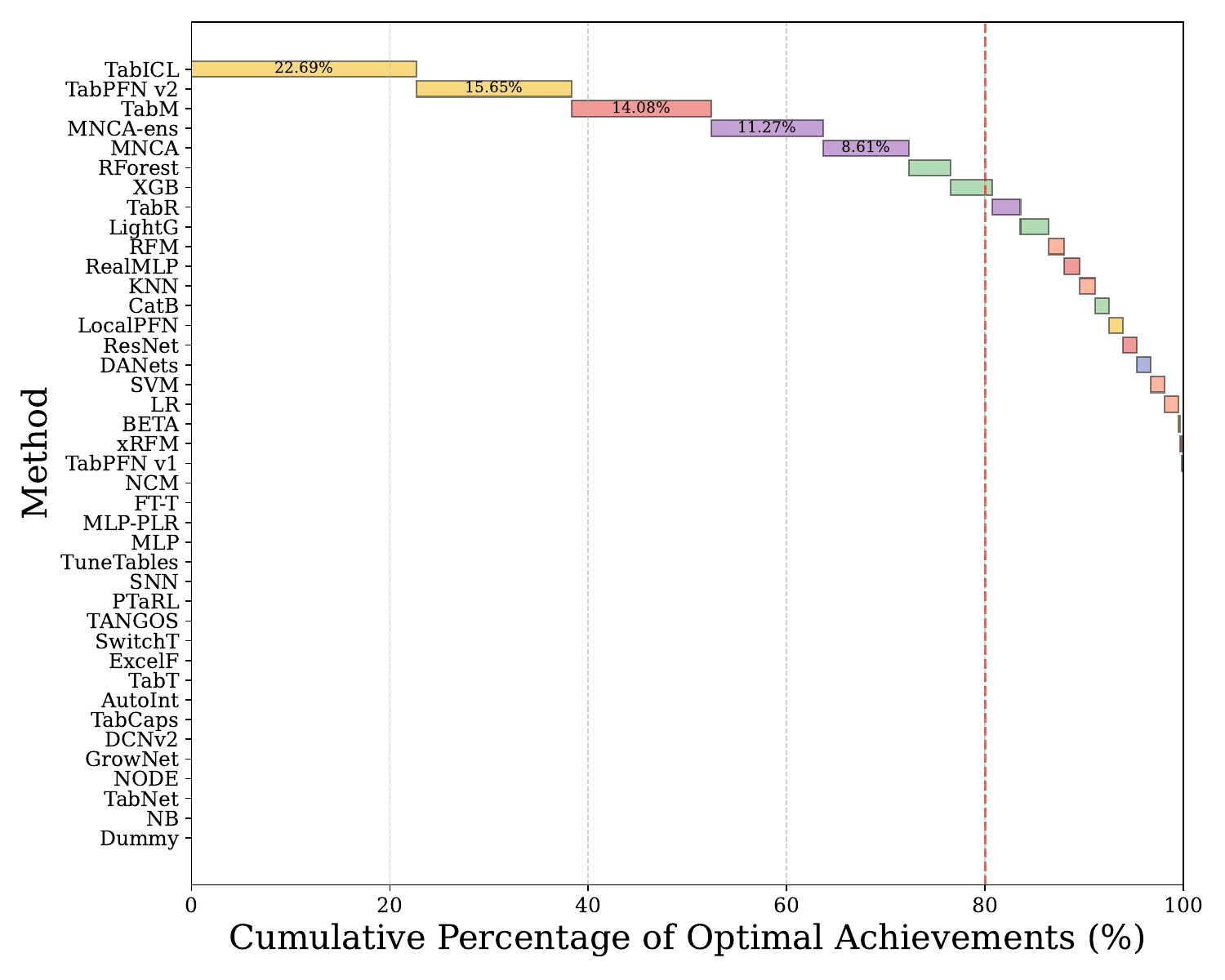}
    \centering
    {\small \mbox{(b) {Multi-Class Classification}}}
    \end{minipage}
    
    \begin{minipage}{0.45\linewidth}
    \includegraphics[width=\textwidth]{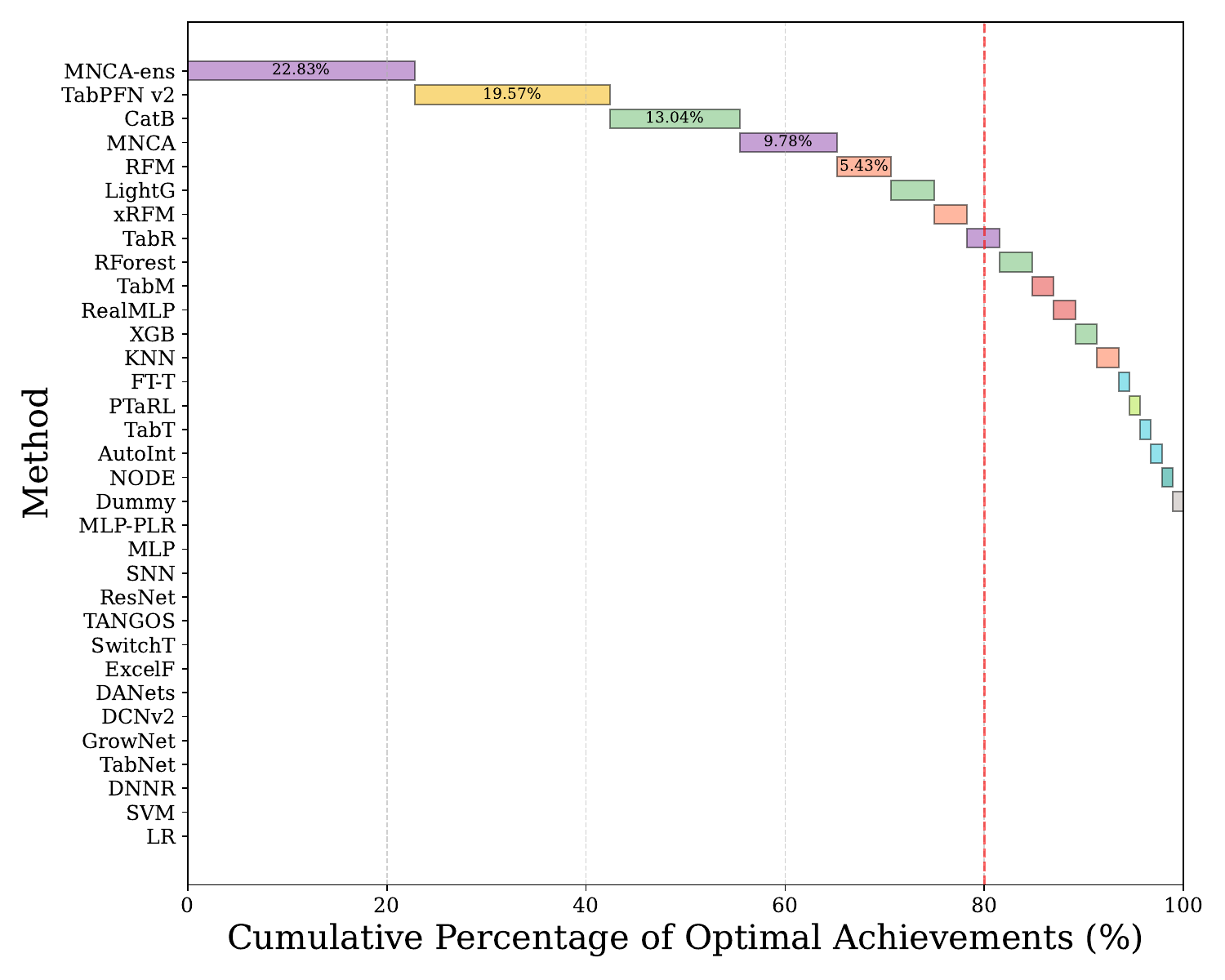}
    \centering
    {\small \mbox{(c) {Regression}}}
    \end{minipage}
    \begin{minipage}{0.45\linewidth}
    \includegraphics[width=\textwidth]{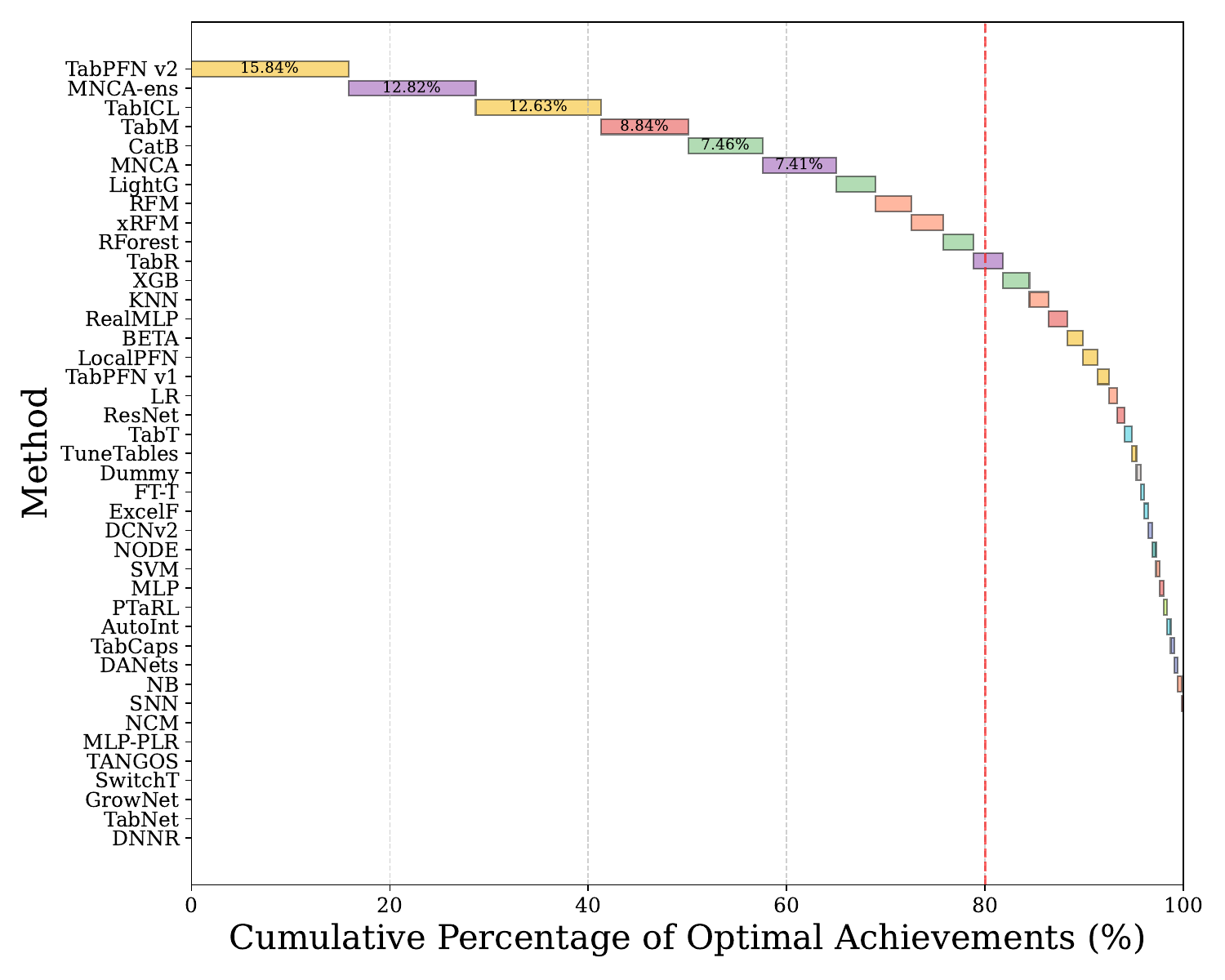}
    \centering
    {\small \mbox{(d) {All Tasks}}}
    \end{minipage}
  \caption{PAMA (Probability of Achieving the Best Accuracy) of various methods in binary classification (a), multi-class classification (b), regression (c), and all tasks (d).
  Each bar segment denotes a tabular method, whose width is the percentage that the method achieves the best performance over a kind of tabular prediction task. The wider the cell, the more often that a method performs well on the tabular prediction task.
  }
  \label{fig:pama}
\end{figure}

\subsection{Probability of Achieving the Best Accuracy}
Since the performance of tabular methods varies across datasets, average ranks and statistical tests may obscure methods that excel in specific scenarios. To complement these aggregate metrics, we evaluate the Probability of Achieving the Best Accuracy (PAMA)~\citep{DelgadoCBA14}, which measures the proportion of datasets on which a method achieves the best performance. This perspective highlights dataset-specific adaptability and identifies methods that frequently dominate.

The results in \autoref{fig:pama} reveal several key findings.  
First, pretrained foundation models show a decisive advantage. TabICL and TabPFN v2 achieve the highest PAMA scores across tasks, winning on a substantial portion of datasets (up to 22.7\% in multi-class classification). Their strong performance underscores the value of pretraining and in-context learning for diverse tabular problems. Importantly, their success is not limited to classification: TabPFN v2 ranks second overall in regression tasks, further demonstrating its generality.

Second, classical ensembles remain highly competitive. CatBoost, LightGBM, and XGBoost frequently appear among the top methods, particularly in regression, where CatBoost and LightGBM achieve some of the highest PAMA scores. Extensions such as RFM/xRFM also perform strongly, often statistically indistinguishable from the top ensembles and pretrained models. These results reinforce earlier findings that tree-based ensembles remain robust baselines, with strong adaptability across heterogeneous datasets.

Third, deep methods vary in their adaptability. ModernNCA and its ensemble variant (MNCA-ens) frequently rank near the top across all tasks, with MNCA-ens achieving the highest PAMA score (22.8\%) in regression. This confirms the strength of neighborhood-based retrieval strategies, particularly for numerical prediction. In contrast, early tree-mimic architectures such as NODE and TabNet rarely achieve top ranks, echoing the earlier statistical test results. Among MLP-based methods, RealMLP consistently achieves non-trivial PAMA scores, outperforming most other DNN variants and demonstrating that careful tuning can elevate simple architectures.

Finally, the PAMA distributions highlight the concentration of top-performing methods. Across binary, multi-class, and regression tasks, fewer than 10 methods account for over 80\% of all best-performing cases (as marked by the dashed red line). This identifies a practical ``shortlist'' of strong candidates—primarily TabICL, TabPFN v2, MNCA-ens, ModernNCA, CatBoost, LightGBM, and TabM—that dominate across most scenarios. Simpler models (\eg, Logistic Regression, KNN) occasionally achieve best results in niche datasets, but their contributions are relatively small and task-specific.

In summary, PAMA provides a complementary view to average rank and statistical tests. While many methods perform competitively on average, only a small subset consistently wins across diverse datasets. Pretrained foundation models clearly lead, followed by strong ensembles and retrieval-based methods, suggesting that future research should focus on enhancing adaptability while preserving efficiency.

\begin{figure}[t]
  \centering
  \begin{minipage}[t]{0.45\linewidth}
    \includegraphics[width=\textwidth]{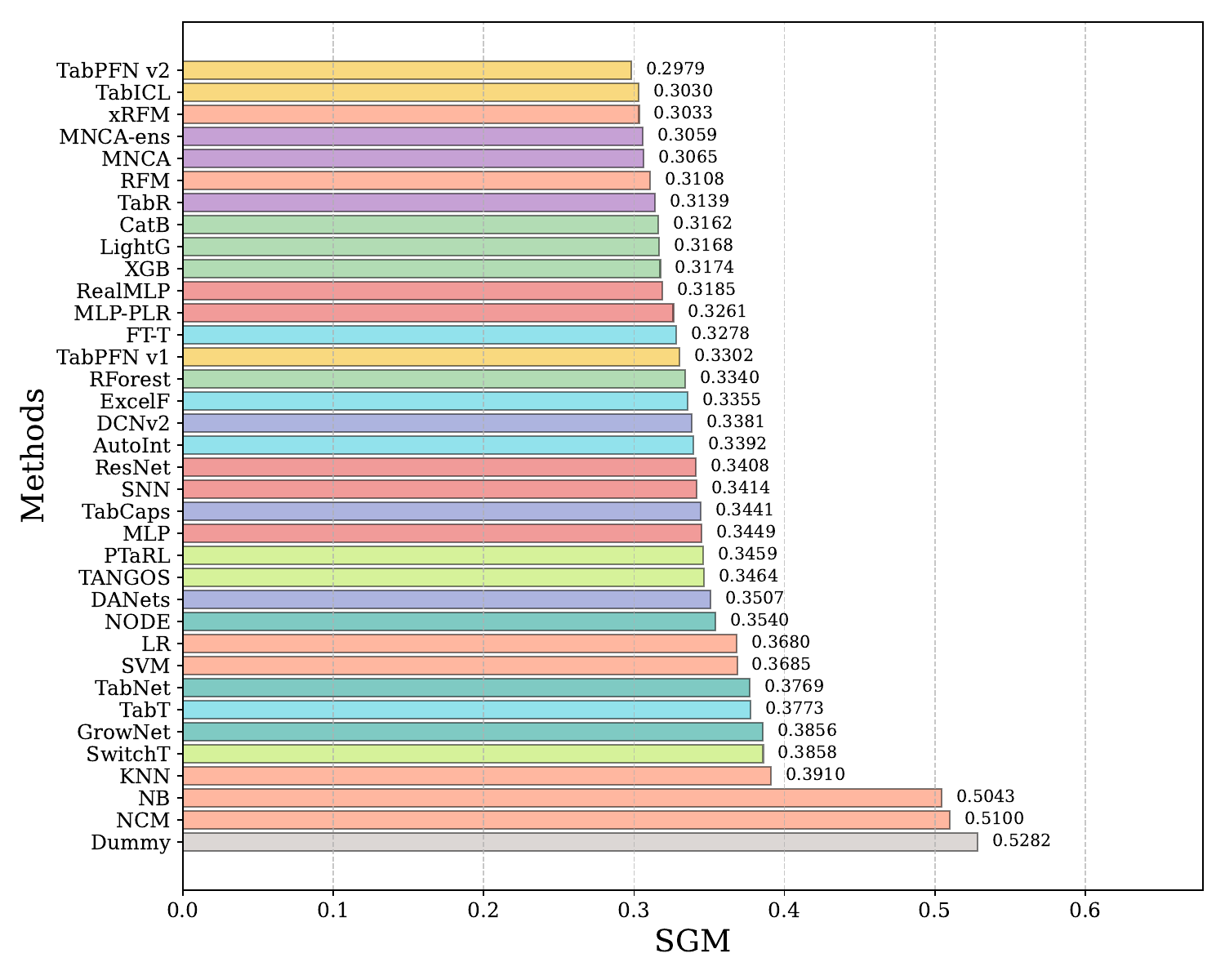}
    \centering
    {\small \mbox{(a) {Binary Classification}}}
    \includegraphics[width=\textwidth]{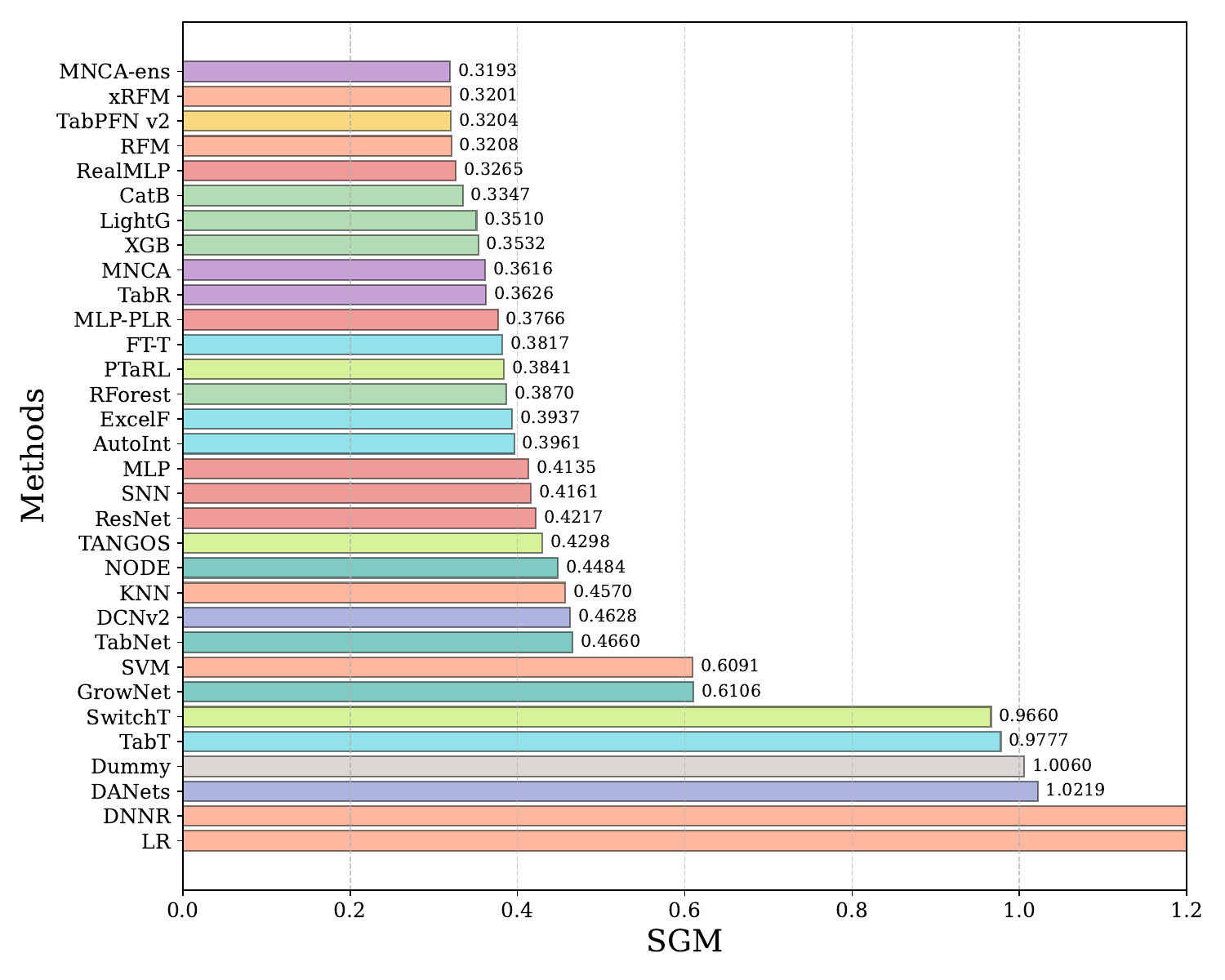}
    \centering
    {\small \mbox{(c) {Regression}}}
  \end{minipage}
  \begin{minipage}[t]{0.45\linewidth}
    \includegraphics[width=\textwidth]{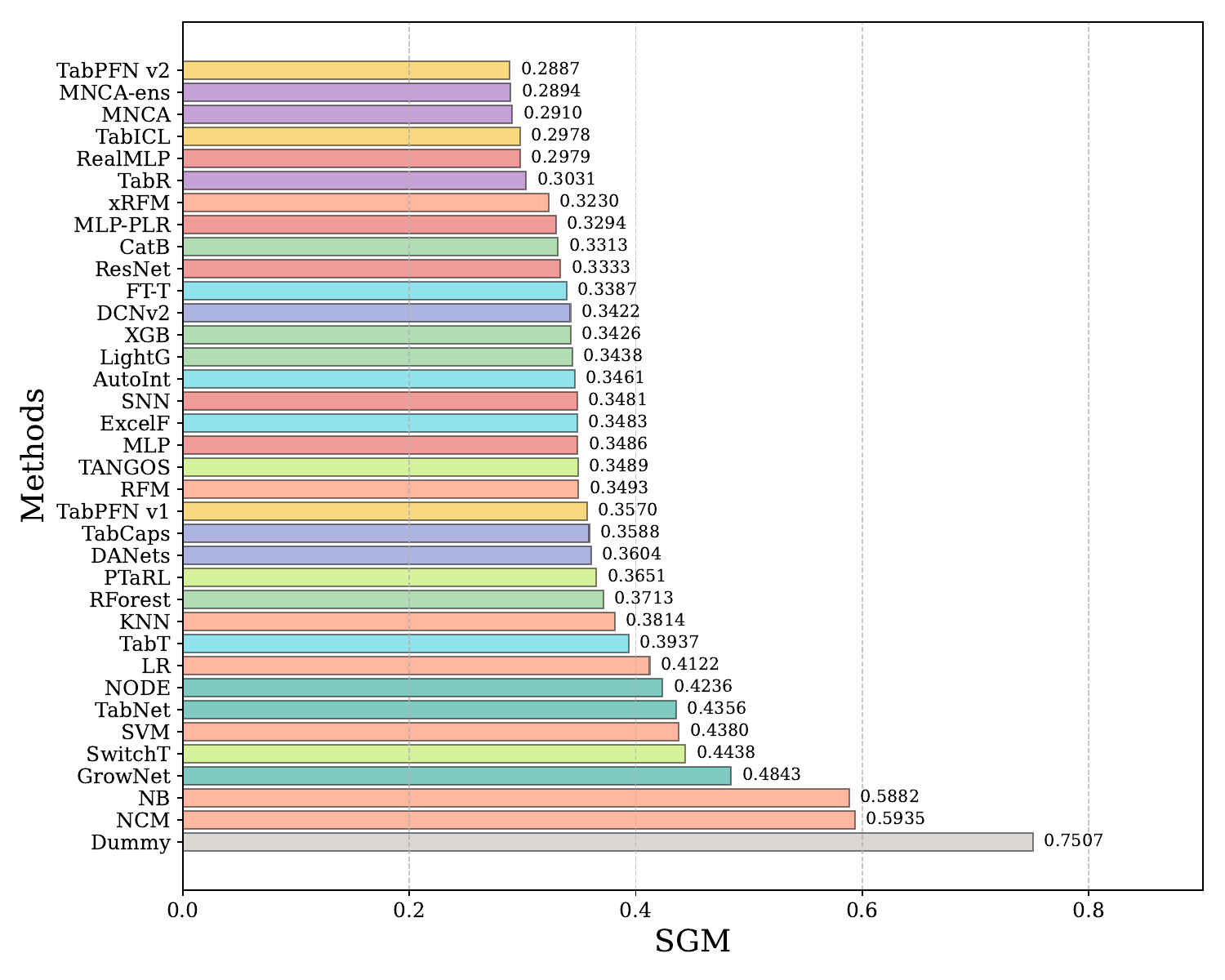}
    \centering
    {\small \mbox{(b) {Multi-Class Classification}}}
  \end{minipage}
  \caption{Aggregated performance across datasets using Shifted Geometric Mean Error (SGM). Per-dataset metrics are classification error (1–accuracy) for classification tasks and normalized RMSE (nRMSE) for regression tasks. Lower values indicate better performance and higher robustness across datasets and random seeds.
  }
  \label{fig:sgm}
\end{figure}

\subsection{Averaged Performance}
To evaluate robustness across datasets and random seeds, we report aggregated metrics: Shifted Geometric Mean Error (SGM) for classification and normalized RMSE (nRMSE) for regression, following~\cite{David2024RealMLP}. The results are shown in \autoref{fig:sgm}.

In binary classification, foundation models (\eg, TabPFN v2, TabICL) achieve the lowest SGM values, closely followed by context-based methods (MNCA-ens, TabR) and strong ensembles (CatBoost, LightGBM). RealMLP and MLP-PLR also perform competitively, further confirming that enhanced MLPs can close much of the historical tree--DNN gap. These results echo earlier average-rank analyses but emphasize that foundation and neighborhood-based models deliver more stable, low-error behavior across seeds.

For multi-class classification, the ranking remains similar, with TabPFN v2 and MNCA-ens again achieving the best SGM, while RealMLP and TabR remain close. Ensembles such as CatBoost and LightGBM still perform strongly but no longer dominate, highlighting the advantage of retrieval-based and pretrained approaches in more complex label structures. Classical methods like LR, KNN, and NB perform poorly under SGM, reinforcing their lack of robustness in high-class scenarios.

Regression shows a slightly different pattern: ensemble-enhanced models (MNCA-ens, xRFM, CatBoost, LightGBM) and RealMLP achieve the lowest nRMSE. TabPFN v2 also ranks among the top performers, suggesting that pretraining contributes to consistent generalization even in continuous targets. In contrast, models such as TabNet, GrowNet, and SwitchTab perform poorly, with high variability across datasets. Linear models (LR) and simple baselines are the weakest, consistent with earlier analyses.

Overall, SGM and nRMSE highlight three key insights. First, pretrained foundation models (TabPFN v2, TabICL) and neighborhood-based ensembles (MNCA-ens, TabR) provide the most stable performance across tasks. Second, gradient boosting ensembles remain reliable, particularly in regression. Third, carefully tuned MLPs (RealMLP, MLP-PLR) consistently rank among the top tier, showing that architectural refinements plus optimization strategies can rival classical ensembles. These findings reinforce conclusions from average-rank and PAMA analyses while underscoring the added stability of foundation and context-based methods under seed-sensitive metrics.

\begin{figure}[t]
  \centering
    \begin{minipage}{0.6\linewidth}
    \includegraphics[width=\textwidth]{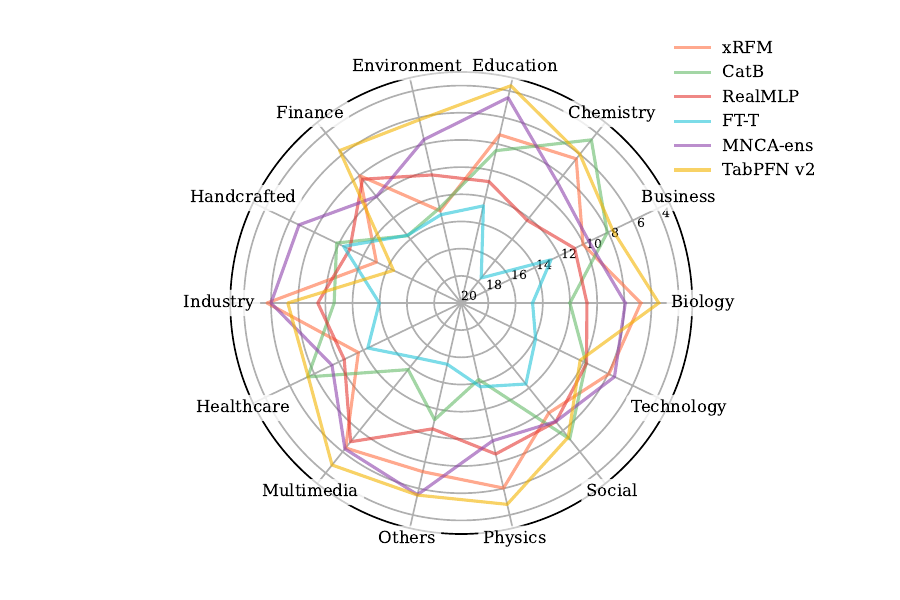}
    \centering
    % {\small \mbox{Comparison across domains}}
    \end{minipage}
  \caption{Comparison of representative tabular prediction methods across 14 application domains. The radar plot shows reversed rank scores, where larger values indicate better average performance in a domain. 
  }
  \label{fig:domain_comparison}
\end{figure}

\begin{figure}[t]
  \centering
    \begin{minipage}{0.45\linewidth}
    \includegraphics[width=\textwidth]{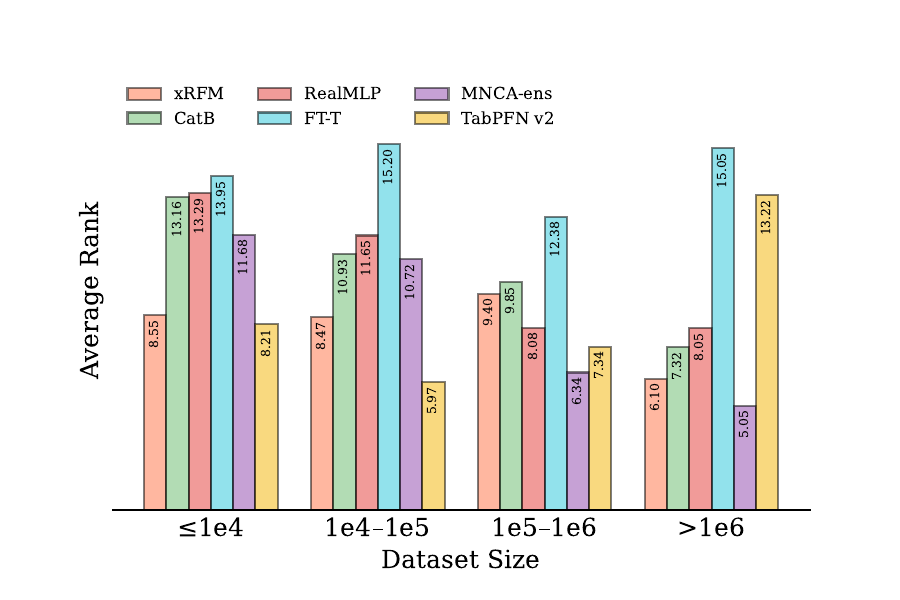}
    \centering
    {\small \mbox{(a) Overall size: $N\times d$}}
    \end{minipage}
   \begin{minipage}{0.45\linewidth}
    \includegraphics[width=\textwidth]{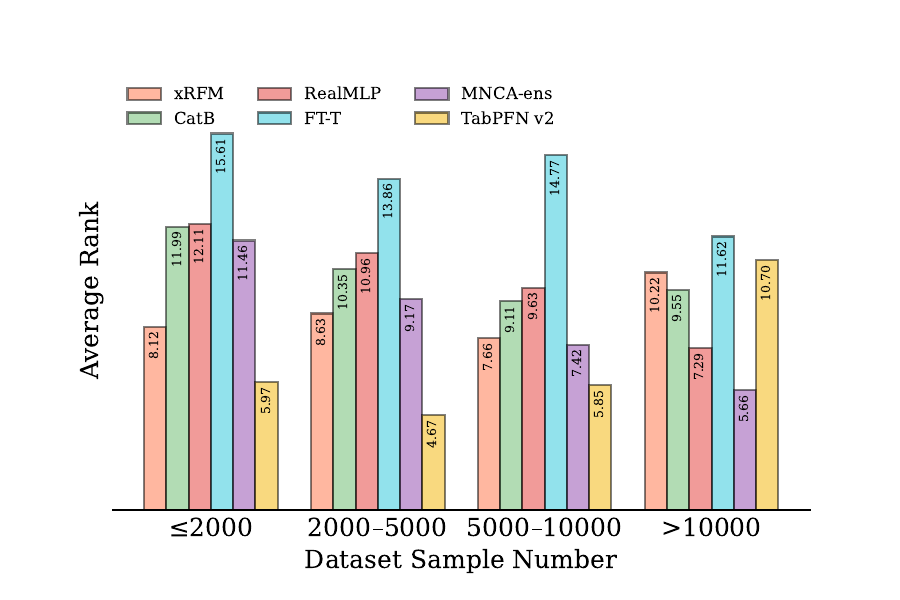}
    \centering
    {\small \mbox{(b) Samples only: $N$}}
    \end{minipage}
    
    \begin{minipage}{0.45\linewidth}
    \includegraphics[width=\textwidth]{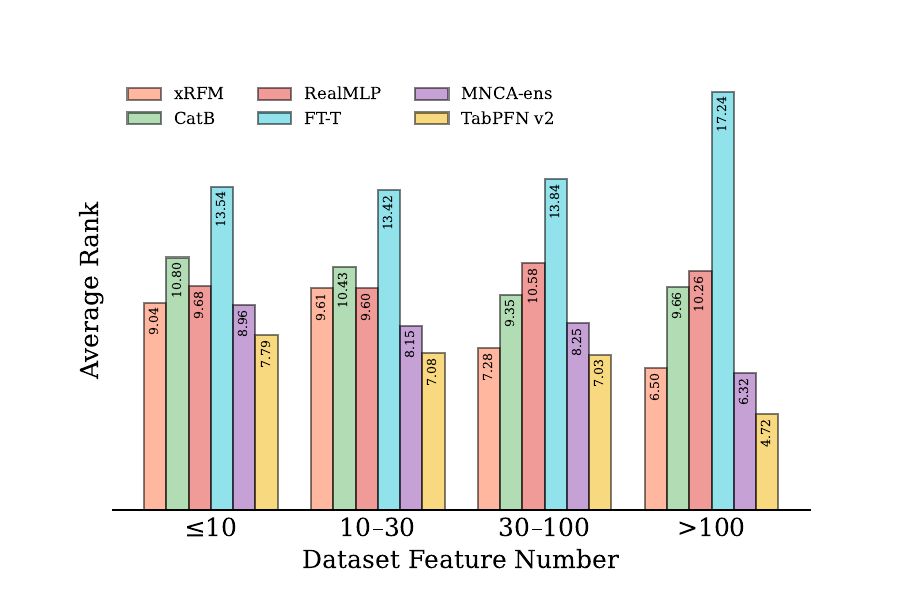}
    \centering
    {\small \mbox{(c) Features only: $d$}}
    \end{minipage}
    \begin{minipage}{0.45\linewidth}
    \includegraphics[width=\textwidth]{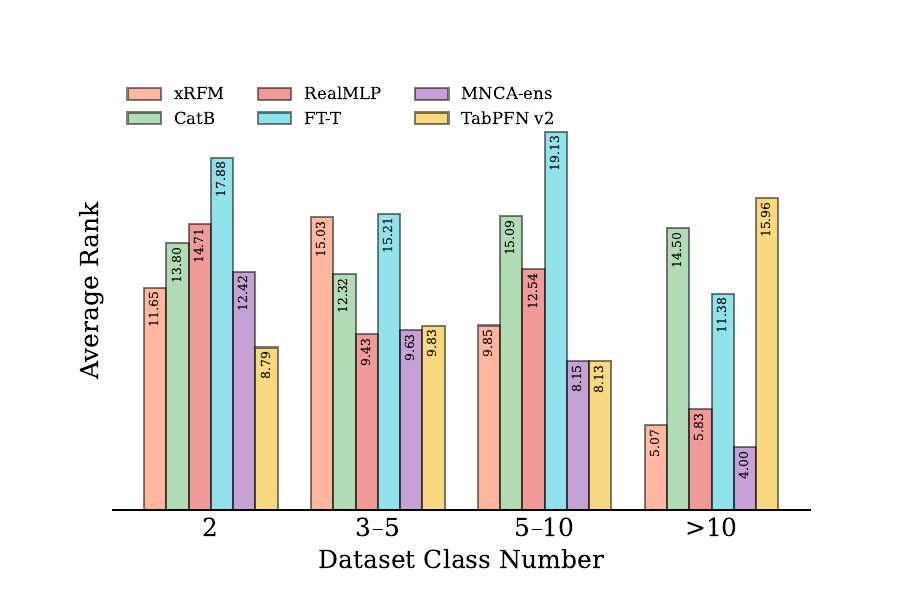}
    \centering
    {\small \mbox{(d) Number of classes: $C$}}
    \end{minipage}
  \caption{Average ranks of representative tabular methods as dataset characteristics vary. 
  (a) summarizes trends with joint scale $N\times d$; (b) and (c) isolate the marginal effects of sample size $N$ and feature dimensionality $d$; (d) varies the number of classes $C$, reflecting the label granularity. 
  }
  \label{fig:size_comparison}
\end{figure}
\subsection{Results across Domains and Dataset Sizes}
To better understand the behavior of tabular methods, we analyze their performance across 14 application domains and four dataset size groups. 
We select six representative models from diverse categories: the tree-based model CatBoost, the token-based model FT-T, the neighborhood-based model MNCA-ens, the pretrained foundation model TabPFN v2, the enhanced MLP variant RealMLP, and the recursive feature model xRFM. \autoref{fig:domain_comparison} presents the reversed average ranks across domains using a spider chart (larger values indicate better performance). \autoref{fig:size_comparison} presents the average ranks across dataset sizes (and other statistics) with bar charts (lower values indicate better performance).

\noindent\textbf{Domain-level analysis shows clear specialization among methods}. Pretrained and neighborhood-based approaches stand out with broad adaptability. TabPFN v2 performs consistently well across many domains, particularly excelling in \texttt{education}, \texttt{multimedia}, and \texttt{social sciences}, reflecting the benefits of pretraining for generalization on heterogeneous tasks. MNCA-ens also achieves strong and steady performance, ranking among the best in \texttt{handcrafted}, \texttt{environmental}, and \texttt{healthcare} datasets, which highlights the robustness of retrieval-based ensemble strategies.
Tree-based ensembles continue to serve as highly competitive baselines. CatBoost is especially effective in \texttt{chemistry} and \texttt{finance}, where categorical structures and feature interactions dominate, while xRFM shows stable results across a wide spectrum of domains, often close to the strongest pretrained and context-based models.
Other deep neural methods demonstrate complementary strengths. RealMLP achieves notable gains in \texttt{finance}, \texttt{physics}, and \texttt{industry}, confirming the effectiveness of enhanced MLP designs in structured domains. FT-T delivers solid adaptability in \texttt{technology} and \texttt{social} datasets, benefiting from tokenization and attention mechanisms, although its advantage over ensembles is less pronounced.

The cross-domain analysis also underscores the variability of model behavior. For instance, in \texttt{biology} and \texttt{healthcare}, ensemble methods maintain strong performance, while in \texttt{multimedia} and \texttt{social} science tasks, foundation and retrieval-based models dominate. This variation suggests that domain alignment is a critical factor in achieving optimal results, and no single approach universally leads across all areas.

\noindent\textbf{Dataset-size and composition analysis reveals fine-grained scalability patterns.}  
As shown in~\autoref{fig:size_comparison}, model performance varies notably when examined along four complementary dimensions—overall dataset size ($N\times d$), number of samples ($N$), feature dimensionality ($d$), and number of classes ($C$).  
\textit{All rank values are computed over the full set of evaluated methods in this study, while the figure visualizes only representative models for clarity.}

Across overall dataset scales (\autoref{fig:size_comparison}a), CatBoost continues to exhibit strong scalability, improving steadily as the dataset size increases. In contrast, RealMLP performs best on small-to-medium datasets but declines slightly as scale grows, highlighting optimization and regularization challenges common to MLP-style models. TabPFN~v2 ranks near the top on medium-to-large datasets, demonstrating that pretraining confers robust generalization in typical-size regimes, though its effectiveness tapers off when data size becomes very large—an observation consistent with its pretraining limits and context-size constraints. MNCA-ens remains consistently strong, benefiting from ensembling over neighborhood-based embeddings, while xRFM shows competitive performance on small-to-medium scales but struggles a bit when dataset size increases.

When isolating the effect of sample count (\autoref{fig:size_comparison}b)), MNCA-ens, CatBoost, and RealMLP all gain from larger $N$, confirming their strong scalability under data abundance. TabPFN~v2, however, shows its best results around $5\text{k}$–$10\text{k}$ samples before flattening out, suggesting that its pretrained inference window constrains further improvements without architectural extension. FT-Transformer remains stable but shows limited scalability advantage. Conversely, xRFM again performs well in the small-data regime ($N\!\leq\!2000$), consistent with its design as a lightweight, backpropagation-free architecture that benefits from smaller sample sizes.

For feature dimensionality (\autoref{fig:size_comparison}c)), both CatBoost and RealMLP retain strong rankings as $d$ grows, showcasing their robustness to redundant or weakly informative features. Interestingly, TabPFN~v2 maintains good performance even when $d>100$, suggesting that its pretraining includes sufficient diversity to generalize beyond low-dimensional regimes. In contrast, FT-Transformer and xRFM exhibit noticeable degradation as $d$ increases, possibly due to insufficient regularization and the growing difficulty of effective feature selection under very high dimensionality.  
We will later show in~\autoref{sec:talent_extension} that such high-dimensional regimes further amplify these differences, where foundation models in particular face performance degradation in extremely wide feature spaces.

When comparing across class cardinalities (\autoref{fig:size_comparison}d)), ensemble-style deep models such as MNCA-ens and RealMLP improve significantly as class numbers increase, demonstrating their flexibility in capturing complex, fine-grained label boundaries. CatBoost remains robust across all $C$, reinforcing its role as a dependable, well-regularized baseline. FT-Transformer shows declining performance beyond $C>10$, possibly due to its token-based design being less effective for large label vocabularies. xRFM maintains moderate performance in low-class scenarios but exhibits limited gains for higher class counts.

\noindent\textbf{Summary across domains and scales.}  
Taken together, the domain- and size-wise analyses reveal that model behavior is shaped jointly by the \emph{structural characteristics} and the \emph{scale} of tabular data. These findings collectively suggest that fine-grained analyses along domain and data-size axes offer more informative insights than aggregate rankings alone. They reveal how different modeling principles respond to variations in feature structure, label granularity, and data scale. Such observations motivate a more nuanced understanding of how classical and deep paradigms complement each other in tabular prediction, laying the groundwork for our subsequent discussion.

\subsection{Revisiting the Tree-DNN Debate}\label{sec:DNN_TREE_debate}
A longstanding question in tabular learning is whether tree-based ensembles or deep neural networks (DNNs) are inherently stronger. Earlier benchmarks generally favored ensembles such as Random Forest, XGBoost, and CatBoost, while deep models struggled to consistently outperform them. This ``tree--DNN divide'' motivated much of the early work on specialized architectures for tabular data~\citep{Grinsztajn2022Why,McElfreshKVCRGW23when}.  

\begin{figure}[t]
  \centering
   \begin{minipage}{0.45\linewidth}
    \includegraphics[width=\textwidth]{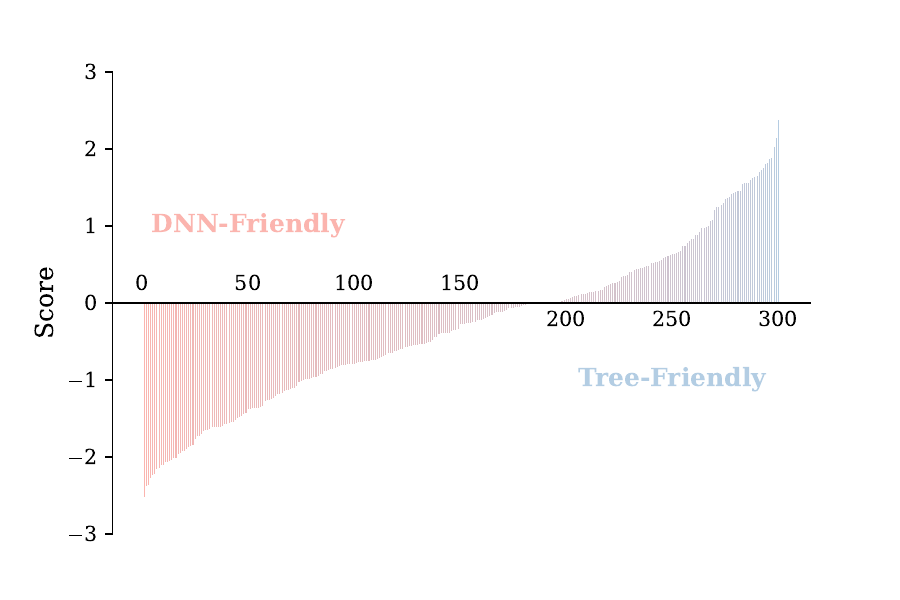}
    \centering
    {\small \mbox{(a) Tree-DNN score w/o TabPFN v2}}
    \end{minipage}
    \begin{minipage}{0.45\linewidth}
    \includegraphics[width=\textwidth]{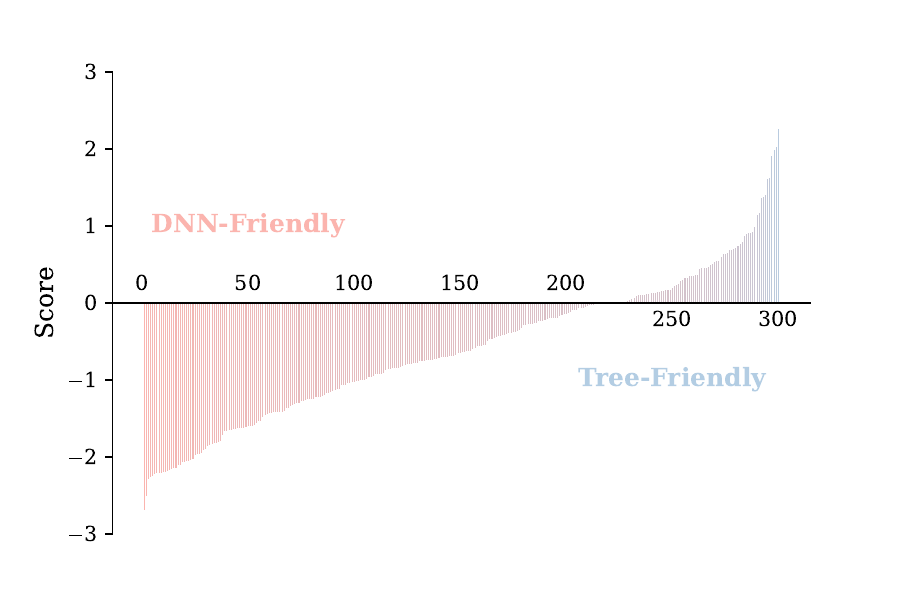}
    \centering
    {\small \mbox{(b) Tree-DNN score w/ TabPFN v2}}
    \end{minipage}
  \caption{Distribution of Tree--DNN scores across 300 datasets. The score is defined as the difference between the best representative tree-based method and the best representative DNN-based method. (a) excludes pretrained foundation models, while (b) includes TabPFN v2 among the DNN-based group. Positive values indicate tree-friendly datasets, negative values indicate DNN-friendly datasets.
  }
  \label{fig:addition_results}
\end{figure}

To quantify this divide, we adopt the Tree--DNN score (\autoref{eq:tree_dnn_score}), defined as the difference between the best-performing tree-based model and the best-performing DNN-based model after normalization. Higher values indicate datasets where ensembles dominate (tree-friendly), while lower values indicate datasets where DNNs dominate (DNN-friendly).  
\begin{equation}
s = \max(\hat{s}_{\rm XGBoost}, \hat{s}_{\rm CatBoost}, \hat{s}_{\rm RForest}, \hat{s}_{\rm LightG}) - \max(\hat{s}_{\rm RealMLP}, \hat{s}_{\rm FT\text{-}T}, \hat{s}_{\rm MNCA}, \hat{s}_{\rm TabM})\;.\label{eq:tree_dnn_score}
\end{equation}
$\hat{s}$ represents the normalized metric, such as accuracy for classification tasks or negative RMSE for regression tasks.

\autoref{fig:addition_results}(a) shows the sorted score distribution without including pretrained models (TabPFN v2). A large portion of datasets remains tree-friendly, confirming that ensembles retain a structural advantage on many tasks. However, a comparable fraction of datasets favors DNNs, reflecting the progress of modern deep tabular methods such as RealMLP and ModernNCA.  

The picture changes once pretrained tabular foundation models are included (\autoref{fig:addition_results}(b)). With TabPFN v2 added to the DNN group, the balance shifts substantially toward DNN-friendly datasets. This highlights the transformative role of foundation models: they narrow, and in many cases invert, the traditional advantage of ensembles by leveraging pretraining and in-context inference. Still, the right tail of the distribution shows that there remain numerous datasets where ensembles achieve clear wins, particularly in regression-heavy or highly categorical settings.  

Overall, the debate has evolved rather than disappeared. Tree-based ensembles remain reliable, statistically competitive baselines across diverse tasks, especially when data structure aligns with their strengths. Yet, pretrained foundation models represent a paradigm shift: they elevate DNNs to state-of-the-art levels across many benchmarks, reducing the universality of ensemble dominance. The results suggest a more nuanced view—trees still matter, but pretrained models are redefining the frontier of tabular learning.

\subsection{Comparisons with Imbalance-Sensitive Criteria}
\begin{figure}[t]
  \centering
   \begin{minipage}{0.45\linewidth}
    \includegraphics[width=\textwidth]{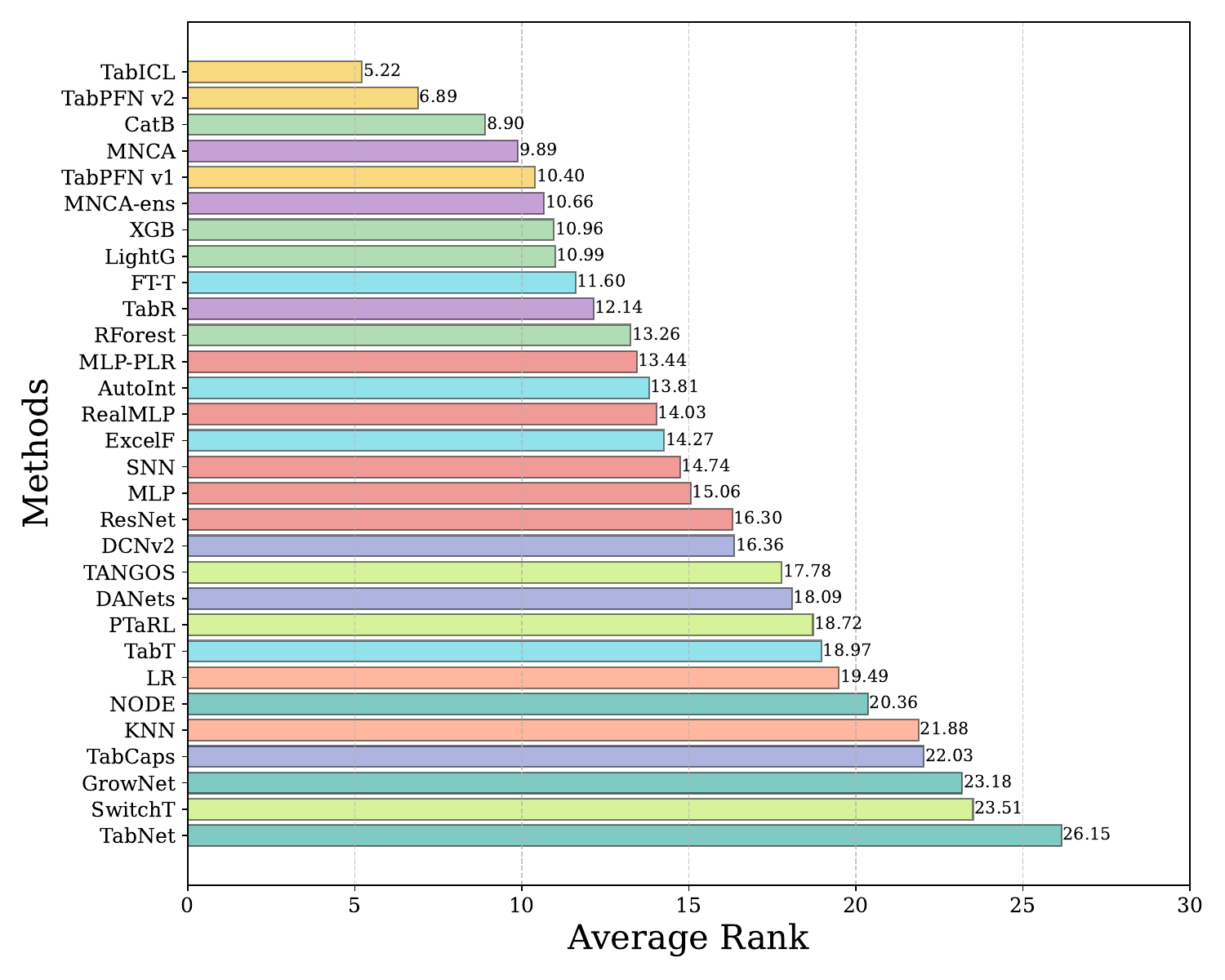}
    \centering
    {\small \mbox{(a) {AUC-based Average Rank}}}
    \end{minipage}
    \begin{minipage}{0.45\linewidth}
    \includegraphics[width=\textwidth]{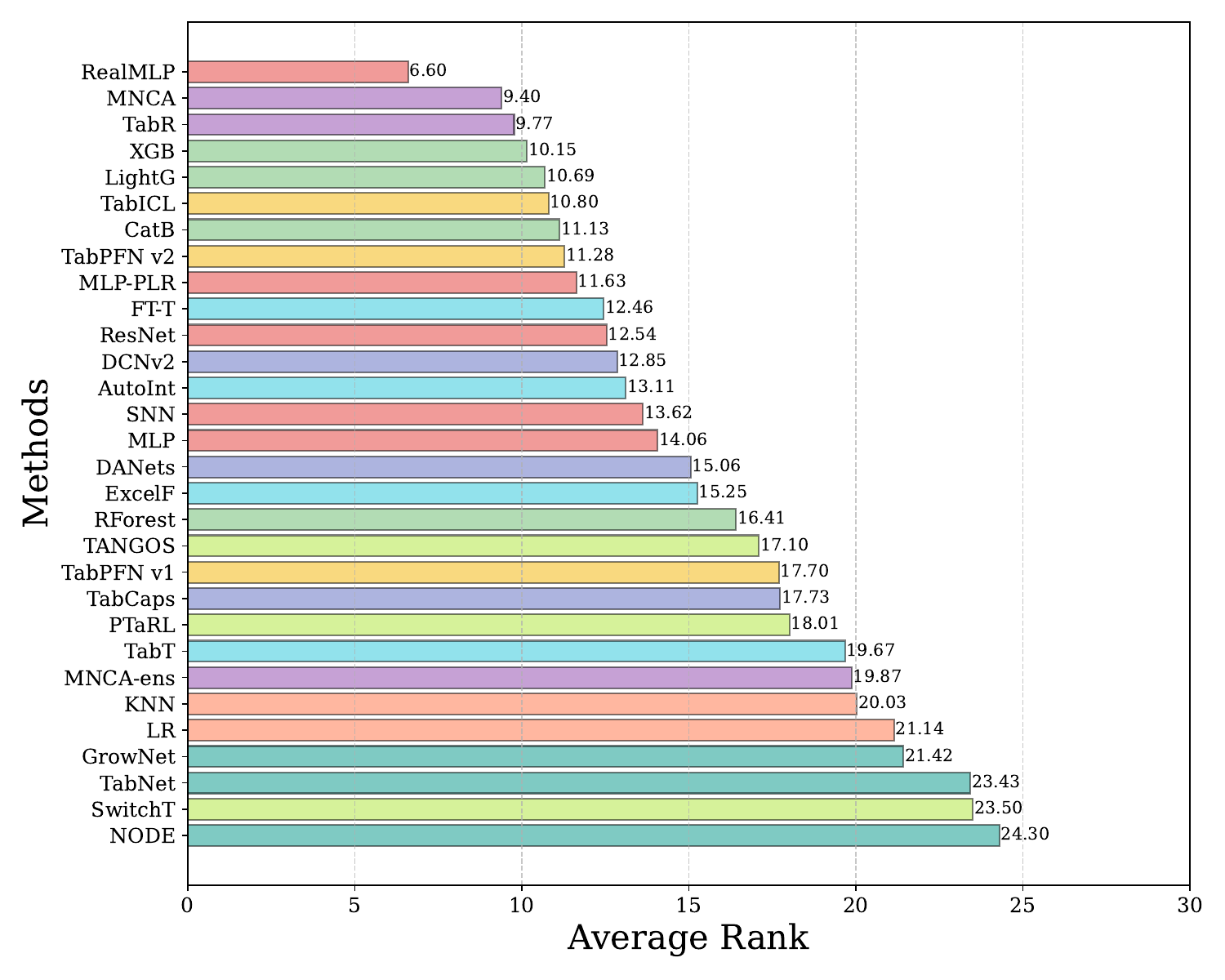}
    \centering
    {\small \mbox{(b) {F1-score-based Average Rank}}}
    \end{minipage}
  \caption{Evaluations of tabular methods on 67 imbalanced classification datasets 
  (31 binary and 36 multi-class tasks with imbalance rates below 0.25). 
  The average ranks are computed using AUC (a) and F1-score (b). 
  Lower values indicate better performance.
  }
  \label{fig:AUC_and_F1}
\end{figure}

Real-world tabular datasets often exhibit class imbalance, making accuracy insufficient as a sole evaluation metric. To complement previous analyses, we assess methods using AUC and F1-score on 67 classification datasets with imbalance rates below 0.25, without applying any additional imbalance-handling strategies. The results are shown in \autoref{fig:AUC_and_F1}.

The rankings reveal both consistency and divergence compared to accuracy-based results. AUC favors ensemble-style approaches and pretrained models: TabICL, TabPFN v2, and CatBoost dominate, with TabICL achieving the lowest average rank. These methods excel at separating minority and majority classes, reflecting their robust decision boundaries under imbalance. ModernNCA and its ensemble variant also remain competitive, consistent with their strong overall performance in earlier benchmarks.

In contrast, F1-score shifts the advantage toward certain DNN-based approaches. RealMLP emerges as the top performer, with ModernNCA, TabR, and CatBoost following closely. These models balance precision and recall more effectively, which is critical when evaluating rare classes. Interestingly, TabICL and TabPFN v2—dominant under AUC—perform slightly less consistently under F1, suggesting that while they are excellent at ranking predictions, they may not optimize precision–recall trade-offs equally well.

Classical ensembles such as Random Forest and XGBoost maintain mid-level performance across both metrics, outperforming most classical baselines but trailing behind modern DNNs and foundation models. Token-based methods like FT-T remain stable, appearing competitive under both AUC and F1, though rarely leading, which aligns with earlier observations from statistical comparisons.

Overall, imbalance-sensitive evaluations confirm the robustness of strong ensembles (CatBoost, LightGBM), advanced DNNs (RealMLP, ModernNCA), and foundation models (TabICL, TabPFN v2). Yet the divergence between AUC and F1 underscores that model choice should align with task-specific goals: AUC-oriented scenarios benefit from ensembles and pretraining, while F1-sensitive contexts may prefer RealMLP or neighborhood-based designs.

\section{Measuring the Heterogeneity of Tabular Data}
\label{sec:hetero_analysis}
One of the central challenges in tabular learning arises from the inherent heterogeneity of tabular datasets~\citep{ZivA22Tabular}. Unlike other modalities such as images or text, where inputs share relatively uniform structures, tabular datasets often combine diverse attribute types--including continuous values, binary indicators, ordinal features, and high-cardinality categorical variables~\citep{Borisov2024Deep}. This diversity poses substantial challenges for deep models, which must simultaneously accommodate heterogeneous statistical properties and learn meaningful interactions across them.

In this section, we build on meta-features that capture dataset properties and systematically examine how their variations reflect the heterogeneity of tabular data. To this end, we introduce a performance-curve prediction task, which learns to forecast the training dynamics of a tabular method from both meta-features and early learning signals. A meta-feature is considered effective if it contributes to accurately predicting these dynamics, thereby indicating its role in shaping model behavior.

By analyzing the predictive relationships between meta-features and training dynamics, we identify which dataset characteristics most strongly influence the effectiveness of deep tabular models. Our results provide insights into how heterogeneity can be measured, which factors are most critical, and how they affect the success or failure of different representative methods. This analysis not only clarifies the limitations of current approaches but also offers guidance for designing more robust and adaptive tabular learning models.

\subsection{Reformulating Meta-Feature Selection as a Dynamics Forecasting Task}
Meta-features capture intrinsic properties of a tabular dataset. To understand which properties matter most for deep tabular learning, we link them to the epoch-wise training dynamics of neural models. Instead of treating meta-feature selection as an isolated procedure, we reformulate it as a forecasting task: predicting validation curves from dataset properties. This formulation allows us to identify which dataset characteristics most strongly influence model behavior.

Formally, given a training set $\sD$, we optimize a deep tabular model $f$ with stochastic gradient descent over~\autoref{eq:objective}. Each epoch is a complete pass through $\sD$, with mini-batches drawn after random permutation of the examples. Assuming the best hyperparameters of $f$ are predetermined, we record validation statistics (\ie, classification accuracy or normalized RMSE for regression) as a sequence $\boldsymbol{a}=[a_1,a_2,\ldots,a_T]\in\bbR_+^T$ over $T$ epochs until early stopping.

We propose to forecast $\boldsymbol{a}$ using two signals: (i) dataset meta-features $\vm_\sD$ that encode structural properties (\eg, number of attributes, joint entropy with the target), and (ii) the initial segment of the validation curve. Specifically, we define a support set $\sS\in\bbR_+^K$ containing the first $K$ values of $\boldsymbol{a}$, and a query set $\sQ\in\bbR_+^{T-K}$ with the remaining values. The task is to learn a mapping
\[
g:\{\vm_\sD, \sS\}\mapsto\sQ,
\]
leveraging both dataset statistics and early validation behavior. By analyzing which meta-features improve forecasting accuracy, we can reveal the key factors shaping the training dynamics of deep tabular methods.

\subsection{Selecting Effective Meta-Features for Heterogeneity Analysis}
To forecast training dynamics from $\sS$, we model the shape of validation curves directly. Let $t$ denote the epoch index and $y$ the performance measure. Because $\sS$ is short, we adopt a learnable approach: predicting the parameters of a curve family from $\{\sS,\vm_\sD\}$, where meta-features provide auxiliary signals that adapt predictions to dataset-specific properties.

All data for this task are drawn from validation curves of MLPs trained with default hyperparameters on our benchmark datasets. We split these curves into 80\% training and 20\% testing, ensuring no dataset overlap between the two.

\noindent{\bf Dynamics Curve Approximation.}
We model validation dynamics with the following curve family:
\begin{equation}
    \boldsymbol{a}_{\theta}(t) = A \log t + B \sqrt{t} + C + D/t, \label{eq:the_law}
\end{equation}
where $t$ is the epoch number, $\boldsymbol{a}(t)$ is the validation performance, and $\theta=\{A,B,C,D\}$ defines the curve. This functional form is empirically selected to capture the characteristic sub-linear growth and asymptotic convergence observed in tabular deep learning validation curves. For accuracy curves, $A$ and $B$ are typically positive, reflecting monotonic improvement. To capture dataset effects, we learn a meta-mapping $h:\{\vm_\sD,\sS\}\mapsto\theta$~\citep{Vinyals2016Matching,Chao2020Meta}. Comparisons with other formulations of the curve family are discussed in the Appendix.

\noindent{\bf Learning Objective.}
We optimize $h$ by minimizing the mean absolute error (MAE) between predicted and observed validation curves:
\begin{equation}
    \min_h \;\sum_{\{\vm_\sD,\sS\}} \sum_{a_t\in\sQ} 
    \ell\!\left(\boldsymbol{a}_{\theta=h(\vm_\sD,\sS)}(t),\,a_t\right). \label{eq:meta-_obj}
\end{equation}
For each dataset, we collect the first five epochs ($\sS$) and meta-features $\vm_\sD$ (see~\autoref{tab:meta_features}), feed them into $h$ (an MLP), and predict $\theta$. The predicted curve extrapolates $\sQ$, and accuracy is assessed by the discrepancy between predictions and ground-truth dynamics. After training, $h$ generalizes to unseen datasets, highlighting the most effective meta-features for forecasting training behavior.

\subsection{The Selected Meta-Features}
We implement $h$ as a four-layer MLP. The input dimension is 24: five dimensions from the first five epochs of $\sS$ and 19 from dataset meta-features (and derived statistics such as \texttt{range.mean}, \texttt{range.std}, etc.). The output is the four parameters of~\autoref{eq:the_law}. We evaluate $h$ on classification and regression curves and find that including meta-features substantially improves prediction accuracy. Additional results are reported in the appendix.

\begin{figure}[t]
\centering
\includegraphics[width=0.8\textwidth]{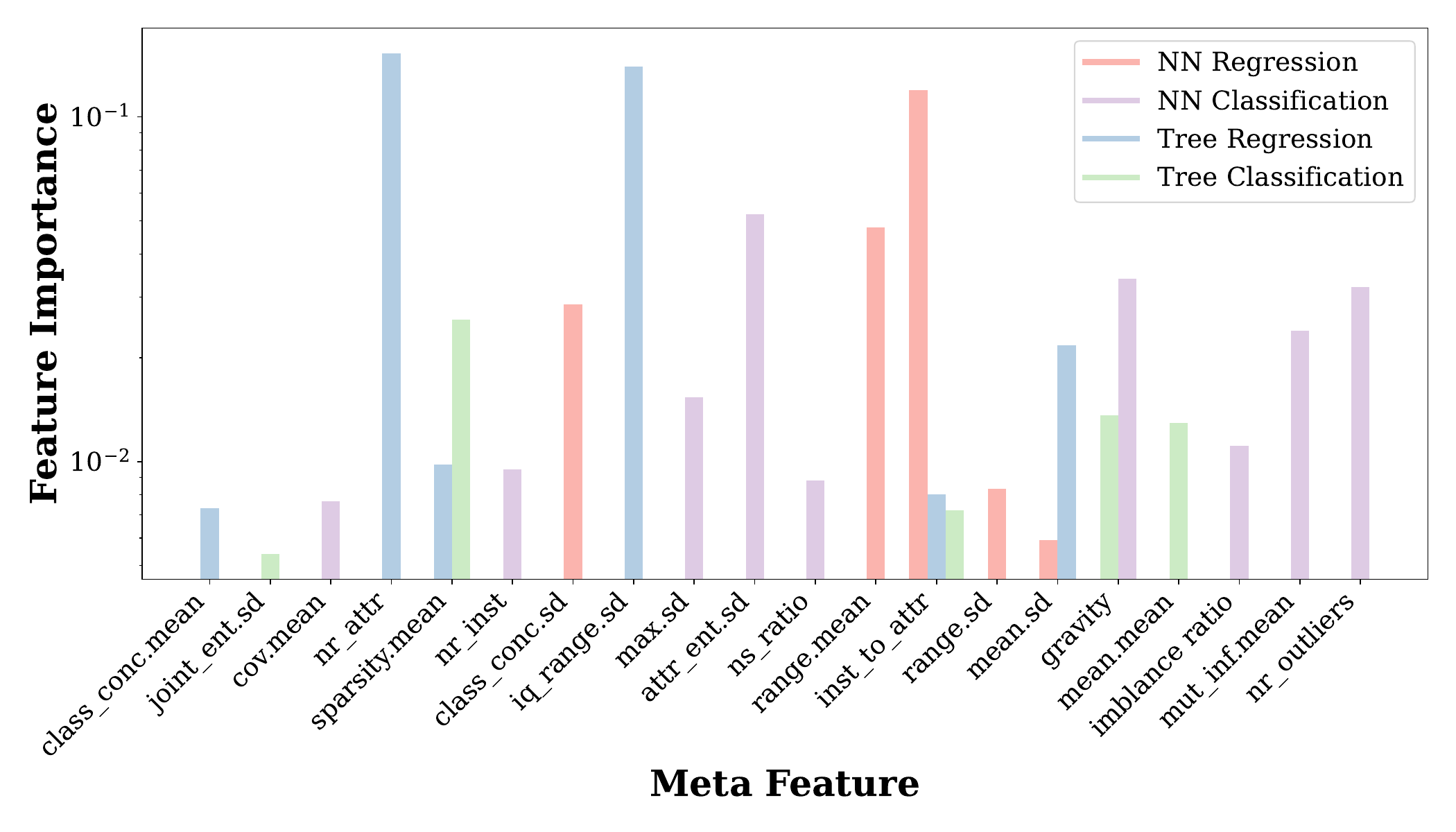}
\caption{Importance of various meta-features for predicting training dynamics across different types of tabular datasets. Legends such as ``NN Regression'' denote selected important meta-features for regression tasks on DNN-friendly datasets.}
\label{fig:feature_importance_full}
\end{figure}
We further analyze which meta-features most strongly influence predictions by examining their weights in $h$. Training datasets are divided into four categories (classification/regression $\times$ tree-/DNN-friendly), where the tree-/DNN-friendly split is determined according to the Tree-DNN score in \autoref{fig:addition_results}(a). For each category, we train a separate predictor and extract the most important meta-features (see~\autoref{fig:feature_importance_full}). Results reveal that: (1) For classification tasks, the \texttt{gravity} meta-feature—measuring the distance between minority and majority class centers—is critical for both tree-based and DNN-based methods. (2) For regression tasks, \texttt{range.mean} (average attribute range) and \texttt{mean.mean} (average attribute mean) are highly predictive. (3) Tree-based methods rely heavily on \texttt{sparsity.mean}, while DNN-based methods are more sensitive to distributional statistics such as \texttt{max.sd}. 

These observations suggest that dataset heterogeneity is well captured by a small set of meta-features encoding complexity, feature variability, and data quality. Concretely:
\begin{itemize}[noitemsep,topsep=0pt,leftmargin=*]
    \item  \texttt{gravity}: Measures the Minkowski distance between the centers of mass of the majority and minority classes. A smaller distance indicates higher overlap between class distributions and, thus, greater classification difficulty.  
    \item \texttt{inst\_to\_attr}: The ratio of the number of instances to the number of attributes, reflecting the balance between samples and features in the dataset.
\end{itemize}

There are four meta-features encoding the heterogeneity of features.
\begin{itemize}[noitemsep,topsep=0pt,leftmargin=*]
    \item  \texttt{sparsity\_mean},  \texttt{sparsity\_std }: Quantify the variability of unique values in numeric features, where sparsity for a feature vector \( v \) is defined as:
     \[
     S(v) = \frac{1.0}{n - 1.0} \cdot \left( \frac{n}{\phi(v)} - 1.0 \right),
     \]
     with \( n \) representing the instance number and \( \phi(v) \) the number of distinct values in \( v \).  
     \item  \texttt{entropy\_mean},  \texttt{entropy\_std}: Reflect the diversity in feature value distributions, where entropy for a feature \( v \) is calculated using:
     \[
     H(v) = -\sum_{k} p_k \cdot \log_{\text{base}}(p_k),
     \]
     where \( p_k \) is the proportion of instances with the \( k \)-th unique value, and the logarithm base equals the number of unique values in \( v \).  
     \item  \texttt{iq\_range\_std}: The standard deviation of interquartile ranges (\( Q_3 - Q_1 \)) across all attributes, capturing variability in feature spread.
     \item  \texttt{range\_mean}: The mean range (\( \max - \min \)) across all attributes.  
     \end{itemize}
     
\noindent There is a meta-feature encoding the data quality of a tabular dataset.
\begin{itemize}[noitemsep,topsep=0pt,leftmargin=*]
    \item \texttt{nr\_outliers} encodes the quality of a tabular dataset. In detail, it is the number of features containing at least one outlier value, where an outlier is defined as a value lying outside 1.5 times the interquartile range (IQR). 
\end{itemize}

\begin{table}[t]
\centering
\caption{Meta-features with the strongest correlation to the performance (average rank) of each representative method. The correlation values indicate the strength of the relationship, with negative values showing better performance as the meta-feature decreases.
}
\label{tab:correlation}
\begin{tabular}{ccc}
\toprule
\textbf{Method} & \textbf{Correlation Value}  & \textbf{Meta-Feature}    \\ 
\midrule
XGBoost              & 0.3809            & \(\texttt{entropy\_mean}\)    \\ 
LightGBM             & -0.3487           &  \(\texttt{entropy\_std}\)    \\
Catboost             & 0.3101            & \(\texttt{entropy\_mean}\)    \\
Random Forest        & -0.2571           & \(\texttt{sparsity\_std}\)    \\
RealMLP              & -0.2819           & \(\texttt{inst\_to\_attr}\)   \\
TabM                 & -0.2059           & \(\texttt{sparsity\_mean}\)   \\
FT-T                 & -0.2671           & \(\texttt{inst\_to\_attr}\)   \\
ModernNCA            & 0.1805            & \(\texttt{entropy\_std}\)     \\
TabPFN v2            & 0.2765            & \(\texttt{inst\_to\_attr}\)   \\ 
\bottomrule
\end{tabular}
\end{table}

These meta-features collectively provide a robust way for assessing dataset characteristics, capturing factors such as class separability, feature variability, and the presence of anomalies. By formalizing these properties, we enable a systematic evaluation of dataset properties and their impact on model performance.  

\subsection{Correlation between Selected Meta-Features and Types of Tabular Methods} 
We analyze the relationship between dataset meta-features and the performance of nine representative methods introduced in~\autoref{sec:benchmark}. By pairing each dataset’s selected meta-features with the corresponding average rank of each method, we identify the meta-feature exhibiting the strongest absolute correlation with that method’s performance. A higher absolute correlation indicates that the property revealed by the meta-feature has a stronger influence on the performance of the corresponding tabular method.

\begin{figure}[t]
  \centering
   \begin{minipage}{0.32\linewidth}
    \includegraphics[width=\textwidth]{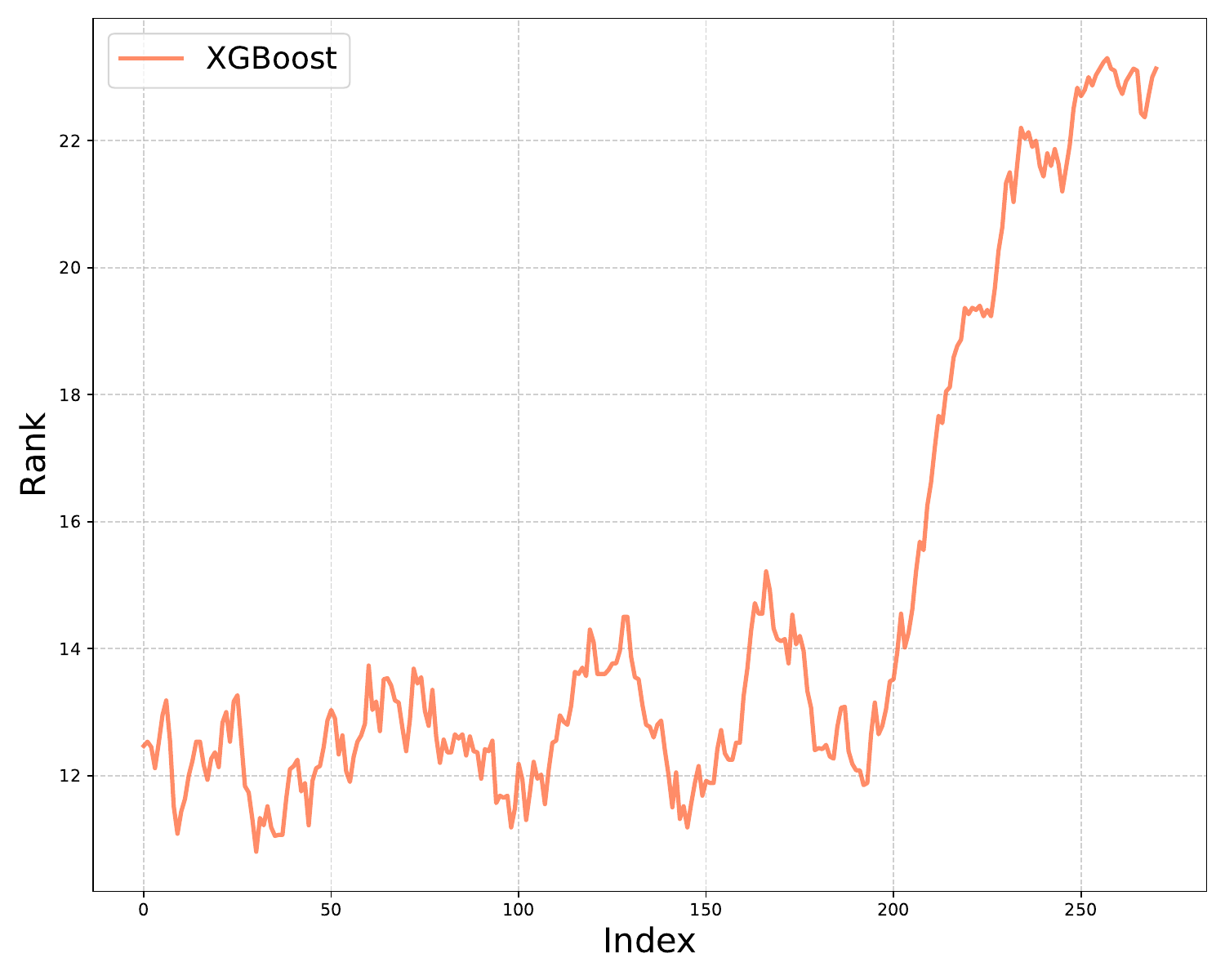}
    \centering
    {\scriptsize \mbox{(a) {XGBoost:\texttt{entropy\_mean}}}}
    \end{minipage}
    \begin{minipage}{0.32\linewidth}
    \includegraphics[width=\textwidth]{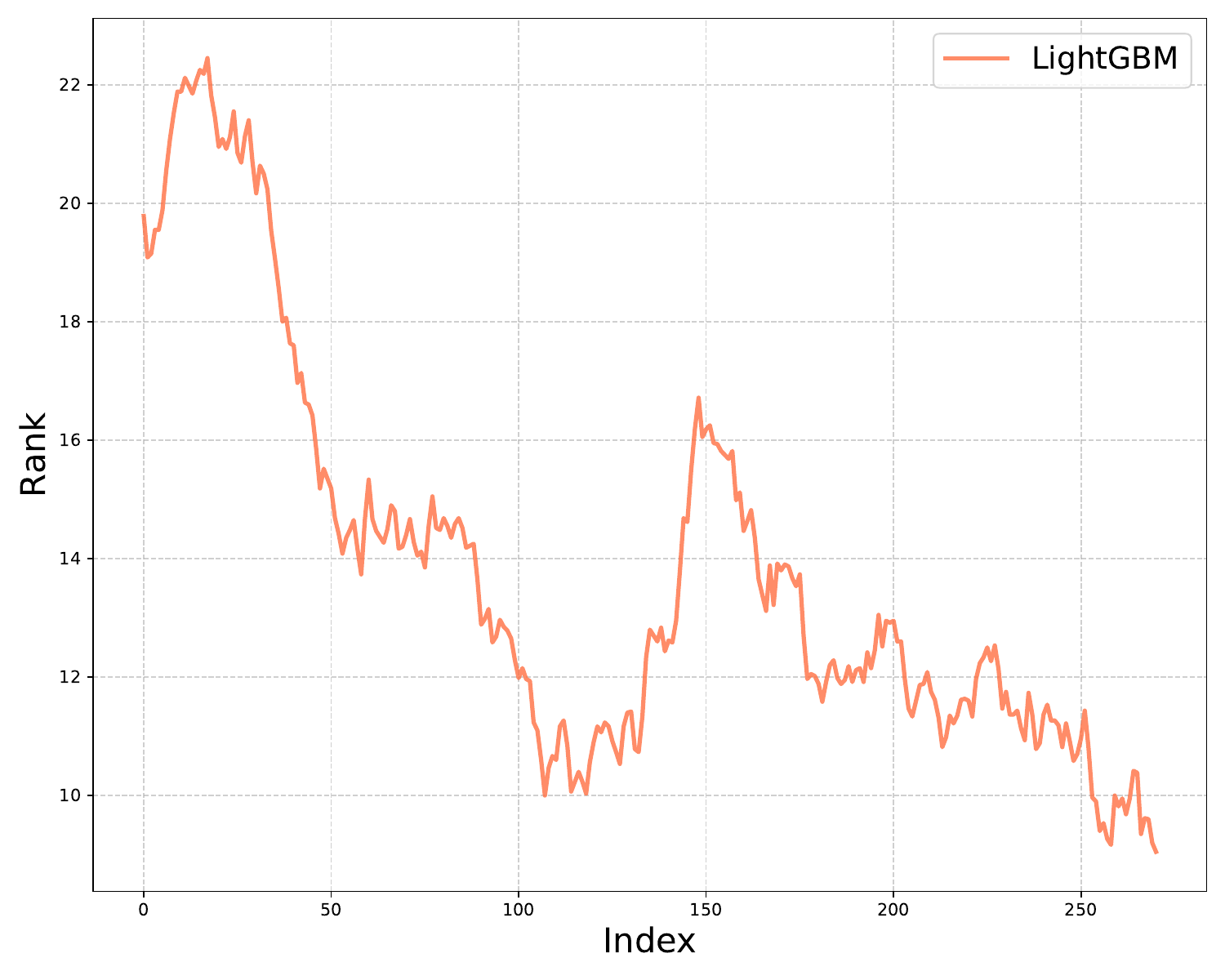}
    \centering
    {\scriptsize \mbox{(b) {LightGBM:\texttt{entropy\_std}}}}
    \end{minipage}
    \begin{minipage}{0.32\linewidth}
    \includegraphics[width=\textwidth]{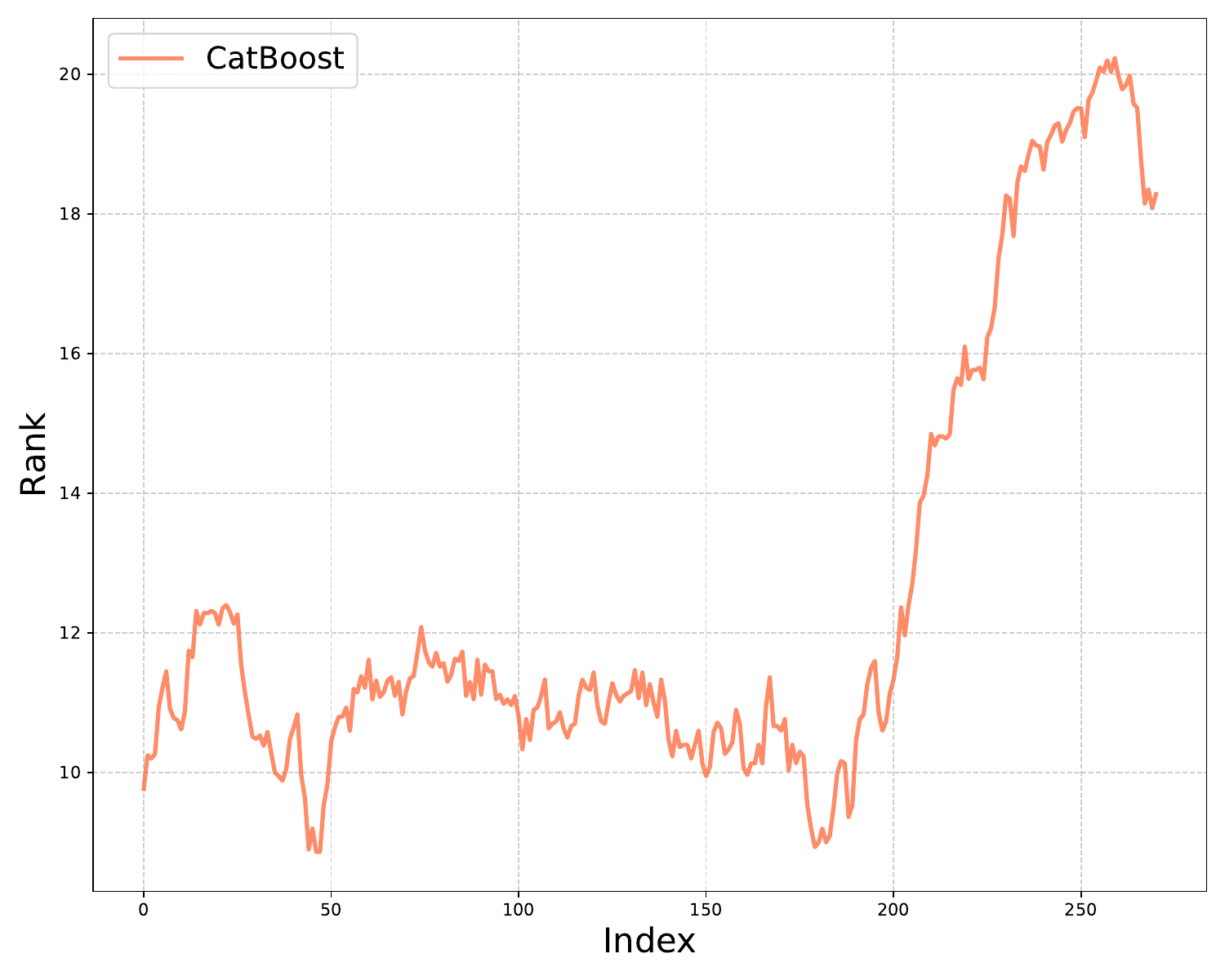}
    \centering
    {\scriptsize \mbox{(c) {Catboost:\texttt{entropy\_mean}}}}
    \end{minipage}
    
    \begin{minipage}{0.32\linewidth}
    \includegraphics[width=\textwidth]{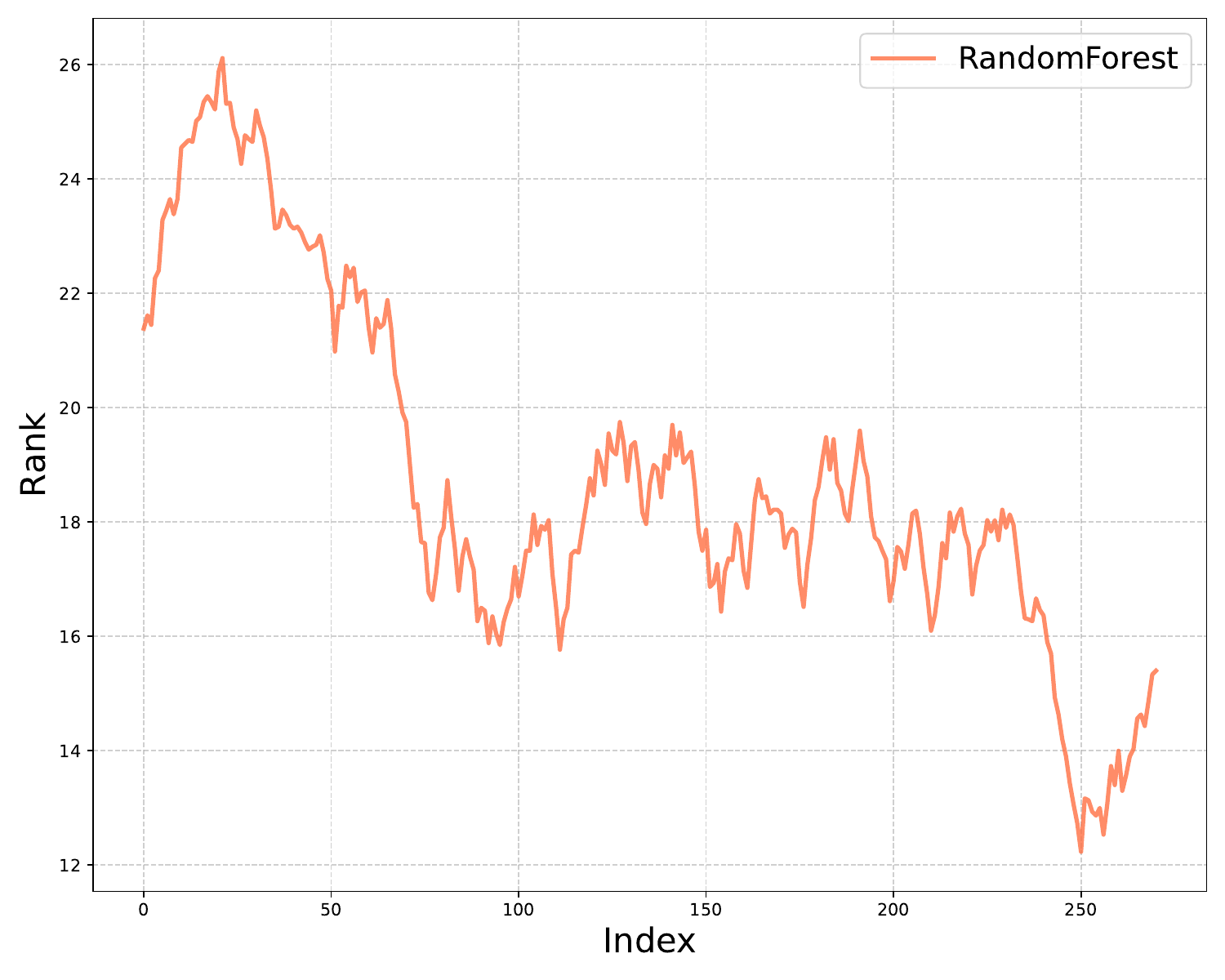}
    \centering
    {\scriptsize \mbox{(d) {Random Forest:\texttt{sparsity\_std}}}}
    \end{minipage}
    \begin{minipage}{0.32\linewidth}
    \includegraphics[width=\textwidth]{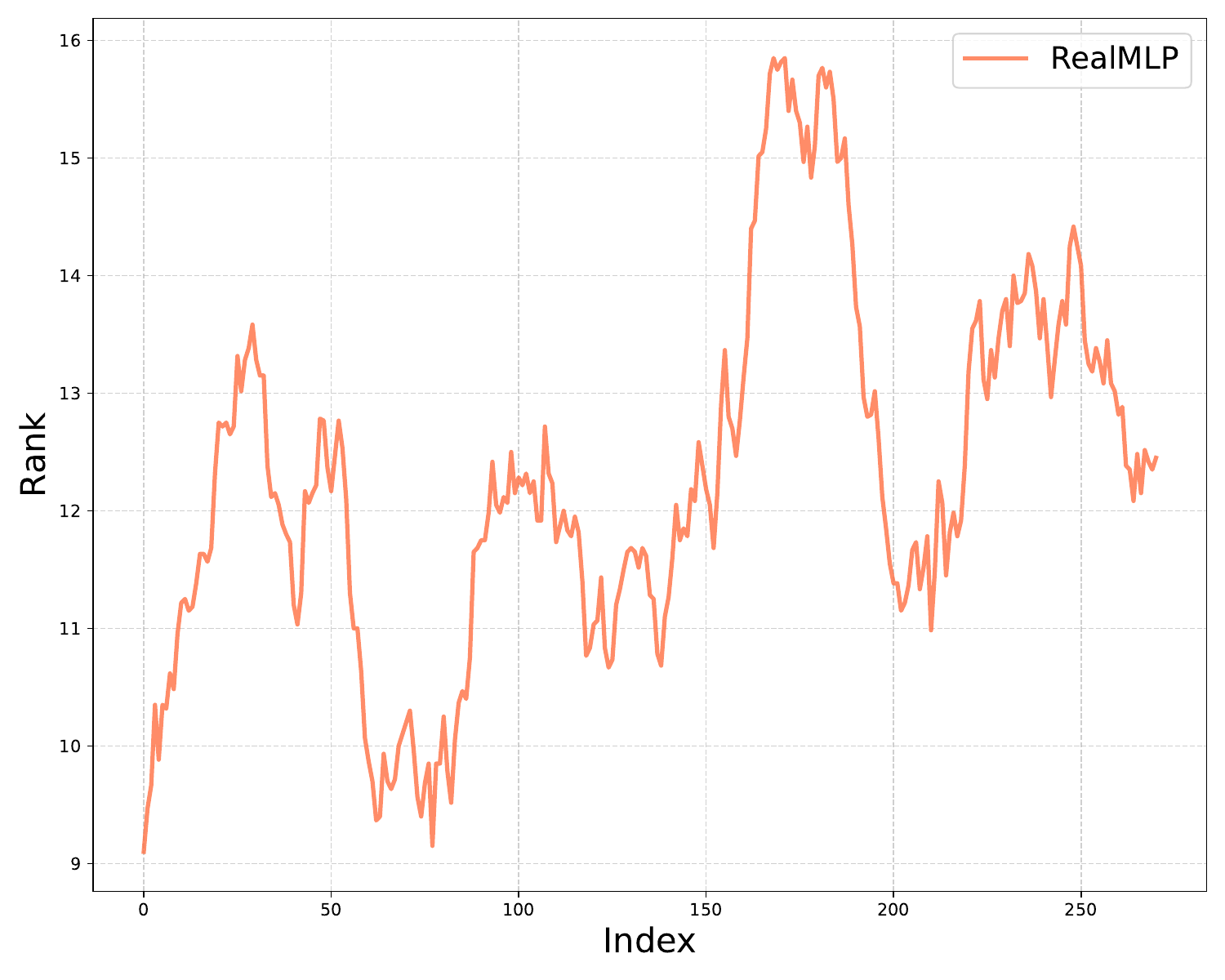}
    \centering
    {\scriptsize \mbox{(e) {RealMLP:\texttt{inst\_to\_attr}}}}
    \end{minipage}
    \begin{minipage}{0.32\linewidth}
    \includegraphics[width=\textwidth]{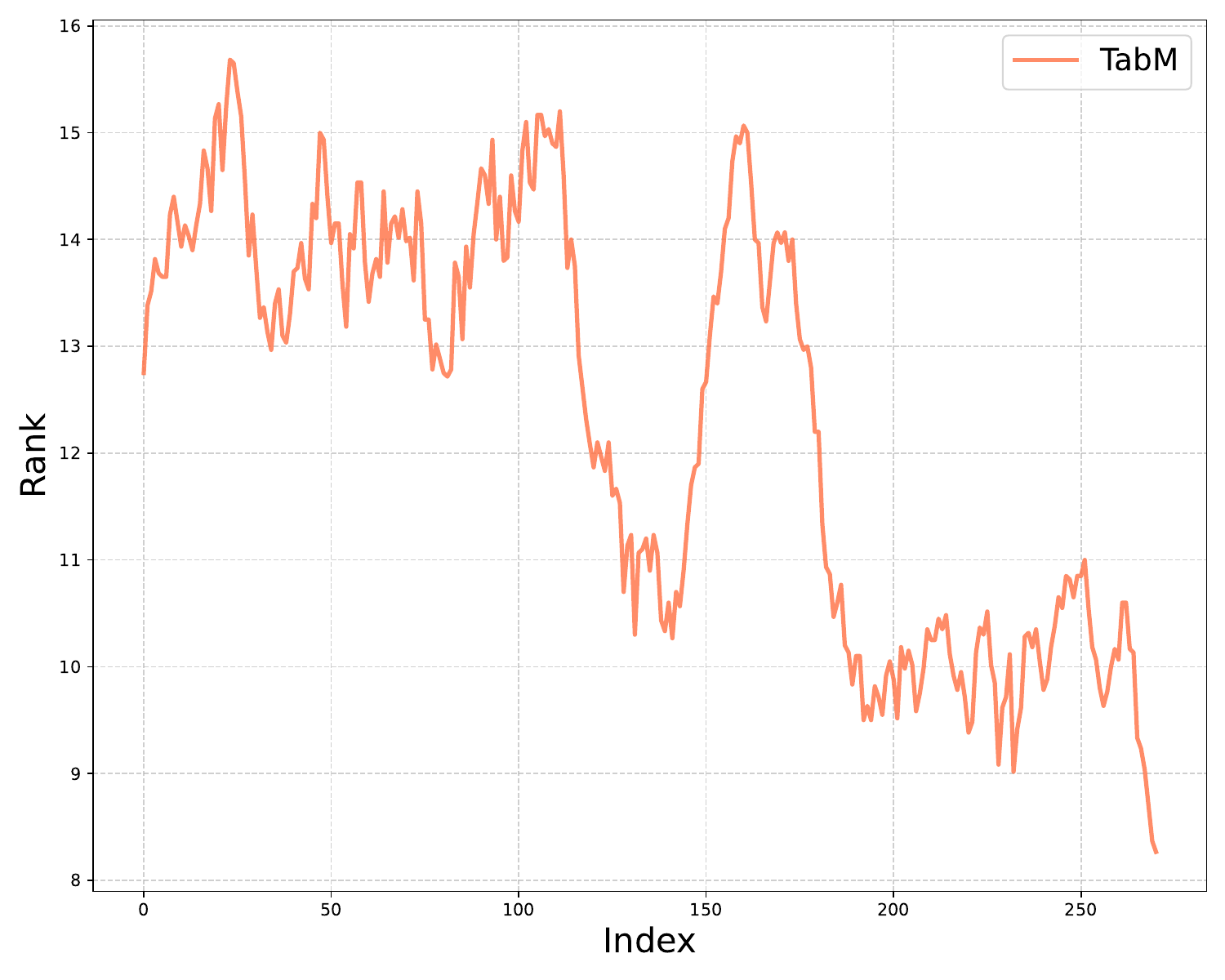}
    \centering
    {\scriptsize \mbox{(f) {TabM:\texttt{sparsity\_mean}}}}
    \end{minipage}
    
    \begin{minipage}{0.32\linewidth}
    \includegraphics[width=\textwidth]{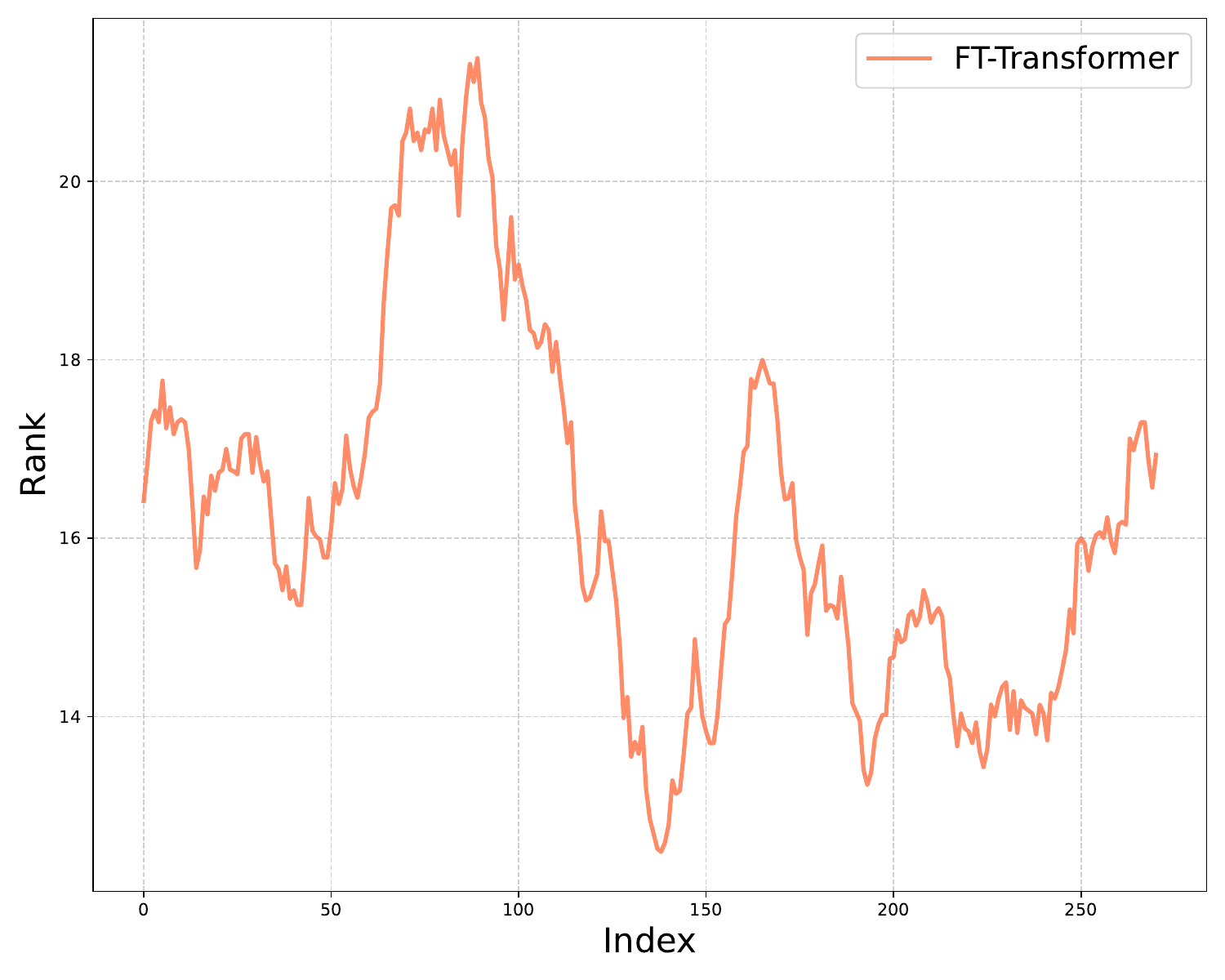}
    \centering
    {\scriptsize \mbox{(g) {FT-Transformer:\texttt{inst\_to\_attr}}}}
    \end{minipage}
    \begin{minipage}{0.32\linewidth}
    \includegraphics[width=\textwidth]{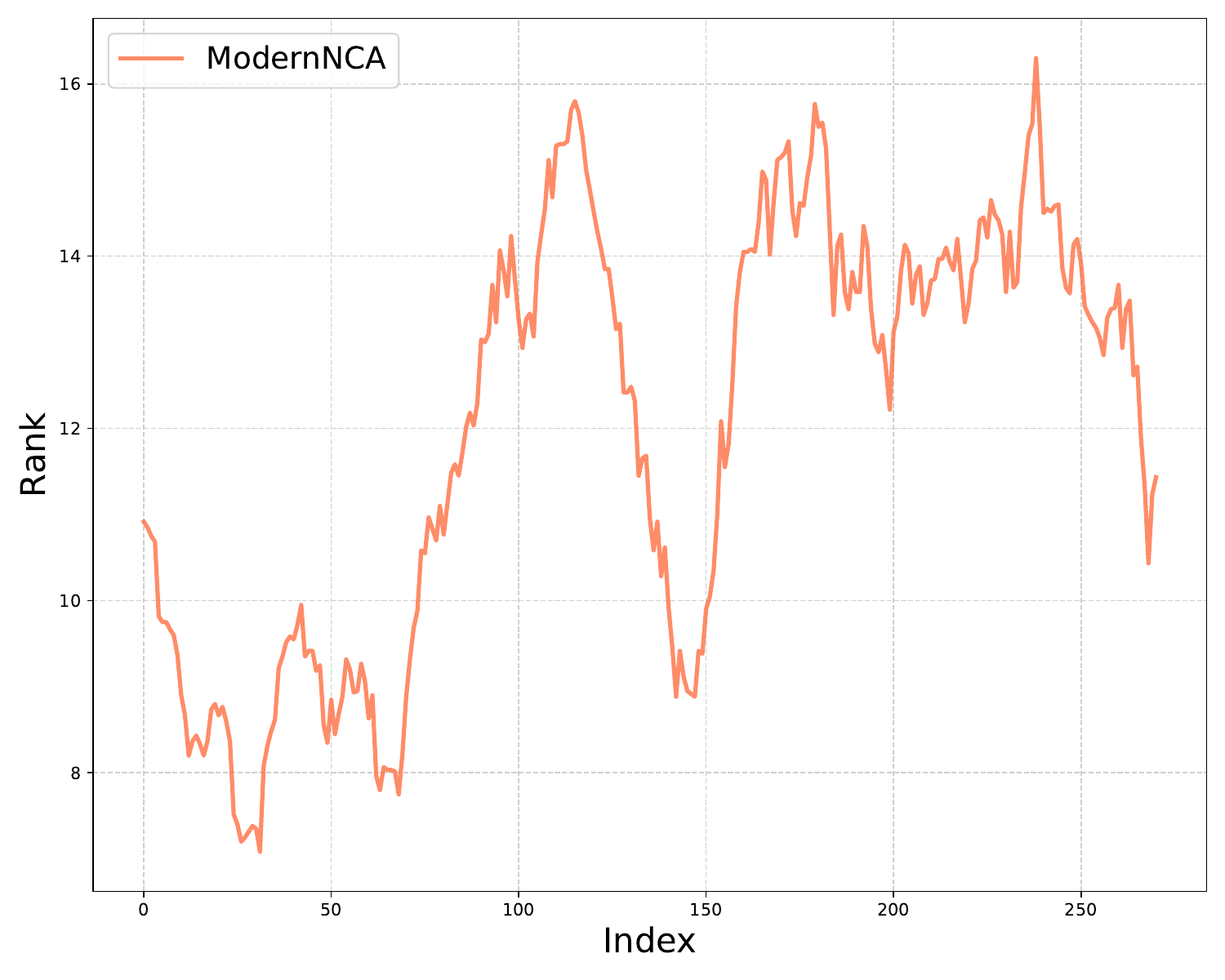}
    \centering
    {\scriptsize \mbox{(h) {ModernNCA:\texttt{entropy\_std}}}}
    \end{minipage}
     \begin{minipage}{0.32\linewidth}
    \includegraphics[width=\textwidth]{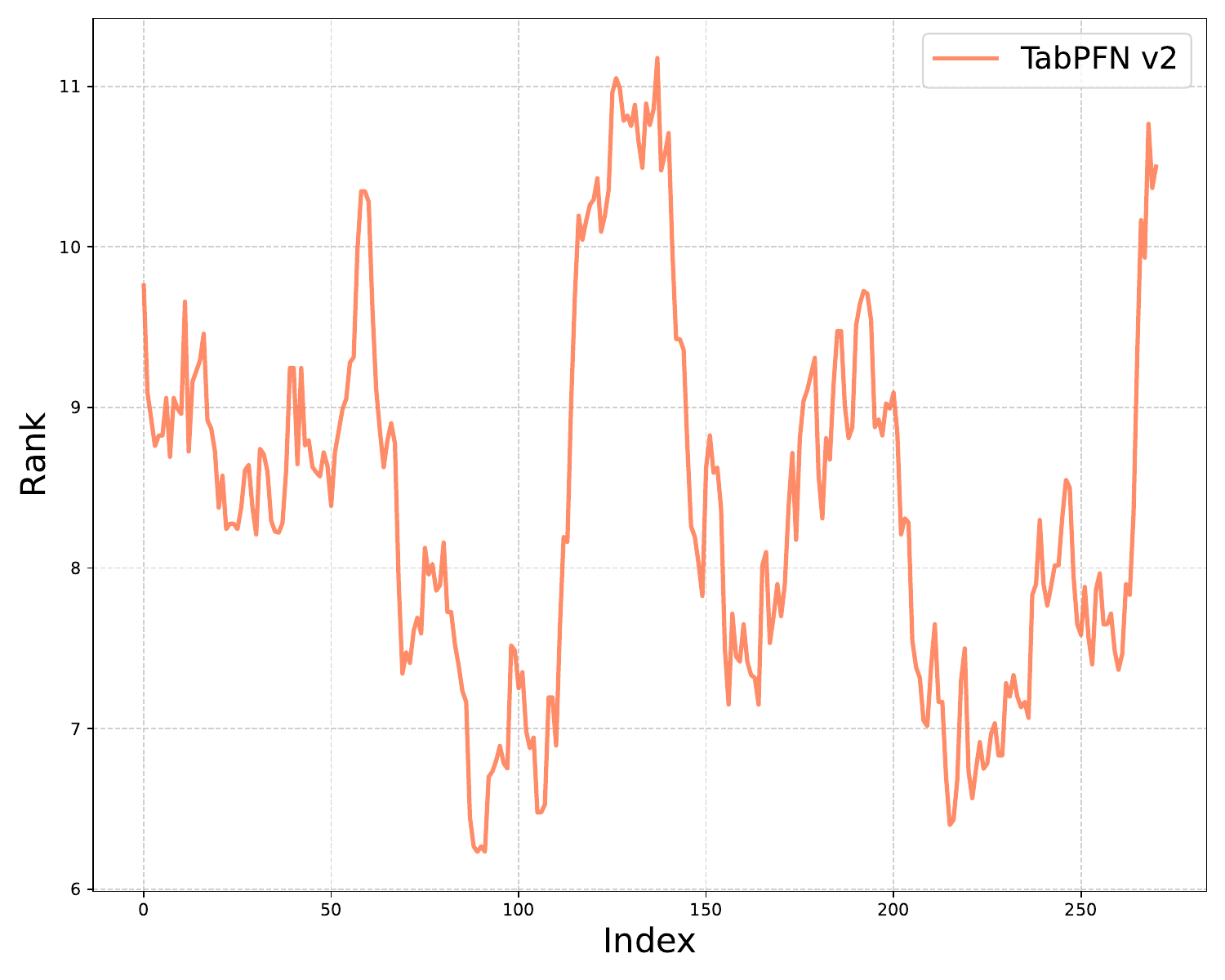}
    \centering
    {\scriptsize \mbox{(i) {TabPFN v2:\texttt{inst\_to\_attr}}}}
    \end{minipage}
  \caption{Dynamics of the average rank for each representative method as shown in \autoref{tab:correlation}, plotted against datasets ordered by their most correlated meta-feature. Each subfigure corresponds to a selected method and its associated meta-feature. The horizontal axis represents datasets ranked by ascending values of the meta-feature, while the vertical axis shows the rank. To enhance readability, the curves have been smoothed.
  }
  \label{fig:dynamics_meta_feature}
\end{figure}
Table~\ref{tab:correlation} summarizes the meta-feature most strongly correlated with the performance of each representative method. 
A negative correlation indicates that higher values of the meta-feature are associated with better performance, while a positive correlation suggests the opposite.

% Specifically, most methods show the strongest correlation with \textit{Feature Heterogeneity} metrics, highlighting the impact of feature distribution variability on performance. In particular, TabM is most sensitive to \(\texttt{sparsity\_mean}\), Random Forest to \(\texttt{sparsity\_std}\), while several deep learning methods—including RealMLP, FT-Transformer, and TabPFN v2—exhibit strong correlations with \(\texttt{inst\_to\_attr}\), a meta-feature representing the ratio of instances to attributes. This underscores the diverse ways in which dataset properties influence the performance of different tabular methods.
Specifically, most methods show the strongest correlation with \textit{Feature Heterogeneity} metrics, highlighting the impact of feature distribution variability on performance. For example, XGBoost and CatBoost are most correlated with \(\texttt{entropy\_mean}\)
, while LightGBM and ModernNCA show their strongest correlations with 
\(\texttt{entropy\_std}\). Random Forest is most sensitive to 
\(\texttt{sparsity\_std}\), whereas TabM relies more on 
\(\texttt{sparsity\_mean}\). Several deep learning methods—including RealMLP, FT-Transformer, and TabPFN v2—exhibit strong correlations with 
\(\texttt{inst\_to\_attr}\)
, a meta-feature representing the ratio of instances to attributes. This diversity underscores that different tabular methods are influenced by distinct dataset properties, even though feature heterogeneity metrics emerge as the most consistent driver across models.

To further illustrate these relationships, we visualize how the rank of different methods changes with respect to their most correlated meta-feature in~\autoref{fig:dynamics_meta_feature}. The horizontal axis shows the sorted values of the most correlated meta-feature, while the vertical axis indicates the average rank for each representative method.

In some cases, the rank changes monotonically. For example, in~\autoref{fig:dynamics_meta_feature}~(b), LightGBM shows a decreasing trend in rank as \(\texttt{entropy\_std}\) increases. Since the standard deviation of entropy captures the variation in feature types within a dataset, this observation suggests that tree-based methods perform better (achieve lower ranks) when the dataset contains a higher diversity of feature types.

Conversely, several deep tabular methods, including RealMLP, FT-Transformer, and TabPFN v2, show a monotonic increase in rank as \(\texttt{inst\_to\_attr}\) increases (see~\autoref{fig:dynamics_meta_feature}~(e), (g), and (i)). This indicates that these methods tend to perform worse when the number of instances relative to the number of features is high, highlighting their sensitivity to dataset size and feature dimensionality. 

TabM and Random Forest exhibit clear correlations with sparsity-related meta-features. Specifically, TabM's rank decreases as \(\texttt{sparsity\_mean}\) decreases (better performance on denser datasets), while Random Forest shows a decreasing trend in rank with lower \(\texttt{sparsity\_std}\) (see~\autoref{fig:dynamics_meta_feature}~(d) and (f)).

Some methods, such as ModernNCA, display moderate sensitivity to feature heterogeneity (\(\texttt{entropy\_std}\)), with rank fluctuating in response to increasing variability (see~\autoref{fig:dynamics_meta_feature}~(h)). These observations collectively highlight the diverse ways in which dataset properties influence the performance of different tabular methods.

\begin{table}[t]
\centering
\caption{Meta-features with the strongest correlation to the performance (average rank) of tree-based methods, DNN-based methods, and their differences. The correlation values indicate the strength of the relationship, with negative values showing better performance as the meta-feature increases.}
\label{tab:correlation_difference}
\begin{tabular}{ccc}
\toprule
\textbf{Category} & \textbf{Correlation Value} & \textbf{Meta-Feature} \\
\midrule
Tree              & -0.3854                    & \(\texttt{entropy\_std}\)    \\
DNN               & -0.3502                    & \(\texttt{inst\_to\_attr}\)  \\
Tree-DNN          & -0.3333                    & \(\texttt{entropy\_std}\)    \\
\bottomrule
\end{tabular}
\end{table}

\subsection{Analysis of the Tree-DNN Performance Gap via Selected Meta-Features} 
% We further analyze which meta-features most significantly influence the performance gap between tree-based and DNN-based methods. Tree-based methods include Random Forest, XGBoost, LightGBM, and CatBoost. We choose representative DNN-based methods with low average ranks, \ie, RealMLP, TabM, ModernNCA, and FT-Transformer.

% For each dataset, we compute the performance gap as the difference in average rank values between tree-based and DNN-based methods. This tree-DNN performance gap quantifies the relative advantage or disadvantage of one category over the other. The meta-features most strongly correlated with the performance of tree-based methods, DNN-based methods, and the tree-DNN gap are listed in~\autoref{tab:correlation_difference}.

We further analyze which meta-features most significantly influence the performance gap between tree-based and DNN-based methods. Tree-based methods include Random Forest, XGBoost, LightGBM, and CatBoost. We choose representative DNN-based methods with low average ranks, \ie, RealMLP, TabM, ModernNCA, and FT-Transformer.

For each dataset, we compute the performance gap as the \textbf{difference in average rank values} between tree-based and DNN-based methods. This tree-DNN performance gap quantifies the relative advantage or disadvantage of one category over the other. The meta-features most strongly correlated with the performance of tree-based methods, DNN-based methods, and the tree-DNN gap are listed in~\autoref{tab:correlation_difference}.

\begin{figure}[t]
  \centering
    \begin{minipage}{0.32\linewidth}
    \includegraphics[width=\textwidth]{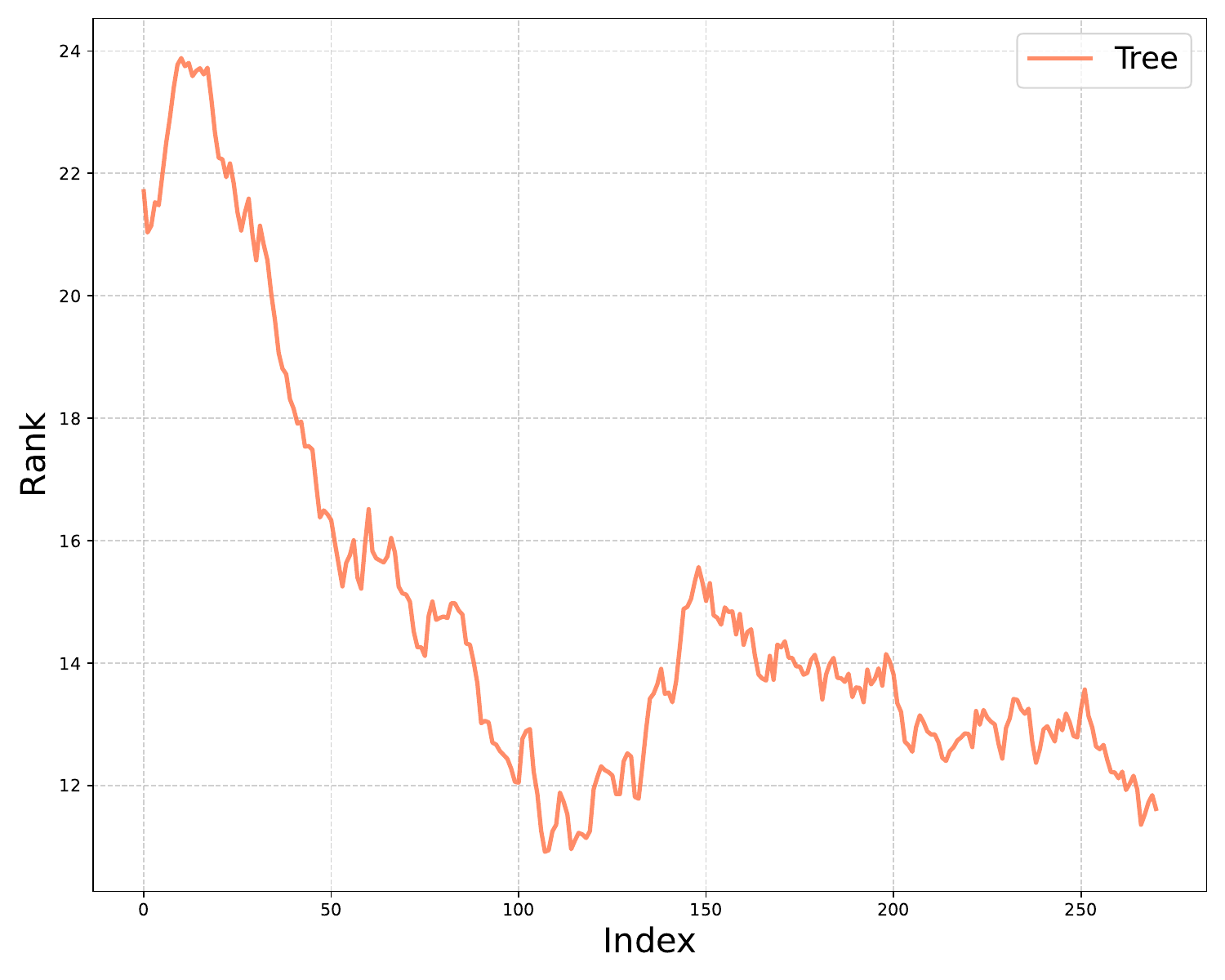}
    \centering
    {\scriptsize \mbox{(a) {Tree: \texttt{entropy\_std}}}}
    \end{minipage}
    \begin{minipage}{0.32\linewidth}
    \includegraphics[width=\textwidth]{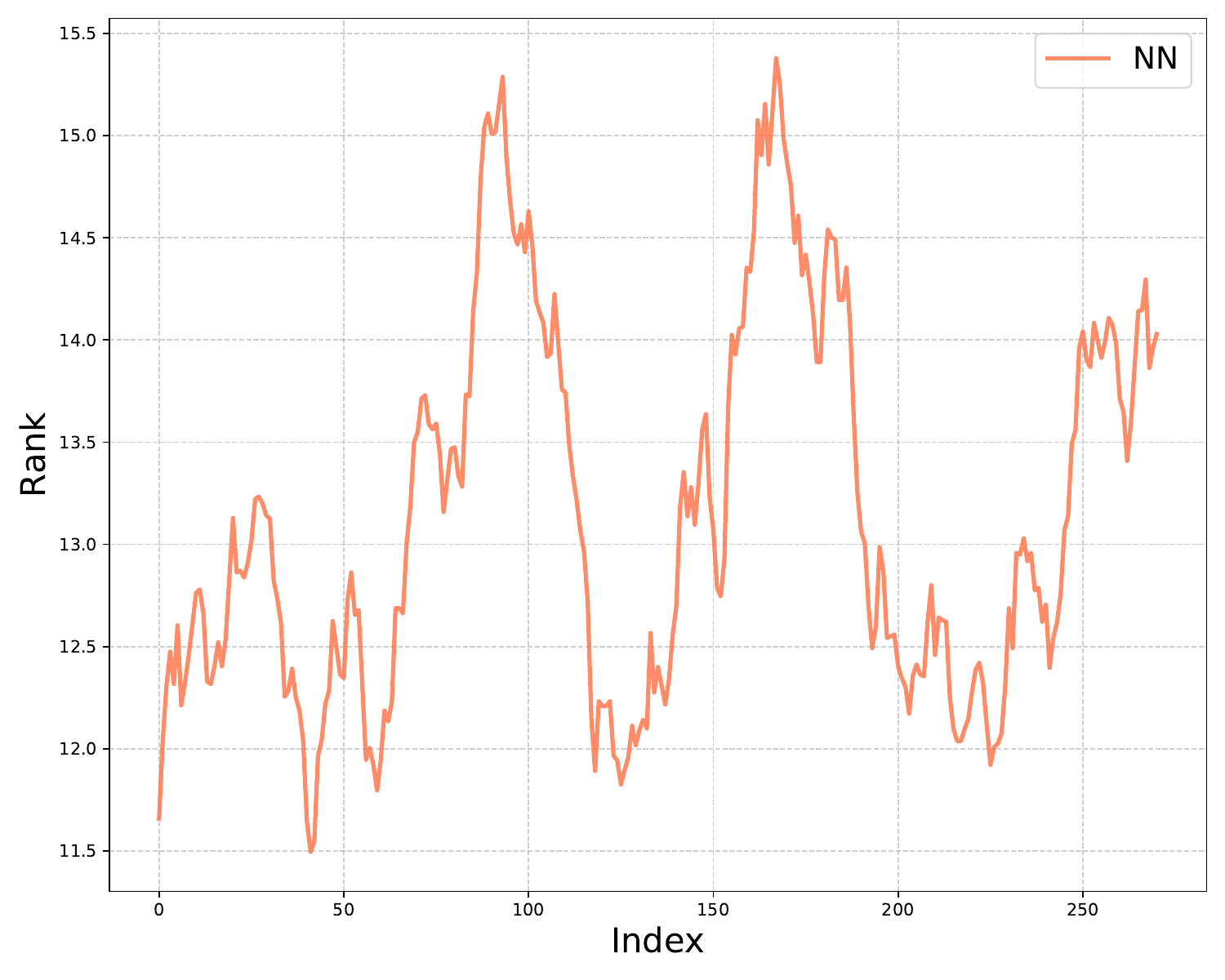}
    \centering
    {\scriptsize \mbox{(b) {DNN: \texttt{inst\_to\_attr}}}}
    \end{minipage}
    \begin{minipage}{0.32\linewidth}
    \includegraphics[width=\textwidth]{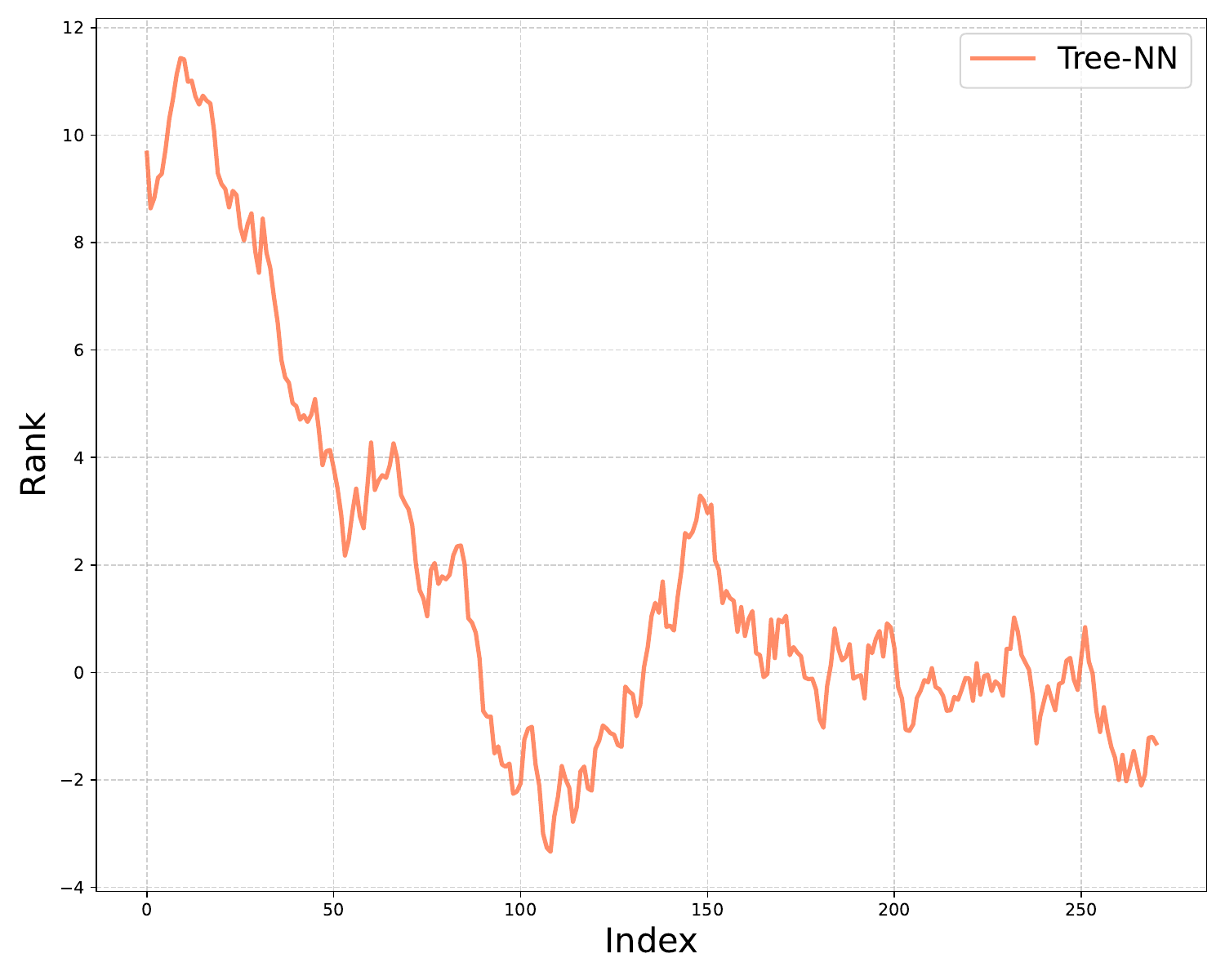}
    \centering
    {\scriptsize \mbox{(c) {Tree-DNN: \texttt{entropy\_std}}}}
    \end{minipage}

    \begin{minipage}{0.32\linewidth}
    \includegraphics[width=\textwidth]{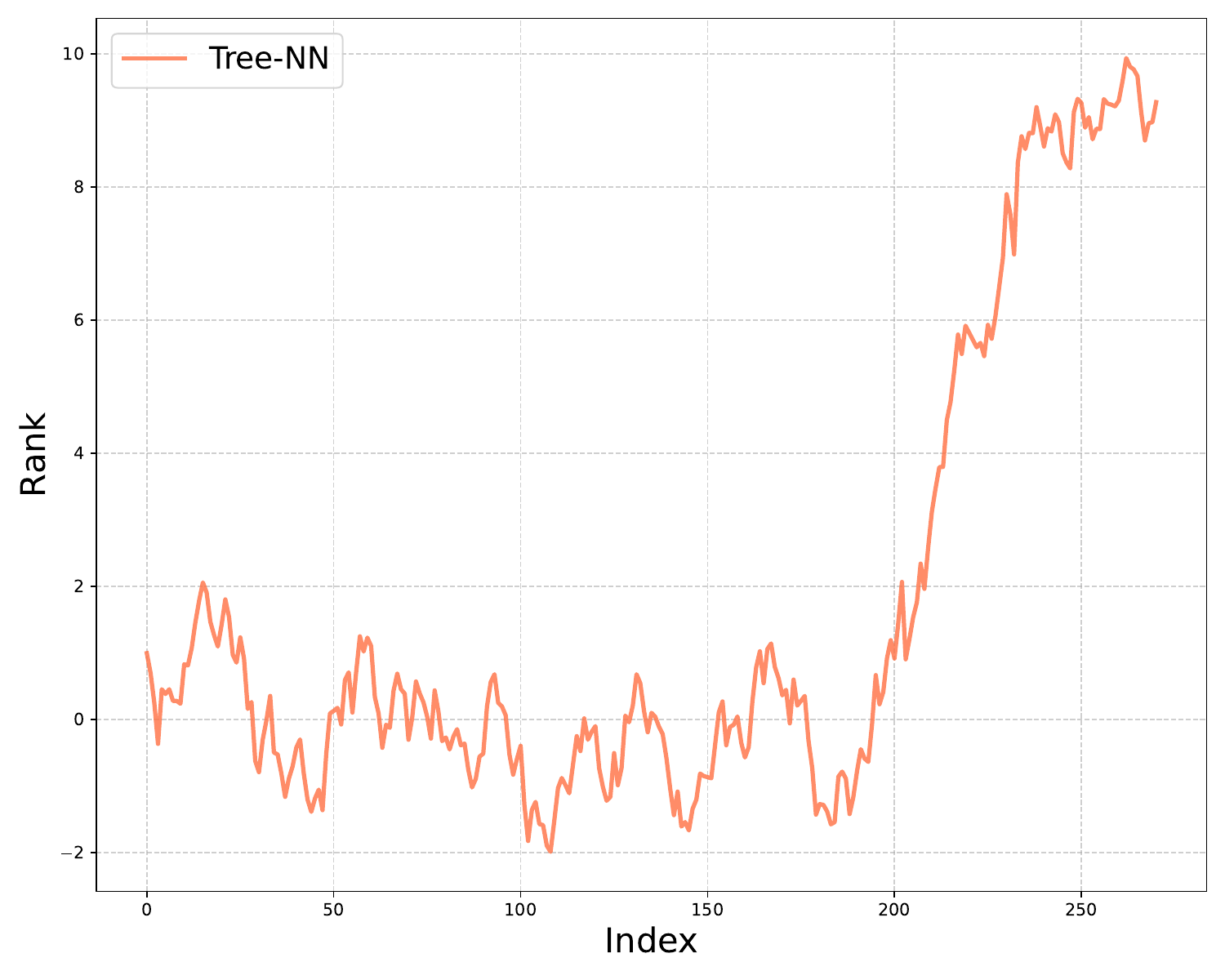}
    \centering
    {\scriptsize \mbox{(d) {Tree-DNN: \texttt{entropy\_mean}}}}
    \end{minipage}
    \begin{minipage}{0.32\linewidth}
    \includegraphics[width=\textwidth]{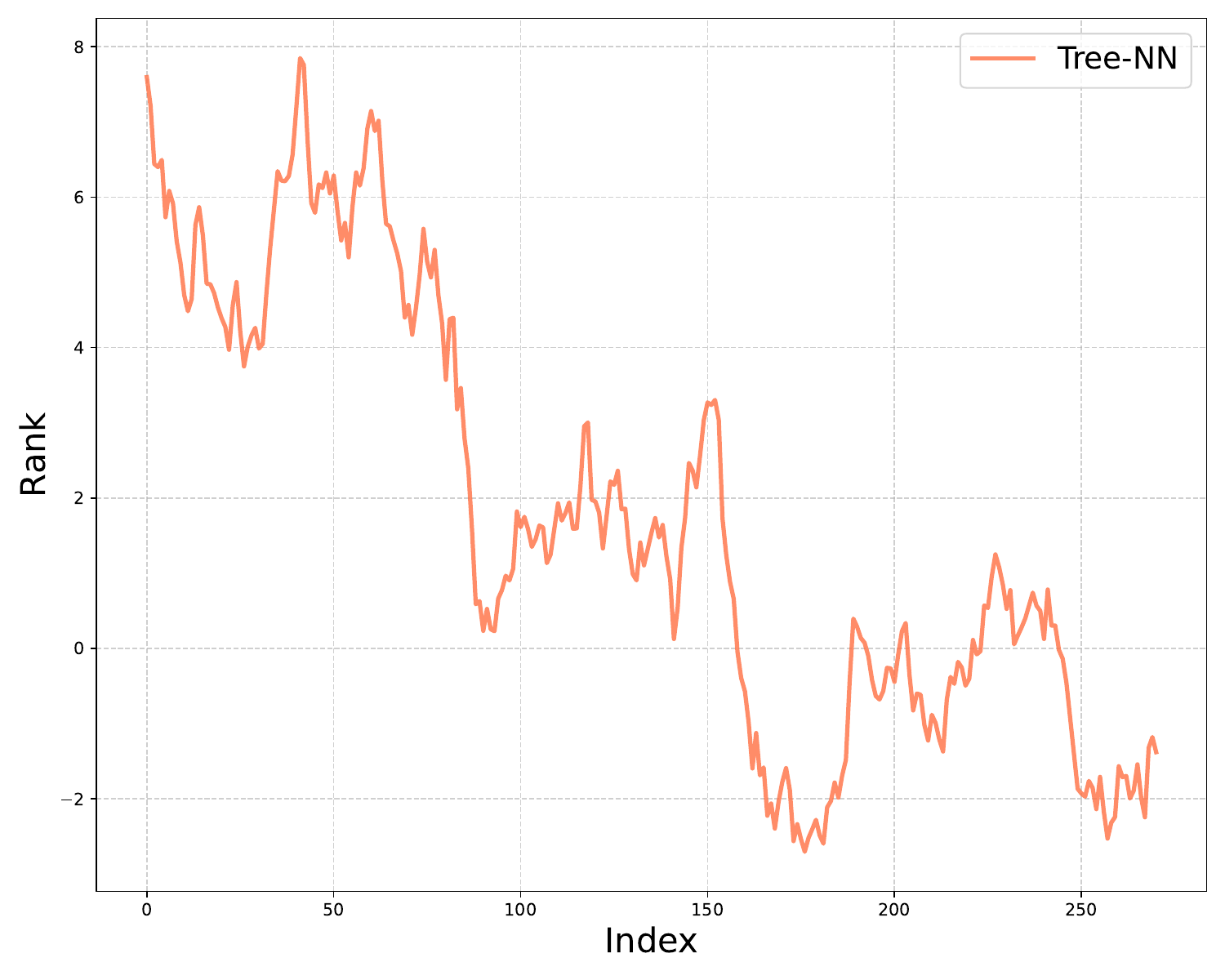}
    \centering
    {\scriptsize \mbox{(e) {Tree-DNN: \texttt{iq\_range\_std}}}}
    \end{minipage}
    \begin{minipage}{0.32\linewidth}
    \includegraphics[width=\textwidth]{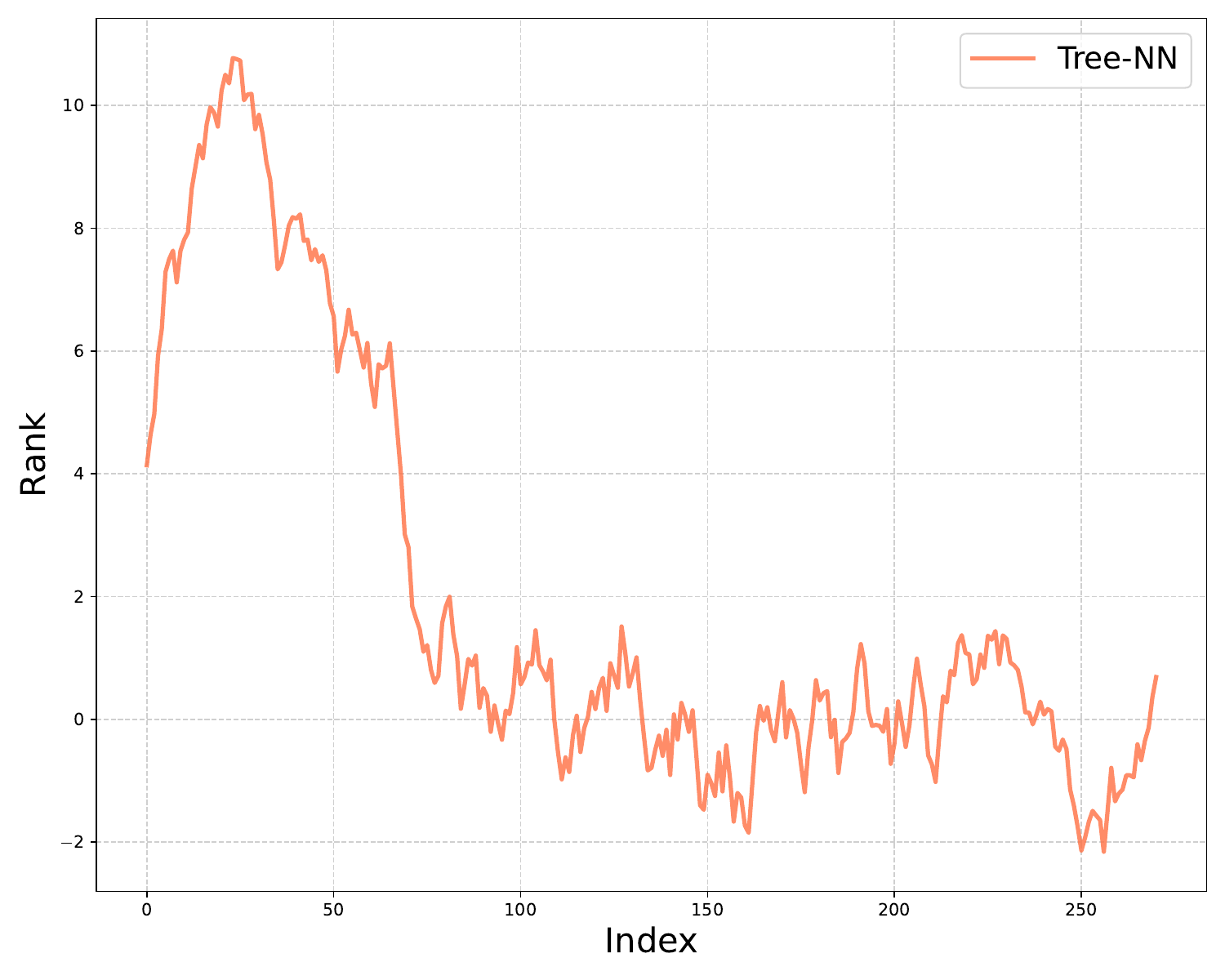}
    \centering
    {\scriptsize \mbox{(f) {Tree-DNN: \texttt{sparsity\_std}}}}
    \end{minipage}
  \caption{(a)-(b): The change of average rank from tree-based methods and DNN-based methods \wrt the most correlated meta-feature \(\texttt{entropy\_std}\), respectively. We further consider the differences in relative performance between tree-based models and DNN-based models, represented by the difference in average rank between the two types of methods. Plots in (c)-(f) highlight the changes of the tree-DNN gap against other meta-feature, such as \texttt{entropy\_std}, \texttt{entropy\_mean}, \texttt{sparsity\_std}, \texttt{iq\_range\_std}, and \texttt{inst\_to\_attr}. To enhance readability, the curves have been smoothed.}
  \label{fig:dynamics_meta_feature_difference}
\end{figure}

Results presented in~\autoref{tab:correlation_difference} establish a strong negative correlation between the Tree-DNN performance gap and Feature Heterogeneity metrics, exemplified by \(\texttt{entropy\_std}\) (Correlation: $-0.3333$). This observation contrasts with the meta-feature \(\texttt{inst\_to\_attr}\)---which has been emphasized in previous studies~\citep{McElfreshKVCRGW23when}---as it exhibits a markedly weaker correlation with the performance differential. These findings suggest that the degree of heterogeneity among dataset features is a more critical determinant driving the relative model superiority. Specifically, datasets characterized by greater feature heterogeneity (\eg, higher variability in feature \(\texttt{entropy}\) or \(\texttt{sparsity}\)) tend to confer an advantage to tree-based methods, likely attributable to their intrinsic ability to effectively handle diverse and non-uniform feature distributions through successive partitioning.

The visualizations in~\autoref{fig:dynamics_meta_feature_difference} provide empirical substantiation for these correlations. Figure~\ref{fig:dynamics_meta_feature_difference}(a) illustrates that as \(\texttt{entropy\_std}\) increases (along the index dimension), the average rank for tree-based methods consistently \textbf{decreases}, indicative of performance improvement. Crucially, Figure~\ref{fig:dynamics_meta_feature_difference}(c), which depicts the Tree-DNN gap against \(\texttt{entropy\_std}\), exhibits a clear negative slope (downward trend). Since the gap is defined as $\text{Rank}_{\text{Tree}} - \text{Rank}_{\text{DNN}}$, a consistently negative value confirms that $\text{Rank}_{\text{Tree}} < \text{Rank}_{\text{DNN}}$, thereby establishing that tree-based models achieve superior performance relative to neural networks as feature heterogeneity intensifies. Conversely, Figure~\ref{fig:dynamics_meta_feature_difference}(b) presents the change in DNN rank with respect to \(\texttt{inst\_to\_attr}\); its overall \textbf{ascending trend} signifies that performance degrades as $\texttt{inst\_to\_attr}$ increases, an outcome consistent with the negative correlation of $-0.3502$ reported in \autoref{tab:correlation_difference}.

\subsection{Performance Comparison across Different Dataset Feature Types} 
\begin{figure}[t] 
\centering 
\includegraphics[width=0.6\textwidth]{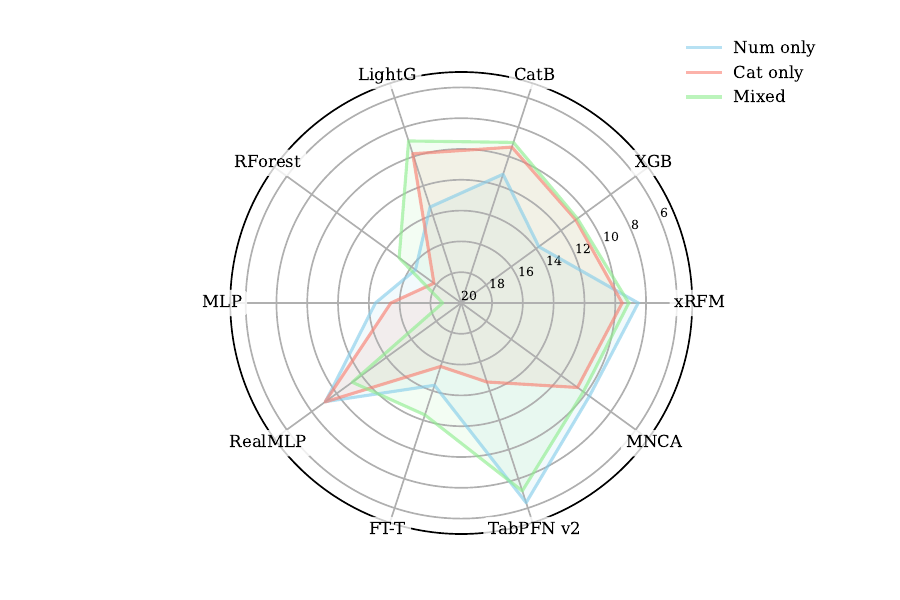}
\caption{Performance comparison of methods across datasets with purely numerical features, purely categorical features, and mixed feature types. Tree-based methods include XGBoost, LightGBM, CatBoost, and Random Forest. DNN-based methods include MLP, ResNet, and TabR. Token-based methods include FT-T, AutoInt, and ExcelFormer.} 
\label{fig:radar} 
\end{figure}

We validate the previous observations using real-world tabular datasets by comparing the performance of various methods across datasets with different feature types: purely numerical features (\textit{No Cat Data}), purely categorical features (\textit{No Num Data}), and mixed feature types (\textit{Mixed Data}). This analysis explores how dataset feature types influence performance, with particular emphasis on heterogeneity metrics such as \texttt{sparsity\_attr}\ or \texttt{entropy\_attr}. The results are visualized in a radar chart in~\autoref{fig:radar}. 

We have several key observations from~\autoref{fig:radar}. First, raw-feature-based DNN methods perform poorly on mixed feature datasets. DNN-based methods such as MLP and RealMLP exhibit their worst performance on datasets with mixed feature types (\textit{Mixed Data}). This observations validates the challenges faced by raw-feature-based neural networks when dealing with datasets characterized by significant heterogeneity in feature distributions. These methods perform better on purely numerical (\textit{No Cat Data}) or purely categorical (\textit{No Num Data}) datasets, where the homogeneity of feature types reduces the learning complexity. 

Tree-based methods like XGBoost, LightGBM, CatBoost, and Random Forest excel on heterogeneous datasets. The results underscore their inherent advantage in handling datasets with heterogeneous feature distributions, including a mix of numerical and categorical features. These methods effectively leverage feature splitting and hierarchical decision-making, making them robust to varying feature types. 

Token-based architectures, exemplified by the FT-Transformer (FT-T), demonstrate performance metrics that align more closely with tree-based models than traditional raw-feature-based DNN methods. This observation suggests that the learned embeddings for categorical and numerical features in FT-T enable a superior capability to manage the challenges inherent in mixed feature datasets. By effectively encoding and projecting features into a unified embedding space, the model likely mitigates the effects of feature variability and heterogeneity, thus facilitating better generalization across heterogeneous datasets. 

Additionally, analysis of the specialized model \textbf{TabPFN v2} reveals distinct performance dependencies on feature types. TabPFN v2 exhibits its strongest performance on datasets composed purely of \textbf{numerical features}, followed by mixed feature datasets, while demonstrating the poorest results on purely \textbf{categorical feature datasets}. This pattern is intrinsically linked to TabPFN v2's pretraining data generation methodology and its inherent strategy for handling categorical features, which typically involves treating them as numerical data after simple preprocessing (such as one-hot or ordinal encoding). Consequently, the development of robust and generalizable strategies for handling \textbf{categorical features} remains a critical challenge for future research in the design of high-performing, universal tabular models.

\section{Lightweight and Stress-Test via {\scshape Talent}-Tiny and {\scshape Talent}-Extension}\label{sec:tiny-bench}
Although the proposed large benchmark facilitates the analysis of deep tabular models, running a single method on all the datasets incurs a high computational burden. In this section, we extract a subset of the benchmark containing 15\% of the full benchmark, \ie, 45 datasets, to enable more efficient tabular analysis. We also collect an extension set with challenging tabular datasets for stress testing. The statistics of all datasets are listed in~\autoref{tab:dataset_basic}.

\subsection{{\scshape Talent}-Tiny: A Compact Benchmark for Efficient and Detailed Evaluations}\label{sec:talent_tiny}

\noindent{\bf Selection strategy}.
As mentioned in~\autoref{sec:benchmark}, {\name} is designed with a large collection of tabular datasets covering diverse characteristics. To curate {\name}-tiny, we apply stricter rules to remove datasets from other modalities, those with inherent distribution shifts or known leakage, and duplicated variants. To ensure representativeness, we also consider the ``evolved'' tree–DNN debate (see~\autoref{sec:DNN_TREE_debate}), selecting datasets where both tree-based and DNN-based methods exhibit diverse behaviors.

We base this selection on the Tree–DNN score (\autoref{eq:tree_dnn_score}), which quantifies a dataset’s preference for representative tree-based versus DNN-based methods. We categorize datasets into three groups: tree-friendly, DNN-friendly, and tie. For each task type (binary classification, multi-class classification, and regression), we partition datasets into groups by size ($N \times d$) to ensure small, medium, and large problems are all represented. From each size group, we select one dataset from each Tree--DNN category (tree, DNN, and tie). When a group has multiple candidates, we prefer datasets with clearer signal-to-noise ratios, higher-quality metadata, and balanced categorical vs. numerical feature compositions.  

To further promote diversity, we refine the pool by enforcing representation across 14 application domains (\eg, \texttt{biology}, \texttt{finance}, and \texttt{healthcare}). In cases where two datasets are similar, we substitute with an alternative to avoid redundancy. This strategy results in 45 datasets: including 15 binary classification, 12 multi-class classification, and 18 regression tasks. The final subset balances dataset size, feature type, domain, and method preference, ensuring {\name}-tiny is compact yet representative for controlled, efficient evaluations.

\begin{figure}[t]
  \centering
  % Binary Classification
  \begin{minipage}{0.45\linewidth}
    \includegraphics[width=\textwidth]{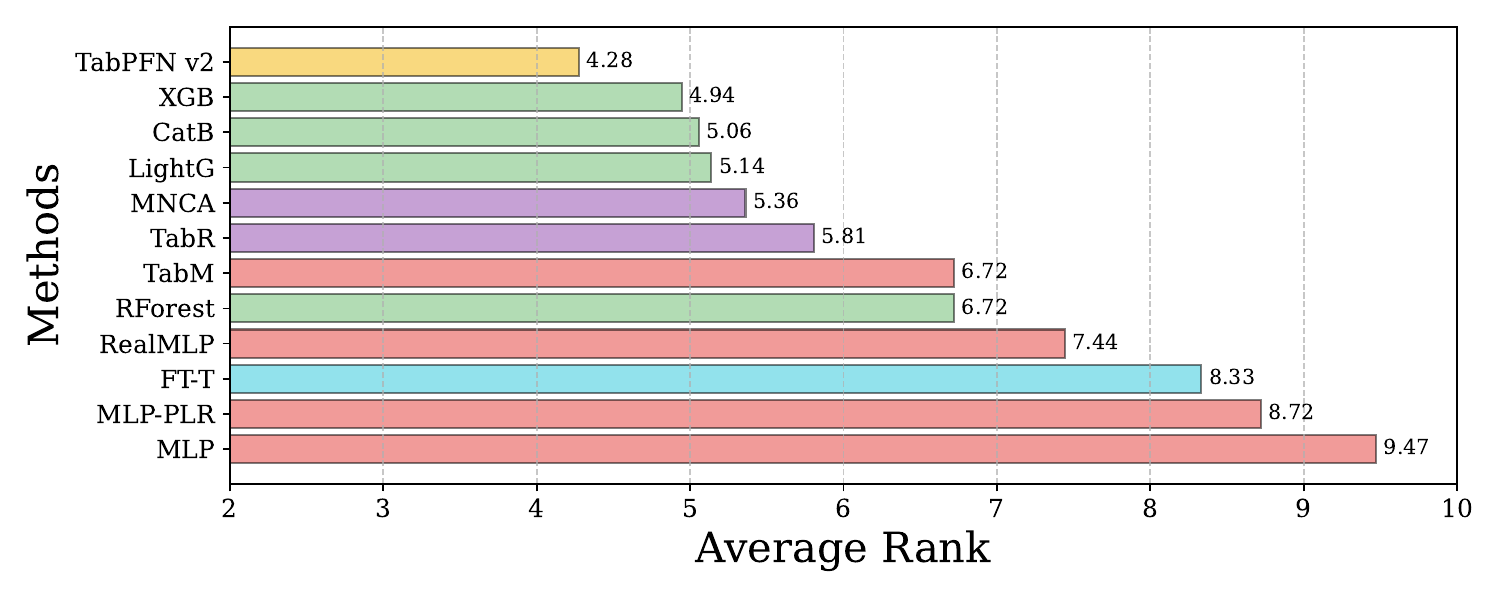}
    \centering
    {\small \mbox{(a1) Binary Classification (Hold-out)}}
  \end{minipage}
  \hfill
  \begin{minipage}{0.45\linewidth}
    \includegraphics[width=\textwidth]{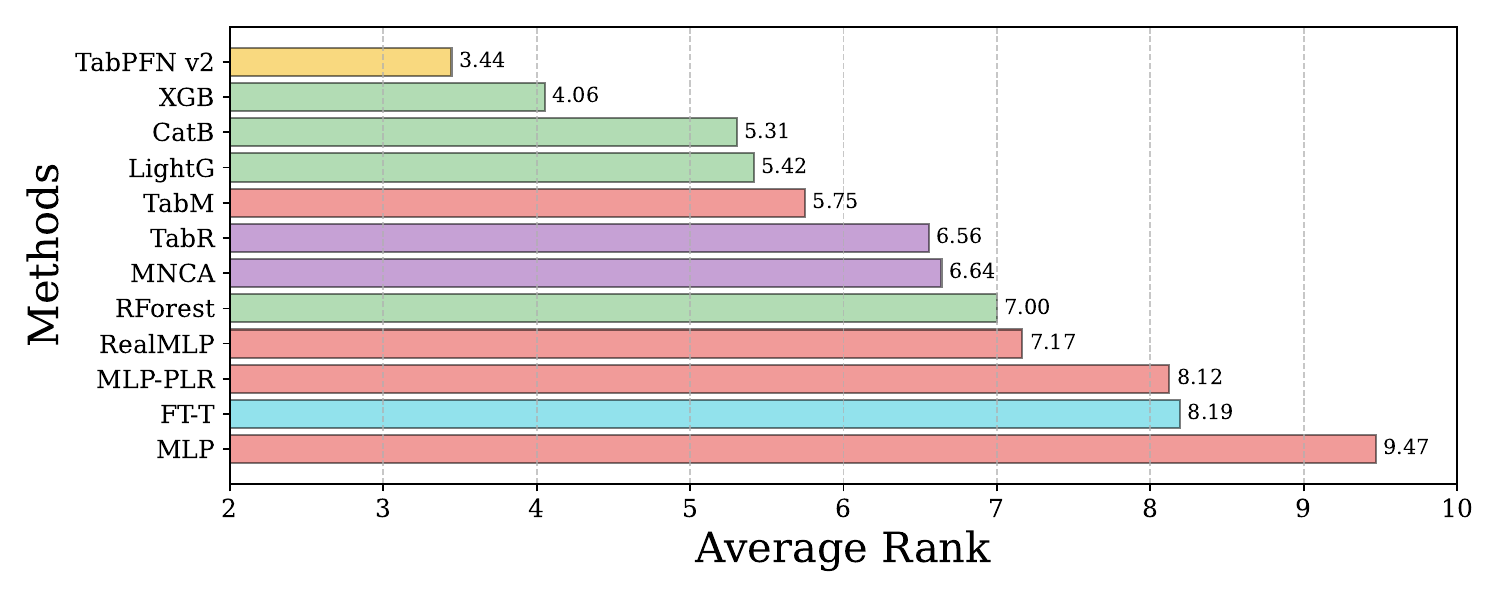}
    \centering
    {\small \mbox{(a2) Binary Classification (CV + Ensemble)}}
  \end{minipage}

  % Multi-Class Classification
  \begin{minipage}{0.45\linewidth}
    \includegraphics[width=\textwidth]{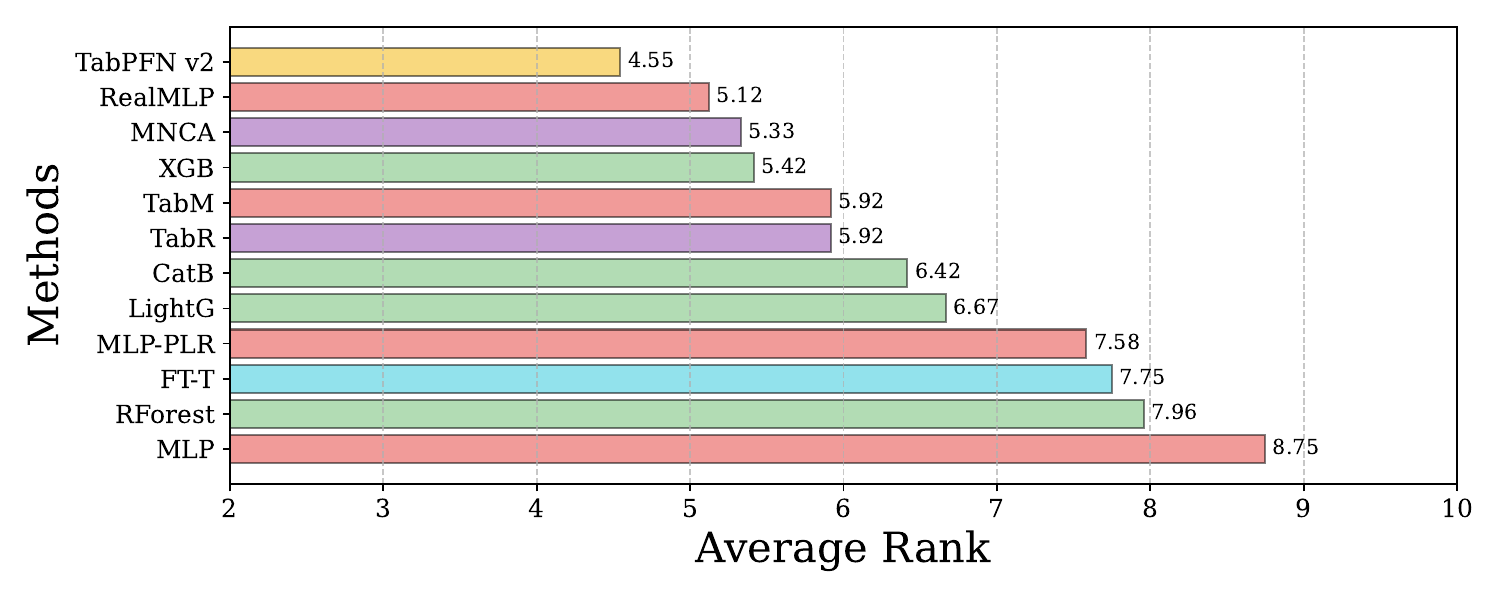}
    \centering
    {\small \mbox{(b1) Multi-Class Classification (Hold-out)}}
  \end{minipage}
    \hfill
  \begin{minipage}{0.45\linewidth}
    \includegraphics[width=\textwidth]{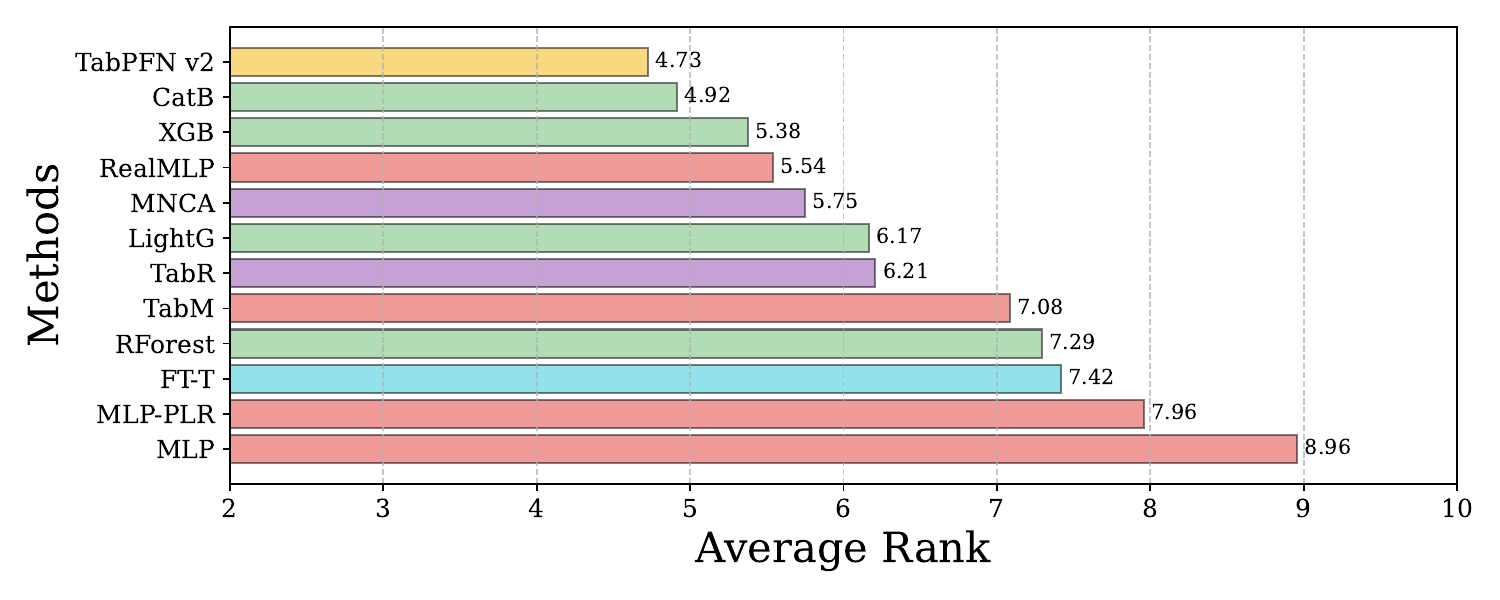}
    \centering
    {\small \mbox{(b2) Multi-Class Classification (CV + Ensemble)}}
  \end{minipage}

  % Regression
  \begin{minipage}{0.45\linewidth}
    \includegraphics[width=\textwidth]{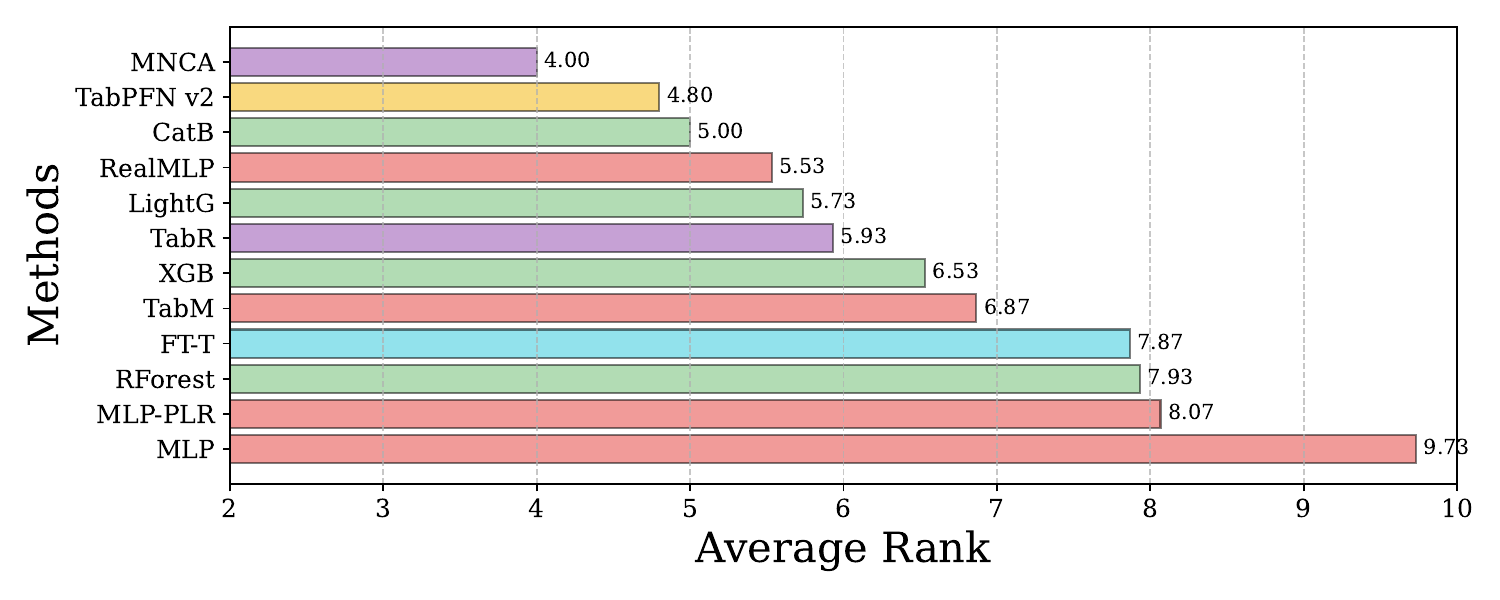}
    \centering
    {\small \mbox{(c1) Regression (Hold-out)}}
  \end{minipage}
    \hfill
  \begin{minipage}{0.45\linewidth}
    \includegraphics[width=\textwidth]{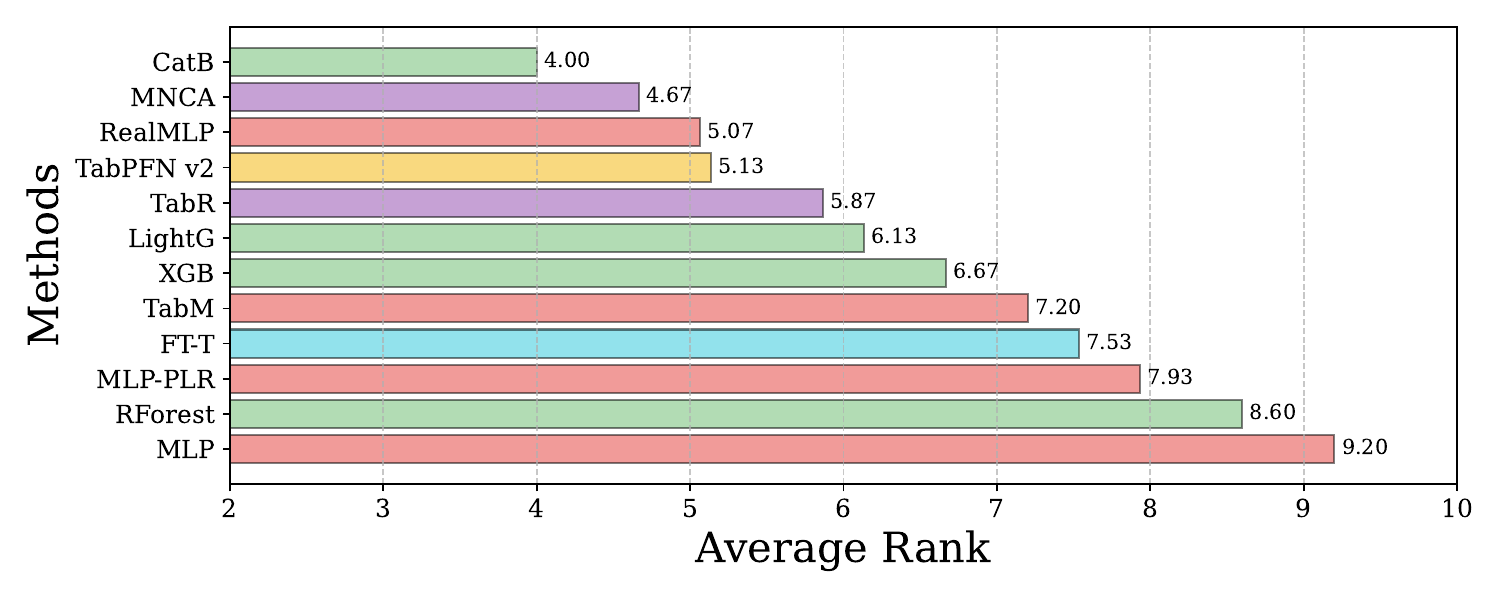}
    \centering
    {\small \mbox{(c2) Regression (CV + Ensemble)}}
  \end{minipage}

  % All Tasks
  \begin{minipage}{0.45\linewidth}
    \includegraphics[width=\textwidth]{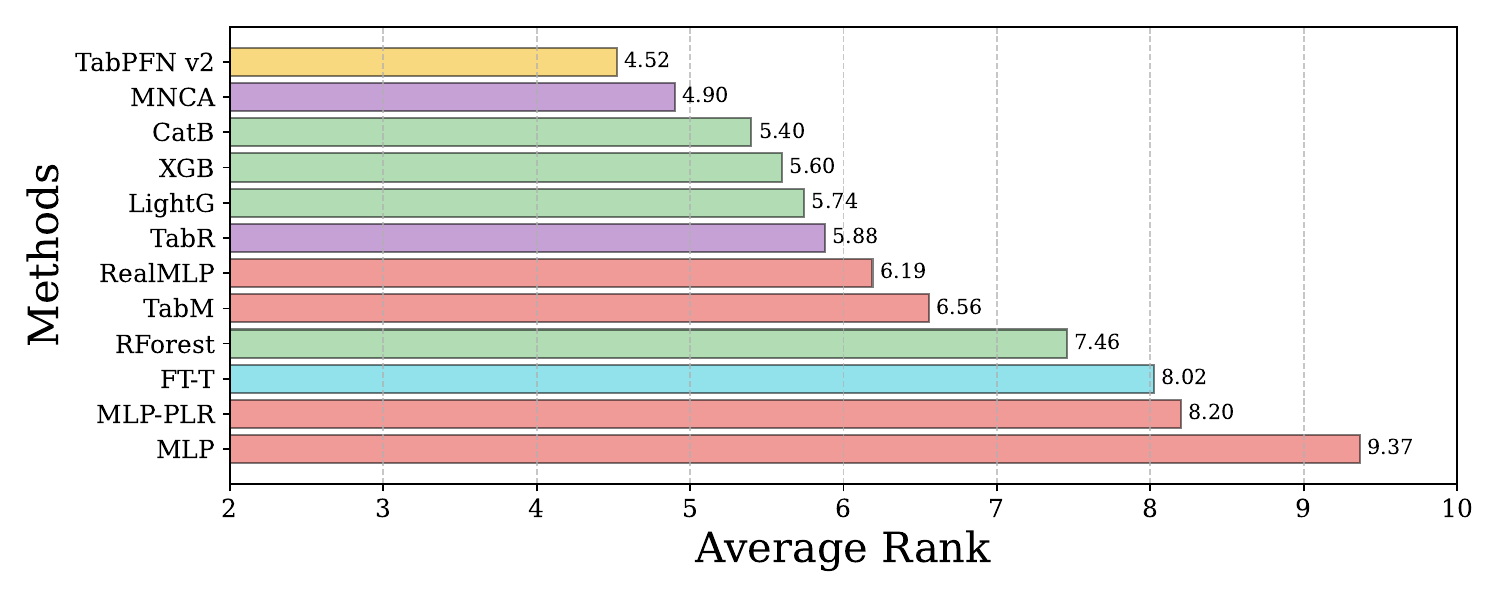}
    \centering
    {\small \mbox{(d1) All Tasks (Hold-out)}}
  \end{minipage}
    \hfill
  \begin{minipage}{0.45\linewidth}
    \includegraphics[width=\textwidth]{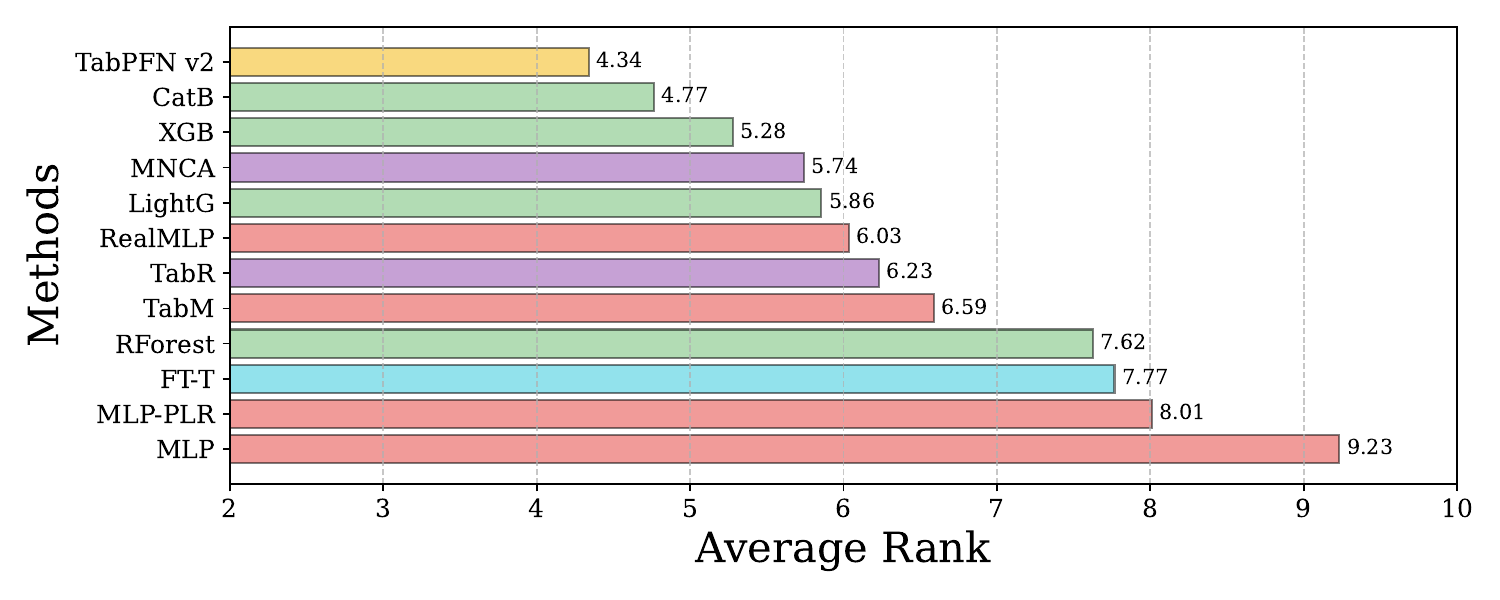}
    \centering
    {\small \mbox{(d2) All Tasks (CV + Ensemble)}}
  \end{minipage}
  \caption{Average rank of representative tabular methods on {\name}-tiny under two evaluation protocols: 
  (left) the original hold-out strategy and (right) cross-validation with ensemble. 
  Ranks are computed based on accuracy for classification tasks and RMSE for regression tasks. 
  Lower rank values indicate stronger performance. While CV+ensemble generally improves absolute performance values, the relative ordering among methods remains stable, validating the practicality of the hold-out strategy for large-scale benchmarks. 
  }
  \label{fig:average_rank_tiny_combined}
\end{figure}

\noindent{\bf Analysis on the necessity for cross-validation}.
Recent studies argue that the hold-out strategy may cause hyperparameters to overfit the validation set~\citep{Tschalzev2024DataCentric,Erickson2025TabArena}, while cross-validation (CV) with ensembling provides more stable evaluation. However, CV greatly increases tuning costs, making it infeasible across 300 datasets in {\name}. Here, we compare both strategies on {\name}-tiny.

\autoref{fig:average_rank_tiny_combined} (left) reports results with the hold-out strategy. The ranking patterns reflect those of the full benchmark, with tree-based ensembles (\eg, CatBoost, LightGBM, XGBoost) showing particularly strong performance in binary classification, and RealMLP and MNCA performing well in multi-class and regression. Importantly, the relative order of methods aligns with large-scale results, indicating that the subset remains representative.

\autoref{fig:average_rank_tiny_combined} (right) shows results with cross-validation plus ensembling. Across most methods, average ranks improve slightly compared to hold-out, confirming that ensembling boosts stability. However, the relative order among methods remains largely unchanged. For example, tree-based methods still dominate in binary classification, while RealMLP and MNCA maintain strong performance in regression and multi-class classification. This validates the use of hold-out in {\name}, given the impractical cost of CV across all datasets.

Interestingly, MNCA does not benefit much from vanilla ensembling, in contrast to the clear gains observed with its dedicated ensemble variant (MNCA-ens) in earlier results. This suggests that some methods require specialized ensemble designs—for MNCA, strategies that increase neighborhood diversity may be particularly important. In contrast, gradient boosting methods and RealMLP benefit naturally from CV+ensemble, showing reduced variance without needing customized ensemble mechanisms.

Overall, {\name}-tiny proves useful for efficient yet representative analysis. The comparison of hold-out versus CV+ensemble indicates that while ensembling stabilizes results, the hold-out strategy provides reliable relative rankings across methods, justifying its adoption in {\name}. At the same time, the divergent behaviors of methods like MNCA highlight that ensemble strategies must sometimes be tailored to model design rather than applied uniformly.

% \begin{table}[t]
% \centering
% \caption{Average rank of MLP with different hyper-parameter search budgets across dataset subsets. Lower is better. HP$k$ denotes $k$ hyper-parameter trials.}
% \label{tab:hpo_trials}
% \begin{tabular}{lccccccc}
% \toprule
% Sample & 1--50 & 51--100 & 101--150 & 151--200 & 201--250 & 251--300 & Overall \\
% \midrule
% MLP-HP50  & 15.63 & 14.86 & 16.76 & 16.81 & 15.67 & 14.96 & 15.78 \\
% MLP-HP100 & 15.36 & 14.63 & 16.52 & 16.49 & 15.62 & 14.24 & 15.52 \\
% MLP-HP150 & 15.26 & 14.93 & 16.49 & 16.19 & 15.24 & 14.05 & 15.37 \\
% MLP-HP200 & 15.41 & 14.75 & 16.85 & 16.18 & 15.26 & 14.24 & 15.45 \\
% \bottomrule
% \end{tabular}
% \end{table}
\begin{table}[t]
\centering
\caption{Performance of MLP models under different hyperparameter tuning trials, measured by average rank across all methods. 
The six groups correspond to subsets divided by dataset sample size, while the last column reports the overall performance. 
Parentheses indicate $p$-values. 
Results show that compared with our standard setting of 100 tuning trials, 50 trials are insufficient for effective tuning, 
whereas increasing the number of trials beyond 100 does not yield further improvement.}
\label{tab:hpo_trials}
\setlength{\tabcolsep}{2.7pt}
{\footnotesize{
\begin{tabular}{cccccccc}
\toprule
 & Group 1 & Group 2 & Group 3 & Group 4 & Group 5 & Group 6 & Overall \\
\midrule
50 trials & +0.63  {\tiny(0.144)} & +0.30  {\tiny(0.147)} & +0.48 {\tiny(0.547)} & +0.64 {\tiny(0.214)} & +0.11 {\tiny(0.919)} & +1.53* {\tiny(0.003)} & +0.61* {\tiny(0.002)} \\
100 trials & 17.54 & 17.10 & 18.57 & 18.32 & 17.36 & 15.94 & 17.47 \\
150 trials & +0.07 {\tiny(0.828)} & $-$0.03 {\tiny(0.879)} & +0.07 {\tiny(0.958)} & $-$0.48 {\tiny(0.522)} & $-$0.54 {\tiny(0.211)} & $-$0.40 {\tiny(0.303)} & $-$0.22 {\tiny(0.262)} \\
200 trials & +0.20 {\tiny(0.750)} & $-$0.29 {\tiny(0.965)} & +0.45 {\tiny(0.547)} & $-$0.65 {\tiny(0.298)} & $-$0.54 {\tiny(0.324)} & 0.00 {\tiny(0.845)} & $-$0.14 {\tiny(0.602)} \\
\bottomrule
\end{tabular}}}
\end{table}

\noindent{\bf Analysis on the number of hyperparameter search.}
We further investigate the influence of the number of hyperparameter search trials, which was set to 100 in our previous experiments following~\cite{GorishniyRKB21Revisiting}. To assess the sensitivity of performance to search effort, we evaluate the MLP method with 50, 100, 150, and 200 trials, and compare their average ranks across subsets of datasets of different sizes.

The results, shown in~\autoref{tab:hpo_trials}, reveal several important trends. On smaller datasets, increasing the number of trials beyond 50 yields diminishing returns: the gap between 50 and 100 trials is noticeable, but further increases to 150 or 200 trials do not lead to consistent improvements. On larger datasets, additional search efforts beyond 100 trials bring marginal but still limited gains. This suggests that 50 trials are insufficient for stable optimization, but 100 trials already provide a near-saturation point for tuning effectiveness. 

Across all dataset groups, the overall differences among 100, 150, and 200 trials are relatively minor, as confirmed by the comparative rankings in the rightmost panel. The performance curves of these three settings are nearly overlapping, indicating that the benefit of exhaustive hyperparameter search is minimal once a certain search budget is reached. Interestingly, while 50 trials consistently rank lower, the rank order of 100, 150, and 200 trials fluctuates slightly across dataset subsets without forming a clear hierarchy. 

These results support the use of 100 trials as a balanced and practical setting in large-scale benchmarks like {\name}, since it offers a strong trade-off between computational efficiency and model competitiveness. Moreover, they highlight that blindly scaling up search budgets does not guarantee better results, especially for tabular tasks where model robustness may dominate over hyperparameter fine-tuning.

\subsection{{\scshape Talent}-Extension: Stress Testing in Challenging Scenarios}
\label{sec:talent_extension}

{\name} provides two complementary layers of coverage.
While the main {\name} benchmark provides broad coverage of typical tabular tasks, real-world data often exhibit more extreme conditions that challenge scalability and model robustness. To explore these regimes, we introduce {\name}-extension, a complementary suite designed to stress-test tabular methods under three specialized yet practically important settings: \emph{high-dimensional feature spaces}, \emph{many-class classification problems}, and \emph{very-large-scale datasets}. These settings expose performance bottlenecks that are not always visible in standard-sized datasets and thus provide a deeper understanding of each model's inductive biases.

\begin{figure}[tb]
  \centering
    \begin{minipage}[t]{0.45\linewidth}
      \includegraphics[width=\textwidth]{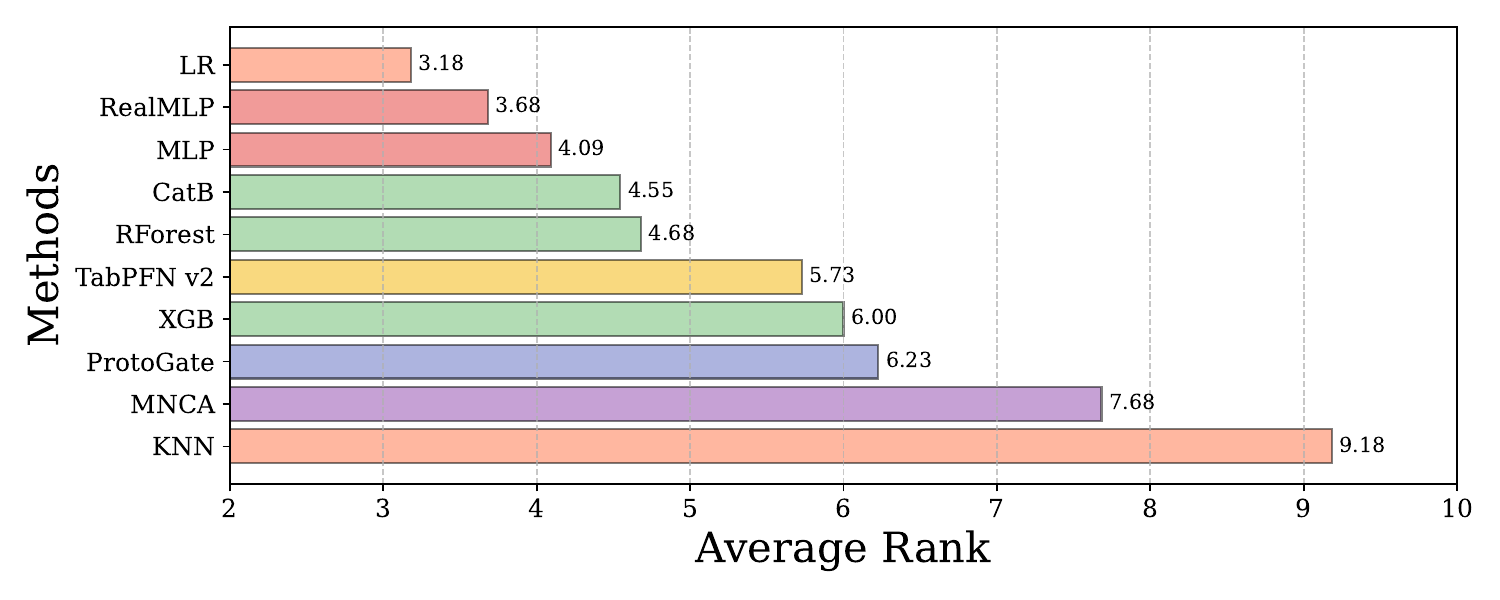}
      \centering
      {\small \mbox{(a) {High-Dimensional: Binary Classification}}}
      
      \includegraphics[width=\textwidth]{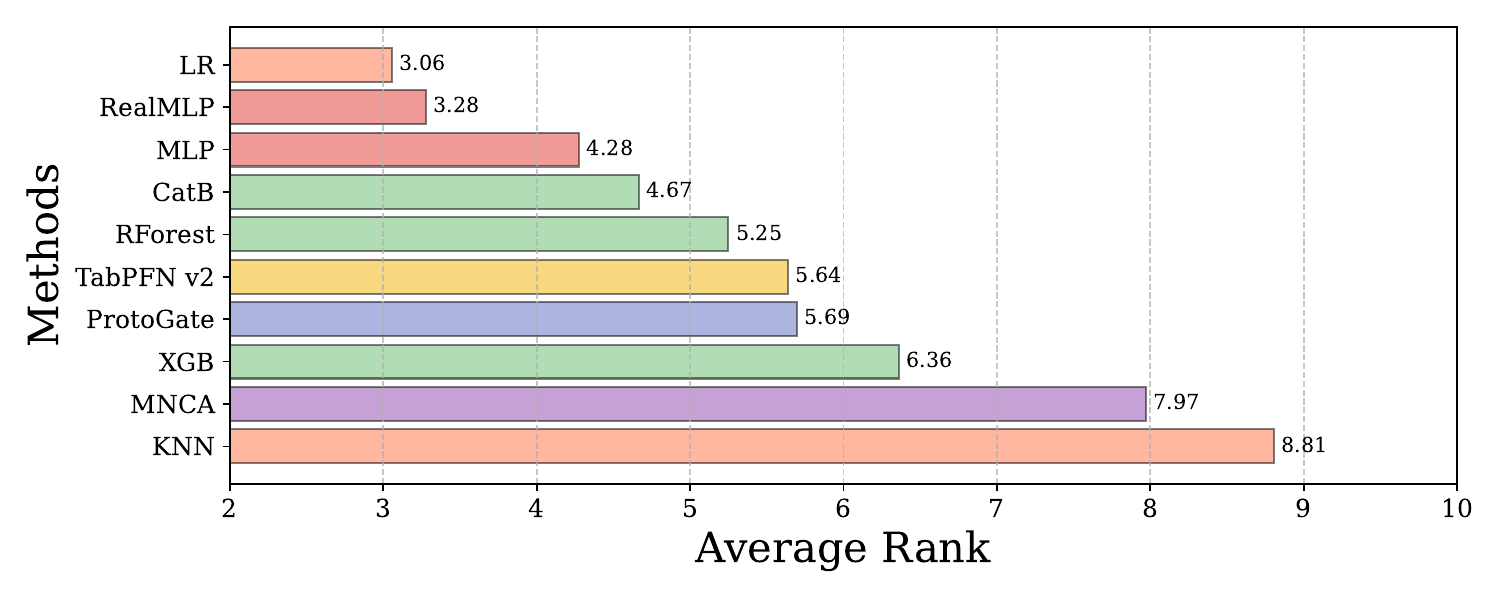}
      \centering
      {\small \mbox{(c) {High-Dimensional: All Tasks}}}
    \end{minipage}
    \begin{minipage}[t]{0.45\linewidth}
      \includegraphics[width=\textwidth]{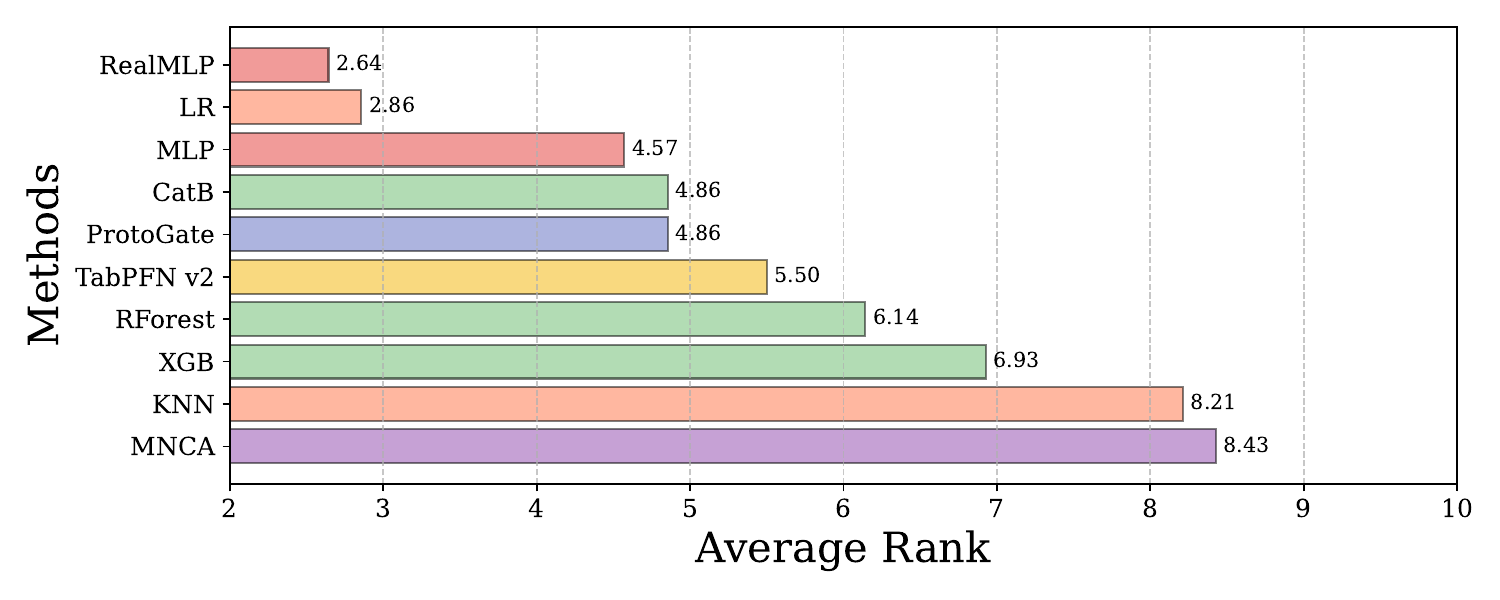}
      \centering
      {\small \mbox{(b) {High-Dimensional: Multi-Class Classification}}}
      
      \includegraphics[width=\textwidth]{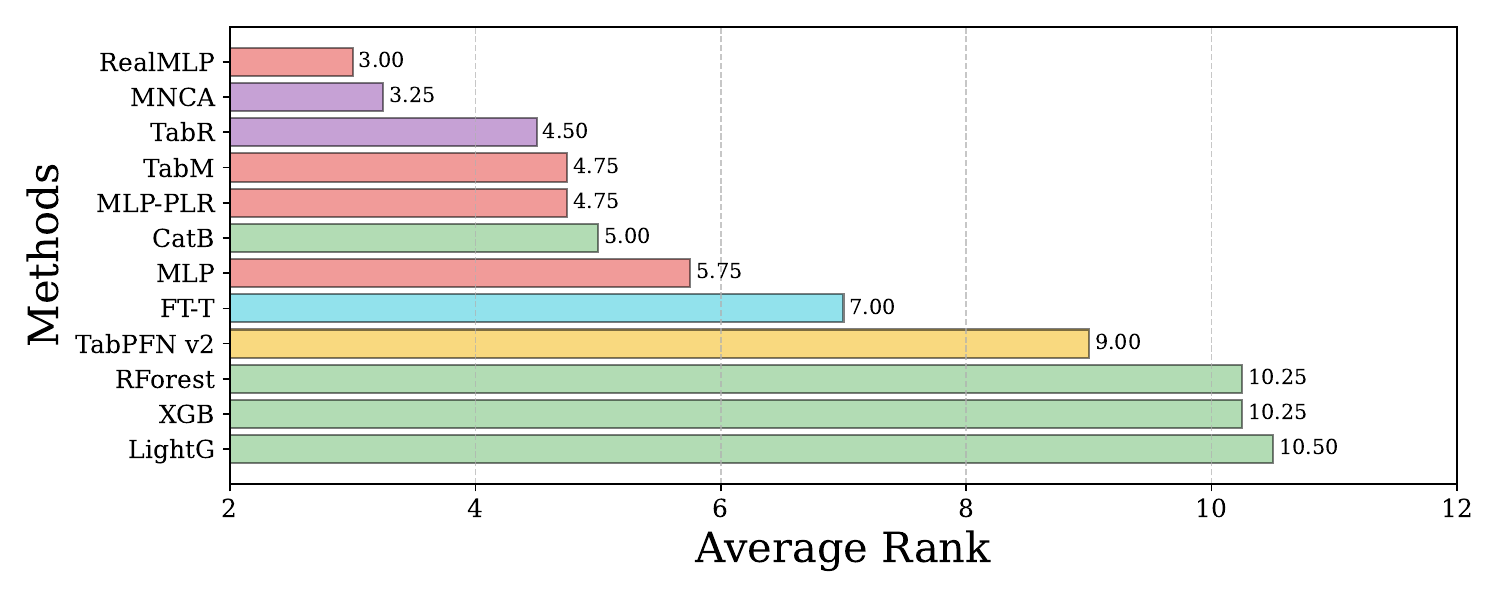}
      \centering
      {\small \mbox{(d) {Many-Class: All Tasks}}}
    \end{minipage}
  \caption{
  Average rank of representative methods on {\name}-extension under high-dimensional and many-class conditions. 
  Subfigures (a)--(c) correspond to high-dimensional datasets, while (d) summarizes the results for many-class classification.
  Ranks are computed using accuracy for classification and RMSE for regression (lower is better).}
  \label{fig:average_rank_hdim}
\end{figure}

\noindent{\bf Dataset groups.} {\name}-extension contains three groups of specialized tabular tasks.
\begin{itemize}[noitemsep,topsep=0pt,leftmargin=*]
    \item \textbf{High-dimensional datasets.} This group contains 18 datasets with feature dimensionality ranging from 2{,}000 to over 20{,}000 (Table~\ref{tab:dataset_basic}). Representative examples include biomedical datasets such as \texttt{colon}, \texttt{glioma}, and \texttt{TOX\_171}, as well as text-derived datasets like \texttt{BASEHOCK} and \texttt{RELATHE}.
    \item \textbf{Many-class datasets.} This group includes 12 datasets with more than ten classes, such as \texttt{orlraws10P} (10 classes, 10{,}304 features) and \texttt{Fashion-MNIST} (10 classes, 784 features). These datasets emphasize the difficulty of learning fine-grained label structures where class-aware objectives and embeddings are critical.
    \item \textbf{Very large-scale datasets.} This group comprises 18 datasets containing hundreds of thousands to millions of instances, such as \texttt{Airlines\_DepDelay} (10M samples), \texttt{Higgs} (1M samples), and \texttt{sf-police-incidents} (2.2M samples). They test the computational scalability of tabular methods under massive data volumes.
\end{itemize}

The evaluation protocol follows the same setup as in {\name}, except for high-dimensional datasets, where limited sample sizes motivate aggregated cross-validation results with default hyperparameters. We evaluate representative methods from classical baselines (Logistic Regression, kNN), tree ensembles (Random Forest, XGBoost, LightGBM, CatBoost), and deep tabular architectures (MLP, RealMLP, FT-Transformer, MNCA, TabM). Some other methods, such as MNCA-ens and TabM, require significantly longer training times in these specialized scenarios, so we omit them from the extended evaluation.

\begin{figure}[t]
  \centering
   \begin{minipage}{0.45\linewidth}
    \includegraphics[width=\textwidth]{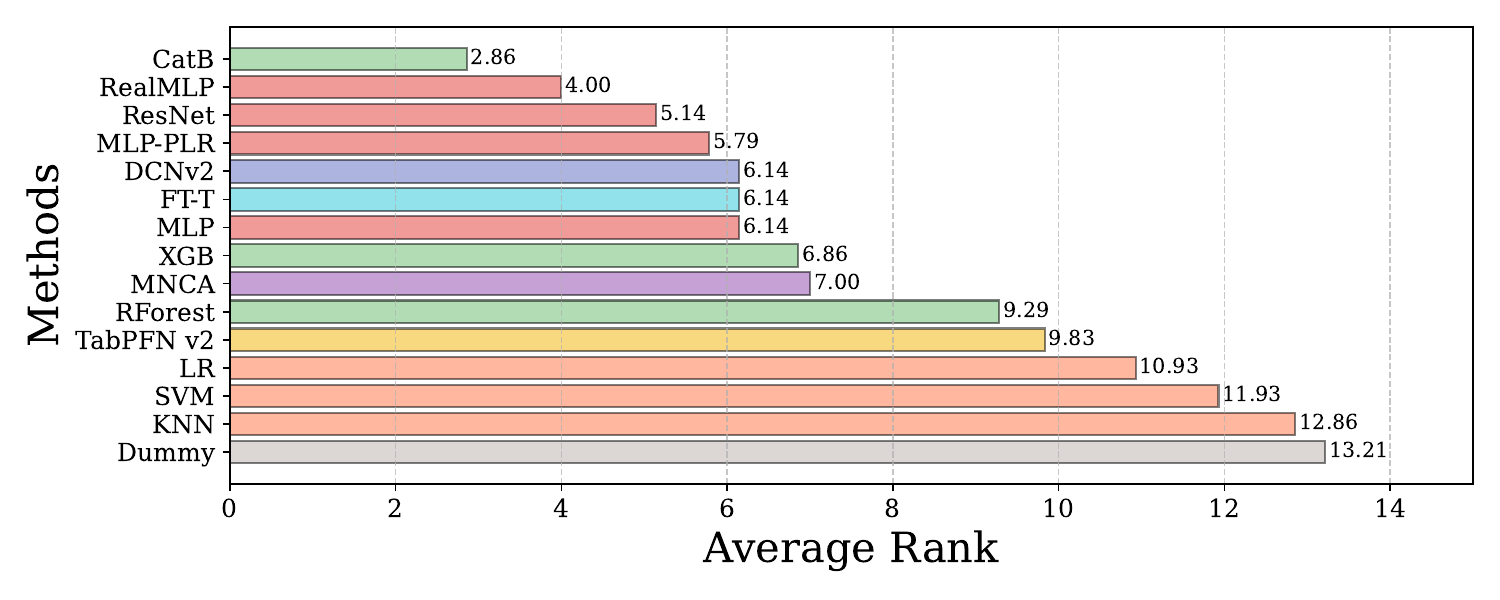}
    \centering
    {\small \mbox{(a) {Binary Classification}}}
    \end{minipage}
    \begin{minipage}{0.45\linewidth}
    \includegraphics[width=\textwidth]{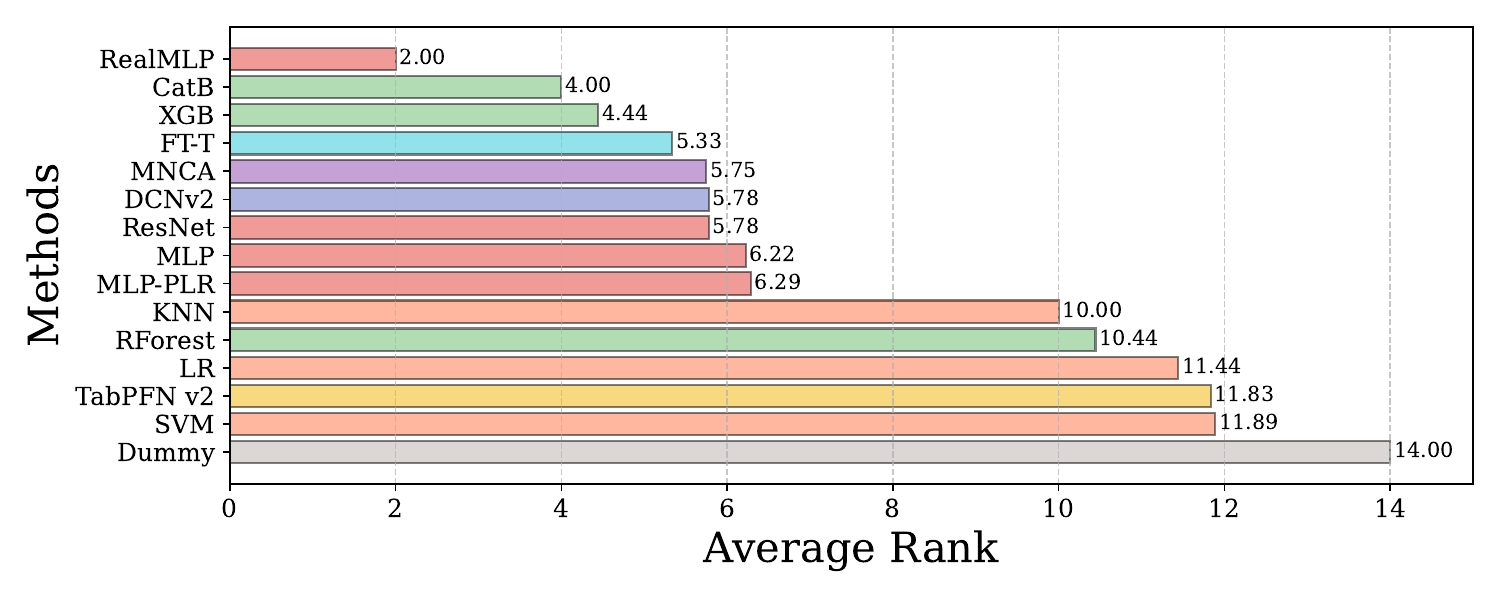}
    \centering
    {\small \mbox{(b) {Multi-Class Classification}}}
    \end{minipage}
    
    \begin{minipage}{0.45\linewidth}
    \includegraphics[width=\textwidth]{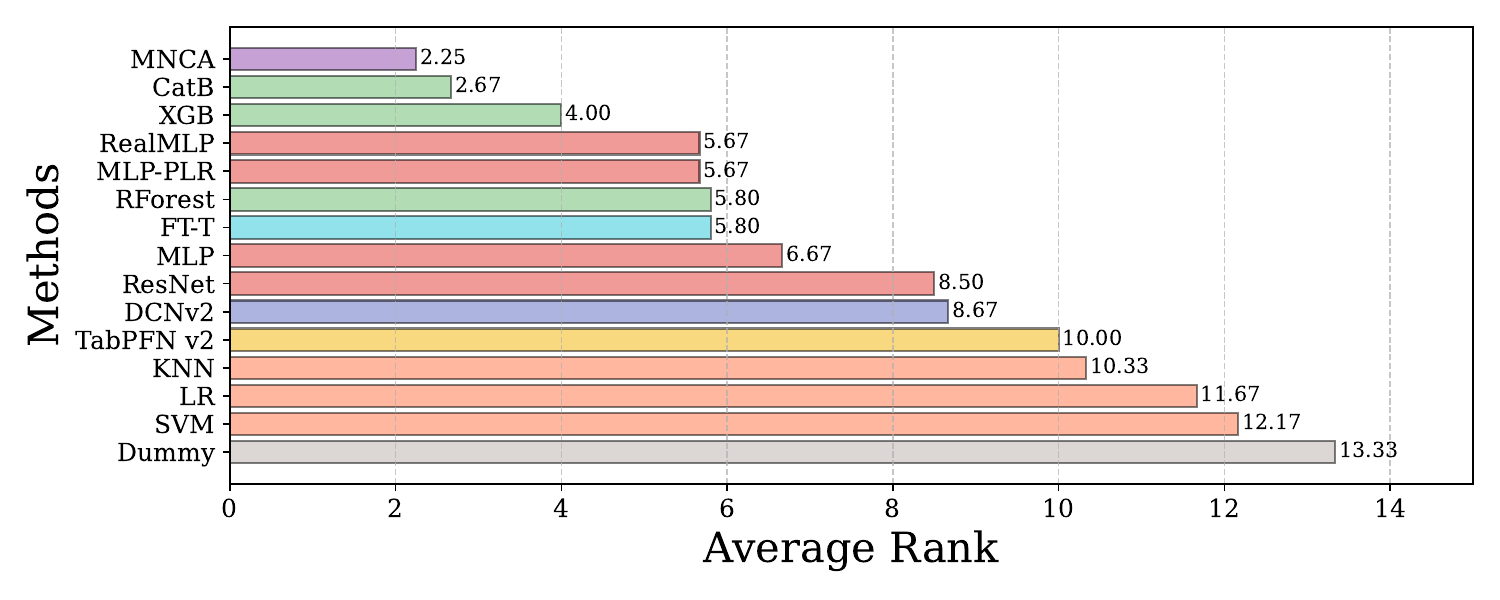}
    \centering
    {\small \mbox{(c) {Regression}}}
    \end{minipage}
    \begin{minipage}{0.45\linewidth}
    \includegraphics[width=\textwidth]{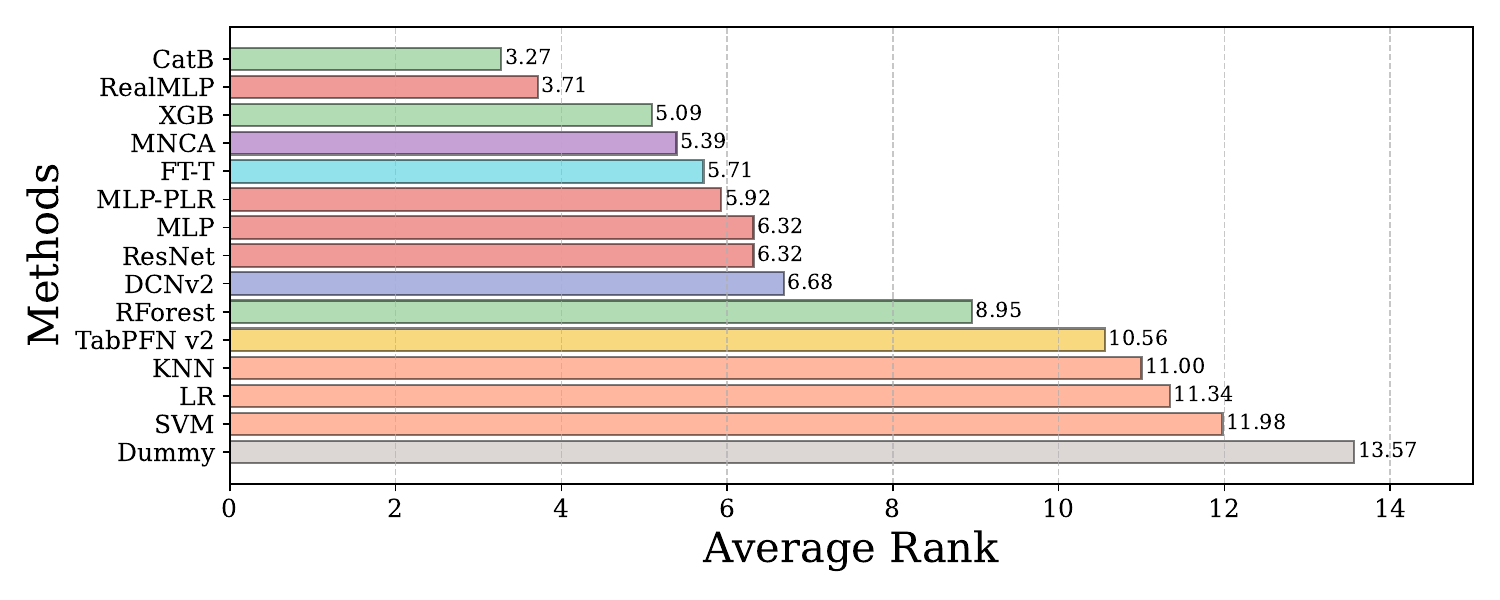}
    \centering
    {\small \mbox{(d) {All Tasks}}}
    \end{minipage}
  \caption{Average rank of representative methods on {\scshape Talent}-extension (very large-scale datasets). 
  Ranks are computed based on accuracy and RMSE for classification and regression tasks, respectively.
  Lower rank values indicate better overall performance.
  }
  \label{fig:average_rank_large}
\end{figure}

\noindent{\bf Results and analysis.}  
The {\name}-extension results reveal several notable shifts relative to the main {\name} benchmark:

\begin{itemize}[noitemsep,topsep=0pt,leftmargin=*]
  \item \textbf{High-dimensional datasets.} On high-dimensional datasets, the results (\autoref{fig:average_rank_hdim}a–c) reveal several notable deviations from the main {\scshape Talent} findings. Surprisingly, Logistic Regression emerges as one of the strongest performers, consistently ranking near the top across binary, multi-class, and aggregated tasks. This indicates that in ultra-sparse, high-dimensional spaces—common in biomedical and text-derived data—simpler linear models with regularization can generalize better than complex nonlinear architectures. 
In contrast, methods like ModernNCA and KNN, which rely on neighborhood retrieval, suffer significant degradation. Their performance drop reflects the \emph{curse of dimensionality}, where distance-based similarity becomes less meaningful as dimensionality grows. 
Tree-based ensembles (\eg, CatBoost, LightGBM, XGBoost) remain stable but are not dominant; their greedy feature-splitting mechanisms struggle when the number of irrelevant or redundant features is large. Interestingly, RealMLP and standard MLPs show resilience—likely due to their ability to perform distributed feature selection through gradient-based learning—but still lag slightly behind linear baselines in some cases. Pretrained foundation models such as TabPFN v2 perform poorly in this setting, suggesting that pretrained priors, while powerful for small to mid-size datasets, transfer poorly when the feature distributions of target data differ drastically from those seen during pretraining. This performance gap points to a distribution mismatch between synthetic pretraining corpora and real high-dimensional domains, where feature semantics differ and sample sparsity limits adaptation.
  
  \item \textbf{Many-class datasets.} In the many-class regime (\autoref{fig:average_rank_hdim}d), the trends diverge from the high-dimensional case. Deep models with strong \emph{representation learning} capacity regain their lead: RealMLP, ModernNCA, and TabR rank at the top, followed closely by TabM and MLP-PLR. Their ability to learn shared embeddings across fine-grained label spaces proves crucial for distinguishing numerous closely related classes. Tree-based ensembles such as CatBoost and XGBoost remain competitive but no longer dominant, suggesting that their partition-based decision structures may not scale efficiently with large label cardinalities. Interestingly, while foundation models (\eg, TabPFN v2) excelled in the main {\name} benchmark, they perform inconsistently here, further underscoring the limitation of in-context pretrained inference when label structures diverge from the few-class distributions prevalent during pretraining. Ensemble-style deep models such as TabM consistently outperform their base DNNs, reaffirming that ensembling remains an effective strategy even in modern deep tabular learning.
  
  \item \textbf{Large-scale datasets.} As shown in \autoref{fig:average_rank_large}, very large datasets (up to millions of rows) produce a performance landscape that differs from the main {\name} results. Classical tree ensembles, especially CatBoost, rank at or near the top across tasks. Beyond their inductive bias for categorical structure, a practical factor likely contributes: highly optimized implementations make it feasible to train and \emph{ensemble many trees} at scale, which compounds accuracy. By contrast, ensembling deep tabular models is far more time-consuming (multiple large models must be trained), so neural methods seldom benefit from the same degree of ensemble amplification under strict compute budgets. 
Modern deep methods still perform well: RealMLP is consistently strong, indicating that well–regularized MLPs scale gracefully. Retrieval/attention models (\eg, MNCA, FT-T) remain competitive but do not close the gap to the best tree ensembles. Pretrained foundation models (\eg, TabPFN v2) lag in this regime, likely due to a distribution/scale mismatch with their pretraining setup and the lack of fine-tuning for massive datasets. 
\end{itemize}

\noindent{\bf Broader observations.}  
Taken together, the {\name}-extension results show both continuity and clear departures from the main {\name} findings. Several methods, such as RealMLP, retain their strengths across regimes, yet the ordering of most methods changes once we move to high dimensionality, many classes, or very large scale. In high‐dimensional problems, simple linear models (\eg, logistic regression) are unexpectedly competitive, and neighborhood/retrieval methods degrade, indicating that feature redundancy and the curse of dimensionality, rather than model expressivity, become the primary bottlenecks. For very large datasets, tree ensembles—most notably CatBoost—regain a clear edge.

Pretrained foundation models remain reasonably robust but do not dominate in these stress settings, indicating limits of pretraining when the target distribution departs from the pretraining regime. It is notable that our use follows the default deployment, and simple divide-and-conquer adaptations of TabPFN v2 have been shown to boost its effectiveness efficiently \citep{Ye2025Closer,Rubachev2025On}. A promising direction is to either endow tabular foundation models with an intrinsic ability to handle stress cases like large-scale datasets, or to develop corresponding lightweight adaptation strategies.

Overall, these stress tests refine the tree–DNN discussion. Foundation models and modern neural architectures have closed much of the gap in typical settings, yet tree ensembles remain hard to beat in very large-scale scenarios, and linear baselines re-emerge in ultra high-dimensional spaces. The evidence points toward hybrid, adaptive pipelines—combining strong trees, scalable MLPs, and pretrained components—as a principled path to robust tabular learning across diverse real-world conditions.
\section{Conclusion}
We presented a large-scale, systematic evaluation and analysis of deep tabular learning using {\name}, a 300+ dataset collection spanning varied sizes, domains, feature compositions, and task types. Across this breadth, method rankings do vary by dataset, but performance consistently concentrates within a small shortlist of models—offering a practical starting point for model selection. We also find that ensembling benefits both tree-based and DNN-based approaches: strong classical ensembles remain competitive, while recent pretrained (foundation) models frequently narrow—though do not fully eliminate—the historical advantage of trees. This refines the ``trees vs. neural networks'' narrative in today’s landscape.
To explain when different families win, we quantified dataset heterogeneity by learning from meta-features and early training dynamics to predict later validation behavior. The analysis highlights the roles of categorical–numerical interplay, sparsity, and entropy variation as key drivers of model advantage. Finally, our two-level design complements the main collection with {\name}-tiny (45 carefully balanced datasets for rapid, reproducible evaluation) and {\name}-extension (high-dimensional, many-class, and very large-scale settings for stress testing). Results on these subsets surface additional distinctions among model families and provide actionable guidance for heterogeneity-aware, ensemble-strengthened tabular learning.

\appendix
\begin{appendices}

\section{Datasets Selection Details}
\label{appendix:data_selection}
This appendix provides detailed information on datasets with certain quality issues and the corresponding adjustments applied in our benchmark construction.

\noindent\textbf{Datasets with mis-labeled task types.}  
We identified 22 datasets whose task types were incorrectly labeled, including 
\texttt{Contaminant-9.0GHz}, \texttt{Contaminant-9.5GHz}, 
\texttt{Contaminant-10.0GHz}, \texttt{Contaminant-10.5GHz}, 
\texttt{Contaminant-11.0GHz}, \texttt{Heart-Disease-Dataset}, 
\texttt{Insurance}, \texttt{Intersectional-Bias}, 
\texttt{Is-this-a-good-customer}, \texttt{KDD}, \texttt{Long}, 
\texttt{Performance-Prediction}, \texttt{Shipping}, 
\texttt{VulNoneVul}, \texttt{Waterstress}, \texttt{compass\_reg}, 
\texttt{credit\_reg}, \texttt{law-school-admission}, 
\texttt{ozone\_level}, \texttt{shill-bidding}, 
\texttt{shrutime}, and \texttt{svmguide3}.  
After reviewing their metadata and label structures, we corrected these datasets to binary classification tasks.

\noindent\textbf{Tabular datasets derived from other modalities.}  
Our benchmark includes 25 datasets where tabular features are extracted from non-tabular sources such as images or audio.  
These include \texttt{Indian\_pines}, \texttt{JapaneseVowels}, 
\texttt{Parkinsons\_Telemonitor}, \texttt{artificial-characters}, 
\texttt{dry\_bean\_dataset}, \texttt{hill-valley}, \texttt{kropt}, 
\texttt{letter}, \texttt{mfeat-factors}, \texttt{mfeat-fourier}, 
\texttt{mfeat-karhunen}, \texttt{mfeat-morphological}, 
\texttt{mfeat-pixel}, \texttt{100-plants-margin}, 
\texttt{100-plants-shape}, \texttt{100-plants-texture}, 
\texttt{optdigits}, \texttt{page-blocks}, \texttt{pendigits}, 
\texttt{phoneme}, \texttt{satellite\_image}, \texttt{satimage}, 
\texttt{segment}, \texttt{semeion}, and \texttt{texture}.  
Although some works~\citep{kohli2024towards,Erickson2025TabArena} exclude such datasets, we retain them because they reflect practical cases where only pre-extracted features are available due to resource or efficiency constraints.

\noindent\textbf{Datasets with known or potential leakage.}  
Prior analyses~\citep{Rubachev2024TabRed,Tschalzev2025Unreflected} have shown that several public tabular datasets contain data leakage, which can distort model comparison outcomes.  
Specifically, leakage has been reported in \texttt{Kaggle\_bike\_sharing\_demand\_challange}, \texttt{Facebook\_Comment\_Volume}, 
\texttt{GesturePhaseSegmentationProcessed}, \texttt{artificial-characters}, 
\texttt{compass}, \texttt{electricity}, \texttt{eye\_movements}, 
\texttt{eye\_movements\_bin}, \texttt{sulfur}, and 
\texttt{Brazilian\_houses\_reproduced}.  
These issues often arise from erroneous preprocessing or from features that directly encode the target variable.  
For instance, the \texttt{sulfur} dataset includes a feature that is a near-duplicate of the target variable, creating a direct leak~\citep{Rubachev2024TabRed}.

We also identify potential leakage in three additional datasets.  
In \texttt{Job\_Profitability}, the target variable \texttt{Jobs\_Gross\_Margin\_Percentage} can likely be inferred from the feature \texttt{Jobs\_Gross\_Margin}.  
In \texttt{CPMP-2015-regression}, the feature \texttt{run status} reveals information about the target runtime.  
In \texttt{estimation\_of\_obesity\_levels}, the inclusion of raw height and weight features makes obesity prediction almost trivial.

\noindent\textbf{Potential leakage when evaluating general tabular models.}
General tabular models are pretrained on multiple real-world or synthetic datasets and are often early-stopped using a separate validation set of real-world datasets. While these models can be efficiently applied to new datasets, the evaluation may lead to potential dataset leakage if the target dataset is part of the pretraining or validation datasets. In such cases, the general model might exhibit inflated performance due to prior exposure to the target dataset.

Specifically, we evaluate the general tabular model TabPFN~\citep{Hollmann2022TabPFN}, which is pretrained on synthetic datasets and early-stopped via its performance on 180 real-world datasets. The checkpoint selection rule of TabPFN potentially creates biases for these datasets. Among the datasets in our benchmark, we found only two of them, \texttt{PizzaCutter3} and \texttt{PieChart3}, that overlap with TabPFN’s validation set. 
For TabPFN v2~\citep{hollmann2025TabPFNv2}, we further examined its validation set and identified \textbf{27 datasets} that overlap with those in our 300-dataset benchmark: \texttt{ada\_prior}, \texttt{allbp}, \texttt{baseball}, \texttt{delta\_ailerons}, \texttt{eye\_movements}, \texttt{eye\_movements\_bin},
\texttt{GAMETES\_Epistasis\_2-Way\_20atts\_0.1H\_EDM-1\_1}, \texttt{hill-valley}, \texttt{JapaneseVowels},
\texttt{jungle\_chess\_2pcs\_raw\_endgame\_complete}, \texttt{led24}, \texttt{longitudinal-survey}, \texttt{page-blocks},
\texttt{ringnorm}, \texttt{rl}, \texttt{thyroid-ann}, \texttt{waveform-5000},
\texttt{debutanizer}, \texttt{delta\_elevators}, \texttt{mauna-loa-atmospheric}, \texttt{puma32H}, \texttt{stock\_fardamento02},
\texttt{treasury}, \texttt{weather\_izmir}, \texttt{wind}.

To maintain comparability with prior studies, these datasets are retained in the general {\name} benchmark.  
However, they are excluded from the stricter {\name}-tiny subset, which focuses on high-quality, leakage-free evaluation.  
This design enables fair historical comparison while supporting rigorous analysis in controlled settings.

% \noindent{\bf Multiple seeds or multiple splits}.
% To comprehensively evaluate (especially a deep) tabular method, multiple random trials with different random seeds are required. There are two strategies to construct the multiple trials. First, \cite{GorishniyRKB21Revisiting} consider the fixed train/val/test splits. With the selected hyper-parameters, 15 random seeds are applied to train a tabular model, and the corresponding performance on the test set are recorded. \cite{McElfreshKVCRGW23when} propose multiple splits, together with different random seeds, but the best hyper-parameters are selected from the first split. We follow \cite{GorishniyRKB21Revisiting} with the fixed splits. We also evaluate the difference between using multiple splits, where methods have similar ranks.

\begin{figure}[t]
  \centering
   \begin{minipage}{0.45\linewidth}
    \includegraphics[width=\textwidth]{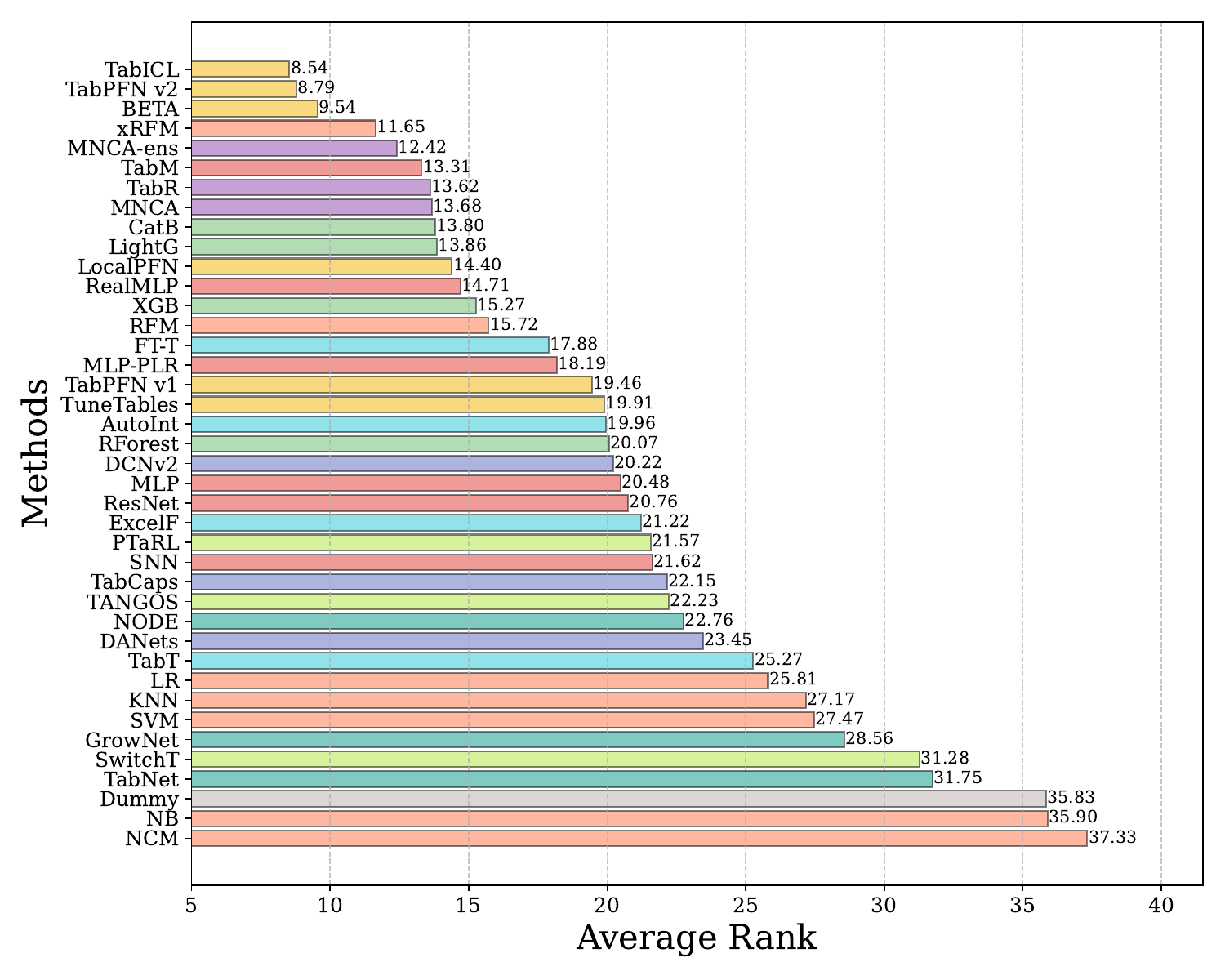}
    \centering
    {\small \mbox{(a) {Binary Classification}}}
    \end{minipage}
    \begin{minipage}{0.45\linewidth}
    \includegraphics[width=\textwidth]{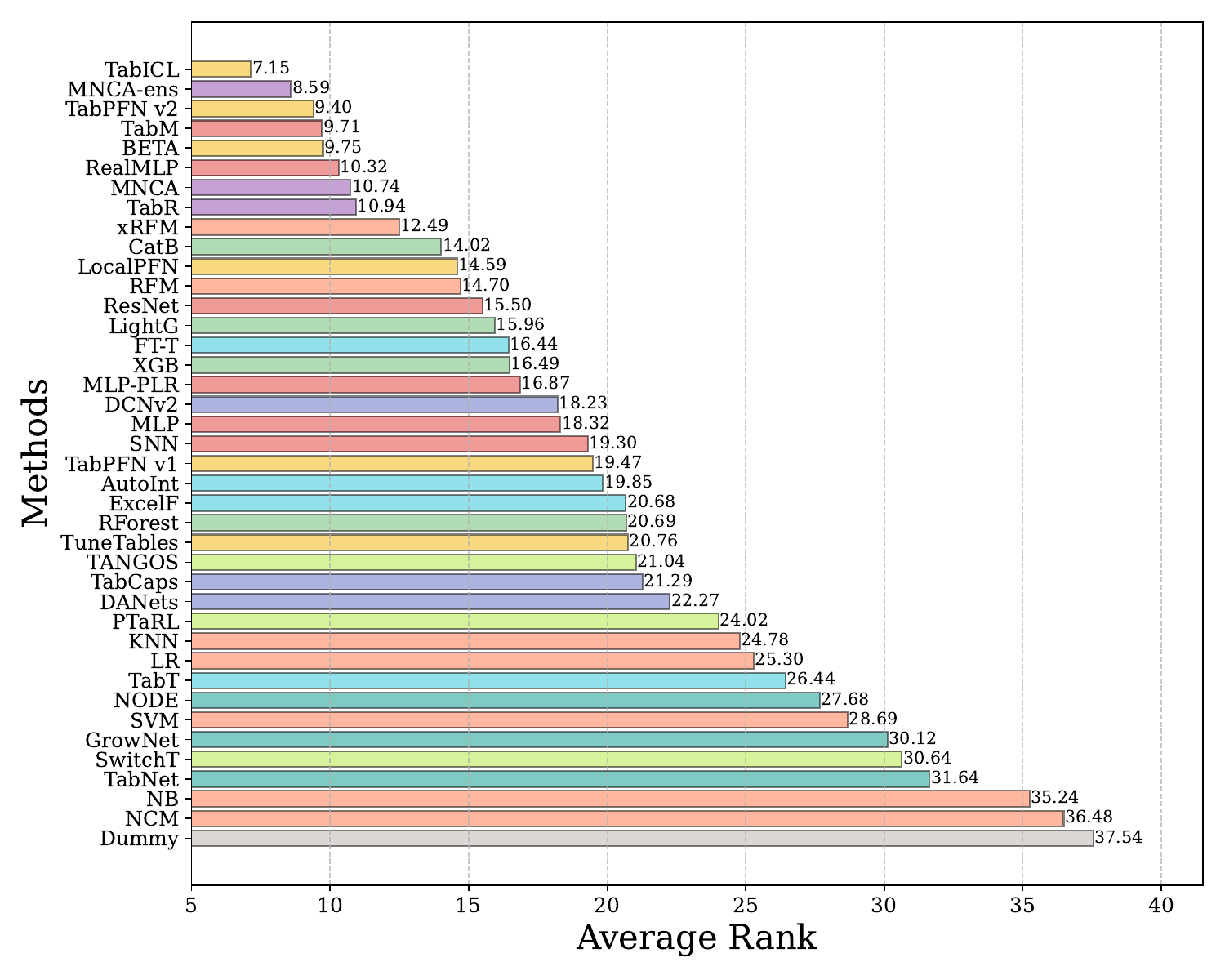}
    \centering
    {\small \mbox{(b) {Multi-Class Classification}}}
    \end{minipage}
    
    \begin{minipage}{0.45\linewidth}
    \includegraphics[width=\textwidth]{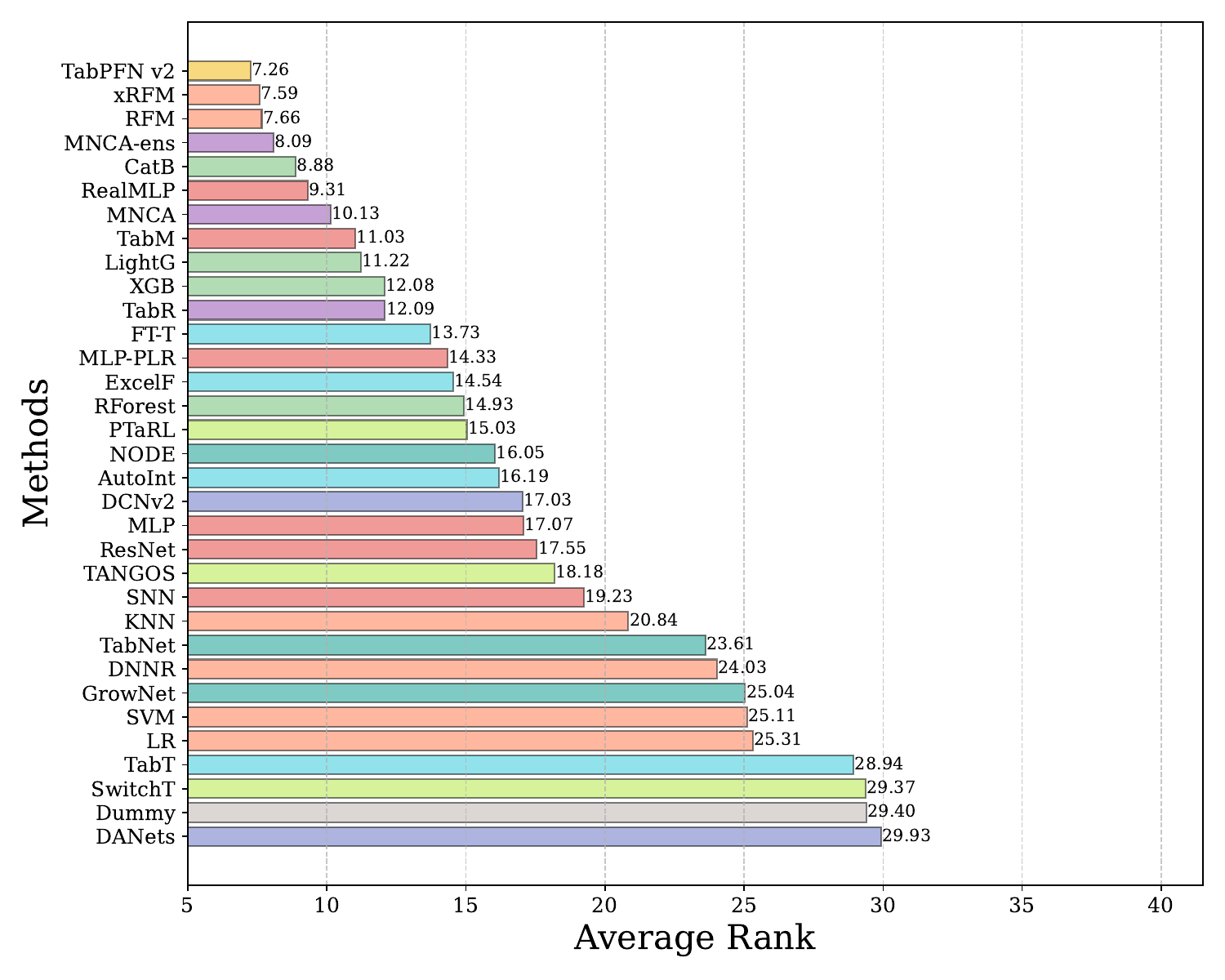}
    \centering
    {\small \mbox{(c) {Regression}}}
    \end{minipage}
    \begin{minipage}{0.45\linewidth}
    \includegraphics[width=\textwidth]{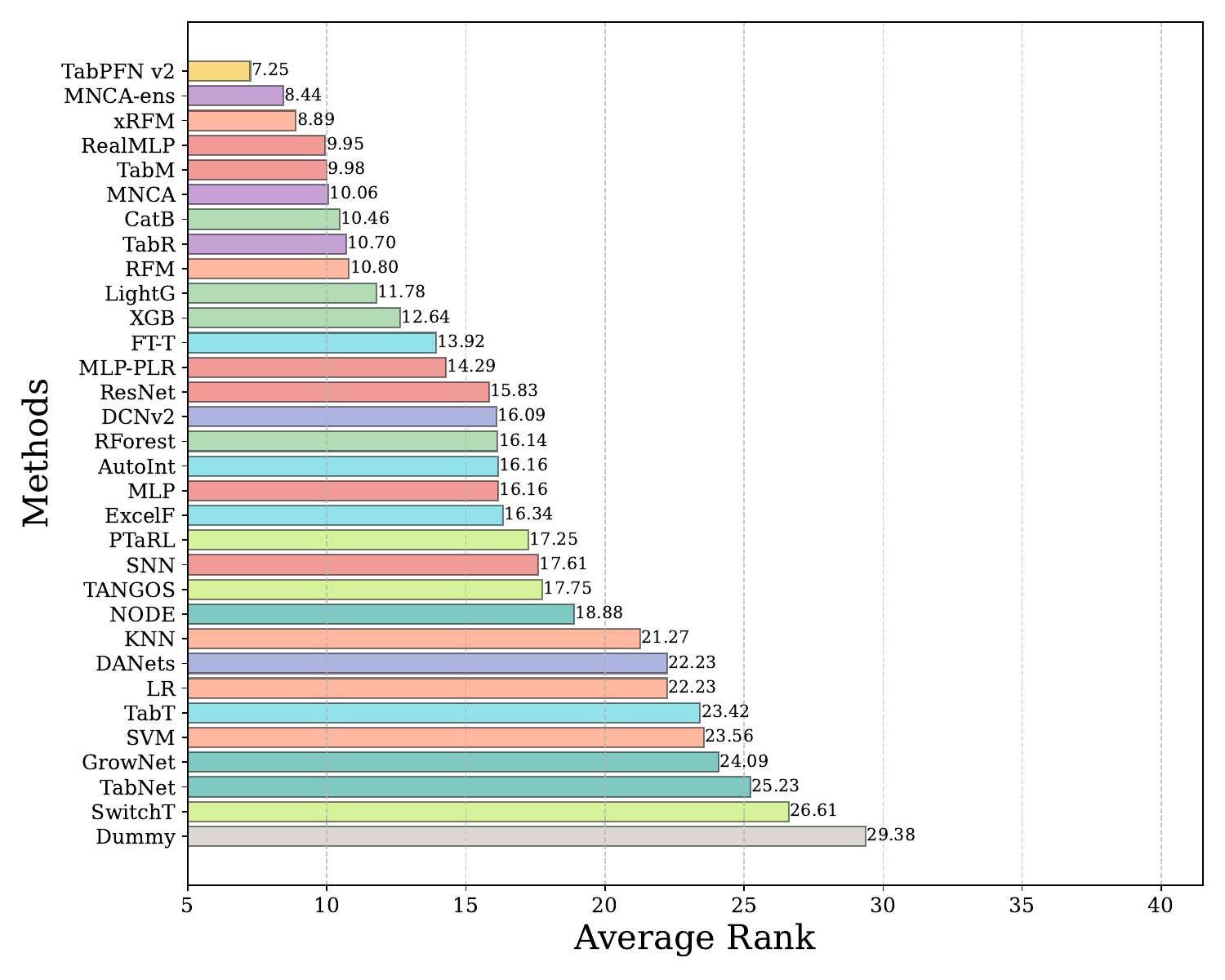}
    \centering
    {\small \mbox{(d) {All Tasks}}}
    \end{minipage}
  \caption{Average rank of tabular methods. We show the average rank of all methods over binary (120 datasets), multi-class (80 datasets), regression (100 datasets), and all 300 datasets. 
  The ranks are calculated based on accuracy and RMSE over classification and regression tasks, respectively.
  The lower the rank value, the better the average performance.
  }
  \label{fig:average_rank}
\end{figure}

\section{Additional Comparison Results}
\subsection{Average Performance and Rankings}
\begin{figure}[t]
  \centering
   \begin{minipage}{0.48\linewidth}
    \includegraphics[width=\textwidth]{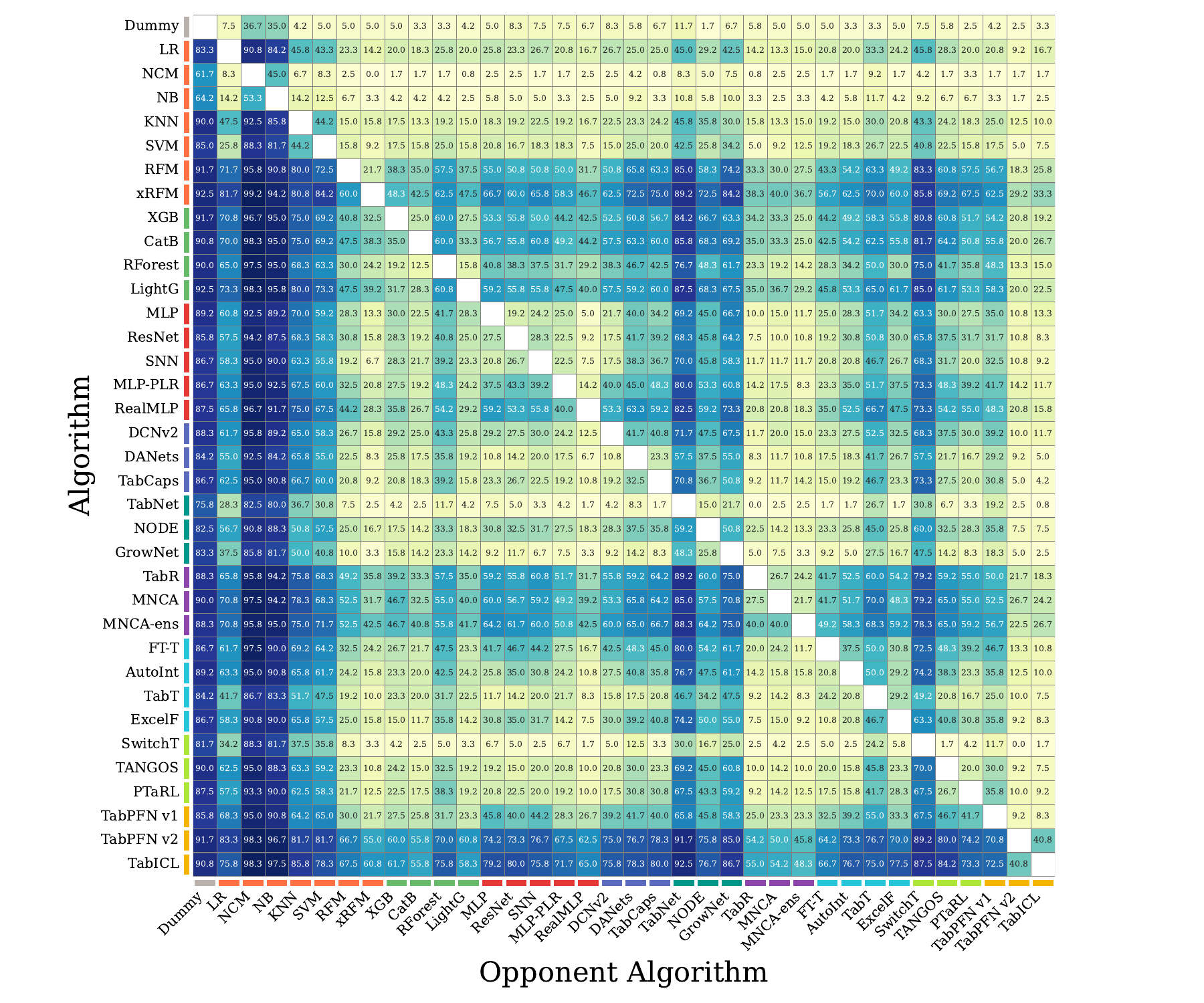}
    \centering
    {\small \mbox{(a) {Binary Classification}}}
    \end{minipage}
    \begin{minipage}{0.48\linewidth}
    \includegraphics[width=\textwidth]{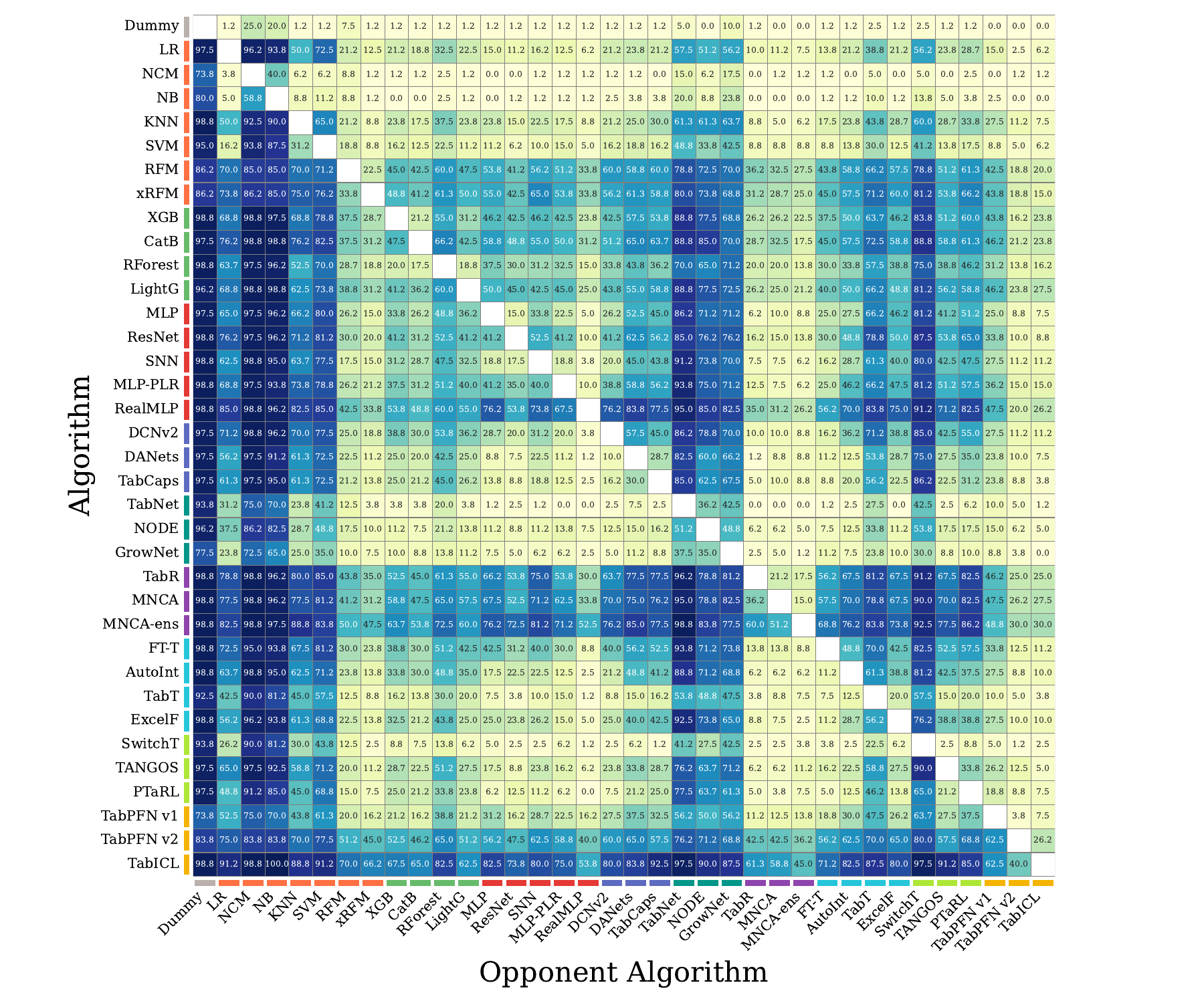}
    \centering
    {\small \mbox{(b) {Multi-Class Classification}}}
    \end{minipage}
    
    \begin{minipage}{0.48\linewidth}
    \includegraphics[width=\textwidth]{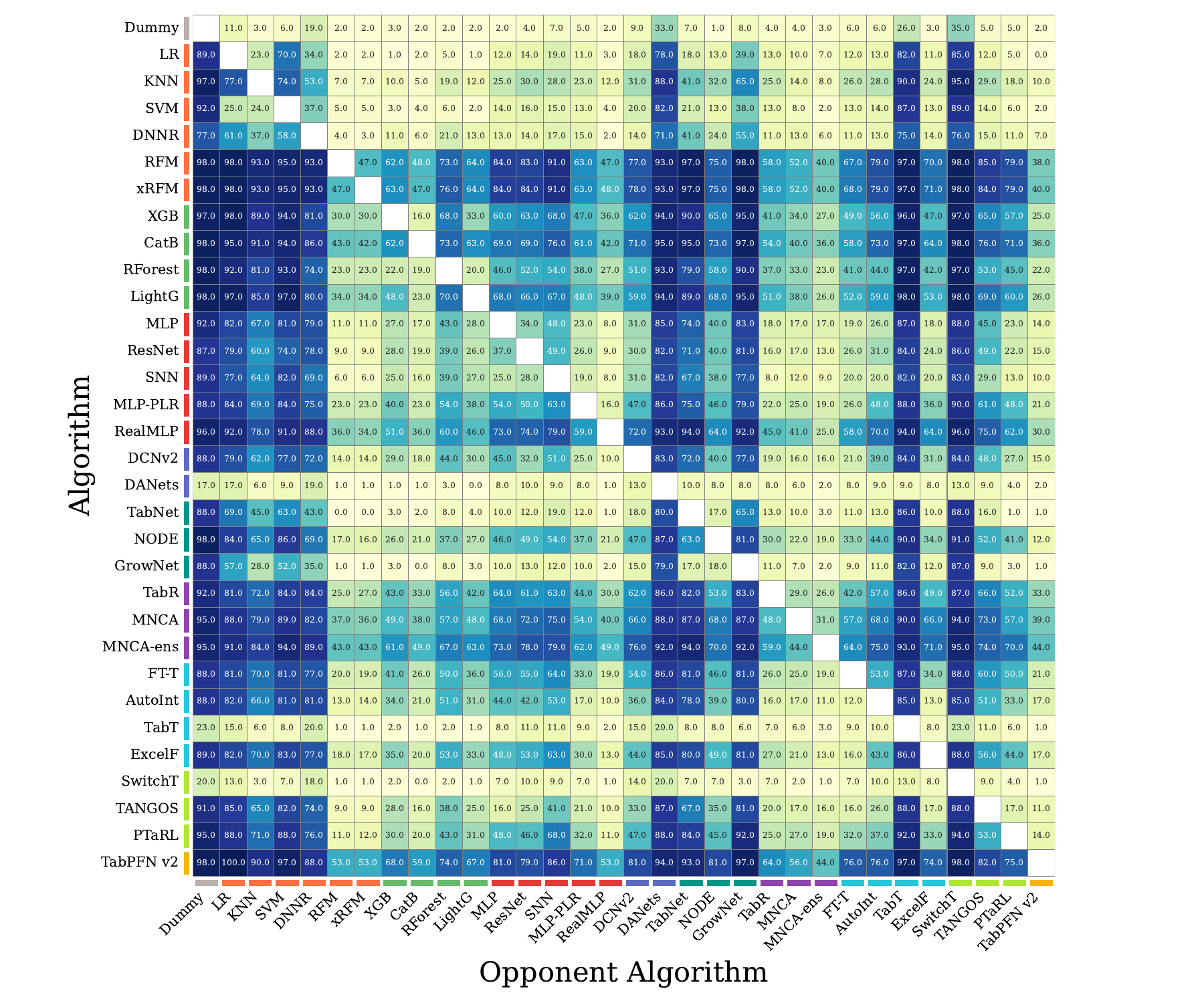}
    \centering
    {\small \mbox{(c) {Regression}}}
    \end{minipage}
    \begin{minipage}{0.48\linewidth}
    \includegraphics[width=\textwidth]{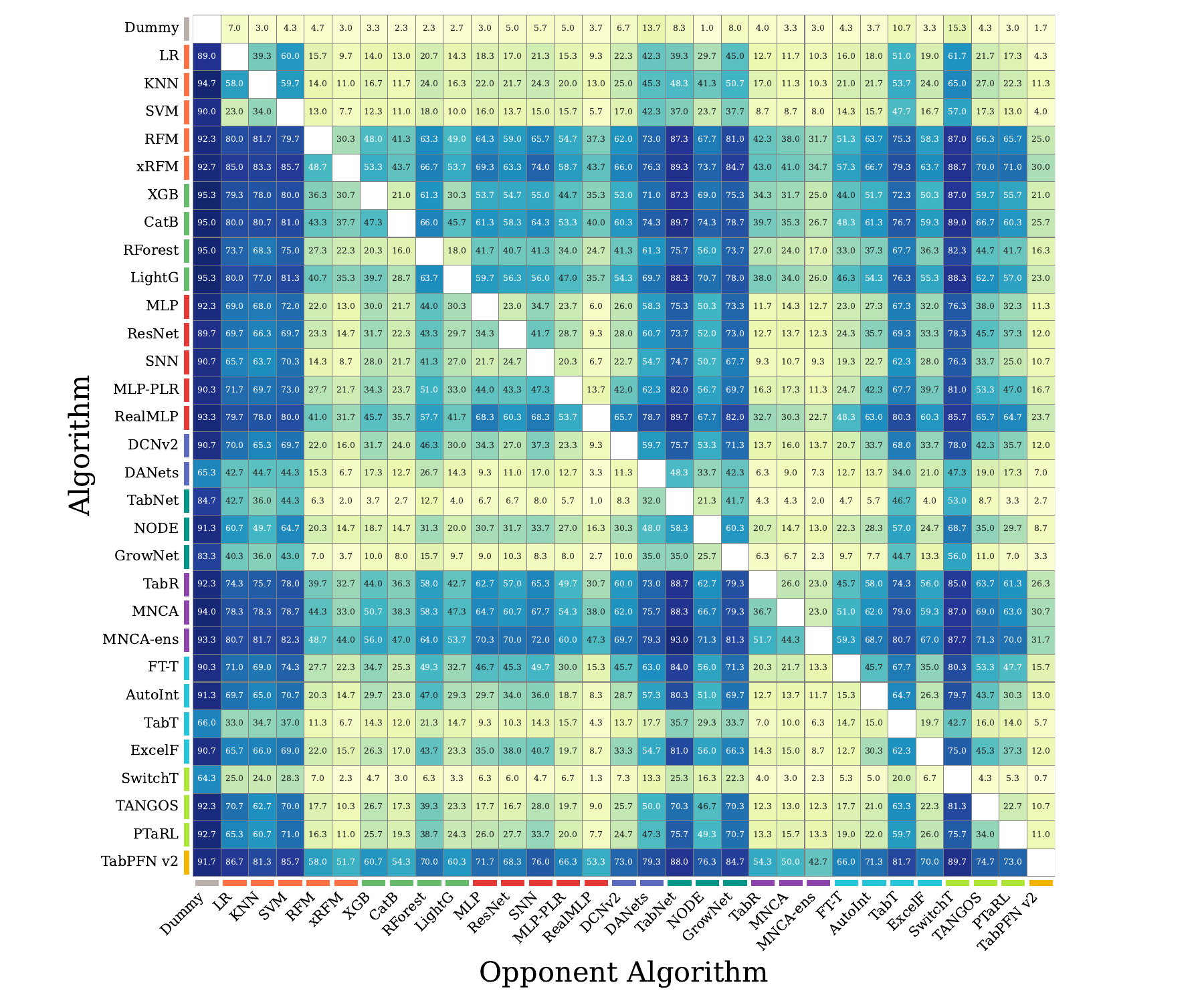}
    \centering
    {\small \mbox{(d) {All Tasks}}}
    \end{minipage}
  \caption{Heatmaps illustrating the statistical comparisons between all pairs of methods based on t-tests with a 95\% confidence interval. The Win/Tie/Lose counts between each pair of methods are also denoted. Darker colors indicate higher counts. 
  }
  \label{fig:heatmap}
\end{figure}

To complement the main statistical comparisons, we report detailed average ranks and pairwise t-test results. 
The average ranks of 40 representative methods across 300 datasets are shown in~\autoref{fig:average_rank}, while pairwise Win/Tie/Lose outcomes are illustrated in~\autoref{fig:heatmap}. 
These provide a finer-grained view of the relative positioning of methods beyond the critical difference diagrams in the main text.

From \autoref{fig:average_rank}, several clear patterns emerge. 
First, pretrained tabular foundation models dominate across settings. 
TabPFN v2 and TabICL consistently achieve the lowest ranks in binary and multi-class classification, confirming the strength of pretraining and in-context learning strategies. 
In regression, TabPFN v2 remains the top performer, followed closely by xRFM, MNCA-ens, and RealMLP. 
Notably, foundation models are the only family that simultaneously excels across all three task types, underlining their broad generalization.

Tree-based ensembles continue to provide strong baselines. 
CatBoost and LightGBM achieve top-tier ranks across both classification and regression tasks, while XGBoost trails slightly but still outperforms most DNNs. 
Recursive Feature Machines (RFM) and its extension xRFM perform particularly well on regression, often rivaling the strongest ensembles and DNNs.

Among deep learning approaches, neighborhood-based methods stand out. 
ModernNCA achieves consistently high ranks across all task types and remains competitive with CatBoost and LightGBM. 
TabR also performs strongly in classification tasks. 
MLP variants show a clear divide: vanilla MLP is weak, but tuned designs such as MLP-PLR and RealMLP achieve much lower ranks, with RealMLP frequently joining the top-performing group. 
Token-based models (\eg, FT-T, ExcelFormer, AutoInt) are stable performers, especially in classification, but their ranks indicate they are not decisively stronger than the best ensembles or neighborhood-based models. 
Tree-mimic models (NODE, TabNet, GrowNet) generally remain in the lower half of the rankings.

\subsection{Pairwise Statistical Comparisons}
To complement the average rank analysis, we further conduct pairwise statistical comparisons between all pairs of methods using t-tests with a 95\% confidence interval. 
The results are visualized in~\autoref{fig:heatmap}, which reports the Win/Tie/Lose counts between each pair of methods across binary, multi-class, regression, and all tasks. Darker colors indicate higher counts, reflecting more consistent superiority in the corresponding comparisons.

The heatmaps confirm many earlier findings: tree-based ensembles (CatBoost, LightGBM, XGBoost) dominate weaker baselines and cluster together as robust, statistically indistinguishable methods. RealMLP and MLP-PLR clearly outperform vanilla MLPs and ResNets, though they tie with strong ensembles in many cases. Token-based approaches (\eg, FT-T, ExcelFormer) show stable but not dominant behavior, often tying with both ensembles and tuned DNNs.

Neighborhood-based methods, especially ModernNCA and MNCA-ens, stand out as frequent winners in classification, rivaling ensembles and confirming the strength of retrieval-based learning. Finally, pretrained foundation models (TabPFN v2, TabICL) achieve the most consistent wins in binary and multi-class classification, while in regression, they remain statistically tied with ensembles and ModernNCA, indicating task-dependent benefits.

Overall, the pairwise comparisons highlight that while pretrained models push state-of-the-art performance, top ensembles and retrieval-based methods remain highly competitive, forming overlapping statistical equivalence groups across many tasks.

\section{Details of the Heterogeneity Analysis}

This section provides additional details complementing our analysis of dataset heterogeneity in~\autoref{sec:hetero_analysis}. We describe the meta-features employed, the recording of training dynamics, the curve families used for modeling validation trajectories, and supplementary results. Finally, we highlight a by-product of this framework: predicting the training dynamics of deep tabular models, which may enable more efficient training in practice.

\subsection{Details of Meta-Features}
The meta-features encode structural and statistical properties of a dataset~\citep{McElfreshKVCRGW23when}. They form the foundation for analyzing how heterogeneity influences model behavior. By incorporating these meta-features, we not only characterize tabular datasets but also predict training dynamics, thereby identifying which dataset factors most strongly shape model performance. We provide the full list of all meta-features in~\autoref{tab:meta_features}.

\begin{table}[t]
\small
\centering
\caption{Meta-features used in the training dynamics prediction task. The first column indicates the selected key meta-features.}
\begin{tabular}{@{}cl|ll@{}}
\toprule
Selected & Meta-Feature & Explanation \\ 
\midrule
&\texttt{attr\_conc}&	The concentration coef. of each pair of distinct attributes.\\
\checkmark & \texttt{class\_conc}& The concentration coefficient between each attribute and class. \\
&\texttt{class\_ent}&	The target attribute Shannon’s entropy.\\
\checkmark & \texttt{inst\_to\_attr}& The ratio between the number of instances and attributes. \\
\checkmark &\texttt{mean}&The mean value of each attribute.\\
& \texttt{sd}&	The standard deviation of each attribute.\\
& \texttt{var}&	The variance of each attribute.\\
\checkmark &\texttt{range}&The range (max - min) of each attribute.\\
\checkmark & \texttt{iq\_range}&The interquartile range (IQR) of each attribute.\\
\checkmark&\texttt{nr\_attr}&The total number of attributes.\\
\checkmark & \texttt{sparsity}&The (possibly normalized) sparsity metric for each attribute.\\
&\texttt{t\_mean}&	The trimmed mean of each attribute.\\
&\texttt{nr\_bin}&	The number of binary attributes.\\
&\texttt{nr\_cat}&	The number of categorical attributes.\\
&\texttt{nr\_num}	& The number of numeric features.\\
&\texttt{nr\_norm}&	The number of attributes normally distributed based in a given method.\\
&\texttt{nr\_cor\_attr}&	The number of distinct highly correlated pair of attributes.\\
\checkmark&\texttt{gravity}&The distance between minority and majority classes' center of mass.\\
&\texttt{nr\_class}&	The number of distinct classes.\\
\checkmark &\texttt{joint\_ent}&The joint entropy between each attribute and class.\\
\checkmark &\texttt{attr\_ent}& Shannon’s entropy for each predictive attribute.\\
\checkmark&\texttt{cov}&The absolute value of the covariance of distinct dataset attribute pairs.\\
& \texttt{eigenvalues}&	The eigenvalues of covariance matrix from dataset.\\
&\texttt{eq\_num\_attr}&	The number of attributes equivalent for a predictive task.\\
\checkmark&\texttt{max}&The maximum value from each attribute.\\
&\texttt{min}&	The minimum value from each attribute.\\
&\texttt{median}&	The median value from each attribute.\\
&\texttt{freq\_class}&	The relative frequency of each distinct class.\\
&\texttt{mad}&	The Median Absolute Deviation (MAD) adjusted by a factor.\\
\checkmark&\texttt{mut\_inf}&The mutual information between each attribute and target.\\
\checkmark&\texttt{nr\_inst}&The number of instances (rows) in the dataset.\\
\checkmark & \texttt{nr\_outliers}&The number of attributes with at least one outlier value.\\
\checkmark & \texttt{ns\_ratio}&The noisiness of attributes.                               \\
\checkmark&\texttt{imblance\_ratio}&The ratio of the number of instances in the minority to the majority class.\\
&\texttt{attr\_to\_inst}&	The ratio between the number of attributes.\\
\bottomrule
\end{tabular}
\label{tab:meta_features}
\end{table}

\subsection{Recording Training Dynamics}
Beyond end-point accuracy or RMSE, we record detailed training dynamics for each dataset–method pair. These include:
\begin{itemize}[noitemsep,topsep=0pt,leftmargin=*]
    \item \textbf{Training logs}: learning rate schedules, batch-wise losses, and intermediate statistics.
    \item \textbf{Performance metrics}: validation/test loss, accuracy, RMSE, as well as secondary metrics such as F1/AUC (classification) and MAE/$R^2$ (regression).
    \item \textbf{Running time}: measured across 15 random seeds, accounting for early stopping at variable epochs.
    \item \textbf{Model size}: for both default and tuned hyperparameters.
\end{itemize}
These logs provide a rich foundation for connecting dataset properties with optimization behavior, thereby supporting heterogeneity analysis.

\subsection{Alternative Curve Families for Validation Dynamics}
To forecast validation trajectories from early training segments, we experimented with several curve families inspired by prior work in vision and language~\citep{Hestness2017Deep,RosenfeldRBS20Constructive,Bahri2021Explaining}. Let $t$ denote epoch and $y$ the performance measure. We consider four functional forms:
\begin{itemize}[noitemsep,topsep=0pt,leftmargin=*]
    \item \textbf{M1}: $y=at^b$ (basic power-law form)~\citep{AlabdulmohsinNZ22Revisiting}.  
    \item \textbf{M2}: $y=at^b+c$ (shifted power-law)~\citep{CortesJSVD93Learning,Hestness2017Deep,RosenfeldRBS20Constructive,Abnar0NS22Exploring}.  
    \item \textbf{M3}: $y=a(t+d)^b+c$ with offset $d$ controlling the onset of improvement.  
    \item \textbf{M4}: $(y-\epsilon_\infty)/((\epsilon_0-y)^a)=bt^c$, where $\epsilon_\infty$ is irreducible error and $\epsilon_0$ random-guess performance.  
\end{itemize}
Parameters are estimated from the initial support set $\sS$ (first epochs). Once fitted, these forms extrapolate the query set $\sQ$, enabling curve reconstruction.

\subsection{Main Results and Analysis}
We implement the meta-mapping $h$ as a four-layer MLP. The input includes both the first 5 validation points and 19 meta-features, and the output is the parameter set $\theta$ defining our curve family.  

We evaluate with Mean Absolute Error (MAE) and Optimal Value Difference (OVD). OVD measures the discrepancy between the optimal values of predicted and true curves (maximum for classification, minimum for regression). Results are reported in \autoref{tab:metrics}.

\begin{table}[t]
\caption{Average MAE and OVD for various curves across test datasets. Both metrics are ``lower is better.''
``Ours'' refers to directly fitting the curves using \autoref{eq:the_law}.
``Ours with MLP'' indicates the method using the learned $h$.}
\label{tab:metrics}
\centering
\begin{tabular}{l|cc}
\toprule
Comparison Methods & MAE & OVD \\
\midrule
M1              &  0.1845   &      0.2936      \\
M2              &  0.8458   &      6.0805      \\
M3              &  0.1331   &      0.1148      \\
M4              &  0.1818   &      0.1794      \\
Direct Fit (ours) &  1.1927   &  1.8860      \\
Meta-learned $h$ (ours)   &  \textbf{0.0748}   &  \textbf{0.0701}    \\
\bottomrule
\end{tabular}
\end{table}

We also attempt to fit parameters for other curve families (M1–M4) using the optimization objective in \autoref{eq:objective}. However, these formulations often face convergence issues, and direct fitting with our curve form produces suboptimal results due to differences in fitting strategy. As shown in \autoref{tab:metrics}, the gap between ``Direct Fit'' and ``Meta-learned $h$'' underscores the necessity of incorporating meta-features: leveraging dataset properties in addition to early dynamics substantially improves curve prediction accuracy.

Importantly, the learned predictor $h$ can accurately extrapolate the remaining validation performance curves from only the first few epochs.

\begin{figure}[t]
  \centering
   \begin{minipage}{0.24\linewidth}
    \includegraphics[width=\textwidth]{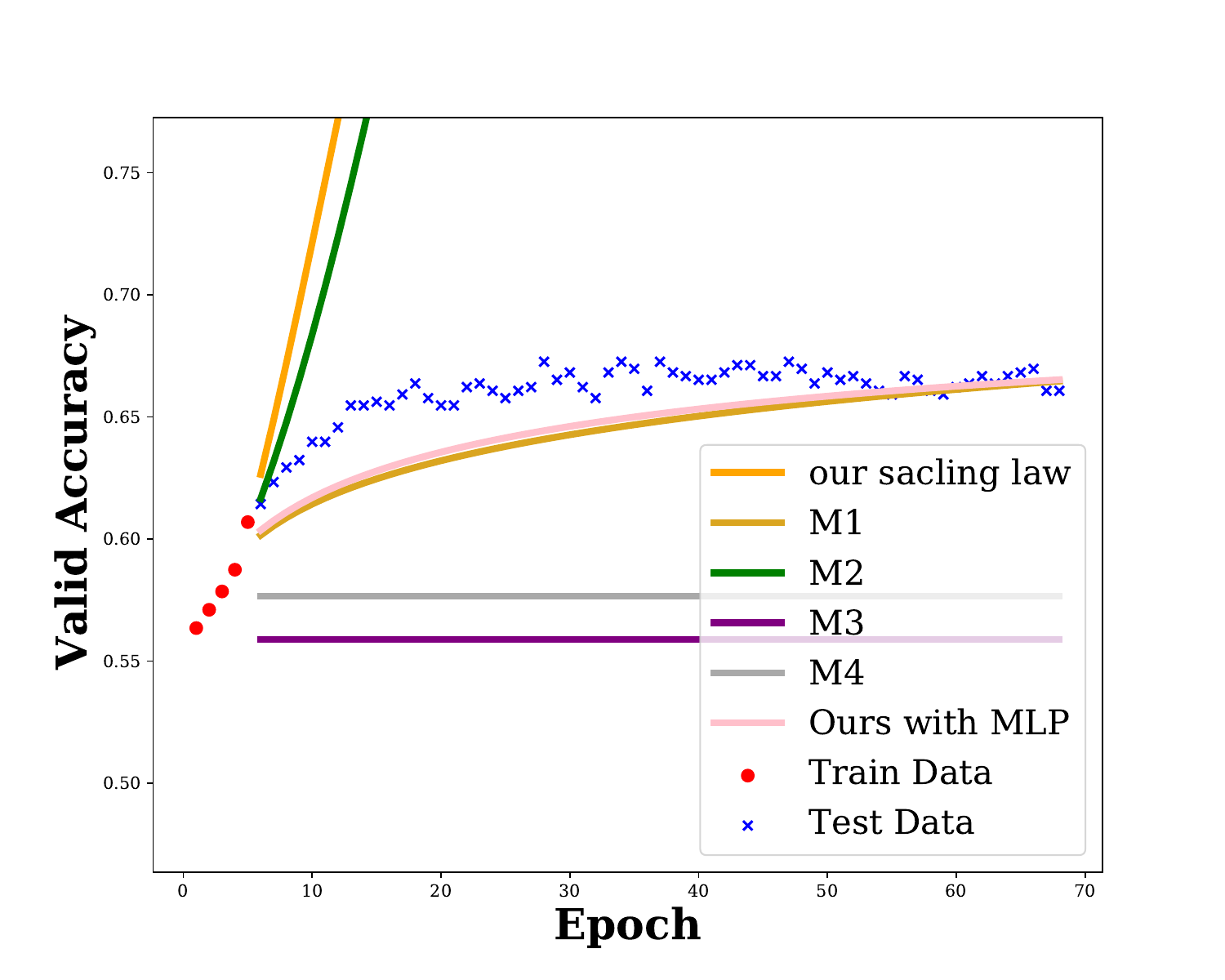}
    \centering
    {\scriptsize \mbox{(a) {abalone}}}
    \end{minipage}
    \begin{minipage}{0.24\linewidth}
    \includegraphics[width=\textwidth]{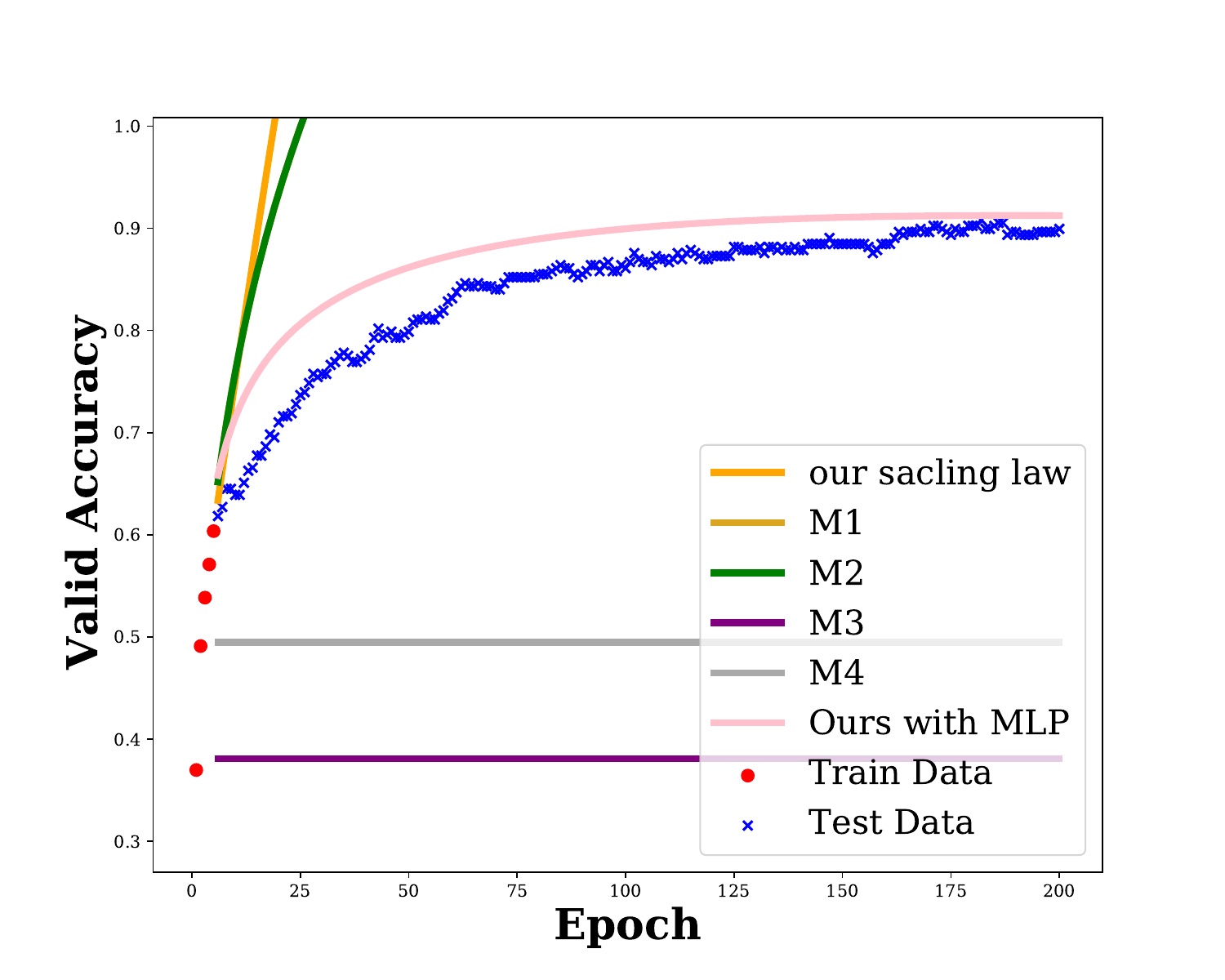}
    \centering
    {\scriptsize \mbox{(b) {estimation\_of\_obesity\_levels}}}
    \end{minipage}
    \begin{minipage}{0.24\linewidth}
    \includegraphics[width=\textwidth]{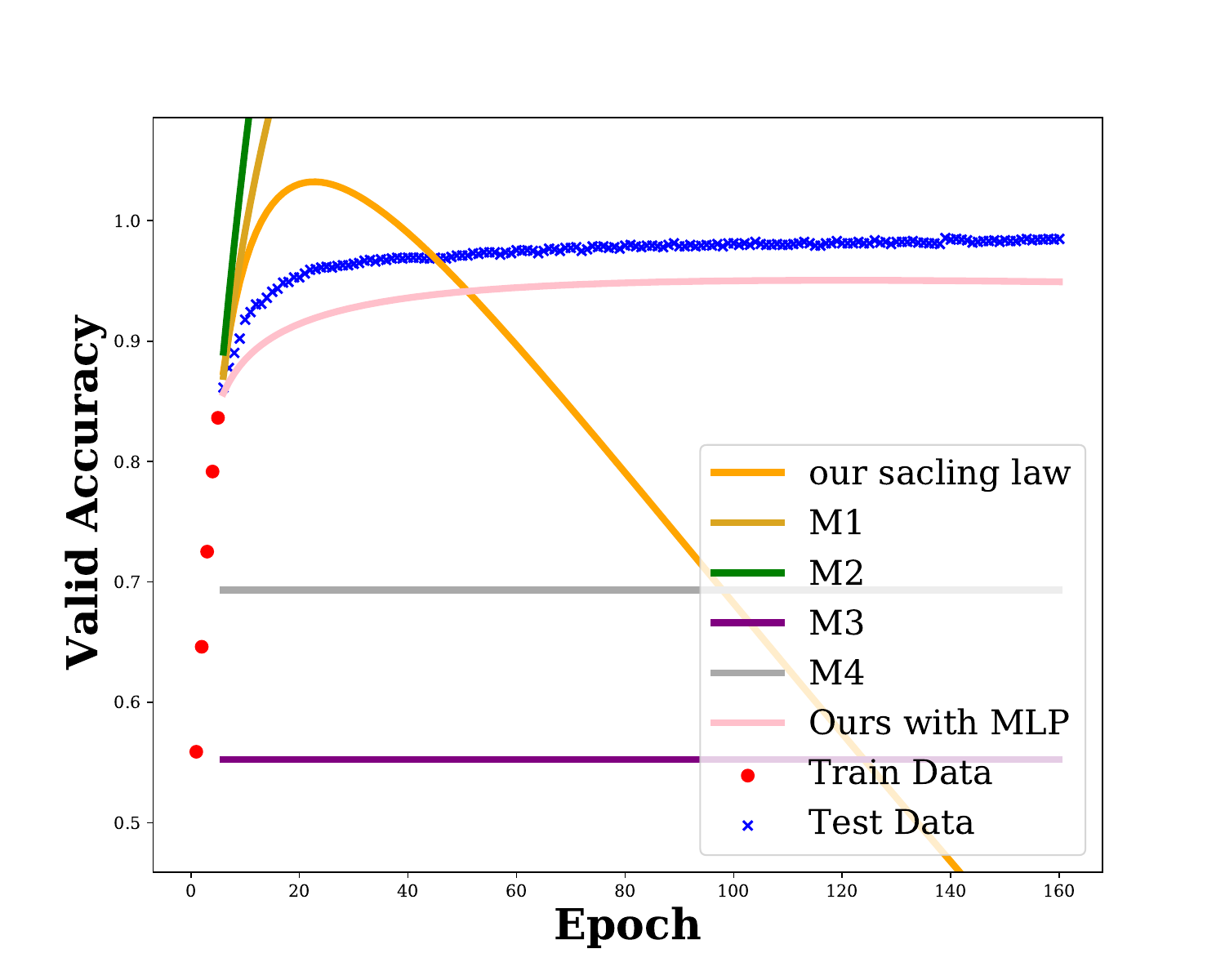}
    \centering
    {\scriptsize \mbox{(c) {JapaneseVowels}}}
    \end{minipage}
    \begin{minipage}{0.24\linewidth}
    \includegraphics[width=\textwidth]{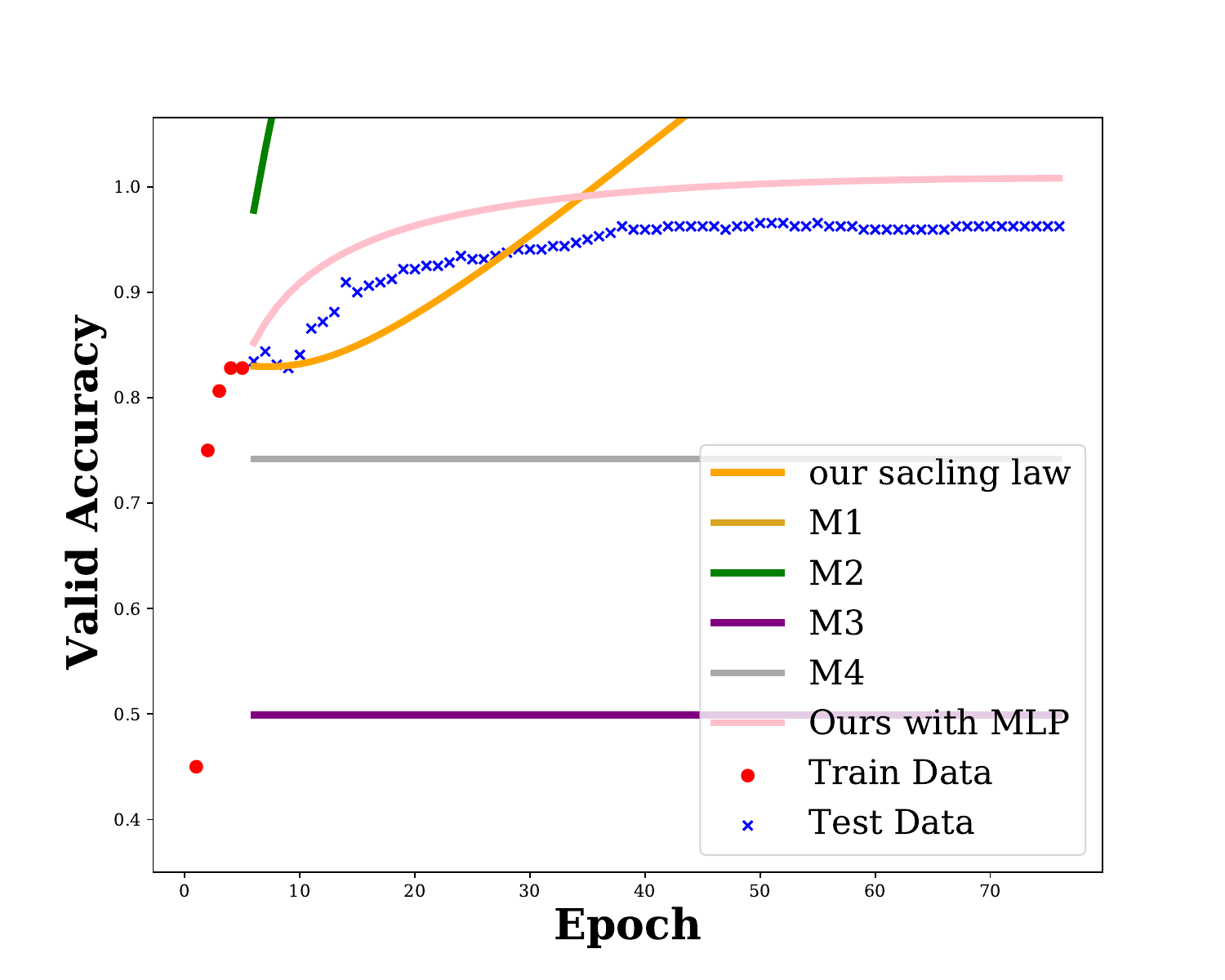}
    \centering
    {\scriptsize \mbox{(d) {mfeat-factors}}}
    \end{minipage}

    \begin{minipage}{0.24\linewidth}
    \includegraphics[width=\textwidth]{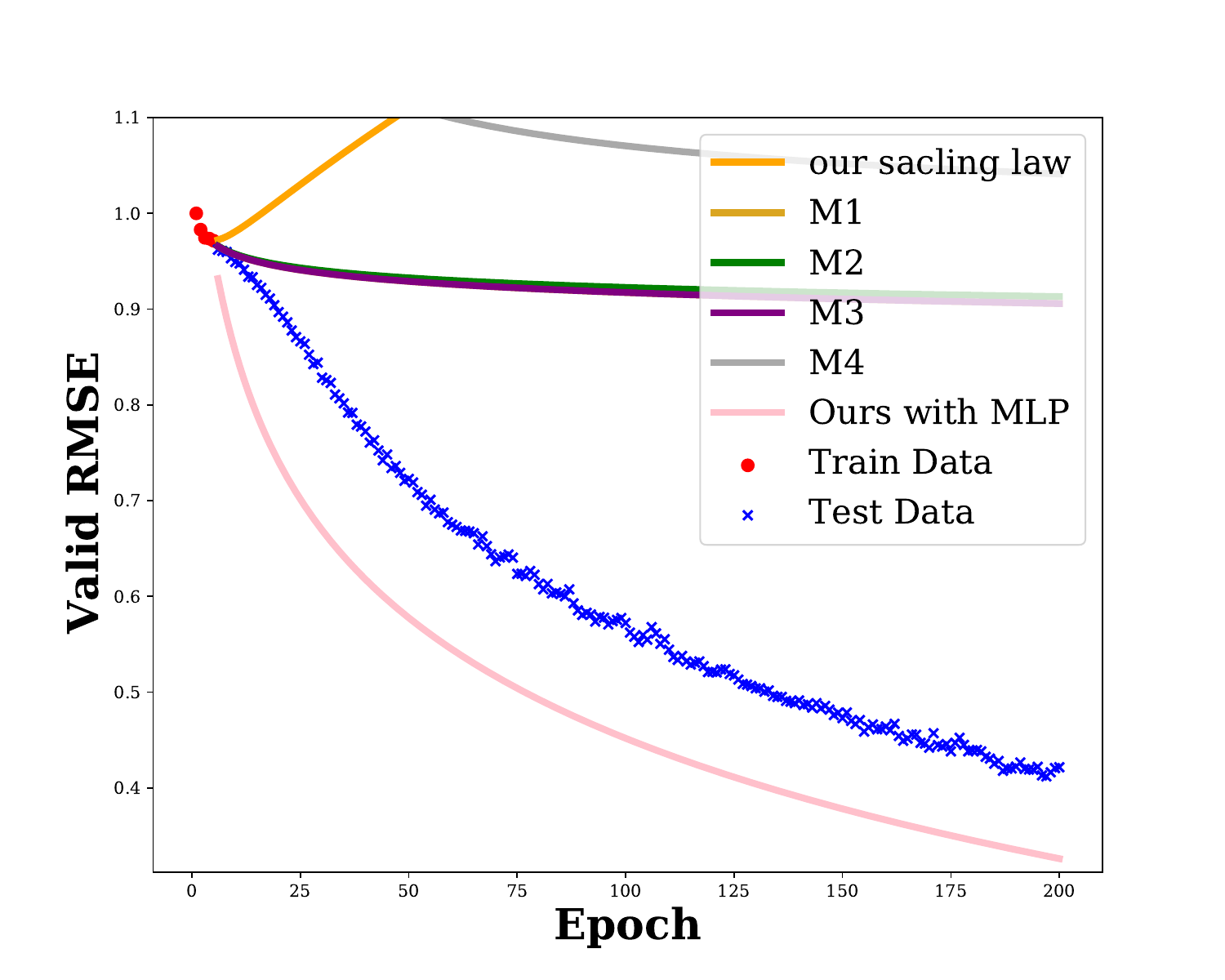}
    \centering
    {\scriptsize \mbox{(e) {3D\_...\_dataset}}}
    \end{minipage}
    \begin{minipage}{0.24\linewidth}
    \includegraphics[width=\textwidth]{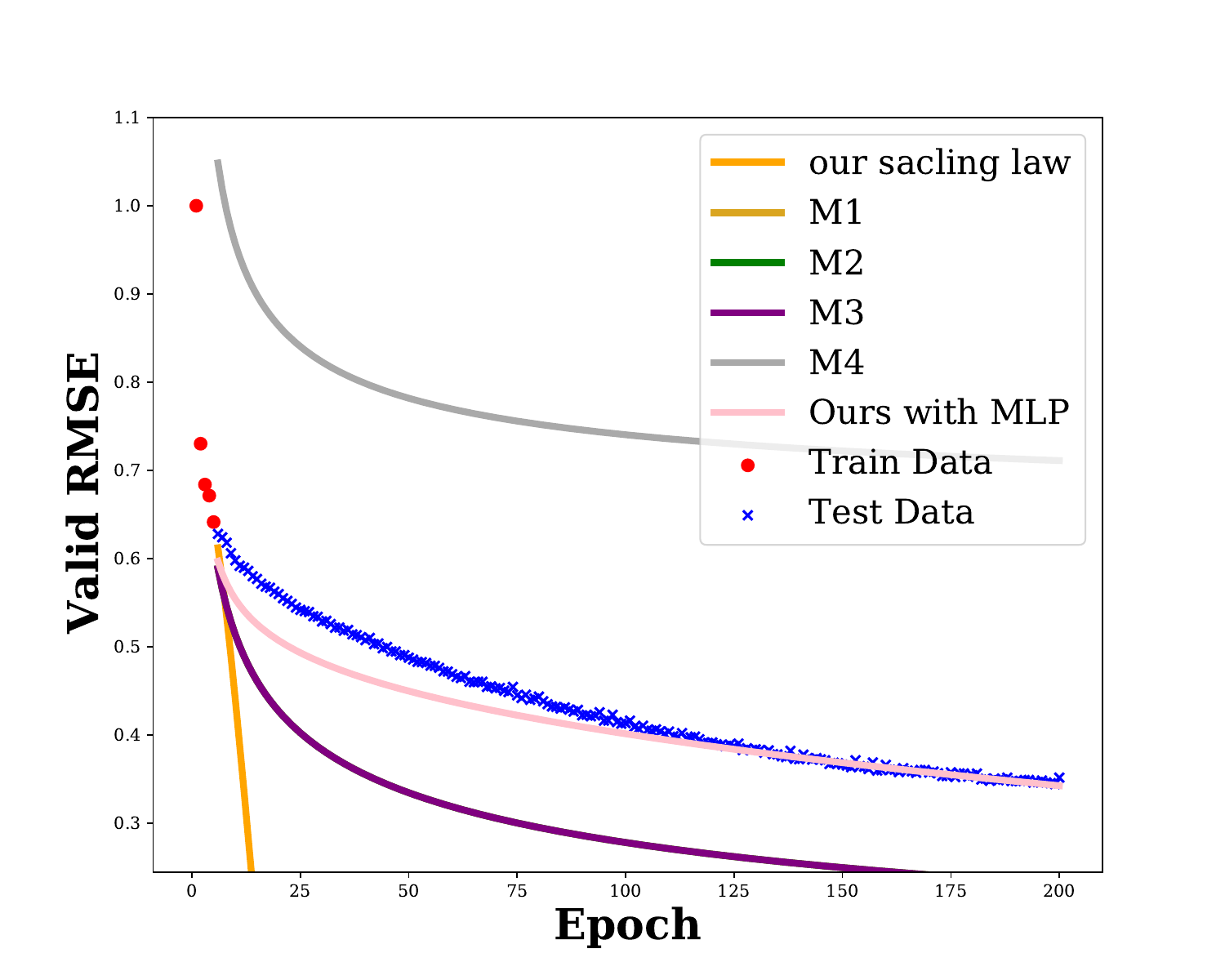}
    \centering
    {\scriptsize \mbox{(f) {Bias\_correction\_r}}}
    \end{minipage}
    \begin{minipage}{0.24\linewidth}
    \includegraphics[width=\textwidth]{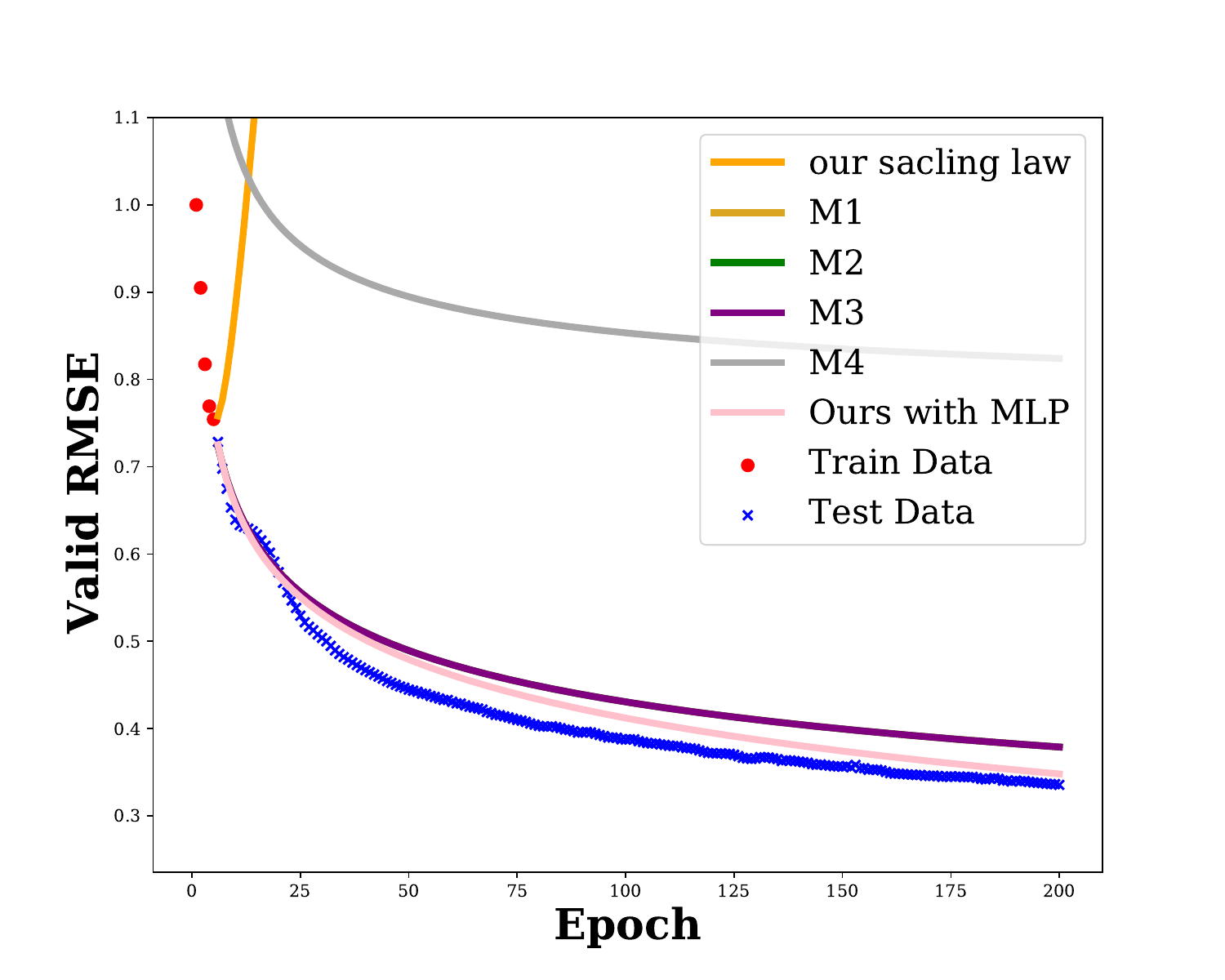}
    \centering
    {\scriptsize \mbox{(g) {Con...-10.5GHz(Urbinati)}}}
    \end{minipage}
    \begin{minipage}{0.24\linewidth}
    \includegraphics[width=\textwidth]{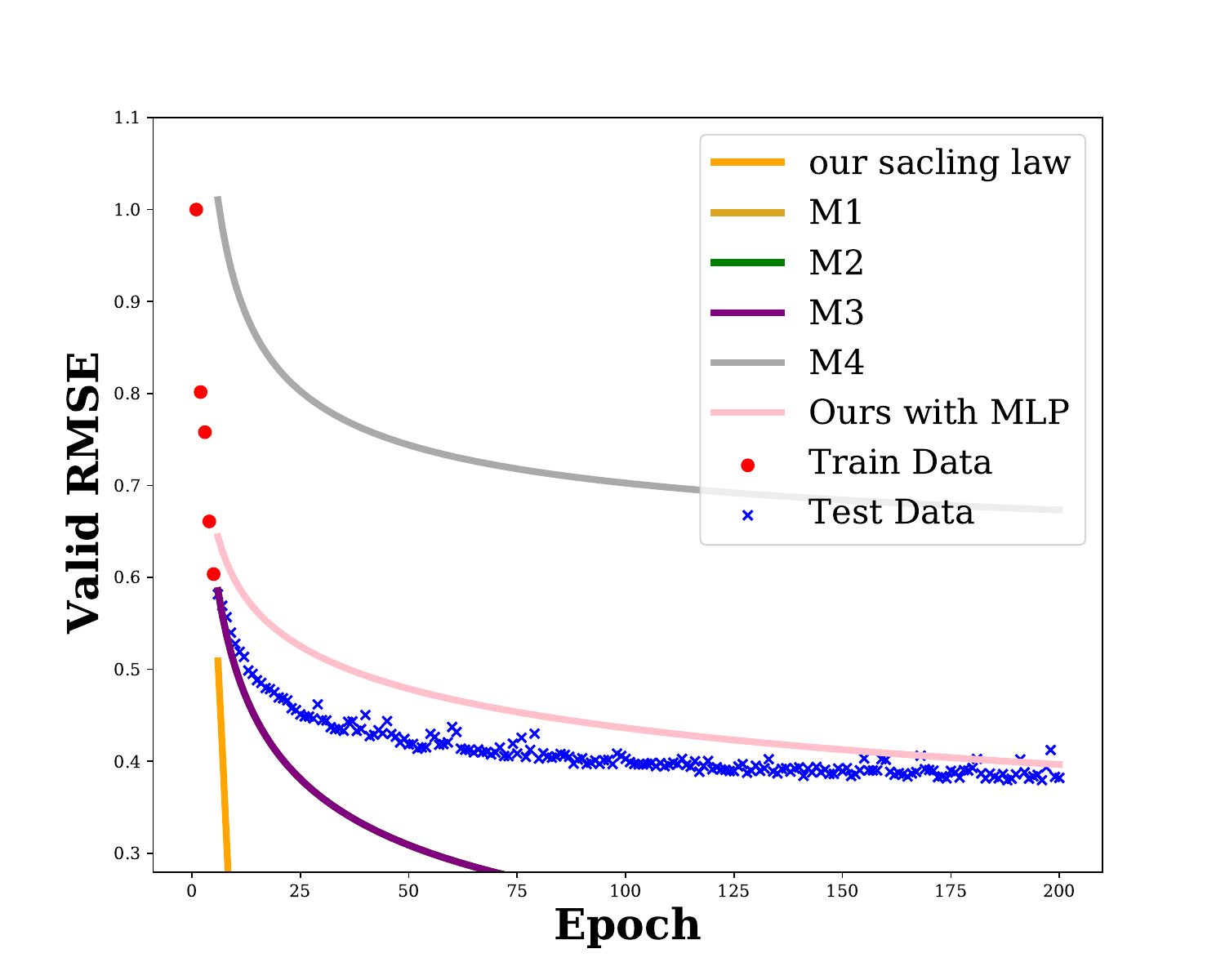}
    \centering
    {\scriptsize \mbox{(h) {MiamiHousing2016}}}
    \end{minipage}

    \begin{minipage}{0.24\linewidth}
    \includegraphics[width=\textwidth]{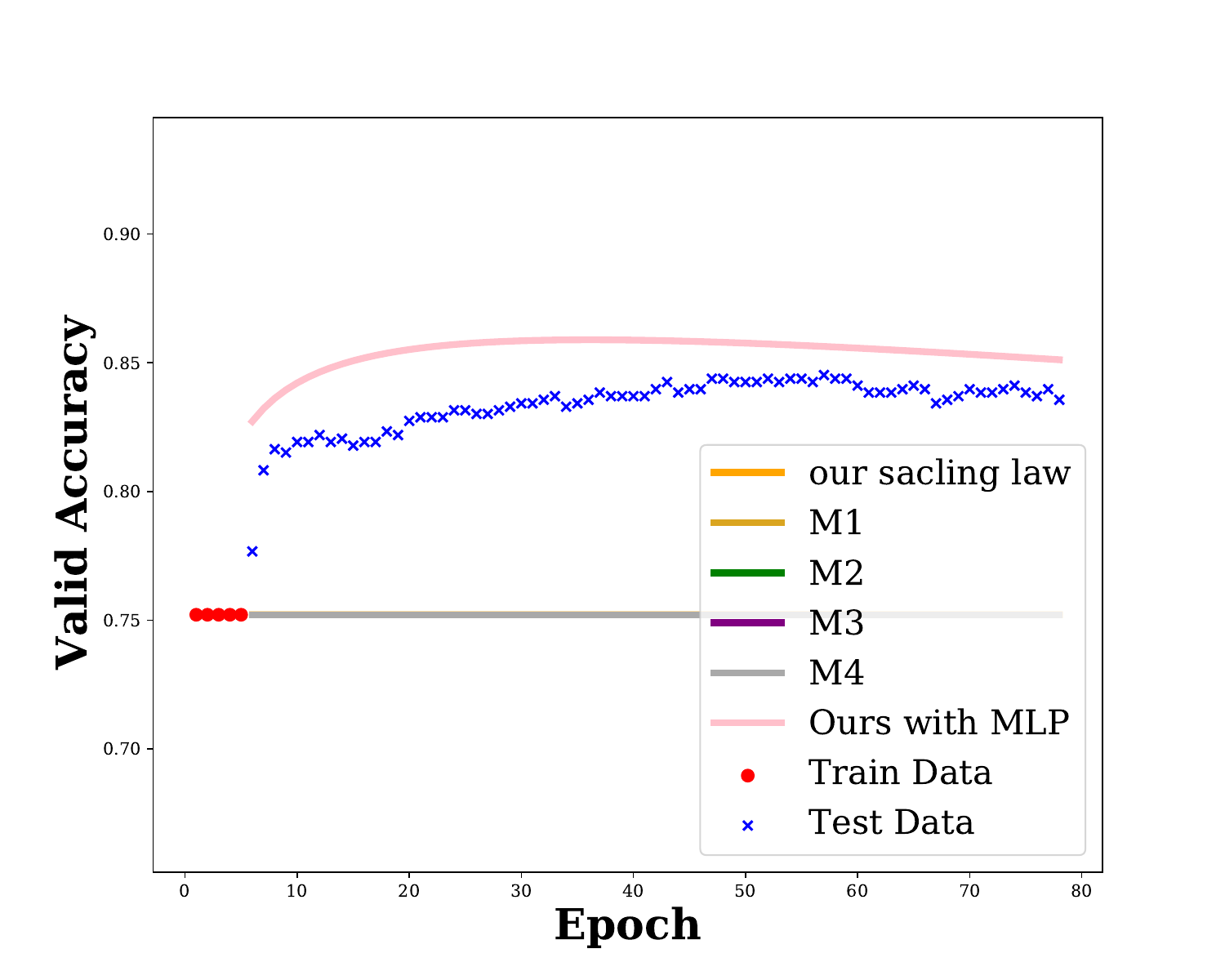}
    \centering
    {\scriptsize \mbox{(i) {ada\_agnostic}}}
    \end{minipage}
    \begin{minipage}{0.24\linewidth}
    \includegraphics[width=\textwidth]{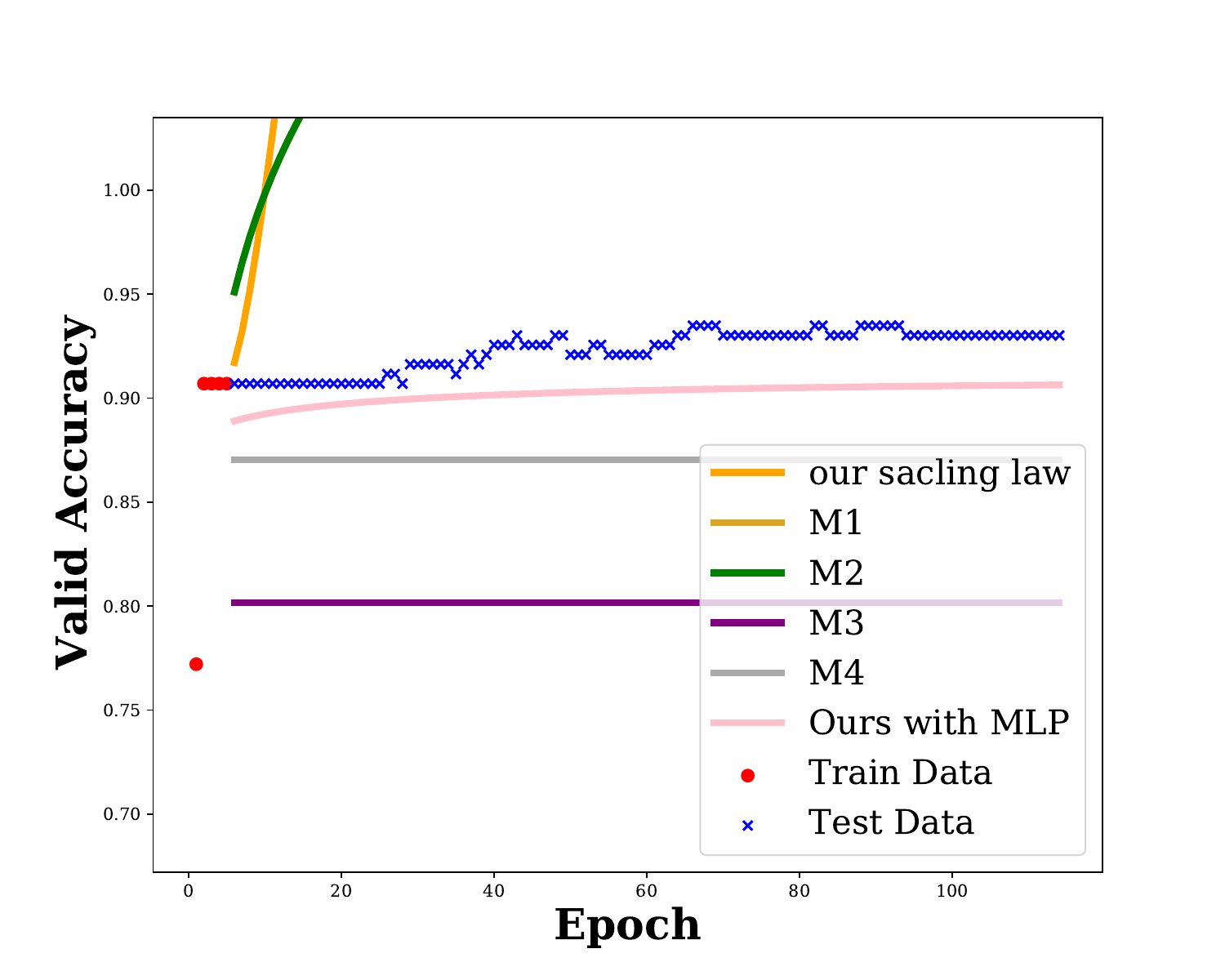}
    \centering
    {\scriptsize \mbox{(j) {baseball}}}
    \end{minipage}
    \begin{minipage}{0.24\linewidth}
    \includegraphics[width=\textwidth]{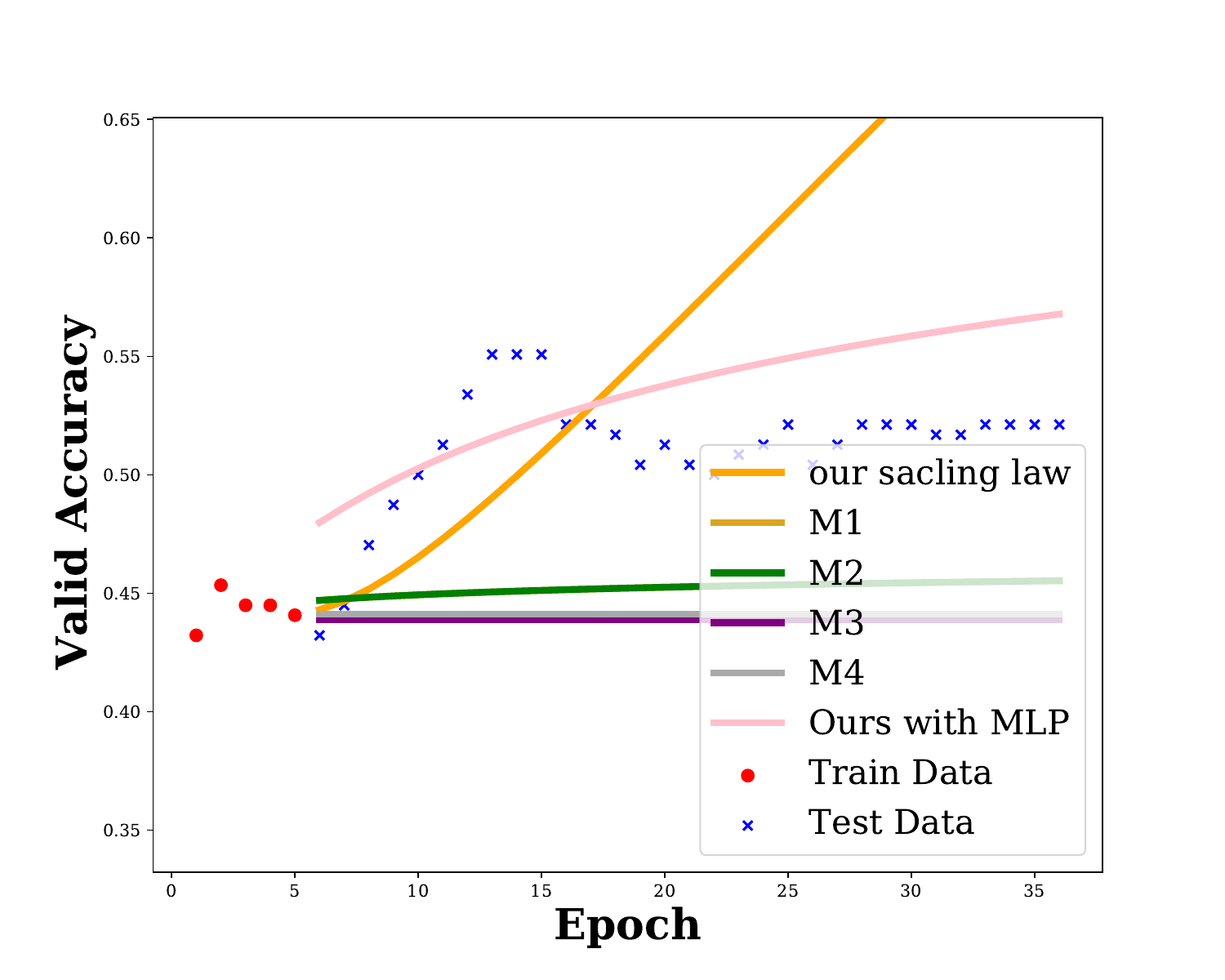}
    \centering
    {\scriptsize \mbox{(k) {cmc}}}
    \end{minipage}
    \begin{minipage}{0.24\linewidth}
    \includegraphics[width=\textwidth]{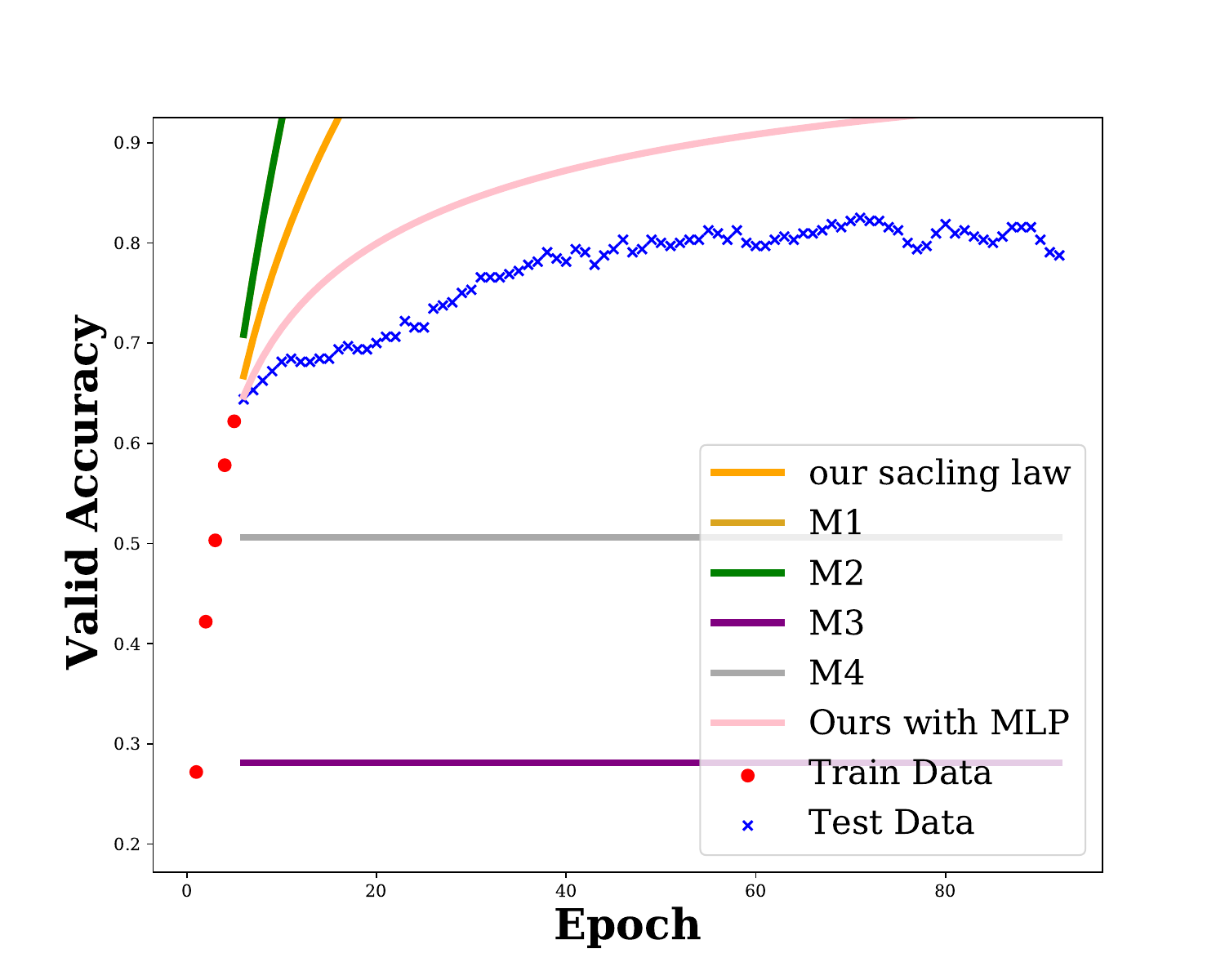}
    \centering
    {\scriptsize \mbox{(l) {mfeat-fourier}}}
    \end{minipage}
    \begin{minipage}{0.24\linewidth}
    \includegraphics[width=\textwidth]{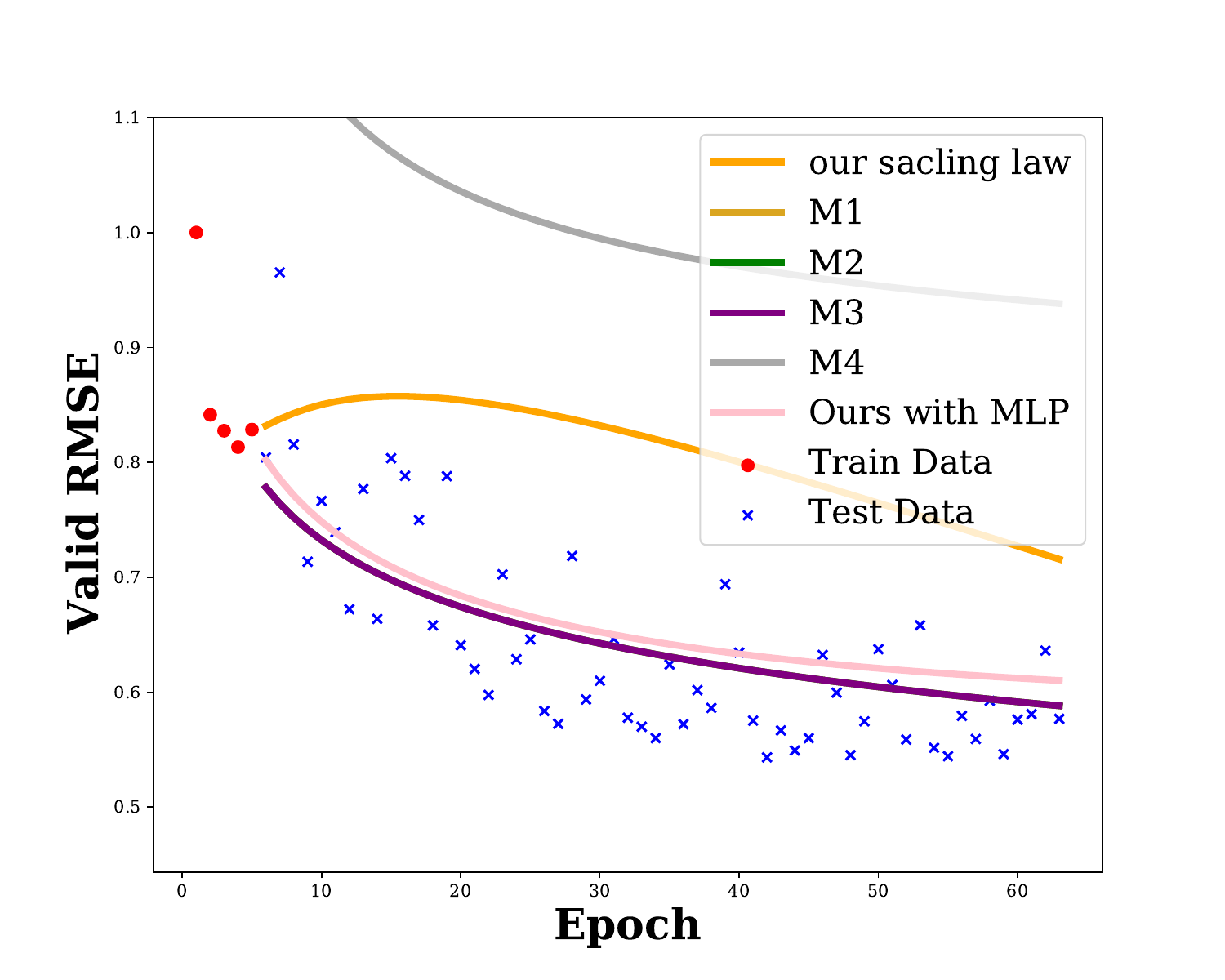}
    \centering
    {\scriptsize \mbox{(m) {Facebook\_...\_Volume}}}
    \end{minipage}
    \begin{minipage}{0.24\linewidth}
    \includegraphics[width=\textwidth]{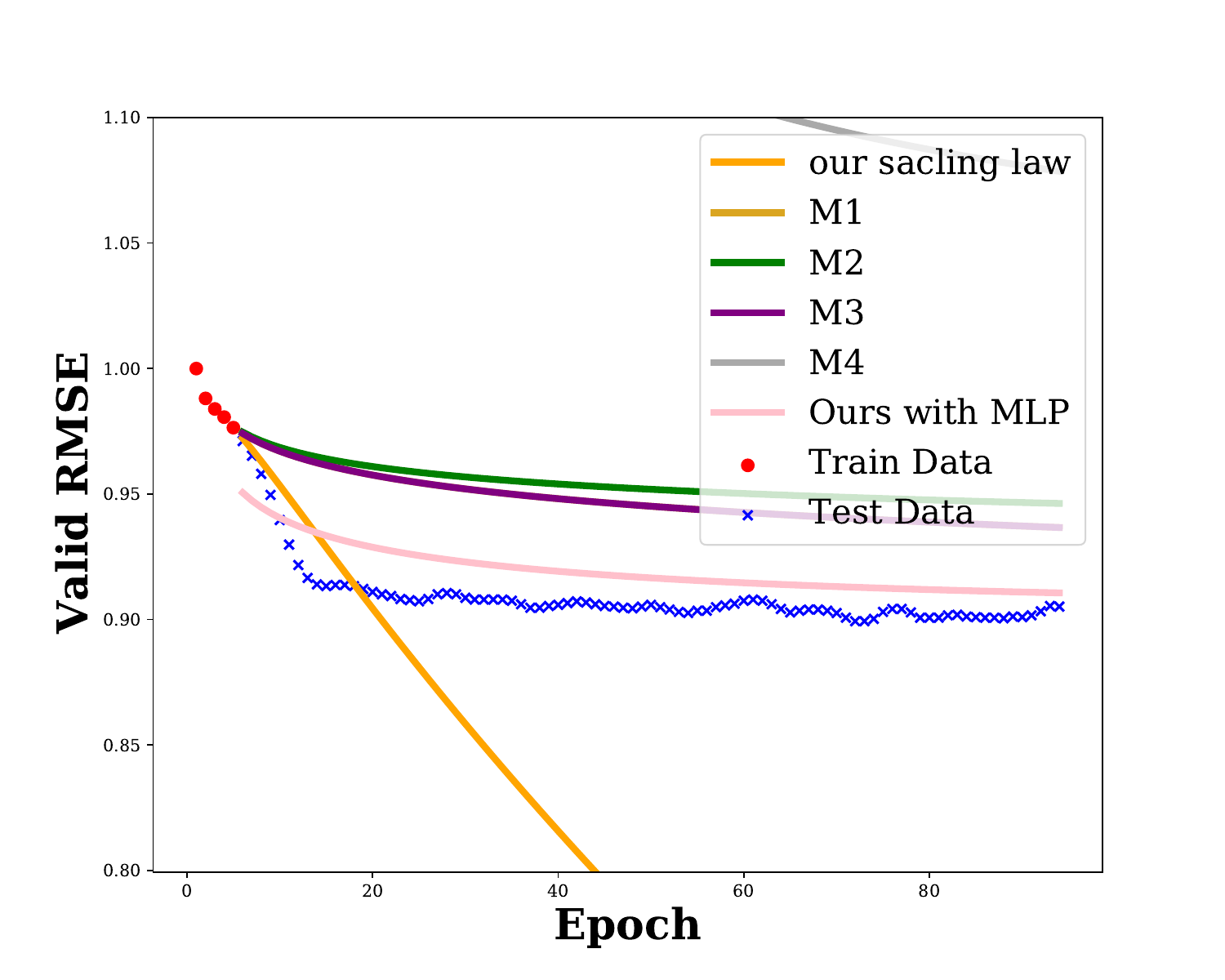}
    \centering
    {\scriptsize \mbox{(n) {Performance-Prediction}}}
    \end{minipage}
    \begin{minipage}{0.24\linewidth}
    \includegraphics[width=\textwidth]{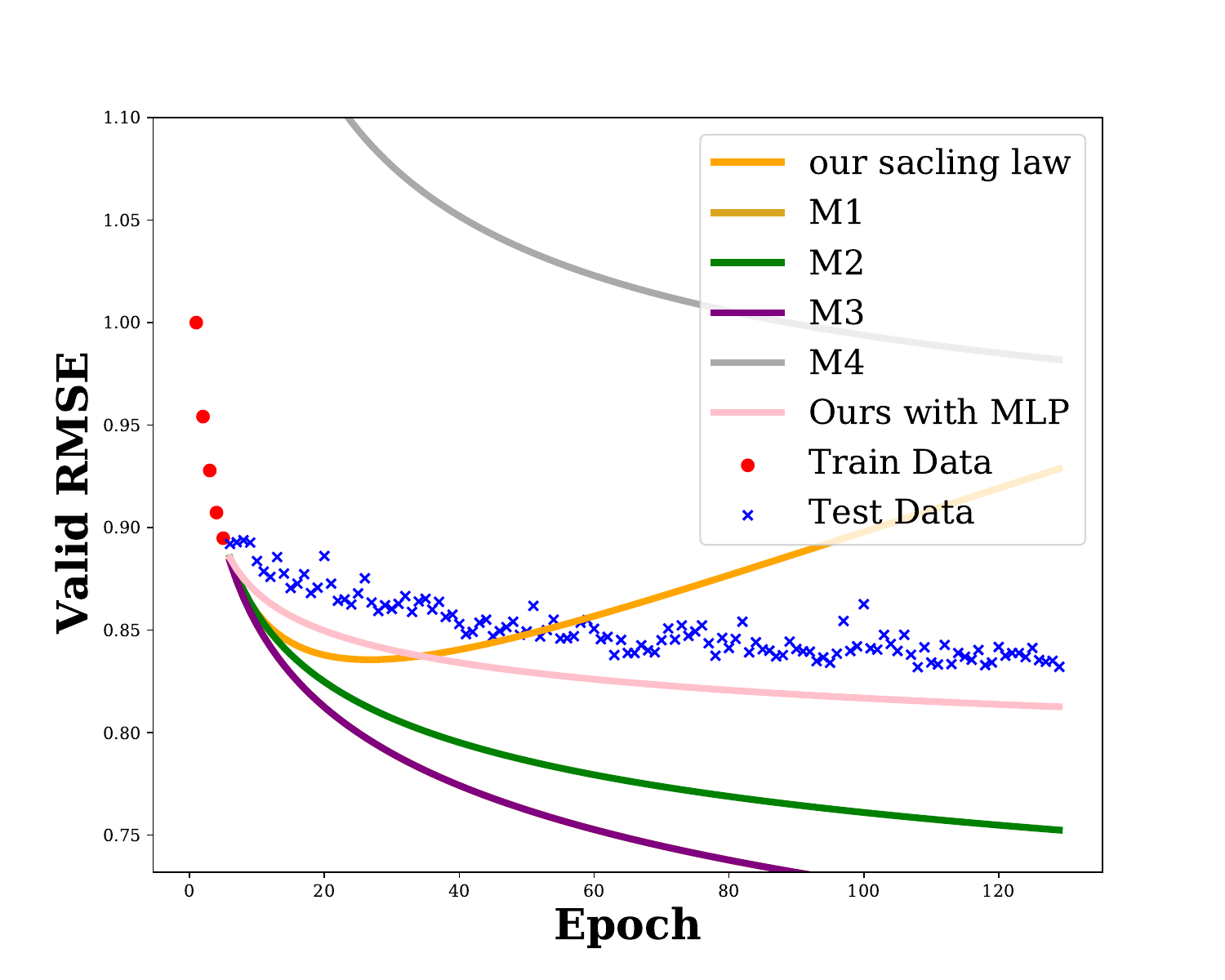}
    \centering
    {\scriptsize \mbox{(o) {wine+quality}}}
    \end{minipage}
    \begin{minipage}{0.24\linewidth}
    \includegraphics[width=\textwidth]{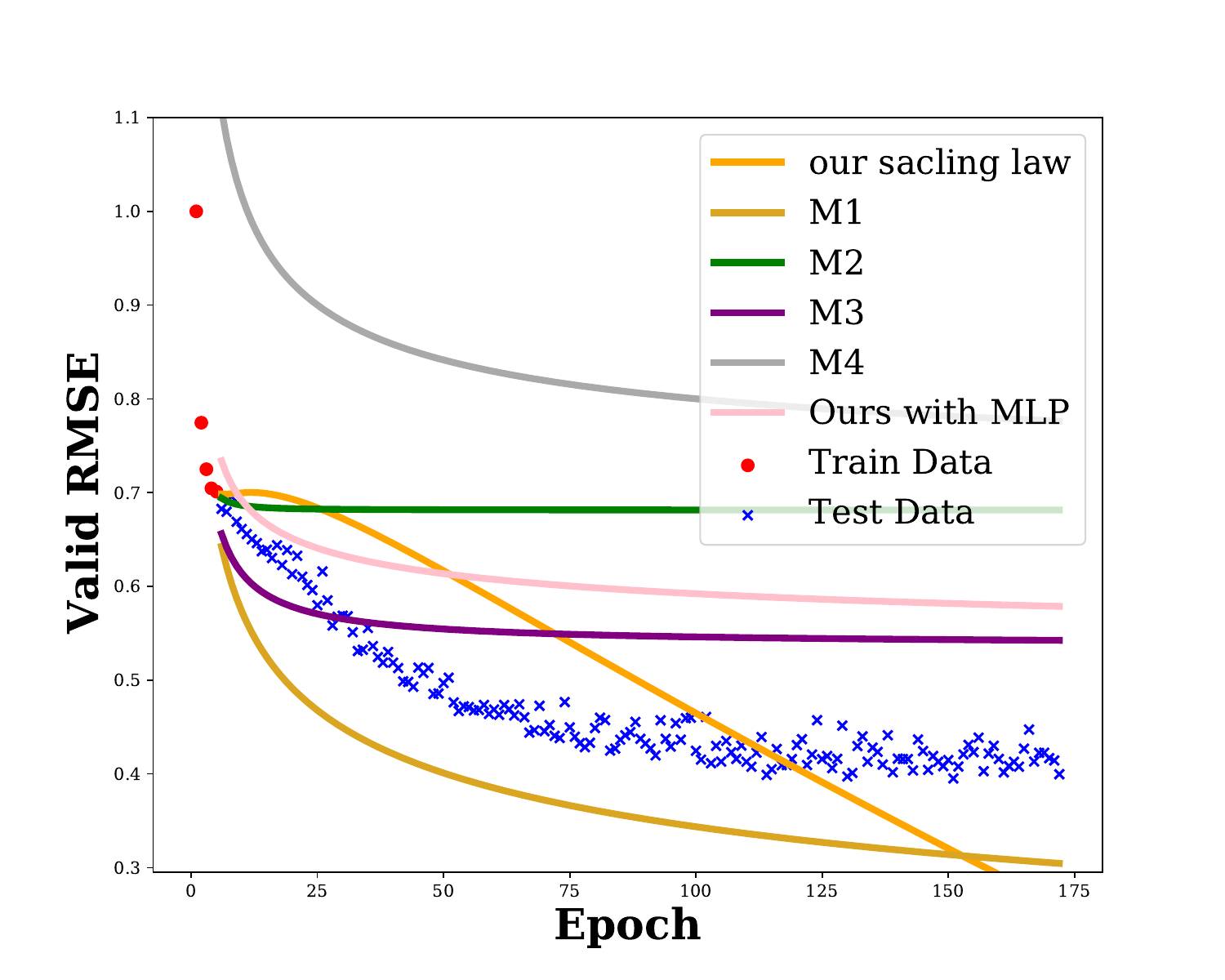}
    \centering
    {\scriptsize \mbox{(p) {Diamonds}}}
    \end{minipage}
  \caption{Visualization of the training dynamics fitting (validation curves of an MLP trained with default hyperparameters) on 16 unseen datasets. The datasets in the first two rows and the last two rows are DNN-friendly and Tree-friendly, respectively. The first and third rows represent classification tasks, while the second and fourth rows represent regression tasks.}
  \label{fig:dynamics_16}
\end{figure}

\autoref{fig:dynamics_16} illustrates qualitative fits for 16 unseen datasets. The predictor $h$ successfully reconstructs both DNN-friendly and tree-friendly learning curves, spanning classification and regression tasks. Compared to baseline curve families, our method delivers more faithful extrapolations, especially when meta-features are included.  

In summary, these results confirm that linking dataset meta-features to training dynamics is effective for characterizing heterogeneity. By capturing how dataset properties shape optimization trajectories, our approach not only predicts validation curves more accurately but also deepens understanding of why models succeed or fail on specific tabular datasets.

\subsection{By-Product: Forecasting Training Dynamics for Efficiency}
While our primary goal is to use dynamic forecasting as a tool for heterogeneity analysis, it also yields a practical by-product: efficient early stopping. Since deep tabular training is often expensive and hyperparameter-sensitive, forecasting later performance from early epochs allows us to prune poor runs. For example, if accuracy plateaus or oscillates early, training can be terminated and resources reallocated~\citep{CortesJSVD93Learning}.  

Thus, although not our main focus, this framework can also guide adaptive training strategies while primarily serving as an analytical tool for understanding heterogeneity in tabular datasets.

\begin{figure}[t]
  \centering
   \begin{minipage}{0.45\linewidth}
    \includegraphics[width=\textwidth]{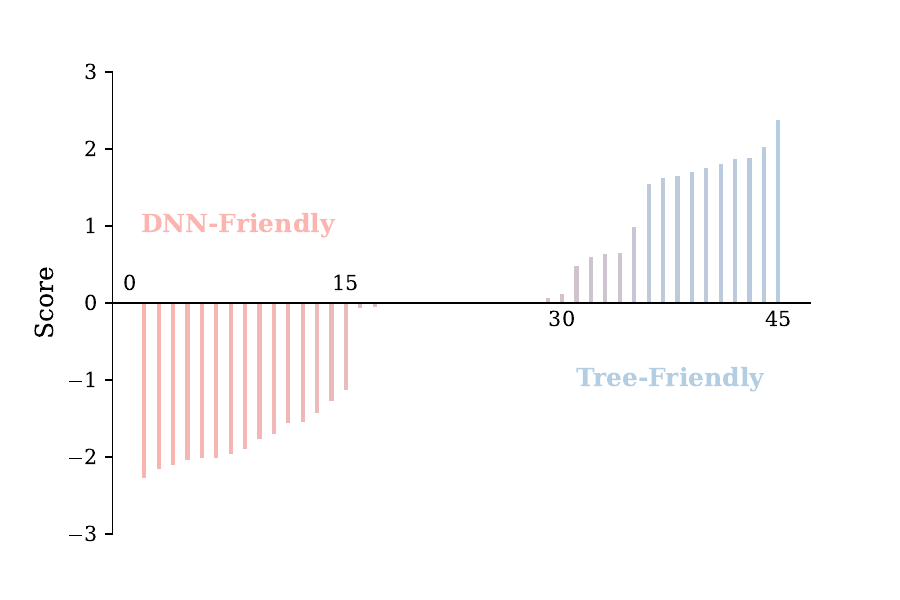}
    \centering
    \end{minipage}
    \caption{Distribution of Tree--DNN scores across the 45 datasets in {\name}-tiny, which illustrates the balanced nature of this curated subset. }
  \label{fig:score_addition_tiny}
\end{figure}

\section{More Details on {\scshape Talent}-Tiny}
\label{app:talent_tiny}
\autoref{fig:score_addition_tiny} illustrates the distribution of Tree--DNN scores across the 45 datasets in {\name}-tiny, a curated subset of {\name} designed for fine-grained analysis.
Each score measures the relative advantage of tree-based versus neural models on the same dataset.
Datasets on the left are DNN-friendly, where deep models such as RealMLP and MNCA perform better, while those on the right are tree-friendly, favoring ensembles like CatBoost.
The overall distribution is approximately symmetric, reflecting that {\name}-tiny maintains a balanced mixture of both categories.
This design facilitates controlled evaluations of model behaviors and helps disentangle algorithmic factors from dataset bias, complementing the large-scale results reported in the main benchmark.

\end{appendices}

\vskip 0.2in
\bibliography{main}

@article{guyon2019analysis,
  title={Analysis of the AutoML challenge series},
  author={Guyon, Isabelle and Sun-Hosoya, Lisheng and Boull{\'e}, Marc and Escalante, Hugo Jair and Escalera, Sergio and Liu, Zhengying and Jajetic, Damir and Ray, Bisakha and Saeed, Mehreen and Sebag, Mich{\`e}le and others},
  journal={Automated Machine Learning},
  volume={177},
  pages= {177--219},
  year={2019}
}

@article{vanschoren2014openml,
  title={OpenML: networked science in machine learning},
  author={Vanschoren, Joaquin and Van Rijn, Jan N and Bischl, Bernd and Torgo, Luis},
  journal={ACM SIGKDD Explorations Newsletter},
  volume={15},
  number={2},
  pages={49--60},
  year={2014}
}

@article{YeZJZ21,
  author       = {Han-Jia Ye and
                  De-Chuan Zhan and
                  Yuan Jiang and
                  Zhi-Hua Zhou},
  title        = {Heterogeneous Few-Shot Model Rectification With Semantic Mapping},
  journal      = {{IEEE} Transactions on pattern analysis and machine intelligence},
  volume       = {43},
  number       = {11},
  pages        = {3878--3891},
  year         = {2021},

}

@inproceedings{ke2017lightgbm,
  title={Lightgbm: A highly efficient gradient boosting decision tree},
  author={Ke, Guolin and Meng, Qi and Finley, Thomas and Wang, Taifeng and Chen, Wei and Ma, Weidong and Ye, Qiwei and Liu, Tie-Yan},
pages        = {3146--3154},
  booktitle ={NIPS},
  year         = {2017}
}

@article{ahmed2010empirical,
  title={An empirical comparison of machine learning models for time series forecasting},
  author={Ahmed, Nesreen K and Atiya, Amir F and Gayar, Neamat El and El-Shishiny, Hisham},
  journal={Econometric reviews},
  volume={29},
  number={5-6},
  pages={594--621},
  year={2010}
}

@inproceedings{richardson2007predicting,
  title={Predicting clicks: estimating the click-through rate for new ads},
  author={Richardson, Matthew and Dominowska, Ewa and Ragno, Robert},
  booktitle={WWW},
  pages        = {521--530},
  year={2007}
}

@inproceedings{Net-DNF,
  author       = {Liran Katzir and
                  Gal Elidan and
                  Ran El-Yaniv},
  title        = {Net-DNF: Effective Deep Modeling of Tabular Data},
  booktitle    = {ICLR},
  year         = {2021}
}

@article{nederstigt2014floppies,
  title={FLOPPIES: a framework for large-scale ontology population of product information from tabular data in e-commerce stores},
  author={Nederstigt, Lennart J and Aanen, Steven S and Vandic, Damir and Frasincar, Flavius},
  journal={Decision Support Systems},
  volume={59},
  pages={296--311},
  year={2014}
}

@article{HassanAHK20,
  author       = {Md. Rafiul Hassan and
                  Sadiq Al-Insaif and
                  Muhammad Imtiaz Hossain and
                  Joarder Kamruzzaman},
  title        = {A machine learning approach for prediction of pregnancy outcome following
                  {IVF} treatment},
  journal      = {Neural Computing and Applications},
  volume       = {32},
  number       = {7},
  pages        = {2283--2297},
  year         = {2020},
}

@inproceedings{chen2016xgboost,
  title={Xgboost: A scalable tree boosting system},
  author={Chen, Tianqi and Guestrin, Carlos},
  booktitle={KDD},
pages        = {785--794},
  year={2016}
}

@inproceedings{Kadra2021Well,
  author       = {Arlind Kadra and
                  Marius Lindauer and
                  Frank Hutter and
                  Josif Grabocka},
  title        = {Well-tuned Simple Nets Excel on Tabular Datasets},
  booktitle    = {NeurIPS},
  pages        = {23928--23941},
  year         = {2021}
}

@inproceedings{PopovMB20Neural,
  author       = {Sergei Popov and
                  Stanislav Morozov and
                  Artem Babenko},
  title        = {Neural Oblivious Decision Ensembles for Deep Learning on Tabular Data},
  booktitle    = {ICLR},
  year         = {2020}
}

@inproceedings{ArikP21TabNet,
  author       = {Sercan {\"{O}}. Arik and
                  Tomas Pfister},
  title        = {TabNet: Attentive Interpretable Tabular Learning},
pages        = {6679--6687},
  booktitle    = {AAAI},
  year         = {2021}
}

@inproceedings{UcarHE21SubTab,
  author       = {Talip Ucar and
                  Ehsan Hajiramezanali and
                  Lindsay Edwards},
  title        = {SubTab: Subsetting Features of Tabular Data for Self-Supervised Representation Learning},
  booktitle    = {NeurIPS},
  pages        = {18853--18865},
  year         = {2021}
}

@inproceedings{Chang0G22NODEGAM,
  author       = {Chun-Hao Chang and
                  Rich Caruana and
                  Anna Goldenberg},
  title        = {{NODE-GAM:} Neural Generalized Additive Model for Interpretable Deep
                  Learning},
  booktitle    = {ICLR},
  year         = {2022}
}

@inproceedings{BahriJTM22Scarf,
  author       = {Dara Bahri and
                  Heinrich Jiang and
                  Yi Tay and
                  Donald Metzler},
  title        = {Scarf: Self-Supervised Contrastive Learning using Random Feature Corruption},
  booktitle    = {ICLR},
  year         = {2022}
}

@article{Rubachev2022revisiting,
  author       = {Ivan Rubachev and
                  Artem Alekberov and
                  Yury Gorishniy and
                  Artem Babenko},
  title        = {Revisiting Pretraining Objectives for Tabular Deep Learning},
  journal      = {CoRR},
  volume       = {abs/2207.03208},
  year         = {2022}
}

@inproceedings{KlambauerUMH17Self,
  author       = {G{\"{u}}nter Klambauer and
                  Thomas Unterthiner and
                  Andreas Mayr and
                  Sepp Hochreiter},
  title        = {Self-Normalizing Neural Networks},
  booktitle    = {NIPS},
  pages        = {971--980},
  year         = {2017}
}

@inproceedings{GorishniyRKB21Revisiting,
  author       = {Yury Gorishniy and
                  Ivan Rubachev and
                  Valentin Khrulkov and
                  Artem Babenko},
  title        = {Revisiting Deep Learning Models for Tabular Data},
pages        = {18932--18943},
  booktitle    = {NeurIPS},
  year         = {2021}
}

@inproceedings{SongS0DX0T19AutoInt,
  author       = {Weiping Song and
                  Chence Shi and
                  Zhiping Xiao and
                  Zhijian Duan and
                  Yewen Xu and
                  Ming Zhang and
                  Jian Tang},
  title        = {AutoInt: Automatic Feature Interaction Learning via Self-Attentive
                  Neural Networks},
pages        = {1161--1170},
  booktitle    = {CIKM},
  year         = {2019}
}

@inproceedings{WangSCJLHC21DCNv2,
  author       = {Ruoxi Wang and
                  Rakesh Shivanna and
                  Derek Zhiyuan Cheng and
                  Sagar Jain and
                  Dong Lin and
                  Lichan Hong and
                  Ed H. Chi},
  title        = {{DCN} {V2:} Improved Deep {\&} Cross Network and Practical Lessons
                  for Web-scale Learning to Rank Systems},
pages        = {1785--1797},
  booktitle    = {WWW},
  year         = {2021}
}

@inproceedings{WangFFW17DCN,
  author       = {Ruoxi Wang and
                  Bin Fu and
                  Gang Fu and
                  Mingliang Wang},
  title        = {Deep {\&} Cross Network for Ad Click Predictions},
  booktitle    = {ADKDD},
  pages        = {12:1--12:7},
  year         = {2017}
}

@article{Badirli2020GrowNet,
  author       = {Sarkhan Badirli and
                  Xuanqing Liu and
                  Zhengming Xing and
                  Avradeep Bhowmik and
                  Sathiya S. Keerthi},
  title        = {Gradient Boosting Neural Networks: GrowNet},
  journal      = {CoRR},
  volume       = {abs/2002.07971},
  year         = {2020}
}

@inproceedings{GuoTYLH17DeepFM,
  author       = {Huifeng Guo and
                  Ruiming Tang and
                  Yunming Ye and
                  Zhenguo Li and
                  Xiuqiang He},
  title        = {DeepFM: {A} Factorization-Machine based Neural Network for {CTR} Prediction},
pages        = {1725--1731},
  booktitle    = {IJCAI},
  year         = {2017}
}

@article{ZivA22Tabular,
  author       = {Ravid Shwartz-Ziv and
                  Amitai Armon},
  title        = {Tabular data: Deep learning is not all you need},
  journal      = {Information Fusion},
  volume       = {81},
  pages        = {84--90},
  year         = {2022}
}

@inproceedings{Hollmann2022TabPFN,
  author       = {Noah Hollmann and
                  Samuel M{\"{u}}ller and
                  Katharina Eggensperger and
                  Frank Hutter},
  title        = {TabPFN: {A} Transformer That Solves Small Tabular Classification Problems
                  in a Second},
  booktitle    = {ICLR},
  year         = {2023}
}

@inproceedings{Chen2024Team,
  author       = {Jiahuan Yan and
				  Jintai Chen and
				  Qianxing Wang and
                  Danny Ziyi Chen and
                  Jian Wu},
  title        = {Team up GBDTs and DNNs: Advancing Efficient and Effective Tabular Prediction with Tree-hybrid MLPs},
  booktitle    = {KDD},
  pages        = {3679--3689},
  year         = {2024}
}

@inproceedings{Gorishniy2022On,
  author       = {Yury Gorishniy and
                  Ivan Rubachev and
                  Artem Babenko},
  title        = {On Embeddings for Numerical Features in Tabular Deep Learning},
 pages = {24991--25004},
  booktitle    = {NeurIPS},
  year         = {2022}
}

@inproceedings{Prokhorenkova2018Catboost,
  author       = {Liudmila Ostroumova Prokhorenkova and
                  Gleb Gusev and
                  Aleksandr Vorobev and
                  Anna Veronika Dorogush and
                  Andrey Gulin},
  title        = {CatBoost: unbiased boosting with categorical features},
pages        = {6639--6649},
  booktitle    = {NeurIPS},
  year         = {2018}
}

@article{Borisov2024Deep,
  author       = {Vadim Borisov and
                  Tobias Leemann and
                  Kathrin Se{\ss}ler and
                  Johannes Haug and
                  Martin Pawelczyk and
                  Gjergji Kasneci},
  title        = {Deep Neural Networks and Tabular Data: {A} Survey},
  journal      = {{IEEE} Transactions Neural Networks and Learning Systems},
  volume       = {35},
  number       = {6},
  pages        = {7499--7519},
  year         = {2024}
}

@inproceedings{ZhangDW16Deep,
  author       = {Weinan Zhang and
                  Tianming Du and
                  Jun Wang},
  title        = {Deep Learning over Multi-field Categorical Data - - {A} Case Study
                  on User Response Prediction},
  booktitle    = {ECIR},
  pages        = {45-57},
  year         = {2016}
}

@inproceedings{Cheng2016Wide,
  author       = {Heng-Tze Cheng and
                  Levent Koc and
                  Jeremiah Harmsen and
                  Tal Shaked and
                  Tushar Chandra and
                  Hrishi Aradhye and
                  Glen Anderson and
                  Greg Corrado and
                  Wei Chai and
                  Mustafa Ispir and
                  Rohan Anil and
                  Zakaria Haque and
                  Lichan Hong and
                  Vihan Jain and
                  Xiaobing Liu and
                  Hemal Shah},
  title        = {Wide {\&} Deep Learning for Recommender Systems},
pages        = {7--10},
  booktitle    = {DLRS},
  year         = {2016}
}

@inproceedings{ChenLWCW22DAN,
  author       = {Jintai Chen and
                  Kuanlun Liao and
                  Yao Wan and
                  Danny Z. Chen and
                  Jian Wu},
  title        = {DANets: Deep Abstract Networks for Tabular Data Classification and
                  Regression},
  pages        = {3930--3938},
  booktitle    = {AAAI},
  year         = {2022}
}

@inproceedings{Dinh2022LIFT,
  author       = {Tuan Dinh and
                  Yuchen Zeng and
                  Ruisu Zhang and
                  Ziqian Lin and
                  Michael Gira and
                  Shashank Rajput and
                  Jy-yong Sohn and
                  Dimitris S. Papailiopoulos and
                  Kangwook Lee},
  title        = {{LIFT:} Language-Interfaced Fine-Tuning for Non-Language Machine Learning
                  Tasks},
 pages = {11763--11784},
  booktitle    = {NeurIPS},
  year         = {2022}
}

@inproceedings{Somepalli2021SAINT,
title={{SAINT}: Improved Neural Networks for Tabular Data via Row Attention and Contrastive Pre-Training},
author={Gowthami Somepalli and Avi Schwarzschild and Micah Goldblum and C. Bayan Bruss and Tom Goldstein},
booktitle={NeurIPS Workshop},
year={2022},
}

@inproceedings{Hegselmann2022TabLLM,
  title={TabLLM: few-shot classification of tabular data with large language models},
  author={Hegselmann, Stefan and Buendia, Alejandro and Lang, Hunter and Agrawal, Monica and Jiang, Xiaoyi and Sontag, David},
  booktitle={AISTATS},
  pages        = {5549--5581},
  year={2023}
}

@inproceedings{Grinsztajn2022Why,
  author       = {L{\'{e}}o Grinsztajn and
                  Edouard Oyallon and
                  Ga{\"{e}}l Varoquaux},
  title        = {Why do tree-based models still outperform deep learning on typical
                  tabular data?},
pages = {507--520},
  booktitle    = {NeurIPS},
  year         = {2022}
}

@inproceedings{Padhi2021Tabular,
  author       = {Inkit Padhi and
                  Yair Schiff and
                  Igor Melnyk and
                  Mattia Rigotti and
                  Youssef Mroueh and
                  Pierre L. Dognin and
                  Jerret Ross and
                  Ravi Nair and
                  Erik Altman},
  title        = {Tabular Transformers for Modeling Multivariate Time Series},
  booktitle    = {ICASSP},
  year         = {2021}
}

@article{Huang2020TabTransformer,
 author       = {Xin Huang and
                  Ashish Khetan and
                  Milan Cvitkovic and
                  Zohar S. Karnin},
  title        = {TabTransformer: Tabular Data Modeling Using Contextual Embeddings},
  journal      = {CoRR},
  volume       = {abs/2012.06678},
  year         = {2020}
}

@inproceedings{Chen2023TabCaps,
  author       = {Jintai Chen and
                  KuanLun Liao and
                  Yanwen Fang and
                  Danny Chen and
                  Jian Wu},
  title        = {TabCaps: A Capsule Neural Network for Tabular Data Classification with BoW Routing},
  booktitle    = {ICLR},
  year         = {2023}
}

@inproceedings{jeffares2023tangos,
  author={Jeffares, Alan and Liu, Tennison and Crabb{\'e}, Jonathan and Imrie, Fergus and van der Schaar, Mihaela},
  title        = {TANGOS: Regularizing Tabular Neural Networks through Gradient Orthogonalization and Specialization},
  booktitle    = {ICLR},
  year         = {2023}
}

@inproceedings{gorishniy2023tabr,
  author={Gorishniy, Yury and Rubachev, Ivan and Kartashev, Nikolay and Shlenskii, Daniil and Kotelnikov, Akim and Babenko, Artem},
  title        = {TabR: Tabular Deep Learning Meets Nearest Neighbors},
  booktitle    = {ICLR},
  year         = {2024}
}

@misc{superconductivty,
  author       = {Hamidieh,Kam},
  title        = {{Superconductivty Data}},
  year         = {2018},
  howpublished = {UCI Machine Learning Repository},
  note         = {{DOI}: https://doi.org/10.24432/C53P47}
}

@inproceedings{vaswani2017attention,
  title={Attention is all you need},
  author={Vaswani, Ashish and Shazeer, Noam and Parmar, Niki and Uszkoreit, Jakob and Jones, Llion and Gomez, Aidan N and Kaiser, {\L}ukasz and Polosukhin, Illia},
  booktitle={NIPS},
  pages        = {5998--6008},
  year={2017}
}

@inproceedings{akiba2019optuna,
  title={Optuna: A next-generation hyperparameter optimization framework},
  author={Akiba, Takuya and Sano, Shotaro and Yanase, Toshihiko and Ohta, Takeru and Koyama, Masanori},
  booktitle={KDD},
  pages = {2623--2631},
  year={2019}
}

@book{goodfellow2016deep,
  title={Deep learning},
  author={Goodfellow, Ian and Bengio, Yoshua and Courville, Aaron},
  year={2016},
  publisher={MIT press}
}

@inproceedings{devlin2018bert,
author       = {Jacob Devlin and
                  Ming-Wei Chang and
                  Kenton Lee and
                  Kristina Toutanova},
  title        = {{BERT:} Pre-training of Deep Bidirectional Transformers for Language Understanding},
pages        = {4171--4186},
  booktitle    = {NAACL-HLT},
  year         = {2019}
}

@inproceedings{simonyan2014very,
  author       = {Karen Simonyan and
                  Andrew Zisserman},
  title        = {Very Deep Convolutional Networks for Large-Scale Image Recognition},
  booktitle    = {ICLR},
  year         = {2015}

}

@inproceedings{Vinyals2016Matching,
  author       = {Oriol Vinyals and
                  Charles Blundell and
                  Tim Lillicrap and
                  Koray Kavukcuoglu and
                  Daan Wierstra},
  title        = {Matching Networks for One Shot Learning},
pages        = {3630--3638},
  booktitle    = {NIPS},
  year         = {2016}
}

@book{bishop2006pattern,
  title={Pattern recognition and machine learning},
  author={Bishop, Christopher},
  year={2006},
  publisher={Springer}
}

@book{Mohri2012FoML,
  author       = {Mehryar Mohri and
                  Afshin Rostamizadeh and
                  Ameet Talwalkar},
  title        = {Foundations of Machine Learning},
  publisher    = {{MIT} Press},
  year         = {2012}
}

@inproceedings{GoldbergerRHS04,
  author       = {Jacob Goldberger and
                  Sam T. Roweis and
                  Geoffrey E. Hinton and
                  Ruslan Salakhutdinov},
  title        = {Neighbourhood Components Analysis},
  booktitle    = {NIPS},
  pages        = {513--520},
  year         = {2004}
}

@inproceedings{FeurerKESBH15,
  author       = {Matthias Feurer and
                  Aaron Klein and
                  Katharina Eggensperger and
                  Jost Tobias Springenberg and
                  Manuel Blum and
                  Frank Hutter},
  title        = {Efficient and Robust Automated Machine Learning},
  booktitle    = {NIPS},
  pages        = {2962--2970},
  year         = {2015}
}

@article{DelgadoCBA14,
  author       = {Manuel Fern{\'{a}}ndez Delgado and
                  Eva Cernadas and
                  Sen{\'{e}}n Barro and
                  Dinani Gomes Amorim},
  title        = {Do we need hundreds of classifiers to solve real world classification
                  problems?},
  journal      = {Journal of Machine Learning Research},
  volume       = {15},
  number       = {1},
  pages        = {3133--3181},
  year         = {2014}
}

@inproceedings{McElfreshKVCRGW23when,
  author       = {Duncan C. McElfresh and
                  Sujay Khandagale and
                  Jonathan Valverde and
                  Vishak Prasad C. and
                  Ganesh Ramakrishnan and
                  Micah Goldblum and
                  Colin White},
  title        = {When Do Neural Nets Outperform Boosted Trees on Tabular Data?},
  booktitle    = {NeurIPS},
pages = {76336--76369},
  year         = {2023}
}

@inproceedings{PTARL,
  author       = {Hangting Ye and Wei Fan and Xiaozhuang Song and Shun Zheng and He Zhao and Dan dan Guo and Yi Chang},
  title        = {PTaRL: Prototype-based Tabular Representation Learning via Space Calibration},
  booktitle    = {ICLR},
  year         = {2024},
}

@inproceedings{Wu2024SwitchTab,
  author       = {Jing Wu and
                  Suiyao Chen and
                  Qi Zhao and
                  Renat Sergazinov and
                  Chen Li and
                  Shengjie Liu and
                  Chongchao Zhao and
                  Tianpei Xie and
                  Hanqing Guo and
                  Cheng Ji and
                  Daniel Cociorva and
                  Hakan Brunzell},
  title        = {SwitchTab: Switched Autoencoders Are Effective Tabular Learners},
  booktitle    = {AAAI},
  pages        = {15924--15933},
  year         = {2024}
}

@inproceedings{JuanZCL16CTR,
  author       = {Yu-Chin Juan and
                  Yong Zhuang and
                  Wei-Sheng Chin and
                  Chih-Jen Lin},
  title        = {Field-aware Factorization Machines for {CTR} Prediction},
  booktitle    = {RecSys},
  pages        = {43--50},
  year         = {2016}
}

@inproceedings{YanLXH14Coupled,
  author       = {Ling Yan and
                  Wu-Jun Li and
                  Gui-Rong Xue and
                  Dingyi Han},
  title        = {Coupled Group Lasso for Web-Scale {CTR} Prediction in Display Advertising},
  booktitle    = {ICML},
  pages        = {802--810},
  year         = {2014}
}

@article{schwartz2007drug,
  title={The drug facts box: providing consumers with simple tabular data on drug benefit and harm},
  author={Schwartz, Lisa M and Woloshin, Steven and Welch, H Gilbert},
  journal={Medical Decision Making},
  volume={27},
  number={5},
  pages={655--662},
  year={2007}
}

@article{subasi2012medical,
  title={Medical decision support system for diagnosis of neuromuscular disorders using DWT and fuzzy support vector machines},
  author={Subasi, Abdulhamit},
  journal={Computers in Biology and Medicine},
  volume={42},
  number={8},
  pages={806--815},
  year={2012}
}

@inproceedings{NaderSL22DNNR,
  author       = {Youssef Nader and
                  Leon Sixt and
                  Tim Landgraf},
  title        = {{DNNR:} Differential Nearest Neighbors Regression},
  booktitle    = {ICML},
  pages        = {16296--16317},
  year         = {2022}
}

@inproceedings{LoshchilovH19AdamW,
  author       = {Ilya Loshchilov and
                  Frank Hutter},
  title        = {Decoupled Weight Decay Regularization},
  booktitle    = {ICLR},
  year         = {2019}
}

@inproceedings{AlabdulmohsinNZ22Revisiting,
  author       = {Ibrahim M. Alabdulmohsin and
                  Behnam Neyshabur and
                  Xiaohua Zhai},
  title        = {Revisiting Neural Scaling Laws in Language and Vision},
  pages = {22300--22312},
  booktitle    = {NeurIPS},
  year         = {2022}
}

@inproceedings{RosenfeldRBS20Constructive,
  author       = {Jonathan S. Rosenfeld and
                  Amir Rosenfeld and
                  Yonatan Belinkov and
                  Nir Shavit},
  title        = {A Constructive Prediction of the Generalization Error Across Scales},
  booktitle    = {ICLR},
  year         = {2020}
}

@article{Hestness2017Deep,
  author       = {Joel Hestness and
                  Sharan Narang and
                  Newsha Ardalani and
                  Gregory F. Diamos and
                  Heewoo Jun and
                  Hassan Kianinejad and
                  Md. Mostofa Ali Patwary and
                  Yang Yang and
                  Yanqi Zhou},
  title        = {Deep Learning Scaling is Predictable, Empirically},
  journal      = {CoRR},
  volume       = {abs/1712.00409},
  year         = {2017}
}

@article{Bahri2021Explaining,
  author       = {Yasaman Bahri and
                  Ethan Dyer and
                  Jared Kaplan and
                  Jaehoon Lee and
                  Utkarsh Sharma},
  title        = {Explaining Neural Scaling Laws},
  journal      = {CoRR},
  volume       = {abs/2102.06701},
  year         = {2021}
}

@inproceedings{Abnar0NS22Exploring,
  author       = {Samira Abnar and
                  Mostafa Dehghani and
                  Behnam Neyshabur and
                  Hanie Sedghi},
  title        = {Exploring the Limits of Large Scale Pre-training},
  booktitle    = {ICLR},
  year         = {2022}
}

@inproceedings{CortesJSVD93Learning,
  author       = {Corinna Cortes and
                  Lawrence D. Jackel and
                  Sara A. Solla and
                  Vladimir Vapnik and
                  John S. Denker},
  title        = {Learning Curves: Asymptotic Values and Rate of Convergence},
  booktitle    = {NIPS},
  pages        = {327--334},
  year         = {1993}
}

@article{Wilstrup2021Symbolic,
  author       = {Casper Wilstrup and
                  Jaan Kasak},
  title        = {Symbolic regression outperforms other models for small data sets},
  journal      = {CoRR},
  volume       = {abs/2103.15147},
  year         = {2021}
}

@inproceedings{CavaOBFVJKM21Contemporary,
  author       = {William G. La Cava and
                  Patryk Orzechowski and
                  Bogdan Burlacu and
                  Fabr{\'{\i}}cio Olivetti de Fran{\c{c}}a and
                  Marco Virgolin and
                  Ying Jin and
                  Michael Kommenda and
                  Jason H. Moore},
  title        = {Contemporary Symbolic Regression Methods and their Relative Performance},
  booktitle    = {NeurIPS Datasets and Benchmarks},
  year         = {2021}
}

@article{Rauf2024TableDC,
  author       = {Hafiz Tayyab Rauf and
                  Andr{\'{e}} Freitas and
                  Norman W. Paton},
  title        = {TableDC: Deep Clustering for Tabular Data},
  journal      = {CoRR},
  volume       = {abs/2405.17723},
  year         = {2024}
}

@inproceedings{Han2022ADBench,
  author       = {Songqiao Han and
                  Xiyang Hu and
                  Hailiang Huang and
                  Minqi Jiang and
                  Yue Zhao},
  title        = {ADBench: Anomaly Detection Benchmark},
  booktitle    = {NeurIPS},
  pages = {32142--32159},
  year         = {2022}
}

@inproceedings{Ye2024ModernNCA,
  author       = {Han-Jia Ye and
                  Huai-Hong Yin and
                  De-Chuan Zhan and
                  Wei-Lun Chao},
  title        = {Revisiting Nearest Neighbor for Tabular Data: A Deep Tabular Baseline Two Decades Later},
  booktitle    = {ICLR},
  year         = {2025}
}

@inproceedings{David2024RealMLP,
  author       = {David Holzm{\"{u}}ller and
                  L{\'{e}}o Grinsztajn and
                  Ingo Steinwart},
  title        = {Better by Default: Strong Pre-Tuned MLPs and Boosted Trees on Tabular
                  Data},
  booktitle    = {NeurIPS},
 pages = {26577--26658},
  year         = {2024}
}

@article{Breiman01RandomForest,
  author       = {Leo Breiman},
  title        = {Random Forests},
  journal      = {Machine Learning},
  volume       = {45},
  number       = {1},
  pages        = {5--32},
  year         = {2001}
}

@article{tibshirani2002diagnosis,
  title={Diagnosis of multiple cancer types by shrunken centroids of gene expression},
  author={Tibshirani, Robert and Hastie, Trevor and Narasimhan, Balasubramanian and Chu, Gilbert},
  journal={Proceedings of the National Academy of Sciences},
  volume={99},
  number={10},
  pages={6567--6572},
  year={2002}
}

@article{friedman2001greedy,
  title={Greedy function approximation: a gradient boosting machine},
  author={Friedman, Jerome H},
  journal={Annals of statistics},
  pages={1189--1232},
  year={2001}
}

@article{friedman2002stochastic,
  title={Stochastic gradient boosting},
  author={Friedman, Jerome H},
  journal={Computational statistics \& data analysis},
  volume={38},
  number={4},
  pages={367--378},
  year={2002}
}

@book{HastieTF09ESL,
  author       = {Trevor Hastie and
                  Robert Tibshirani and
                  Jerome H. Friedman},
  title        = {The Elements of Statistical Learning: Data Mining, Inference, and
                  Prediction, 2nd Edition},
  publisher    = {Springer},
  year         = {2009}
}

@inproceedings{kohli2024towards,
  title={Towards Quantifying the Effect of Datasets for Benchmarking: A Look at Tabular Machine Learning},
  author={Kohli, Ravin and Feurer, Matthias and Eggensperger, Katharina and Bischl, Bernd and Hutter, Frank},
  booktitle={ICLR Workshop},
  year={2024}
}

@inproceedings{Tschalzev2024DataCentric,
  author       = {Andrej Tschalzev and
                  Sascha Marton and
                  Stefan L{\"{u}}dtke and
                  Christian Bartelt and
                  Heiner Stuckenschmidt},
  title        = {A Data-Centric Perspective on Evaluating Machine Learning Models for
                  Tabular Data},
  booktitle={NeurIPS},
 pages = {95896--95930},
  year={2024}
}

@inproceedings{JiangYW00Z24Tabular,
  author       = {Jun-Peng Jiang and
                  Han-Jia Ye and
                  Leye Wang and
                  Yang Yang and
                  Yuan Jiang and
                  De-Chuan Zhan},
  title        = {Tabular Insights, Visual Impacts: Transferring Expertise from Tables
                  to Images},
  booktitle    = {ICML},
 pages =          {21988--22009},
  year         = {2024}
}

@inproceedings{DengDSLL009ImageNet,
  author       = {Jia Deng and
                  Wei Dong and
                  Richard Socher and
                  Li-Jia Li and
                  Kai Li and
                  Li Fei-Fei},
  title        = {ImageNet: {A} large-scale hierarchical image database},
  booktitle    = {CVPR},
  pages        = {248--255},
  year         = {2009}
}

@article{MaciaBOH13Learner,
  author       = {N{\'{u}}ria Maci{\`{a}} and
                  Ester Bernad{\'{o}}-Mansilla and
                  Albert Orriols-Puig and
                  Tin Kam Ho},
  title        = {Learner excellence biased by data set selection: {A} case for data
                  characterisation and artificial data sets},
  journal      = {Pattern Recognition},
  volume       = {46},
  number       = {3},
  pages        = {1054--1066},
  year         = {2013}
}

@inproceedings{Chen2023Excel,
  author       = {Jintai Chen and
                  Jiahuan Yan and
                  Qiyuan Chen and
                  Danny Ziyi Chen and
                  Jian Wu and
                  Jimeng Sun},
  title        = {Can a Deep Learning Model be a Sure Bet for Tabular Prediction?},
  booktitle    = {KDD},
  pages        = {288--296},
  year         = {2024}
}

@inproceedings{XuSCV19TabGAN,
  author       = {Lei Xu and
                  Maria Skoularidou and
                  Alfredo Cuesta-Infante and
                  Kalyan Veeramachaneni},
  title        = {Modeling Tabular data using Conditional {GAN}},
  booktitle    = {NeurIPS},
  pages        = {7333--7343},
  year         = {2019}
}

@inproceedings{SvirskyL24Interpretable,
  author       = {Jonathan Svirsky and
                  Ofir Lindenbaum},
  title        = {Interpretable Deep Clustering for Tabular Data},
  booktitle    = {ICML},
pages={47314--47330},
  year         = {2024}
}

@inproceedings{VeroBV24CuTS,
  author       = {Mark Vero and
                  Mislav Balunovic and
                  Martin T. Vechev},
  title        = {CuTS: Customizable Tabular Synthetic Data Generation},
  booktitle    = {ICML},
pages={49408--49433},
  year         = {2024}
}

@inproceedings{HansenSSP23Reimagining,
  author       = {Lasse Hansen and
                  Nabeel Seedat and
                  Mihaela van der Schaar and
                  Andrija Petrovic},
  title        = {Reimagining Synthetic Tabular Data Generation through Data-Centric
                  {AI:} {A} Comprehensive Benchmark},
  booktitle    = {NeurIPS},
 pages = {33781--33823},
  year         = {2023}
}

@article{HouGXQ23Incremental,
  author       = {Chenping Hou and
                  Shilin Gu and
                  Chao Xu and
                  Yuhua Qian},
  title        = {Incremental Learning for Simultaneous Augmentation of Feature and
                  Class},
  journal      = {{IEEE} Transactions on pattern analysis and machine intelligence},
  volume       = {45},
  number       = {12},
  pages        = {14789--14806},
  year         = {2023}
}

@article{HouFZH23Adaptive,
  author       = {Chenping Hou and
                  Ruidong Fan and
                  Ling-Li Zeng and
                  Dewen Hu},
  title        = {Adaptive Feature Selection With Augmented Attributes},
  journal      = {{IEEE} Transactions on pattern analysis and machine intelligence},
  volume       = {45},
  number       = {8},
  pages        = {9306--9324},
  year         = {2023}
}

@article{XuTZHH23Label,
  author       = {Chao Xu and
                  Hong Tao and
                  Jing Zhang and
                  Dewen Hu and
                  Chenping Hou},
  title        = {Label Distribution Changing Learning with Sample Space Expanding},
  journal      = {Journal of Machine Learning Research},
  volume       = {24},
  pages        = {36:1--36:48},
  year         = {2023}
}

@inproceedings{ShenkarW22Anomaly,
  author       = {Tom Shenkar and
                  Lior Wolf},
  title        = {Anomaly Detection for Tabular Data with Internal Contrastive Learning},
  booktitle    = {ICLR},
  year         = {2022}
}

@inproceedings{YinQZW024MCM,
  author       = {Jiaxin Yin and
                  Yuanyuan Qiao and
                  Zitang Zhou and
                  Xiangchao Wang and
                  Jie Yang},
  title        = {{MCM:} Masked Cell Modeling for Anomaly Detection in Tabular Data},
  booktitle    = {ICLR},
  year         = {2024}
}

@inproceedings{Rubachev2024TabRed,
  author       = {Ivan Rubachev and
                  Nikolay Kartashev and
                  Yury Gorishniy and
                  Artem Babenko},
  title        = {TabReD: {A} Benchmark of Tabular Machine Learning in-the-Wild},
  booktitle    = {ICLR},
  year         = {2025}
}

@article{zhou2024core,
    author = {Zhou, Zhi-Hua},
    title = {Learnability with time-sharing computational resource concerns},
    journal = {National Science Review},
    volume = {11},
    number = {10},
    pages = {nwae204},
    year = {2024}
}

@article{M3,
  author       = {John Schulman and
                  Filip Wolski and
                  Prafulla Dhariwal and
                  Alec Radford and
                  Oleg Klimov},
  title        = {Proximal Policy Optimization Algorithms},
  journal      = {CoRR},
  volume       = {abs/1707.06347},
  year         = {2017}
}

@article{Demsar06Statistical,
  author       = {Janez Demsar},
  title        = {Statistical Comparisons of Classifiers over Multiple Data Sets},
  journal      = {Journal of Machine Learning Research},
  volume       = {7},
  pages        = {1--30},
  year         = {2006}
}

@inproceedings{Yury2024TabM,
  author       = {Yury Gorishniy and
                  Akim Kotelnikov and
                  Artem Babenko},
  title        = {{TabM}: Advancing Tabular Deep Learning with Parameter-Efficient Ensembling},
  booktitle    = {ICLR},
  year         = {2025}
}

@inproceedings{Marton2024GRANDE,
  author       = {Sascha Marton and
                  Stefan L{\"{u}}dtke and
                  Christian Bartelt and
                  Heiner Stuckenschmidt},
  title        = {{GRANDE:} Gradient-Based Decision Tree Ensembles for Tabular Data},
  booktitle    = {ICLR},
  year         = {2024}
}

@inproceedings{Jiang2024ProtoGate,
  author       = {Xiangjian Jiang and
                  Andrei Margeloiu and
                  Nikola Simidjievski and
                  Mateja Jamnik},
  title        = {ProtoGate: Prototype-based Neural Networks with Global-to-local Feature
                  Selection for Tabular Biomedical Data},
  booktitle    = {ICML},
  pages={21844--21878},
  year         = {2024}
}

@inproceedings{Xu2024BiSHop,
  author       = {Chenwei Xu and
                  Yu-Chao Huang and
                  Jerry Yao-Chieh Hu and
                  Weijian Li and
                  Ammar Gilani and
                  Hsi-Sheng Goan and
                  Han Liu},
  title        = {BiSHop: Bi-Directional Cellular Learning for Tabular Data with Generalized
                  Sparse Modern Hopfield Model},
  booktitle    = {ICML},
pages={55048--55075},
  year         = {2024}
}

@inproceedings{Thomas2024LocalPFN,
  author       = {Valentin Thomas and
                  Junwei Ma and
                  Rasa Hosseinzadeh and
                  Keyvan Golestan and
                  Guangwei Yu and
                  Maksims Volkovs and
                  Anthony L. Caterini},
  title        = {Retrieval {\&} Fine-Tuning for In-Context Tabular Models},
  booktitle    = {NeurIPS},
  pages = {108439--108467},
  year         = {2024}
}

@inproceedings{Feuer2024TuneTable,
  author       = {Benjamin Feuer and
                  Robin Tibor Schirrmeister and
                  Valeriia Cherepanova and
                  Chinmay Hegde and
                  Frank Hutter and
                  Micah Goldblum and
                  Niv Cohen and
                  Colin White},
  title        = {TuneTables: Context Optimization for Scalable Prior-Data Fitted Networks},
  booktitle    = {NeurIPS},
 pages = {83430--83464},
  year         = {2024}
}

@inproceedings{DBLP:conf/icml/Chen2023Trompt,
  author       = {Kuan-Yu Chen and
                  Ping-Han Chiang and
                  Hsin-Rung Chou and
                  Ting-Wei Chen and
                  Tien-Hao Chang},
  title        = {Trompt: Towards a Better Deep Neural Network for Tabular Data},
  booktitle    = {ICML},
  pages        = {4392--4434},
  year         = {2023}
}

@inproceedings{BonetMGI2024HyperFast,
  author       = {David Bonet and
                  Daniel Mas Montserrat and
                  Xavier Gir{\'{o}}-i-Nieto and
                  Alexander G. Ioannidis},
  title        = {HyperFast: Instant Classification for Tabular Data},
  booktitle    = {AAAI},
  pages        = {11114--11123},
  year         = {2024}
}

@article{Zhou2023TabToken,
  author       = {Qi-Le Zhou and
                  Han-Jia Ye and
                  Leye Wang and
                  De-Chuan Zhan},
  title        = {Unlocking the Transferability of Tokens in Deep Models for Tabular
                  Data},
  journal      = {CoRR},
  volume       = {abs/2310.15149},
  year         = {2023}
}

@article{Chao2020Meta,
  author       = {Wei-Lun Chao and
                  Han-Jia Ye and
                  De-Chuan Zhan and
                  Mark E. Campbell and
                  Kilian Q. Weinberger},
  title        = {Revisiting Meta-Learning as Supervised Learning},
  journal      = {CoRR},
  volume       = {abs/2002.00573},
  year         = {2020}
}

@article{WainbergAF16Are,
  author       = {Michael Wainberg and
                  Babak Alipanahi and
                  Brendan J. Frey},
  title        = {Are Random Forests Truly the Best Classifiers?},
  journal      = {Journal of Machine Learning Research},
  volume       = {17},
  pages        = {110:1--110:5},
  year         = {2016}
}

@inproceedings{Qu2025TabICL,
  author       = {Jingang Qu and
                  David Holzm{\"{u}}ller and
                  Ga{\"{e}}l Varoquaux and
                  Marine Le Morvan},
  title        = {TabICL: {A} Tabular Foundation Model for In-Context Learning on Large
                  Data},
  booktitle    = {ICML},
  year         = {2025}
}

@article{hollmann2025TabPFNv2,
  title={Accurate predictions on small data with a tabular foundation model},
  author={Hollmann, Noah and M{\"u}ller, Samuel and Purucker, Lennart and Krishnakumar, Arjun and K{\"o}rfer, Max and Hoo, Shi Bin and Schirrmeister, Robin Tibor and Hutter, Frank},
  journal={Nature},
  volume={637},
  number={8045},
  pages={319--326},
  year={2025}
}

@article{Wang2023UniPredict,
  author       = {Ruiyu Wang and
                  Zifeng Wang and
                  Jimeng Sun},
  title        = {UniPredict: Large Language Models are Universal Tabular Predictors},
  journal      = {CoRR},
  volume       = {abs/2310.03266},
  year         = {2023}
}

@inproceedings{Yak2024IngesTables,
  author       = {Scott Yak and
				  Yihe Dong and
				  Javier Gonzalvo and 
				  Sercan Ö. Arık},
  title        = {IngesTables: Scalable and Efficient Training of LLM-Enabled Tabular Foundation Models},
  booktitle    = {Table Representation Learning Workshop at NeurIPS 2023},
  year         = {2023}
}

@article{Erickson2025TabArena,
  author       = {Nick Erickson and
                  Lennart Purucker and
                  Andrej Tschalzev and
                  David Holzm{\"{u}}ller and
                  Prateek Mutalik Desai and
                  David Salinas and
                  Frank Hutter},
  title        = {TabArena: {A} Living Benchmark for Machine Learning on Tabular Data},
  journal      = {CoRR},
  volume       = {abs/2506.16791},
  year         = {2025}
}

@book{Szeliski2022Computer,
  author       = {Richard Szeliski},
  title        = {Computer Vision - Algorithms and Applications, Second Edition},
  publisher    = {Springer},
  year         = {2022}
}

@article{Wolpert1996NFL,
  author       = {David H. Wolpert},
  title        = {The Lack of {A} Priori Distinctions Between Learning Algorithms},
  journal      = {Neural Computation},
  volume       = {8},
  number       = {7},
  pages        = {1341--1390},
  year         = {1996}
}

@article{Jiang2025Survey,
  author       = {Jun-Peng Jiang and
                  Si-Yang Liu and
                  Hao-Run Cai and
                  Qi-Le Zhou and
                  Han-Jia Ye},
  title        = {Representation Learning for Tabular Data: {A} Comprehensive Survey},
  journal      = {CoRR},
  volume       = {abs/2504.16109},
  year         = {2025}
}

@inproceedings{Ye2024Towards,
  author       = {Chao Ye and
                  Guoshan Lu and
                  Haobo Wang and
                  Liyao Li and
                  Sai Wu and
                  Gang Chen and
                  Junbo Zhao},
  title        = {Towards Cross-Table Masked Pretraining for Web Data Mining},
  booktitle    = {WWW},
  pages        = {4449--4459},
  year         = {2024}
}

@inproceedings{Gardner2024Tabular8B,
  author       = {Josh Gardner and
                  Juan C. Perdomo and
                  Ludwig Schmidt},
  title        = {Large Scale Transfer Learning for Tabular Data via Language Modeling},
  booktitle    = {NeurIPS},
 pages = {45155--45205},
  year         = {2024}
}

@inproceedings{Yan2024Making,
  author       = {Jiahuan Yan and
                  Bo Zheng and
                  Hongxia Xu and
                  Yiheng Zhu and
                  Danny Z. Chen and
                  Jimeng Sun and
                  Jian Wu and
                  Jintai Chen},
  title        = {Making Pre-trained Language Models Great on Tabular Prediction},
  booktitle    = {ICLR},
  year         = {2024}
}

@inproceedings{Wen2024GTL,
  author       = {Xumeng Wen and
                  Han Zhang and
                  Shun Zheng and
                  Wei Xu and
                  Jiang Bian},
  title        = {From Supervised to Generative: {A} Novel Paradigm for Tabular Deep
                  Learning with Large Language Models},
  booktitle    = {SIGKDD},
  pages        = {3323--3333},
  year         = {2024}
}

@inproceedings{liu2025BETA,
  author       = {Si-Yang Liu and
                  Han-Jia Ye},
  title        = {TabPFN Unleashed: {A} Scalable and Effective Solution to Tabular Classification
                  Problems},
  booktitle      = {ICML},
  year         = {2025}
}

@article{Zhang2025Limix,
  author       = {Xingxuan Zhang and
				 Gang Ren and 
				 Han Yu and
				 Hao Yuan and
				Hui Wang and
				Jiansheng Li and
				Jiayun Wu and
				Lang Mo and
				Li Mao and
				Mingchao Hao and
				Ningbo Dai and
				Renzhe Xu and
				Shuyang Li and
				Tianyang Zhang and
				Yue He and
				Yuanrui Wang and
				Yunjia Zhang and
				Zijing Xu and
				Dongzhe Li and
				Fang Gao and
				Hao Zou and
				Jiandong Liu and
				Jiashuo Liu and
				Jiawei Xu and
				Kaijie Cheng and
				Kehan Li and
				Linjun Zhou and
				Qing Li and
				Shaohua Fan and
				Xiaoyu Lin and
				Xinyan Han and
				Xuanyue Li and
				Yan Lu and
				Yuan Xue and
				Yuanyuan Jiang and
				Zimu Wang and
				Zhenlei Wang and
				Peng Cui},
  title        = {LimiX: Unleashing Structured-Data Modeling Capability for Generalist Intelligence},
  journal      = {CoRR},
  volume       = {abs/2509.03505},
  year         = {2025}
}

@article{Beaglehole2025xRFM,
  author       = {Daniel Beaglehole and
                  David Holzm{\"{u}}ller and
                  Adityanarayanan Radhakrishnan and
                  Mikhail Belkin},
  title        = {{xRFM}: Accurate, scalable, and interpretable feature learning models
                  for tabular data},
  journal      = {CoRR},
  volume       = {abs/2508.10053},
  year         = {2025}
}

@article{Hancock2020Categorical,
  author       = {John T. Hancock and
                  Taghi M. Khoshgoftaar},
  title        = {Survey on categorical data for neural networks},
  journal      = {Journal of Big Data},
  volume       = {7},
  number       = {1},
  pages        = {28},
  year         = {2020}
}

@article{Radhakrishnan2023RFM,
  author       = {Adityanarayanan Radhakrishnan and
                  Daniel Beaglehole and
                  Parthe Pandit and
                  Mikhail Belkin},
  title        = {Mechanism of feature learning in deep fully connected networks and kernel machines that recursively learn features},
  journal      = {CoRR},
  volume       = {abs/2212.13881v3},
  year         = {2023}
}

@article{Garg2025RealTabPFN,
  author       = {Anurag Garg and
                  Muhammad Ali and
                  Noah Hollmann and
                  Lennart Purucker and
                  Samuel M{\"{u}}ller and
                  Frank Hutter},
  title        = {Real-TabPFN: Improving Tabular Foundation Models via Continued Pre-training With Real-World Data},
  journal      = {CoRR},
  volume       = {abs/2507.03971},
  year         = {2025}
}

@inproceedings{Wen2020BatchEnsemble,
  author       = {Yeming Wen and
                  Dustin Tran and
                  Jimmy Ba},
  title        = {BatchEnsemble: an Alternative Approach to Efficient Ensemble and Lifelong
                  Learning},
  booktitle    = {ICLR},
  year         = {2020}
}

@inproceedings{Zeng2025TabFlex,
  author       = {Yuchen Zeng and
                  Tuan Dinh and
                  Wonjun Kang and
                  Andreas C. Mueller},
  title        = {TabFlex: Scaling Tabular Learning to Millions with Linear Attention},
  booktitle    = {ICML},
  year         = {2025}
}

@inproceedings{Mueller2020MotherNet,
  author       = {Andreas C. Mueller and
                  Carlo Curino and
                  Raghu Ramakrishnan},
  title        = {MotherNet: Fast Training and Inference via Hyper-Network Transformers},
  booktitle    = {ICLR},
  year         = {2025}
}

@article{Ma2024TabDPT,
  author       = {Junwei Ma and
                  Valentin Thomas and
                  Rasa Hosseinzadeh and
                  Hamidreza Kamkari and
                  Alex Labach and
                  Jesse C. Cresswell and
                  Keyvan Golestan and
                  Guangwei Yu and
                  Maksims Volkovs and
                  Anthony L. Caterini},
  title        = {TabDPT: Scaling Tabular Foundation Models},
  journal      = {CoRR},
  volume       = {abs/2410.18164},
  year         = {2024}
}

@inproceedings{Cai2025Temporal,
  author       = {Hao-Run Cai and
                  Han-Jia Ye},
  title        = {Understanding the Limits of Deep Tabular Methods with Temporal Shift},
  booktitle    = {ICML},
  year         = {2025}
}

@inproceedings{Tschalzev2025Unreflected,
    title={Unreflected Use of Tabular Data Repositories Can Undermine Research Quality},
    author={Andrej Tschalzev and Lennart Purucker and Stefan L{\"u}dtke and Frank Hutter and Christian Bartelt and Heiner Stuckenschmidt},
    booktitle={ICLR},
    year={2025},
}

@article{Ye2025Closer,
  author       = {Han-Jia Ye and
                  Si-Yang Liu and
                  Wei-Lun Chao},
  title        = {A Closer Look at TabPFN v2: Understanding Its Strengths and Extending Its Capabilities},
  journal      = {CoRR},
  volume       = {abs/2502.17361},
  year         = {2025}
}

@article{Rubachev2025On,
  author       = {Ivan Rubachev and
                  Akim Kotelnikov and
                  Nikolay Kartashev and
                  Artem Babenko},
  title        = {On Finetuning Tabular Foundation Models},
  journal      = {CoRR},
  volume       = {abs/2506.08982},
  year         = {2025}
}

@inproceedings{Breugel2024Position,
  author       = {Boris van Breugel and
                  Mihaela van der Schaar},
  title        = {Position: Why Tabular Foundation Models Should Be a Research Priority},
  booktitle    = {ICML},
  year         = {2024}
}

\end{document}